\documentclass[runningheads]{llncs}
\usepackage{eccv}

\usepackage{eccvabbrv}

\usepackage{graphicx}
\usepackage{booktabs}

\usepackage[accsupp]{axessibility}  %

\usepackage{graphicx}
\usepackage{graphbox}
\usepackage{caption}
\usepackage{rotating}           %
\usepackage{pdflscape}

\usepackage{mathtools}
\usepackage{mathrsfs}
\usepackage{bm}
\usepackage{bbm}
\usepackage{nicefrac}

\usepackage{booktabs}
\usepackage{multirow}
\usepackage{tabularx}
\usepackage{array}
\usepackage{colortbl}
\usepackage{hhline}
\usepackage{arydshln}          %
\usepackage{diagbox}
\usepackage{siunitx}    %

\usepackage{enumitem}

\usepackage{bm} 

\usepackage{algorithm}
\usepackage{algorithmic}

\usepackage{graphicx}
\usepackage{subcaption}
\usepackage{float}
\usepackage{wasysym}
\usepackage{url}
\usepackage{xspace}

\usepackage{thmtools}
\usepackage{thm-restate}

\usepackage{float}
\usepackage{footmisc}
\usepackage{makecell}

\newcommand{\x}{\bm{x}}

\newcommand\myldots{\ifmmode\ldots\else\makebox[0.5em][c]{.\hfil.\hfil.}\thinspace\fi}
\newcommand\mycdots{\ifmmode\cdots\else\makebox[0.5em][c]{.\hfil.\hfil.}\thinspace\fi}

\usepackage{xcolor}

\usepackage{hyperref}

\usepackage{orcidlink}

\begin{document}

\title{D2PO: Optimizing Diffusion Samplers via Dynamic Preference}

\author{Jinkyu Kim$^*$ \inst{1} \and%
Jinyoung Choi\thanks{indicates equal contribution.}\inst{3} \and%
Bohyung Han\inst{1, 2}}%

\authorrunning{J.~Kim et al.}

\institute{
${}^1$ECE \& ${}^2$IPAI, Seoul National University, Korea \\
  ${}^3$AIGS, Ulsan National Institute of Science and Technology, Korea  }

\maketitle

\begin{abstract}

We propose D2PO (Dynamic Direct Preference Optimization), a principled framework for optimizing diffusion sampling policies with respect to timestep schedules and classifier-free guidance (CFG) weights.
Our work is motivated by a fundamental limitation of existing student-teacher regression frameworks; low-NFE student samplers are trained to mimic high-NFE teachers, often sacrificing high-frequency texture fidelity while preserving coarse global structures, thereby misaligning the sampler with perceptual quality.
D2PO addresses this challenge by reformulating sampler optimization as a preference-based alignment problem, leveraging the Direct Preference Optimization (DPO) framework.
To make DPO applicable to diffusion samplers, we model the sampling policy as an energy-based model (EBM), transforming preference comparisons into tractable energy differences.
We further introduce a novel energy formulation derived directly from the pretrained score network, enabling preference evaluation in perturbed spaces that jointly capture structural consistency and fine-grained details.
Moreover, we introduce dynamic preferences, where the preferred samples used for alignment progressively improve as the sampling policies are learned.
This self-improving mechanism replaces rigid static teacher supervision with an iterative, preference-guided refinement process, providing progressively stronger alignment signals.
Extensive experiments demonstrate that D2PO aligns diffusion samplers with perceptual quality more faithfully, unlocking the full potential of high-quality teachers and consistently outperforming conventional regression-based schedulers under low-NFE constraints.


\end{abstract}

\section{Introduction}
\label{sec:intro}

Diffusion Probabilistic Models (DPMs)~\cite{sohl2015deep,ho2020denoising,song2021scorebased} have achieved unprecedented fidelity in high-resolution image synthesis, text-to-image generation~\cite{rombach2022high,dhariwal2021diffusion}, and video generation~\cite{ho2022video,singer2022make,zhou2022magic,wang2023modelscope}. 
However, this performance comes at a substantial computational cost. 
DPMs are inherently iterative, requiring many function evaluations (NFE) during sampling, which makes high-quality generation expensive and limits practical deployment.

A broad range of approaches has been explored to mitigate this bottleneck, including accelerated numerical solvers~\cite{dockhorn2022genie,lu2022fastode,lu2023dpm,liu2022pseudo,zhao2023unipc,zhang2023lookahead,choi2025rx-dpm}, few-step knowledge distillation~\cite{salimans2022progressive,song2023consistency,kim2024consistency,zheng2024trajectory,salimans2024moment,yin2024onestep,yin2024improved,zhou2024score,zhou2025adversarial}, architectural modifications~\cite{ma2024learning,ye2024training}, and training-time improvements~\cite{kingma2021variational,vahdat2021score,xiao2021tackling,kang2024ogdm}. 
More recently, directly optimizing the sampling policy parameters---such as timestep schedules~\cite{li2023autodiffusion,watson2021learning,tong2024learning,sabour2024align,xue2024accelerating,frankels4s}, classifier-free guidance weights~\cite{galashov2025learn}, and high-order solver coefficients~\cite{frankels4s,zhang2024iia,wang2026image}---has emerged as a critical direction for acceleration.

\begin{figure}[t]
  \centering
  
  \begin{subfigure}{\linewidth}
  \centering
  \footnotesize ``Two birds that are sitting in a marsh area.''

  \vspace{2mm}

  \begin{minipage}{0.23\linewidth}
    \includegraphics[width=\linewidth]{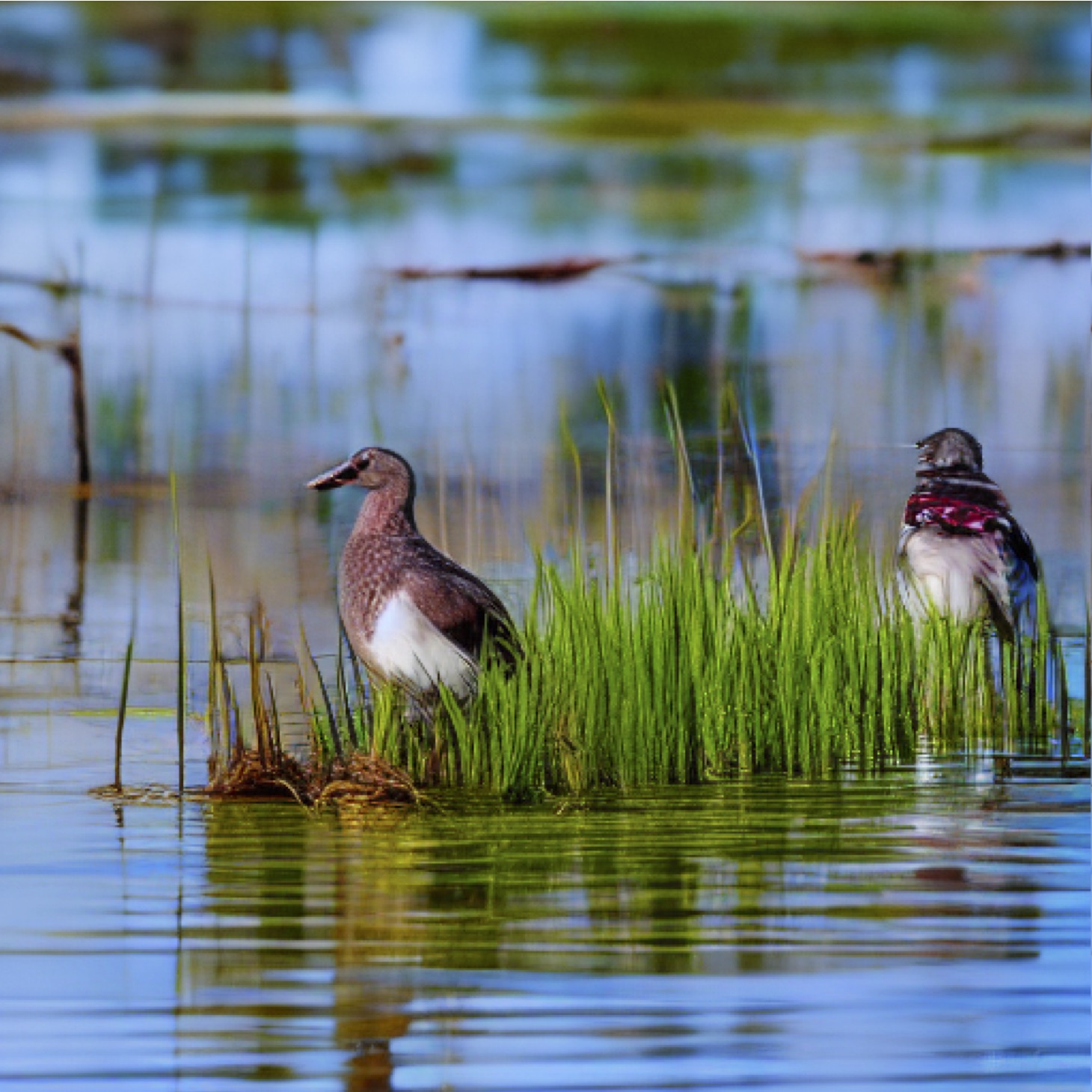}
    \centering\footnotesize $\Delta=1$
  \end{minipage}
  \hspace{1mm}  %
  \begin{minipage}{0.23\linewidth}
    \includegraphics[width=\linewidth]{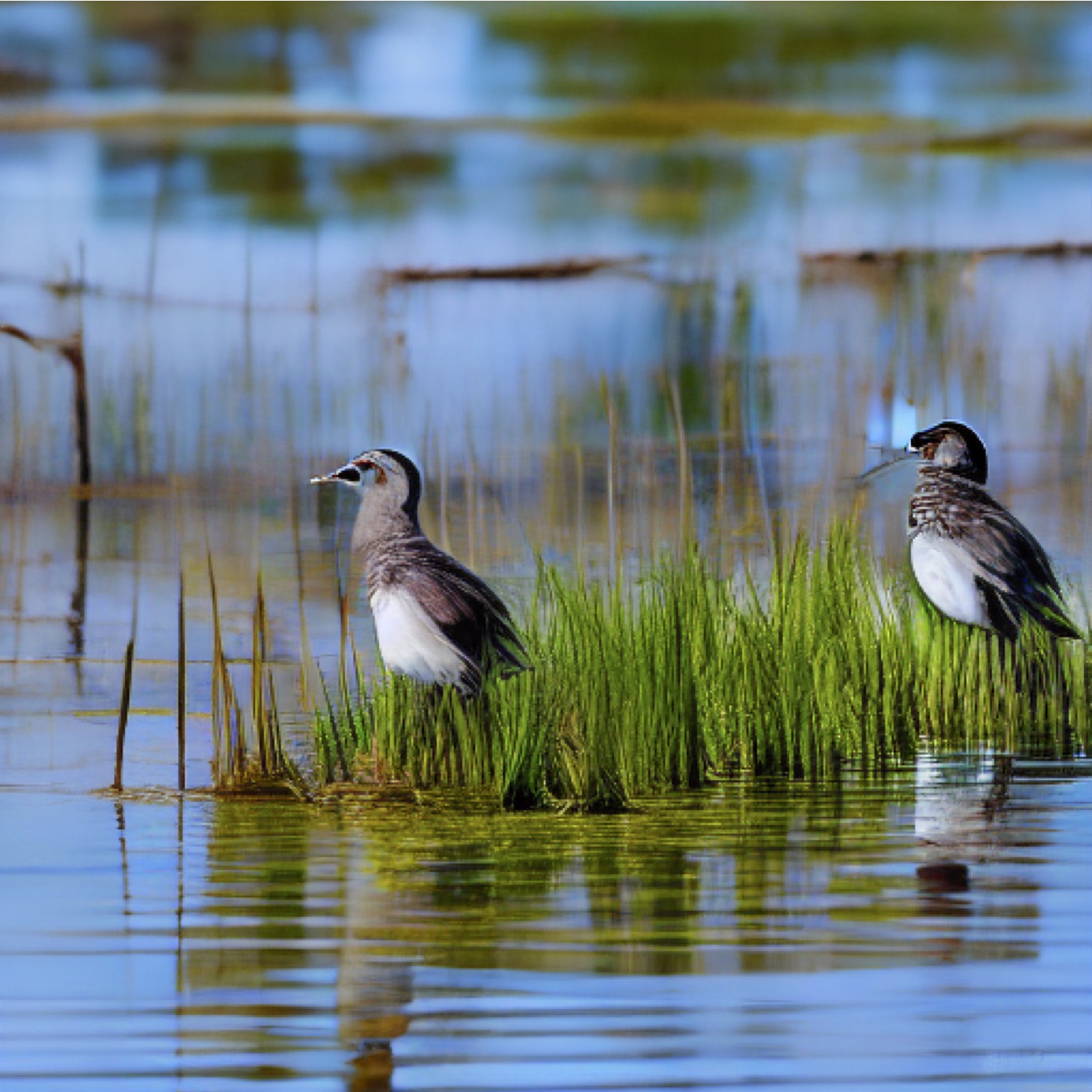}
    \centering\footnotesize $\Delta=2$
  \end{minipage}
  \hspace{1mm}
  \begin{minipage}{0.23\linewidth}
    \includegraphics[width=\linewidth]{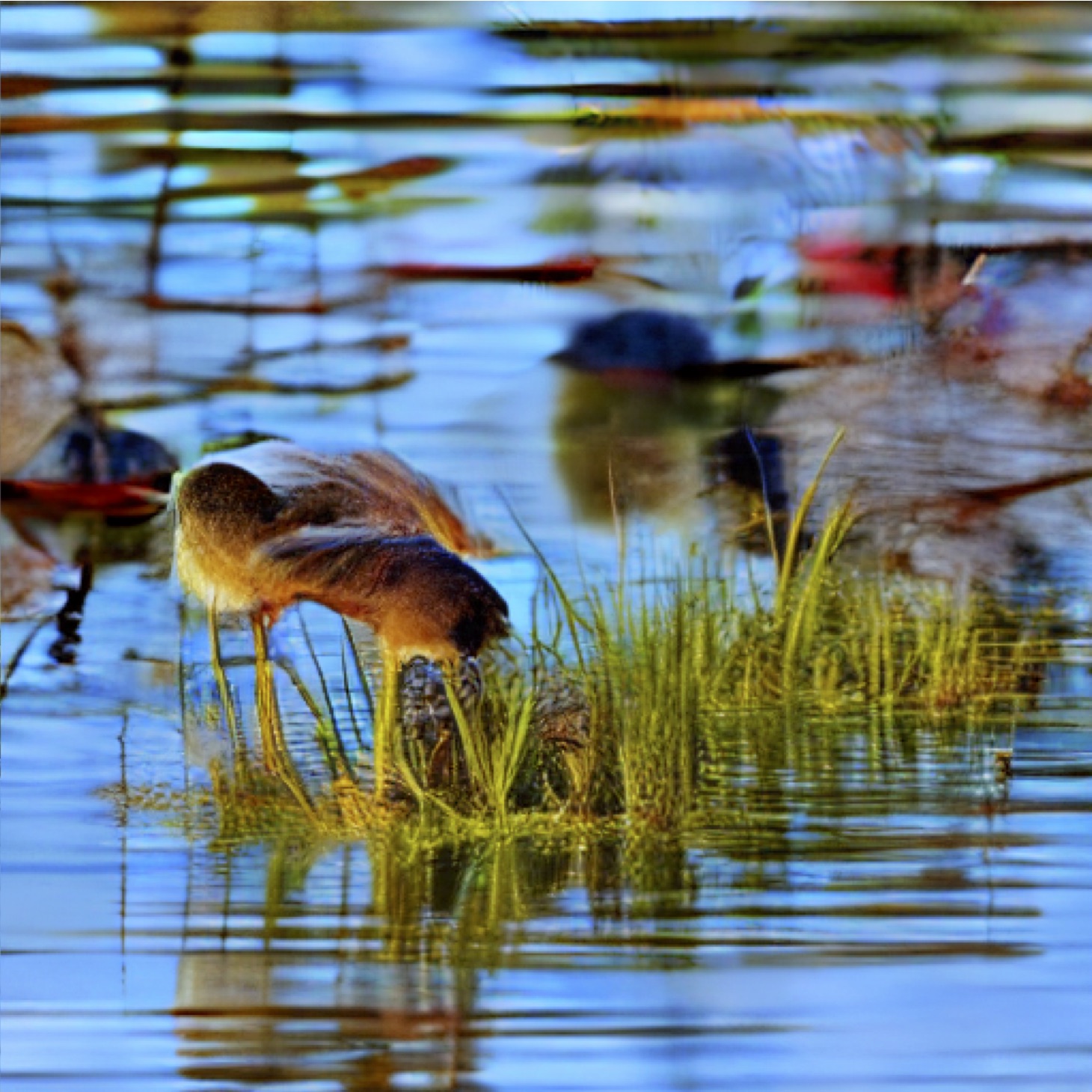}
    \centering\footnotesize $\Delta=3$
  \end{minipage}
  \hspace{1mm}
  \begin{minipage}{0.23\linewidth}
    \includegraphics[width=\linewidth]{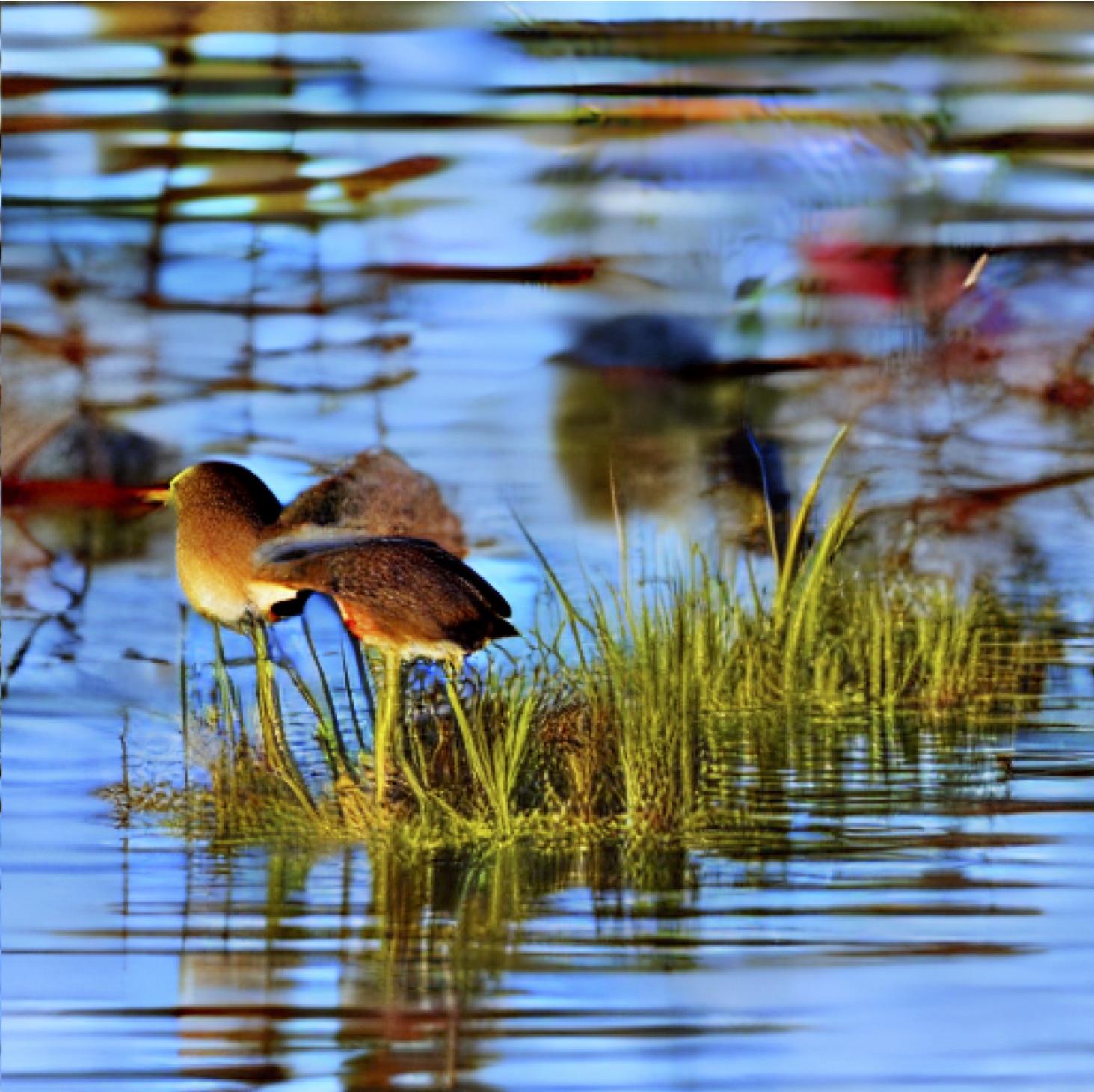}
    \centering\footnotesize $\Delta=4$
  \end{minipage}
\end{subfigure}

 \vspace{2mm}
     \begin{subfigure}{\linewidth}
    \centering
    \centering
    \footnotesize ``There is a small bus with several people standing next to it."
     \vspace{2mm}

  \begin{minipage}{0.23\linewidth}
    \includegraphics[width=\linewidth]{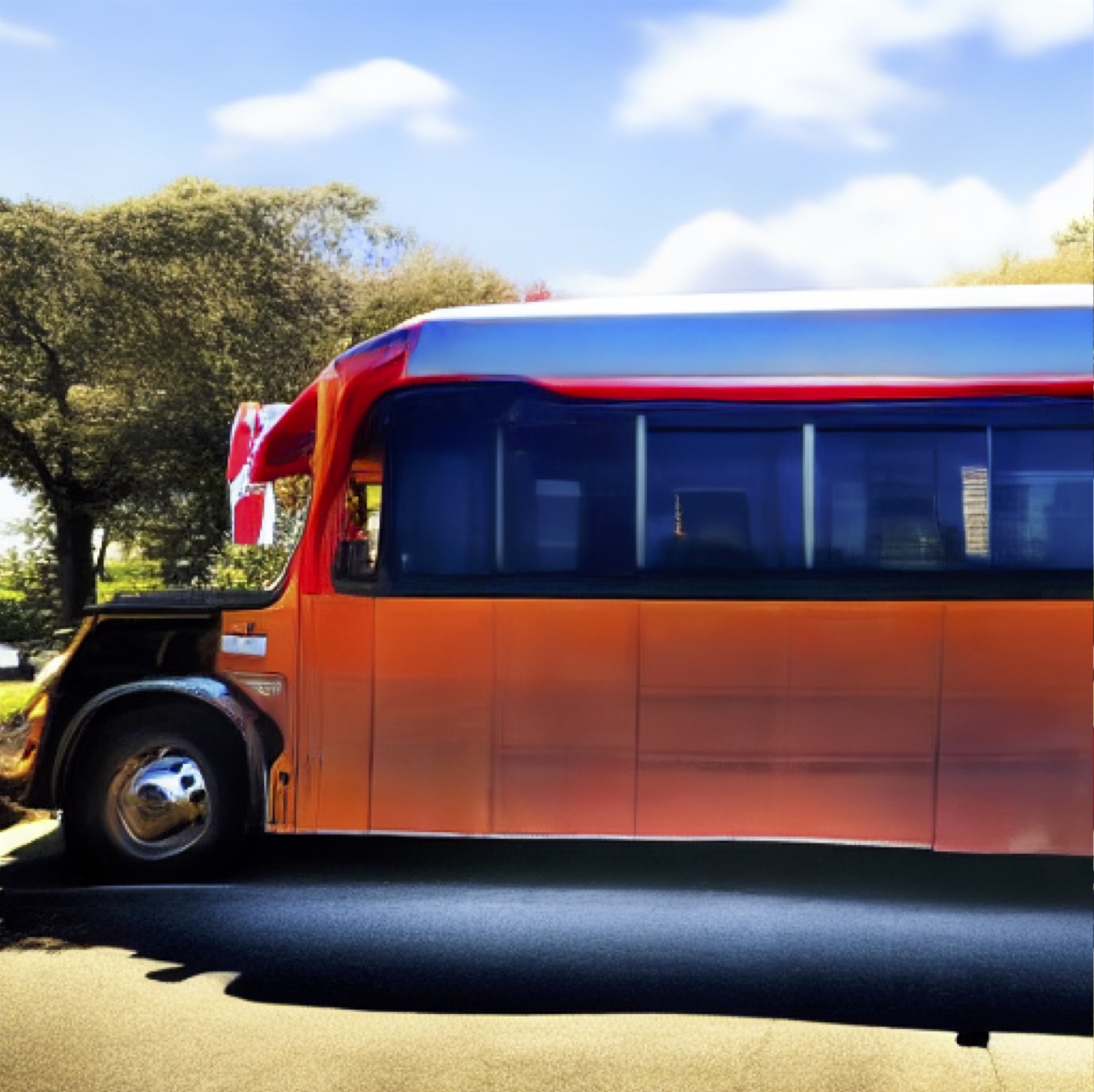}
    \centering\footnotesize $\Delta=1$
  \end{minipage}
  \hspace{1mm}  %
  \begin{minipage}{0.23\linewidth}
    \includegraphics[width=\linewidth]{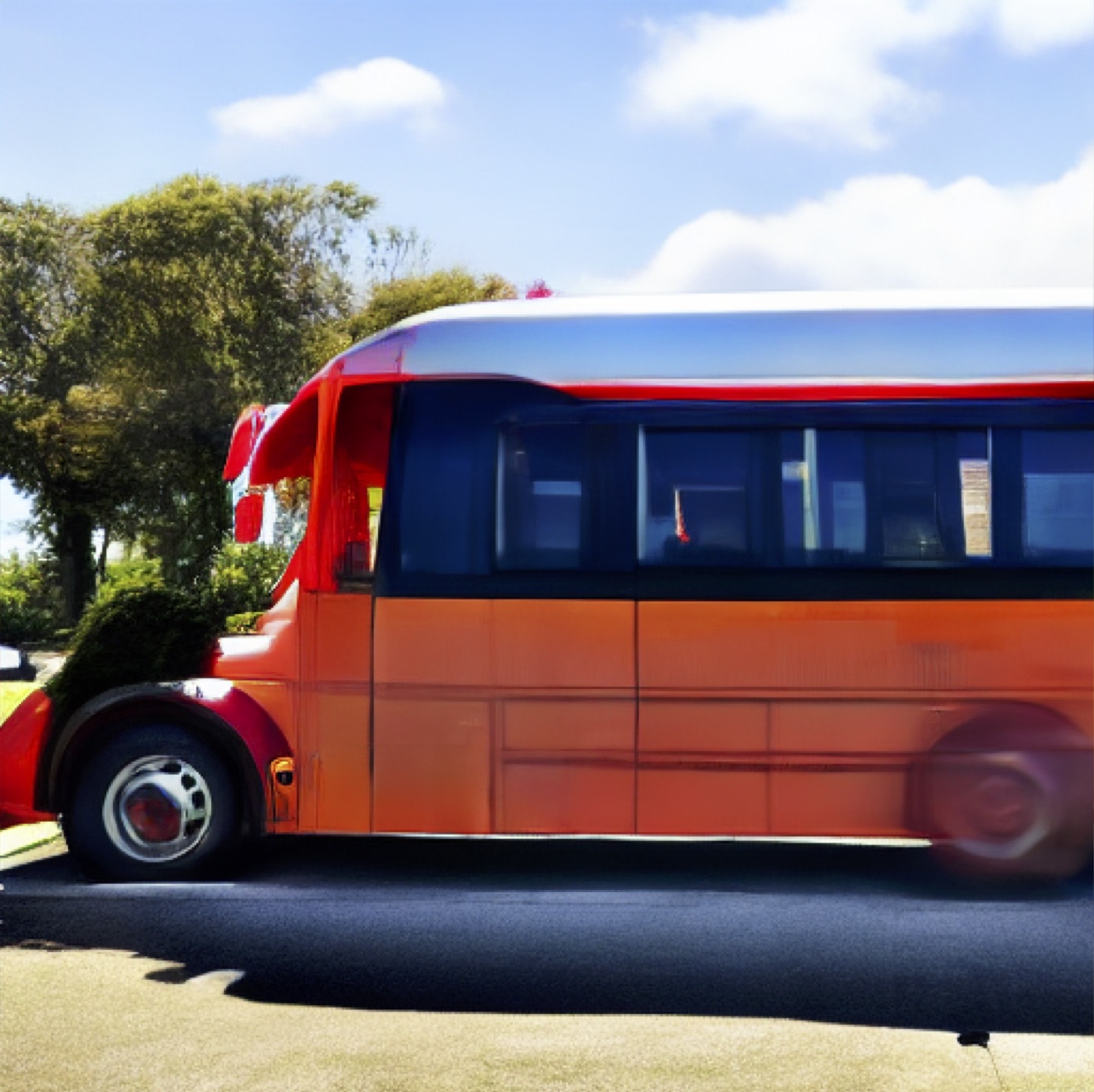}
    \centering\footnotesize $\Delta=2$
  \end{minipage}
  \hspace{1mm}
  \begin{minipage}{0.23\linewidth}
    \includegraphics[width=\linewidth]{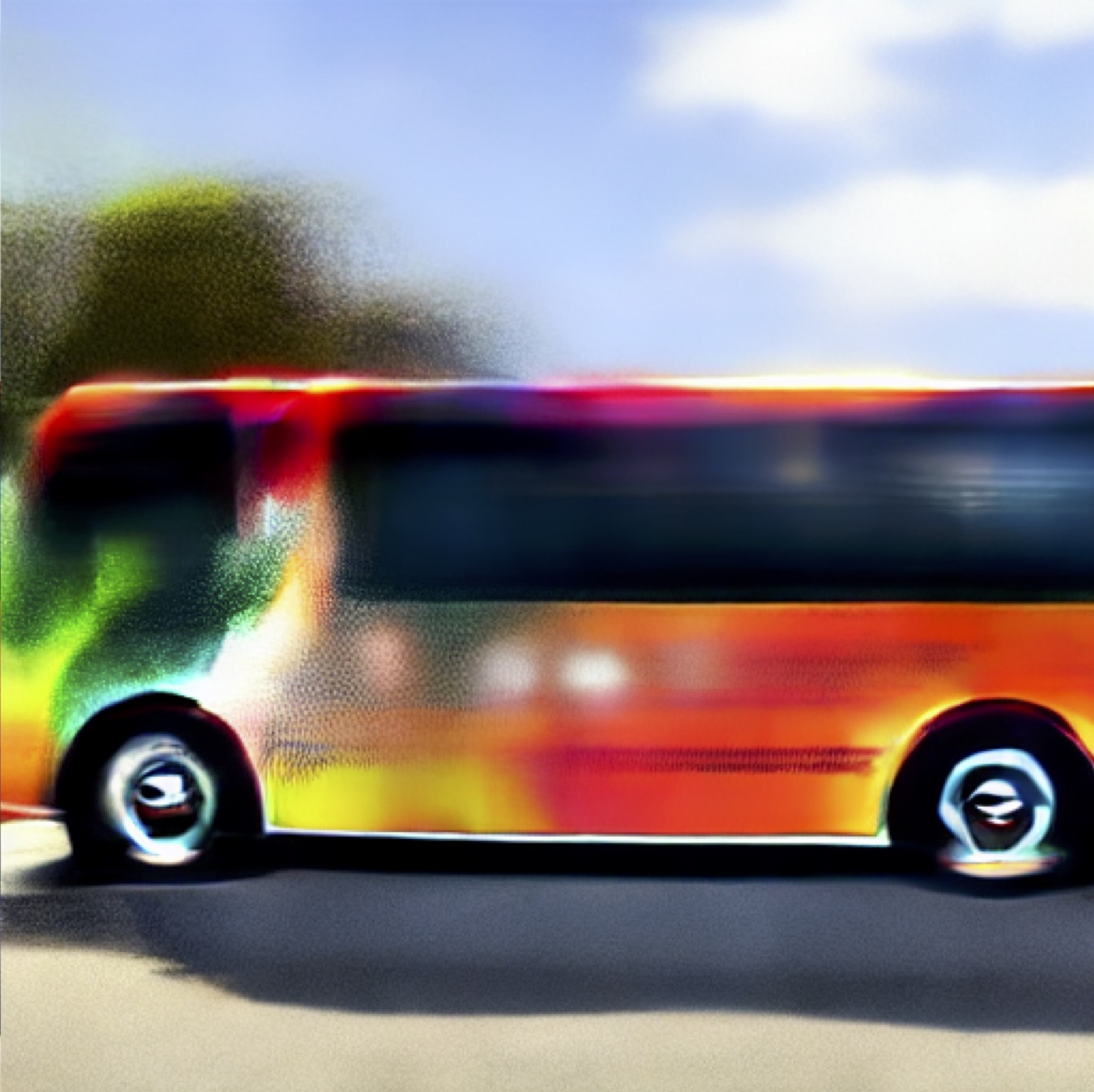}
    \centering\footnotesize $\Delta=3$
  \end{minipage}
  \hspace{1mm}
  \begin{minipage}{0.23\linewidth}
    \includegraphics[width=\linewidth]{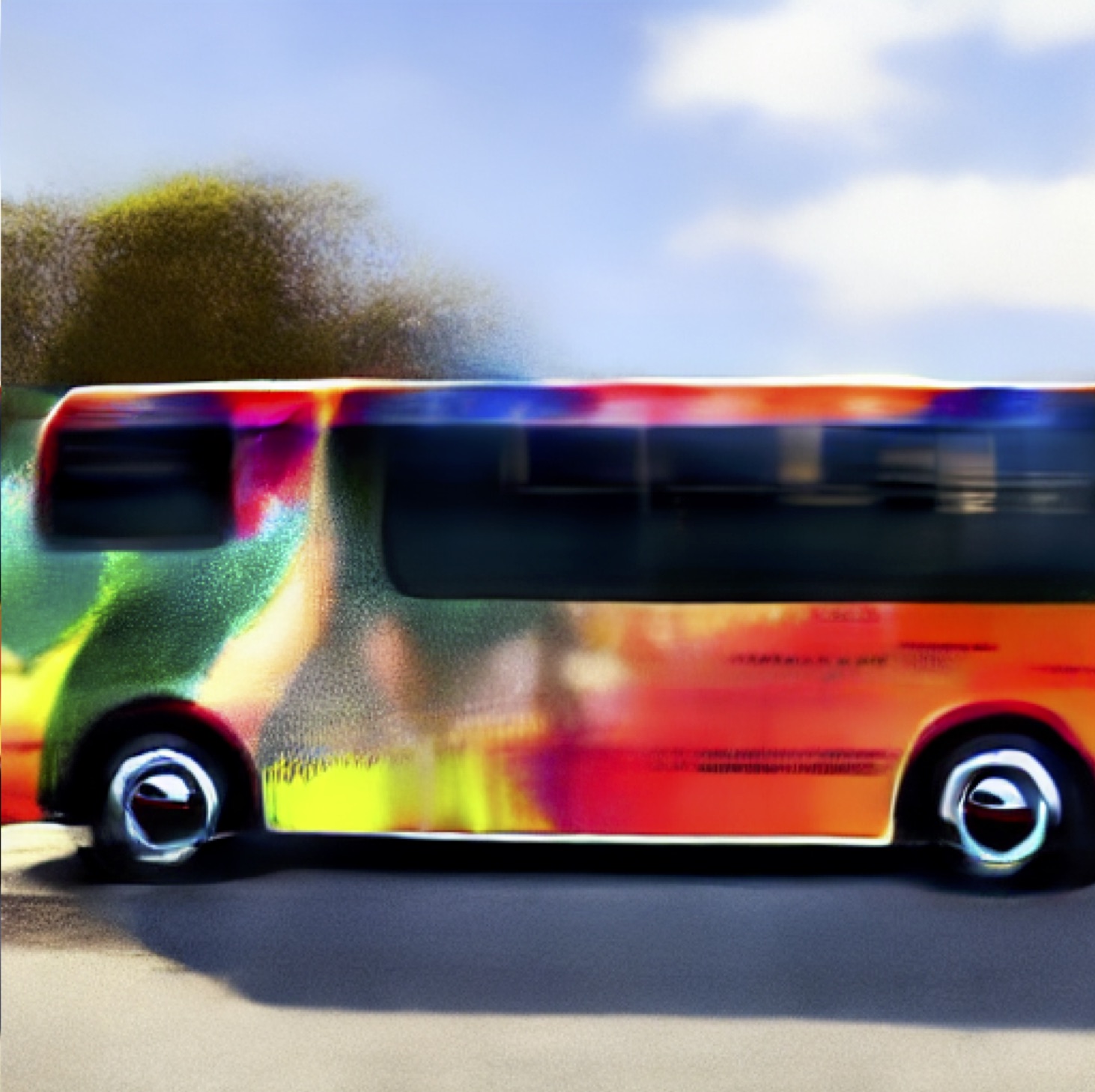}
    \centering\footnotesize $\Delta=4$
  \end{minipage}
\end{subfigure}

  \caption{
Qualitative evidence of the performance bottleneck in LD3~\cite{tong2024learning}. %
 All images are generated by the same model with $\text{NFE}=4$.
The columns show the impact of increasing the NFE gap ($\Delta = T - S$) {between the teacher ($T$) and the student ($S$).}
While a small gap ($\Delta=1$) yields high-quality outputs, larger gaps (up to $\Delta=4$) lead to severe artifacts, demonstrating LD3's inability to leverage high-fidelity teachers.
}
 \vspace{-3mm}
  \label{fig:imitation_gap}
\end{figure}

In optimizing these sampler parameters, prior works typically adopt either a distribution or an instance perspective. 
Specifically, one line of work optimizes distribution-level metrics, such as FID~\cite{li2023autodiffusion} or KID~\cite{watson2022learning} over large sample sets, but such population-level objectives yield weak, high-variance gradients for the low-dimensional sampler parameters.
Alternatively, instance-wise distillation methods~\cite{tong2024learning,frankels4s} regress a low-NFE student sampler onto the trajectories or outputs of a fixed high-NFE teacher via $\ell_2$ or LPIPS~\cite{zhang2018unreasonable} losses. 
Although effective when the student-teacher gap is moderate, this regression paradigm exhibits a structural limitation when aggressive acceleration is desired.

{When the NFE gap between the student and teacher becomes substantial---a common scenario when striving for maximum acceleration---this regression-based objective forces the student sampler to approximate a high-fidelity trajectory within its restricted capacity, compromising generation quality.}
This structural constraint often suppresses high-frequency textures and fine-grained details, preventing the student from fully benefiting from stronger teachers. 
We empirically validate this phenomenon in \cref{fig:imitation_gap}; as the teacher NFE increases while the student NFE remains fixed, the perceptual quality degrades, even for a state-of-the-art scheduler (LD3~\cite{tong2024learning}).
This degradation directly reflects the structural rigidity of fixed-teacher regression, where the student is forced to prioritize low-level alignment over perceptual quality, failing to discover more effective sampling paths.

To address this limitation, we reinterpret sampler optimization as a preference-based alignment problem rather than a regression-to-teacher task. 
We introduce D2PO (Dynamic Direct Preference Optimization), a framework inspired by DPO but adapted to diffusion sampling policies. %
Applying DPO to diffusion samplers is non-trivial because the marginalized log-probability is intractable.
To obtain a tractable surrogate, we model the policy-induced distribution as an Energy-Based Model (EBM). 
We define the energy using a novel score-based distance that measures discrepancies between samples %
leveraging the pretrained diffusion score model. 
By comparing score predictions across multiple noise levels, this metric captures both structural and high-frequency differences that conventional perceptual metrics fail to reflect.

D2PO replaces the static teacher framework with a dynamic reference mechanism that evolves alongside the student policy.
Specifically, at each training step, the preference pair is constructed by executing the current policy under two different computational budgets: the losing sample is generated using a fast, sparse timestep schedule, while the winning sample is obtained via a denser, more refined schedule of the same policy.
Instead of imitating an external, immutable target, the student is encouraged to align with its own high-quality, dense-schedule approximation.
This dynamic preference loop eliminates the fixed error floor inherent in static distillation and implicitly drives the sampler to minimize discretization errors, thereby promoting highly accurate and self-improving sampling trajectories.

Our contributions are summarized as follows:
\begin{itemize}
  \item[$\bullet$] We propose D2PO, a preference-based framework for optimizing diffusion samplers, establishing a tractable alignment objective by modeling the deterministic policy as an energy-based surrogate.
  \item[$\bullet$] We formulate a novel score-based energy metric derived from the pretrained score network, providing a multi-scale learning signal that captures fine-grained textural and structural details beyond conventional perceptual losses.
  \item[$\bullet$] We introduce a dynamic preference mechanism that replaces static teacher supervision with a refinement-based target, enabling continual self-improvement without being bounded by a fixed residual error.
  \item[$\bullet$] We comprehensively validate that D2PO learns superior sampling policies, outperforming state-of-the-art distillation-based baselines under various experimental settings.
  \end{itemize}

\section{Related Work}
\label{sec:related_work}

\subsection{Optimizing diffusion sampling parameters}
\label{subsec:opt_sampler}
Since the trajectory of time steps profoundly impacts generation quality under a fixed computational budget, substantial research has focused on finding optimal sampling schedules.
Early heuristic approaches, such as EDM~\cite{karras2022elucidating}, employ polynomial spacing to densify steps near the clean data manifold, while Watson \etal~\cite{watson2021learning} introduce a dynamic programming framework to search for optimal discrete schedules that maximize log-likelihood.
Analytic-DPM~\cite{bao2022analytic} improves efficiency by deriving training-free, optimal analytical forms for reverse variances directly from the pretrained score network, while obtaining the corresponding optimal trajectory via dynamic programming~\cite{watson2021learning}.
To automate and generalize schedule optimization, AutoDiffusion~\cite{li2023autodiffusion} employs an evolutionary search targeted at minimizing FID, while DDSS~\cite{watson2022learning} optimizes sampler parameters via direct sample-quality feedback such as KID~\cite{binkowski2018demystifying}.

Another line of work derives analytical error bounds or geometric properties of ODE/SDE trajectories to optimize time discretization.
Methods such as those by Chen \etal~\cite{chen2024adaptive,chen2024trajectory}, AYS~\cite{sabour2024align}, and Xue \etal~\cite{xue2024accelerating} dynamically adjust step sizes based on trajectory curvature or upper bounds of solver errors.
More recently, LD3~\cite{tong2024learning} adopts a relaxed matching objective to learn discretized trajectories through student-teacher regression.

Beyond timestep optimization, recent literature explores tuning other sampling parameters to further accelerate inference.
For instance, Galashov \etal~\cite{galashov2025learn} learn time-dependent CFG weights via a self-consistency objective.
Similarly, S4S~\cite{frankels4s} optimizes solver coefficients at each step using teacher-student matching.
Extending this direction, ConsistencySolver~\cite{wang2026image} employs a learnable high-order solver to dynamically predict optimal integration coefficients.%

\subsection{Aligning pretrained models with preferences}
\label{subsec:align}

Driven by the limitations of predefined training objectives, aligning generative models directly with pairwise human or AI preferences has emerged as a dominant paradigm.
This approach originated in large language models via Reinforcement Learning from Human Feedback (RLHF)~\cite{ouyang2022training}, which optimizes policies using a separate reward model.
To simplify this multi-stage pipeline, Direct Preference Optimization (DPO)~\cite{rafailov2024direct} integrates the reward implicitly into the classification loss, enabling stable and direct policy updates.
Subsequent self-play frameworks like SPIN~\cite{chen2024self} further remove the need for preference annotations, generating negatives from the model itself and contrasting them with SFT responses.

Recently, this preference alignment paradigm has been actively adapted to text-to-image diffusion and flow-matching models to enhance visual quality and aesthetic appeal.
Standard post-training methods, including DPO-style formulations~\cite{wallace2024diffusion,liang2025aesthetic,yang2024using,yuan2024self} and online reinforcement learning variants~\cite{black2023training,fan2024reinforcement,liu2025flow}, predominantly focus on fine-tuning the foundational weights of the denoiser or velocity networks.
While effective, optimizing high-dimensional model parameters is computationally expensive and risks degrading the quality of outputs.
In contrast, D2PO keeps the generative backbone frozen and exclusively optimizes the low-dimensional sampling policy, offering a highly lightweight, orthogonal, and complementary solution to existing weight-tuning approaches.

\section{Preliminaries}
\label{sec:preliminaries}

This section briefly reviews score-based diffusion models and Direct Preference Optimization (DPO), which provide the theoretical foundation for our dynamic sampler optimization framework.

\subsection{Diffusion probabilistic models}
\label{sec:dpm_pre}

We consider score-based diffusion models~\cite{ho2020denoising,song2021scorebased}, which transform a data distribution $p_{\mathrm{real}}(\x_0)$ into a tractable prior through a gradual noising process. 
Specifically, we adopt the Variance Preserving (VP) stochastic differential equation (SDE), which is given by
\begin{align}
d\x_t = -\frac{1}{2}\beta(t)\x_t\, dt + \sqrt{\beta(t)}\, d\mathbf{w}_t,
\label{eq:fwd_sde}
\end{align}
where $\mathbf{w}_t$ denotes a standard Wiener process and $t \in [0,T]$. 
This forward process admits a closed-form marginal:
\begin{align}
p_t(\x_t \mid \x_0) 
= \mathcal{N}\!\left(\x_t; \alpha_t \x_0, \sigma_t^2 \mathbf{I}\right),
\end{align}
where
\begin{align}
\alpha_t = \exp\!\left(-\frac{1}{2}\!\int_0^t \beta(s)\, ds\right)
\qquad \text{and} \qquad
\sigma_t^2 = 1 - \alpha_t^2.
\end{align}

To construct the reverse process, a neural network $s_\theta(\x_t,t)$ is trained to approximate the score function of the marginal distribution, \ie, $s(\x_t, t) = \nabla_{\x_t}\log p_t(\x_t)$.
This is achieved via denoising score matching (DSM), which minimizes
\begin{align}
\label{eq:dsm_loss}
\mathcal{L}_{\mathrm{DSM}}
=
\int_0^T 
\lambda(t)\,
\mathbb{E}_{\x_0 \sim p_{\mathrm{real}},\, \x_t \sim p_t(\cdot|\x_0)}
\left[
\bigl\|
s_\theta(\x_t,t)
-
\nabla_{\x_t}\log p_t(\x_t|\x_0)
\bigr\|_2^2
\right]
dt,
\end{align}
where
\begin{align}
\nabla_{\x_t}\log p_t(\x_t|\x_0)
=
-\frac{\x_t - \alpha_t \x_0}{\sigma_t^2}.
\end{align}

Once $s_\theta$ is trained, samples are generated by solving the corresponding reverse-SDE from $t=T$ to $t=0$, which is given by
\begin{align}
d\x_t
=
\Bigl[
-\frac{1}{2}\beta(t)\x_t
-
\beta(t)\, s_\theta(\x_t,t)
\Bigr] dt
+
\sqrt{\beta(t)}\, d\bar{\mathbf{w}}_t,
\label{eq:rev_sde}
\end{align}
where $\bar{\mathbf{w}}_t$ is a reverse-time Wiener process.

This score-based formulation is equivalent to the noise-prediction parameterization $\epsilon_\theta(\x_t,t)$ in DDPM~\cite{ho2020denoising}. 
The two representations are related by
\begin{align}
s_\theta(\x_t,t)
=
-\frac{\epsilon_\theta(\x_t,t)}{\sigma_t}.
\end{align}

\subsection{Direct preference optimization}
\label{sec:dpo_pre}

We build upon Direct Preference Optimization (DPO)~\cite{rafailov2024direct}, a framework for aligning generative policies with preference data. 
DPO provides a closed-form solution to the KL-regularized reward maximization problem commonly used in RLHF~\cite{ouyang2022training}, which is defined as
\begin{align}
\max_{\pi}
\;
\mathbb{E}_{\mathbf{x} \sim \pi(\cdot|c)}
\bigl[
r(\mathbf{x}, c)
\bigr]
-
\beta
D_{\mathrm{KL}}
\!\left(
\pi(\cdot|c)
\,\|\, 
\pi_{\mathrm{ref}}(\cdot|c)
\right),
\label{eq:rlhf_obj}
\end{align}
where $\pi_{\mathrm{ref}}$ is a reference policy and $\beta (>0)$ controls the strength of regularization.

Rather than explicitly learning a reward model $r(\mathbf{x},c)$ under the context $c$ and performing reinforcement learning, DPO leverages preference pairs $(\mathbf{x}_w, \mathbf{x}_l)$, where $\mathbf{x}_w$ is preferred over $\mathbf{x}_l$.
Specifically, by analyzing the optimal solution of Eq.~\eqref{eq:rlhf_obj}, the reward difference between two samples can be expressed via the log-likelihood ratio of the optimal policy relative to the reference policy. %
This leads to a logistic classification objective on preference pairs as
\begin{align}
\label{eq:dpo_general}
\mathcal{L}_{\mathrm{DPO}}(\phi)
=
-
\mathbb{E}_{\mathcal{D}}
\left[
\log
\sigma
\!\left(
\beta
\Bigl(
\log \frac{\pi_\phi(\mathbf{x}_w|c)}{\pi_{\mathrm{ref}}(\mathbf{x}_w|c)}
-
\log \frac{\pi_\phi(\mathbf{x}_l|c)}{\pi_{\mathrm{ref}}(\mathbf{x}_l|c)}
\Bigr)
\right)
\right],
\end{align}
where $\sigma(\cdot)$ denotes the sigmoid function.
This formulation eliminates the need for an explicit reward model, directly updating the policy $\pi_\phi$ to maximize the relative log-likelihood of preferred samples over unpreferred ones while remaining anchored to the reference policy.

\section{D2PO: Dynamic Direct Preference Optimization}
\label{sec:d2po}

\subsection{Problem formulation}
\label{sub:problem}

We aim to optimize the sampling policy of a pretrained diffusion model  $s_\theta$ by learning a set of sampler parameters $\phi = \{\mathcal{S}, \boldsymbol{\omega}\}$,  where $\mathcal{S}$ and $\boldsymbol{\omega}$ denote the timestep schedule and  the per-step classifier-free guidance (CFG) weights, respectively.
Given a prompt $c$ and an initial noise $\x_T \sim \mathcal{N}(\mathbf{0},\mathbf{I})$, our sampler, \ie, ODE solver, deterministically generates an image as
\begin{equation}
\x_\phi = \Phi_\phi(\x_T, c; \theta),
\end{equation}
where $\Phi_\phi$ denotes a fixed numerical solver parameterized by $\phi$.
Although $\Phi_\phi$ is deterministic, it induces a conditional distribution over generated images through the randomness of the initial noise $\x_T$:
\begin{equation}
q_\phi(\x | c) = \int \delta(\x - \Phi_\phi(\x_T,c ; \theta))\, p(\x_T)\, d\x_T,
\end{equation}
where $\delta(\cdot)$ denotes the Dirac delta function.

Our goal is to align $q_\phi(\x | c)$ with perceptually preferred outputs using pairwise preference tuples $(c, \x_T, \x_w, \x_l)$,  where $\x_w$ is preferred over $\x_l$.
A central challenge is that $q_\phi(\x | \x_T, c) = \delta(\x - \x_\phi)$ is degenerate, rendering its likelihood non-differentiable and direct preference optimization ill-posed.
We address this by introducing a tractable surrogate formulation.

\subsection{Energy-based surrogate for deterministic policies}
\label{subsec:ebm}

To apply Direct Preference Optimization (DPO) to our sampler, we replace $q_\phi(\x | \x_T, c)$ with a smooth surrogate policy $\pi_\phi(\x | c, \x_T)$, defined as an Energy-Based Model (EBM):
\begin{equation}
\pi_\phi(\x | c, \x_T)
=
\frac{1}{Z(\phi,c,\x_T)}
\exp\!\left(-\alpha E(\x; \phi, c, \x_T)\right),
\label{eq:ebm}
\end{equation}
where $Z$ is the partition function and $\alpha~(>0)$ a temperature parameter.
We define the energy as a distance to the sampler output, which is given by
\begin{equation}
E(\x; \phi, c, \x_T) \equiv d(\x, \x_\phi),
\label{eq:energy_def_policy}
\end{equation}
where $d(\cdot, \cdot)$ is a predefined distance metric.

This surrogate assigns high probability to images close to the sampler's output $\x_\phi$ and smoothly decays as the proximity decreases.
Such functional relaxations are commonly employed to bypass the non-differentiability of objective functions for optimization.
For example, score-based models apply Gaussian perturbations---mathematically forming EBMs with $\ell^2$ energy---to resolve undefined gradients. 
Soft Actor-Critic~\cite{haarnoja2018soft} and the Gumbel--Softmax~\cite{jang2017categorical} use probabilistic relaxations on discrete policies and operations to enable backpropagation. 
Our surrogate plays a similar role for generative optimization: it replaces the non-differentiable Dirac delta function with a smooth landscape whose mode coincides with $\x_\phi$, thereby binding updates to the true generative process while keeping the objective differentiable.

Importantly, when computing the log-probability ratio between two candidates sharing the same context $(c,\mathbf{x}_T)$, the identical partition function $Z(\phi,c,\mathbf{x}_T)$ cancels out, allowing the ratio to simplify exactly as follows:
\begin{align}
\log \frac{\pi_\phi(\x_w | c, \x_T)}  {\pi_\phi(\x_l | c, \x_T)} = -\alpha \left( E(\x_w;\phi, c, \x_T) - E(\x_l;\phi, c, \x_T) \right).
\end{align}
By applying this to both the student sampler $\phi$ and the reference sampler $\phi_\text{ref}$ and substituting this expression into the DPO objective~(\cref{eq:dpo_general}), we derive the final D2PO objective:
\begin{align}
\mathcal{L}_{\mathrm{D2PO}}(\phi) = - \mathbb{E}_{\mathcal{D}} \left[ \log \sigma \left( \beta \Delta_\phi(\x_w,\x_l) \right) \right],
\end{align}
where
\begin{align}
\Delta_\phi(\x_w,\x_l) = \big( d(\x_w,\x_{\phi_{\mathrm{ref}}}) - d(\x_w,\x_\phi) \big) - \big( d(\x_l,\x_{\phi_{\mathrm{ref}}}) - d(\x_l,\x_\phi) \big), \nonumber
\end{align}
and $\phi_{\mathrm{ref}}$ denotes the reference policy.
The temperature parameter $\alpha$ is implicitly absorbed into the scaling factor $\beta$ for simplicity.

\subsection{Score-based distance}
\label{sub:dips}

The effectiveness of D2PO depends on the choice of the distance function $d(\cdot,\cdot)$ used in the energy definition of the surrogate policy $\pi_\phi$.
A na\"ive choice would adopt a predefined metric such as $\ell_2$ or LPIPS, or train a separate network to approximate the energy.
{However, such choices fail to exploit the rich representations already encoded within the pretrained diffusion model.%

\vspace{-3mm}
\subsubsection{Score-induced energy.}
Our key insight is to derive the energy directly from the pretrained score network $s_\theta(\x_t,t)$.
Recall that $s_\theta$ approximates the data score $\nabla_{\x_t} \log p_t(\x_t)$ at noise level $t$.
Since the score characterizes the geometry of the data distribution, it naturally quantifies sample likelihood.

A sample that lies on the true data manifold should be locally consistent with this score field.
Conversely, a sample out of the data distribution fails to align with the score trajectory.
We therefore define the ideal score-induced energy of a sample as the degree of its misalignment with the learned score geometry as follows:
\begin{align}
E(\x;\phi) \equiv \int_{0}^T w(t)\, \left\| s_\theta(\x_t, t) - s_\theta(\x_{\phi,t}, t) \right\|_2^2\, dt,
\label{eq:energy_ideal}
\end{align}
where $w(t)$ is a weighting function over noise levels, and $\x_t$ and {$\x_{\phi,t}$} denote the perturbed versions of $\x$ and $\x_\phi$ at noise level $t$.

We evaluate the score discrepancy over the perturbed data distributions rather than the clean data distribution ($t=0$) for both theoretical and practical reasons.
In standard score-based generative modeling~\cite{song2021scorebased,yin2024onestep,wang2023prolificdreamer}, the clean data score is unavailable and pretrained diffusion models do not directly learn this clean score.
Instead, they are trained to approximate the scores of perturbed distributions $p_t$ across a continuous spectrum of noise levels.
These noisy score fields encode the multi-scale geometry of the data manifold, capturing coarse semantic structures at large $t$ and fine-grained details at small $t$.
Leveraging these perturbed distributions is therefore tractable and consistent with the objective of the pretrained diffusion model.

\vspace{-3mm}
\subsubsection{Noise-prediction distance.}

{
To transform the ideal energy formulation in Eq.~\eqref{eq:energy_ideal} into a practical optimization objective, we reframe the score discrepancy via the noise-prediction error as follows:
\begin{equation}
d_\theta(\mathbf{x}, \mathbf{x}_\phi; t) = \left\| \epsilon_\theta(\mathbf{x}_t, t) - \epsilon_\theta(\mathbf{x}_{\phi, t}, t) \right\|_2^2
\label{eq:d_theta}
\end{equation}
which leverages the implicit relation, $s_\theta = -\epsilon_\theta / \sigma_t$. 
Directly substituting this noise-prediction distance for the score-based distance in Eq.~\eqref{eq:energy_ideal} under a uniform weighting, however, introduces a scale mismatch across different noise levels.
Specifically, because the score discrepancy equals the noise-prediction distance up to a scaling factor of $\sigma_t^{-2}$, the integrand tends to diverge numerically as $t \to 0$, causing the low-noise terms to dominate the overall energy.
To resolve this imbalance and stabilize the optimization, we follow the established practice in DDPM~\cite{ho2020denoising} by adopting the weighting function $w(t) = \sigma_t^2$, which cancels the $\sigma_t^{-2}$ factor; the score-induced energy in Eq.~\eqref{eq:energy_ideal} reduces to the total noise-prediction distance over the continuous trajectory, which is given by
\begin{equation}
E(\mathbf{x}; \phi) = \int_0^T d_\theta(\mathbf{x}, \mathbf{x}_\phi; t) \, dt.
\label{eq:energy_reduced}
\end{equation}
As evaluating this continuous integral is computationally expensive, we approximate it in practice via Monte Carlo sampling.
}

\vspace{-3mm}
\subsubsection{Practical D2PO objective.}
Substituting {the weighted score-based distance} for the surrogate policy $\pi_\phi$ and applying the DPO objective between the student policy $\pi_\phi$ and the reference policy $\pi_{\phi_{\mathrm{ref}}}$, we obtain the following objective:
\begin{align}
\mathcal{L}_{\mathrm{D2PO}}(\phi) =
-\mathbb{E}_{\mathcal{D}}
\Bigg[
\log \sigma 
 \Big(\mathbb{E}_{t\sim\mathcal{U}(0,T)}
\Big[
\beta \Delta_\phi(\x_w,\x_l; t)
\Big]
\Big)
\Bigg],
\label{eq:d2po_loss_practical}
\end{align}
where
\begin{align}
\Delta_\phi(\x_w,\x_l; t) \hspace{-0.5mm} = \hspace{-0.5mm} \big( d_\theta(\x_w,\x_{\phi_{\mathrm{ref}}};t) \hspace{-0.5mm} - \hspace{-0.5mm} d_\theta(\x_w,\x_\phi;t) \big)
\hspace{-0.5mm} - \hspace{-0.5mm} \big( d_\theta(\x_l,\x_{\phi_{\mathrm{ref}}};t) \hspace{-0.5mm} - \hspace{-0.5mm} d_\theta(\x_l,\x_\phi;t) \big). \nonumber
\end{align}
A key benefit of the proposed {noise-prediction distance} lies in its evaluation over multi-level noise.
The score field captures coarse semantic structure at large noise levels $t$ and fine-grained detail at small $t$, and the weighting $w(t)=\sigma_t^2$ aggregates these scales into a single well-conditioned signal over $t$.
This yields a significantly richer preference signal compared to perceptual metrics such as LPIPS.

\vspace{-2mm}
\subsection{Dynamic preference}
\label{sub:dynamic}

{To understand D2PO and its dynamic optimization mechanism, it is essential to identify three key components: the \emph{student sampler}, the \emph{reference sampler}, and the \emph{winning sampler}, which are parameterized by $\phi$, $\phi_{\text{ref}}$, and $\phi'$, respectively.}

{The student sampler, governed by the target parameters $\phi = \{ \mathcal{S}, \boldsymbol{\omega} \}$, represents the policy we aim to optimize. 
To establish a preference comparison for DPO training, we utilize its output $\boldsymbol{x}_\phi$ as a baseline rather than relying on an external target. 
Specifically, the losing sample $\boldsymbol{x}_l$ is synthesized by applying a degradation operator $\mathcal{G}$ (e.g., a low-pass filter) to the student output, expressed as $\boldsymbol{x}_l = \mathcal{G}(\mathrm{sg}[\boldsymbol{x}_\phi])$, where $\mathrm{sg}[\cdot]$ denotes the stop-gradient operator.}

{Unlike standard DPO which employs a static reference model, D2PO dynamically updates the reference parameters $\phi_{\text{ref}} = \{\mathcal{S}_{\text{ref}}, \boldsymbol{\omega}_{\text{ref}}\}$.
Inspired by SPIN~\cite{chen2024self}, the reference timestep schedule $\mathcal{S}_{\text{ref}}$ is synchronized by copying the student schedule $\mathcal{S}$ at the end of each epoch. 
Meanwhile, the reference CFG weights $\boldsymbol{\omega}_{\text{ref}}$ are adjusted at each training step via an Exponential Moving Average (EMA) with a momentum parameter $\lambda$, i.e., $\boldsymbol{\omega}_{\text{ref}} \leftarrow \lambda \boldsymbol{\omega}_{\text{ref}} + (1-\lambda) \boldsymbol{\omega}.$}

{Designing a dynamic winning sampler is a core contribution of D2PO. 
Instead of introducing a pre-computed, fixed teacher, we formulate a dynamic teacher sampler whose implied distribution, {$\pi_{\phi'}$}, is generated relative to the current student parameters $\phi$ at each training step. 
For instance, if the student schedule $\mathcal{S}$ dictates a coarse trajectory with $N$ timesteps, we construct the dynamic teacher's schedule $\mathcal{S}'$ by refining $\mathcal{S}$ with additional intermediate timesteps, yielding a denser $2N$-step trajectory (e.g., via linear interpolation). 
The dynamic teacher sampler, parameterized by $\phi'$, is then induced by the same numerical solver operating under this finer-grained schedule $\mathcal{S}'.$}

{This dynamic framework provides a more robust learning signal than a static teacher policy ($\pi^{\text{fix}}$).
By design, the dynamic teacher represents a higher-fidelity trajectory derived from the student's current parameters.
Consequently, the D2PO loss penalizes the discrepancy between the student's coarse numerical path and this refined counterpart.
This formulation encourages the student sampler to yield a trajectory that remains consistent under step-size refinement, which is achieved when the discrete path closely approximates the true continuous-time trajectory.
Ultimately, rather than tracking an arbitrary external target, the student effectively learns to minimize its own discretization error.}

\subsection{Theoretical analysis}
\label{sub:analysis}

We provide a theoretical justification for the efficacy of the dynamic teacher mechanism in D2PO, thereby reducing discretization error.

\vspace{-3mm}
\subsubsection{Setup}

Let $\pi$ denote the true continuous-time policy representing the target distribution.
We define $\pi_\phi$ as the student policy induced by the parameters $\phi$ under a discrete numerical schedule with $N$ timesteps.
The dynamic teacher corresponds to a refined policy $\pi_{\phi'}$ evaluated on a finer discretization schedule (e.g., $2N$ timesteps).
To quantify discrepancies between policies, we employ a metric $\rho(\cdot,\cdot)$ that satisfies the triangle inequality.
The true error of the student policy relative to the continuous-time target is defined as
\begin{align}
\epsilon^{\text{true}}_{\phi} = \rho(\pi_\phi, \pi).
\end{align}

\vspace{-3mm}
\subsubsection{Dynamic teacher}

The dynamic DPO objective minimizes the discrepancy between the student and its refined counterpart:
\begin{align}
\mathcal{L}_{\text{dyn}} = \rho(\pi_\phi, \pi_{\phi'}).
\end{align}
Assuming the underlying numerical solver exhibits a convergence order of $k > 0$~\cite{suli2003introduction}, the dynamic teacher $\pi_{\phi'}$ constructed via a $2\times$ refinement yields a reduced true error relative to the continuous-time target, which is given by
\begin{align}
\epsilon^{\text{true}}_{\phi'}
=
\rho(\pi_{\phi'}, \pi)
\approx
\frac{1}{2^k}
\,\epsilon^{\text{true}}_{\phi}.
\end{align}
By applying the  triangle inequality, $\rho(\pi_\phi, \pi) \le \rho(\pi_\phi, \pi_{\phi'}) + \rho(\pi_{\phi'}, \pi)$, we establish a lower bound on the dynamic loss:
\begin{align}
\mathcal{L}_{\text{dyn}} \ge \left| \epsilon^{\text{true}}_{\phi} - \epsilon^{\text{true}}_{\phi'} \right|
\approx
\left( 1 - \frac{1}{2^k} \right) \epsilon^{\text{true}}_{\phi}.
\end{align}
Consequently, $\mathcal{L}_{\text{dyn}}$ serves as a non-trivial surrogate that upper-bounds (and scales proportionally with) the student's true error. 
Minimizing $\mathcal{L}_{\text{dyn}}$ enforces trajectory consistency across different discretization granularities, thereby driving a systematic reduction in discretization error. 
Since $\epsilon^{\text{true}}_{\phi'} = O(2^{-k})$, the residual error of the teacher vanishes progressively under refinement, aligning $\mathcal{L}_{\text{dyn}}$ more closely with the true optimization objective as training proceeds.

\vspace{-3mm}
\subsubsection{Fixed teacher}

The conventional fixed-teacher objective minimizes%
\begin{align}
\mathcal{L}_{\text{fix}} = \rho(\pi_\phi, \pi^{\text{fix}}),
\end{align}
where $\pi^{\text{fix}}$ denotes a static teacher policy that is independent of the student parameters $\phi$.
Let the intrinsic error of this fixed teacher be
\begin{align}
\epsilon^{\text{fix}} = \rho(\pi^{\text{fix}}, \pi),
\end{align}
which remains constant with respect to $\phi$.
Applying the triangle inequality yields the following bounds on the empirical loss:
\begin{align}
\left| \epsilon^{\text{true}}_{\phi} - \epsilon^{\text{fix}} \right|
\le
\mathcal{L}_{\text{fix}}
\le
\epsilon^{\text{true}}_{\phi} + \epsilon^{\text{fix}}.
\end{align}
Even if $\mathcal{L}_{\text{fix}}$ is optimized to its global minimum ($\mathcal{L}_{\text{fix}} = 0$), the resulting student policy is bounded by $\epsilon^{\text{true}}_{\phi} = \epsilon^{\text{fix}}$. 
Therefore, minimizing $\mathcal{L}_{\text{fix}}$ cannot reduce the true error below this asymptotic error floor. 
D2PO bypasses this performance bottleneck because the dynamic teacher evolves alongside the student, preventing the optimization from stagnation at a fixed residual error floor.

\section{Experiment}
\label{sec:experiment}

\subsection{Experimental setup}
We comprehensively evaluate D2PO across multiple generation tasks, architectures, and datasets. 
For text-to-image synthesis, we use the pre-trained Stable Diffusion v1.5 model~\cite{rombach2022high} using prompts from the COCO~\cite{lin2014microsoft} dataset.  
To demonstrate the generalizability of our approach, we apply this exact same sampler optimization to ImageNet (256$\times$256) generation in the latent space, as well as to the sampling process of the flow-matching-based InstaFlow model. 
Crucially, we keep all pre-trained model parameters strictly frozen; our method exclusively optimizes the sampling policy, consisting of the {continuously parameterized timestep schedule} $\mathcal{S}$ and the per-step CFG weights $\boldsymbol{\omega}$ for several advanced ODE solvers, including iPNDM~\cite{zhang2023fast}, UniPC~\cite{zhao2023unipc} and DPM-Solver++~\cite{lu2023dpm}.
We compare D2PO against state-of-the-art discretization methods, including DMN~\cite{xue2024accelerating}, GITS~\cite{chen2024trajectory}, and  LD3~\cite{tong2024learning}.
For evaluation, we measure distributional fidelity using FID~\cite{heusel2017gans}, and further assess text-to-image perceptual quality using HPSv2~\cite{wu2023human} and Aesthetic scores~\cite{Schuhmann2022LaionAesthetics}\footnote{Baselines for text-to-image tasks are re-evaluated on newly generated samples since HPSv2 and Aesthetic scores are omitted in the original papers. For ImageNet, we copy the FID values reported in the LD3~\cite{tong2024learning} paper.}.

\begin{table}[t]
\centering
\caption{
Quantitative comparison of sampler optimization methods on text-to-image synthesis using Stable Diffusion v1.5 across different ODE solvers.
}
\label{tab:sd15_comparison}
\vspace{-2mm}
\setlength{\tabcolsep}{2.2pt}
\scalebox{0.85}{
\begin{tabular}{c l c c c c c c c c c}
\toprule
& & \multicolumn{3}{c}{\textbf{iPNDM}} & \multicolumn{3}{c}{\textbf{UniPC}} & \multicolumn{3}{c}{\textbf{DPM-Solver++}} \\
\cmidrule(lr){3-5} \cmidrule(lr){6-8} \cmidrule(lr){9-11}
{Steps} & {Method} 
& {HPS $\uparrow$} & {Aesthetic $\uparrow$} & {FID $\downarrow$} 
& {HPS $\uparrow$} & {Aesthetic $\uparrow$} & {FID $\downarrow$} 
& {HPS $\uparrow$} & {Aesthetic $\uparrow$} & {FID $\downarrow$} \\
\midrule
\multirow{4}{*}{4} 
& DMN~\cite{xue2024accelerating}   & 0.2030 & 5.0936 & 21.39 & 0.2030 & 5.1084 & 22.03 & 0.1979 & 5.0978 & 24.33 \\
& GITS~\cite{chen2024trajectory}  & 0.2128 & 5.1413 & 18.12 & 0.2109 & 5.1592 & 20.14 & 0.2096 & 5.1519 & 19.86 \\
& LD3~\cite{tong2024learning}     & 0.2191 & 5.1756 & 17.60 & 0.2180 & 5.1755 & 18.34 & 0.2191 & 5.1736 & 17.46 \\
& D2PO                            & \textbf{0.2237} & \textbf{5.2024} & \textbf{15.69} & \textbf{0.2185} & \textbf{5.1761} & \textbf{16.97} & \textbf{0.2216} & \textbf{5.1854} & \textbf{16.84} \\
\midrule
\multirow{4}{*}{5} 
& DMN~\cite{xue2024accelerating}   & 0.2146 & 5.1514 & 17.35 & 0.2175 & 5.1732 & 16.99 & 0.2108 & 5.1549 & 19.17 \\
& GITS~\cite{chen2024trajectory}  & 0.2274 & 5.1926 & 14.07 & 0.2268 & 5.1928 & 15.46 & 0.2260 & 5.2058 & 15.29 \\
& LD3~\cite{tong2024learning}     & 0.2346 & 5.2463 & 13.59 & 0.2355 & \textbf{5.2591} & \textbf{13.89} & 0.2344 & 5.2378 & \textbf{13.27} \\
& D2PO                            & \textbf{0.2385} & \textbf{5.2701} & \textbf{13.38} & \textbf{0.2369} & 5.2776 & 14.47 & \textbf{0.2374} & \textbf{5.2740} & 14.16 \\
\midrule
\multirow{4}{*}{6} 
& DMN~\cite{xue2024accelerating}   & 0.2283 & 5.2107 & 13.66 & 0.2322 & 5.2371 & 13.74 & 0.2267 & 5.2188 & 14.51 \\
& GITS~\cite{chen2024trajectory}  & 0.2382 & 5.2407 & \textbf{12.33} & 0.2404 & 5.2461 & \textbf{12.38} & 0.2397 & 5.2610 & \textbf{12.41} \\
& LD3~\cite{tong2024learning}     & 0.2375 & 5.2565 & 13.10 & 0.2402 & 5.2689 & 12.90 & 0.2386 & 5.2445 & 12.62 \\
& D2PO                            & \textbf{0.2482} & \textbf{5.3309} & 13.54 & \textbf{0.2441} & \textbf{5.3192} & 14.38 & \textbf{0.2458} & \textbf{5.3237} & 14.00 \\
\midrule
\multirow{4}{*}{7} 
& DMN~\cite{xue2024accelerating}   & 0.2428 & 5.2778 & \textbf{11.89} & 0.2474 & 5.3011 & \textbf{12.12} & 0.2399 & 5.2734 & 13.02 \\
& GITS~\cite{chen2024trajectory}  & 0.2394 & 5.2368 & 12.16 & 0.2379 & 5.2065 & 12.91 & 0.2355 & 5.2174 & 13.16 \\
& LD3~\cite{tong2024learning}     & 0.2434 & 5.2763 & 12.41 & 0.2421 & 5.2660 & 12.77 & 0.2455 & {5.2795} & \textbf{12.09} \\
& D2PO                            & \textbf{0.2513} & \textbf{5.3257} & 12.71 & \textbf{0.2499} & \textbf{5.3472} & 13.61 & \textbf{0.2502} & \textbf{5.3458} & 13.52 \\
\bottomrule
\end{tabular}}
\end{table}

\begin{table*}[t]
    \centering
    
    \begin{minipage}[t]{0.36\linewidth}
        \centering
        \caption{
            Quantitative comparison on ImageNet-256 (latent space) using the 3rd-order (3M) iPNDM solver. 
            Baseline results are reported from the LD3~\cite{tong2024learning} paper. 
        }
        \label{tab:imagenet_ipndm}
        \setlength{\tabcolsep}{2.0pt}
        \scalebox{0.85}{
            \begin{tabular}{l c c c c}
                \toprule
                & \multicolumn{4}{c}{Steps} \\
                \cmidrule(lr){2-5}
                {Method} & {4} & {5} & {6} & {7} \\
                \midrule
                Uniform & 13.86 & 7.80 & 6.03 & 5.35 \\
                GITS~\cite{chen2024trajectory} & 56.00 & 43.56 & 19.33 & 10.33 \\
                DMN~\cite{xue2024accelerating} & 10.15 & 7.33 & 7.25 & 7.40 \\
                LD3~\cite{tong2024learning} & 9.19 & 6.03 & 5.09 & \textbf{4.68} \\
                D2PO & \textbf{7.28} & \textbf{5.48} & \textbf{4.80} & 4.70 \\
                \bottomrule
            \end{tabular}
        }
    \end{minipage}
    \hfill
    \begin{minipage}[t]{0.6\linewidth}
        \centering
        \caption{
            Quantitative comparison on the InstaFlow~\cite{liu2024instaflow} model using the prompts from the COCO dataset.
            Higher HPS and Aesthetic scores (\(\uparrow\)) are better, while lower FID scores (\(\downarrow\)) are better.
        }
        \label{tab:instaflow_comparison}
        \setlength{\tabcolsep}{4pt} %
        \scalebox{0.85}{ %
            \begin{tabular}{c l ccc}
                \toprule
                {Steps} & {Method} & {HPS \(\uparrow\)} & {Aesthetic \(\uparrow\)} & {FID \(\downarrow\)} \\
                \midrule
                \multirow{3}{*}{2} 
                & Uniform & 0.1865 & 5.0060 & 44.27 \\
                & LD3~\cite{tong2024learning} & 0.1708 & 4.6270 & 63.55 \\
                & D2PO & \textbf{0.1872} & \textbf{5.0924} & \textbf{40.68} \\
                \midrule
                \multirow{3}{*}{4} 
                & Uniform & 0.2189 & 5.1222 & 16.85 \\
                & LD3~\cite{tong2024learning} & 0.2086 & 5.0712 & 23.21 \\
                & D2PO & \textbf{0.2197} & \textbf{5.1415} & \textbf{15.50} \\
                \midrule
                \multirow{3}{*}{6} 
                & Uniform & 0.2317 & {5.1847} & 13.68 \\
                & LD3~\cite{tong2024learning} & 0.2299 & 5.1740 & 15.60 \\
                & D2PO & \textbf{0.2342} & \textbf{5.1855} & \textbf{12.70} \\
                \bottomrule
            \end{tabular}
        }
    \end{minipage}
    
\end{table*}

\vspace{-1mm}
\begin{table}[t]
\centering
\caption{
Ablation study of D2PO on text-to-image synthesis using Stable Diffusion v1.5 and the iPNDM solver using prompts from the COCO dataset. 
We report performance at \# steps=4 and 5 after removing or altering key components: dynamic preference, score-based energy, and the reference timestep update strategy.
}
\label{tab:ablation_study_no_hps}
\vspace{-1mm}
\setlength{\tabcolsep}{20pt}
\scalebox{0.85}{
\begin{tabular}{c l ccc}
\toprule
{Steps} & {Method} & {Aesthetic \(\uparrow\)} & {FID \(\downarrow\)} \\
\midrule
\multirow{4}{*}{4} 
& {D2PO (Full)} & \bfseries 5.2024 & \bfseries 15.69 \\
& $\quad$ w/o dynamic preference                  & 5.1810 & 16.91 \\
& $\quad$ w/o score-based energy                  & 5.1796 & 17.88 \\
& $\quad$ w/ EMA reference timestep               & 5.1937 & 15.75 \\

\midrule
\multirow{4}{*}{5} 
& {D2PO (Full)} & \bfseries 5.2701 & \bfseries  13.38\\
& $\quad$ w/o dynamic preference                  & 5.2615 & 13.70 \\
& $\quad$ w/o score-based energy                  & 5.2630 & 14.59 \\
& $\quad$ w/ EMA reference timestep               & 5.2639 & 13.42 \\

\bottomrule
\end{tabular}}
\vspace{-1mm}
\end{table}

\vspace{-1mm}
\subsection{Main results}
Our quantitative results on text-to-image synthesis using Stable Diffusion v1.5 are presented in \cref{tab:sd15_comparison}, where we apply D2PO to three representative ODE solvers---iPNDM, UniPC, and DPM-Solver++.
For evaluation, we generate 30k samples using text prompts from COCO dataset~\cite{lin2014microsoft}, following the standard~\cite{tong2024learning, frankels4s}.
Overall, the results demonstrate that D2PO achieves superior performance compared to existing baselines, asserting its effectiveness and robustness across multiple advanced solvers.

At a low number of steps (4 and 5), D2PO functions as a {superior error corrector}.
In this regime, D2PO achieves state-of-the-art perceptual quality—showing the highest Aesthetic scores across all solvers and the highest HPS scores for iPNDM.
Furthermore, when using 4 steps, D2PO surpasses all baselines in FID across all three solvers.
This directly validates our hypothesis. Regression-based methods sacrifice high-frequency details, in part because their loss metrics, \eg, LPIPS, fail to capture these errors.
Our novel score-based distance metric captures both structural and textural discrepancies by measuring score differences at various noise levels.
This high sensitivity to fine-grained texture loss allows D2PO to correct the foundational flaws of the baseline, simultaneously improving both quality and fidelity.

At a higher number of steps (6 and 7), as the baseline's severe discretization error is reduced, the expected quality-diversity trade-off~\cite{brock2018large, kingma2018glow, ho2022classifier} emerges, and D2PO's behavior shifts to its primary goal of {preference alignment}.
While D2PO maintains its significant lead in perceptual scores (HPS/Aesthetic), its FID score becomes comparable or slightly higher than the baselines.
This shift is the expected signature of successful alignment.
D2PO's objective function, which aligns with its dynamic, higher-fidelity $2N$ teacher, optimizes the sampler towards a distribution that maximizes perceptual quality.
This distribution is distinct from the {average} of the real data distribution, which FID measures.
This result demonstrates that D2PO is not failing; rather, it is successfully harnessing the full potential of the model by optimizing for its intended perceptual objective, which is to reduce its own discretization error.

\begin{figure}[!tb] 
  \centering
  \begin{subfigure}{\linewidth} 
  \caption*{Steps = 4: ``Two dogs curled up asleep on a couch."}
    \includegraphics[width=0.24\linewidth]{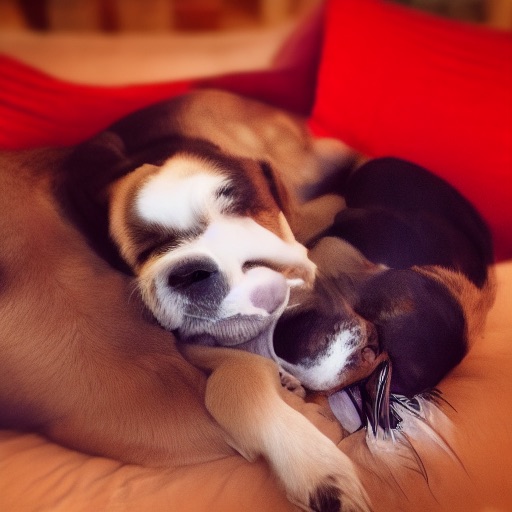}\hfill  
    \includegraphics[width=0.24\linewidth]{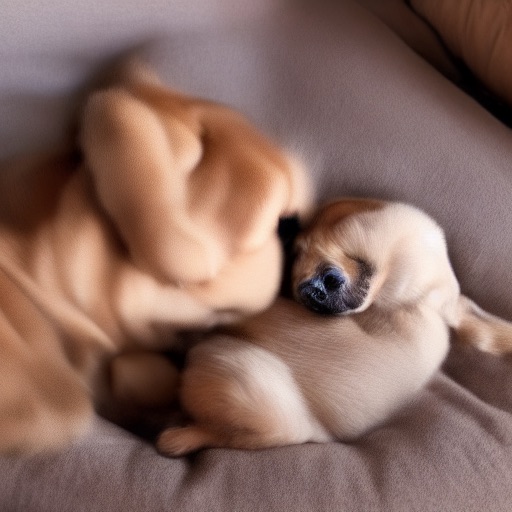}\hfill 
    \includegraphics[width=0.24\linewidth]{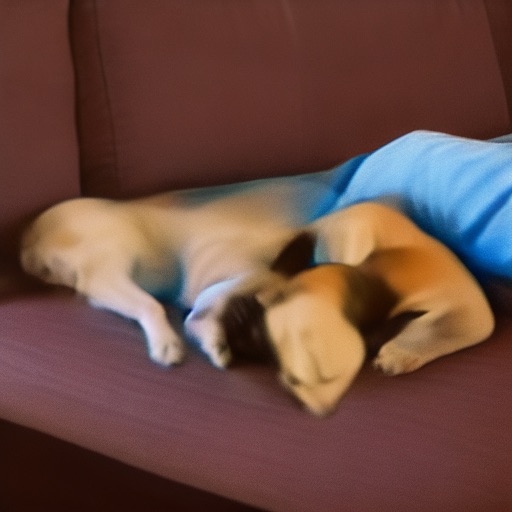}\hfill
    \includegraphics[width=0.24\linewidth]{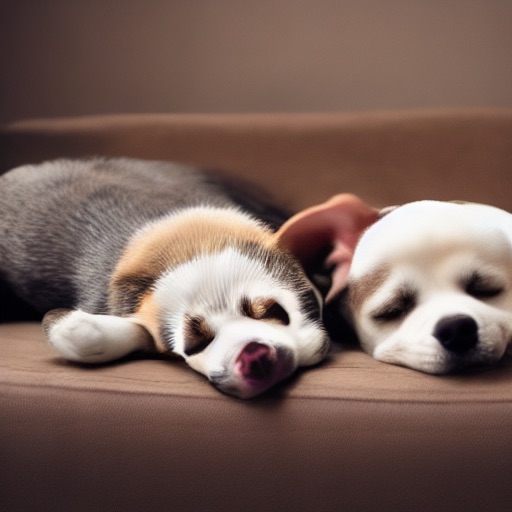}
    \vspace{1mm}
  \end{subfigure}
  
  \begin{subfigure}{\linewidth} 
  \caption*{Steps = 5: ``A man riding a snow board on top of a snow covered slope."}
    \includegraphics[width=0.24\linewidth]{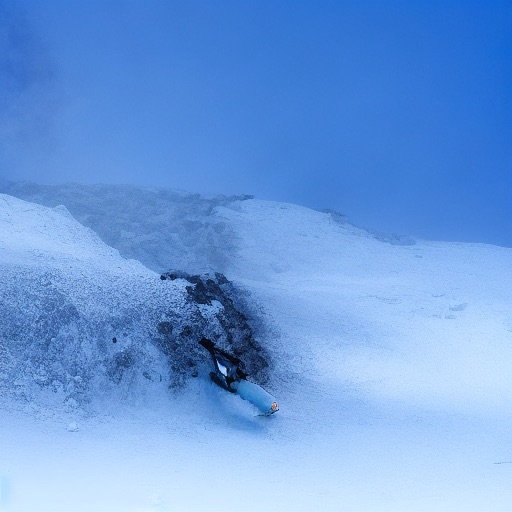}\hfill  
    \includegraphics[width=0.24\linewidth]{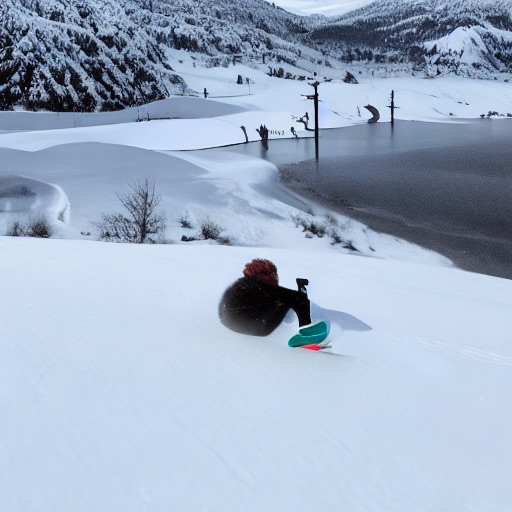}\hfill 
    \includegraphics[width=0.24\linewidth]{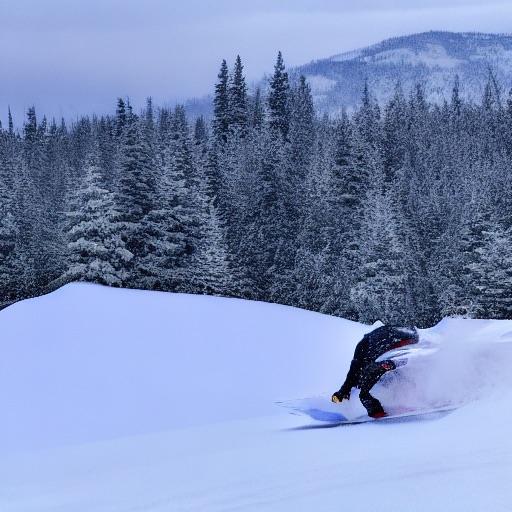}\hfill
    \includegraphics[width=0.24\linewidth]{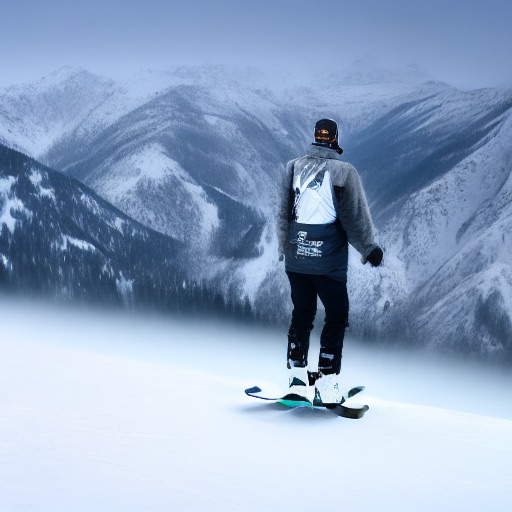}
    \vspace{1mm}
  \end{subfigure}

  \begin{subfigure}{\linewidth} 
  \caption*{Steps = 6: ``A male tennis player in white shorts is playing tennis."}
    \includegraphics[width=0.24\linewidth]{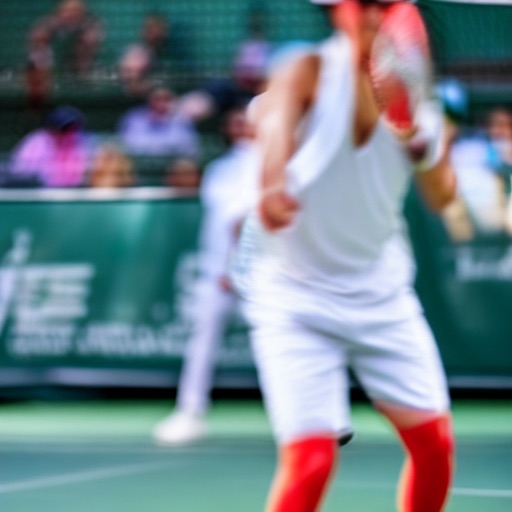}\hfill  
    \includegraphics[width=0.24\linewidth]{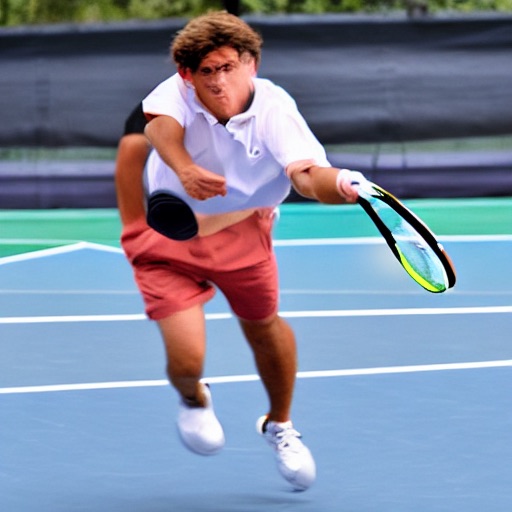}\hfill 
    \includegraphics[width=0.24\linewidth]{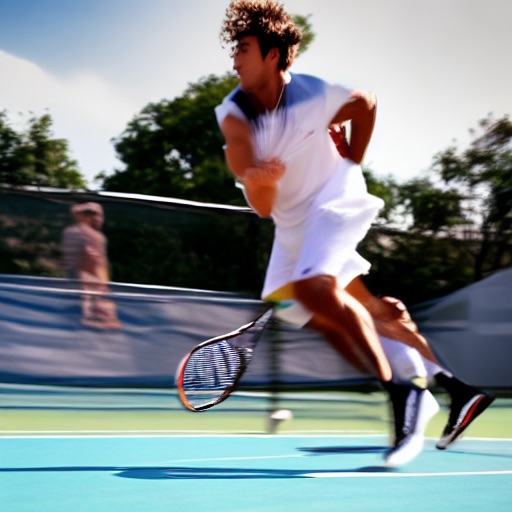}\hfill
    \includegraphics[width=0.24\linewidth]{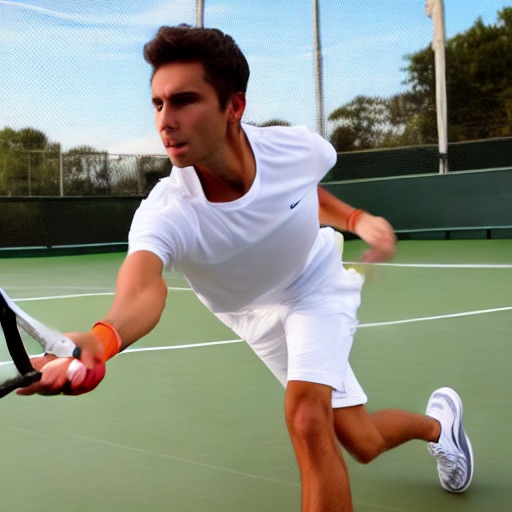}
    \vspace{1mm}
  \end{subfigure}

  \begin{subfigure}{\linewidth} 
  \caption*{Steps = 7: ``A little boy about to hit a baseball during a game."}
    \includegraphics[width=0.24\linewidth]{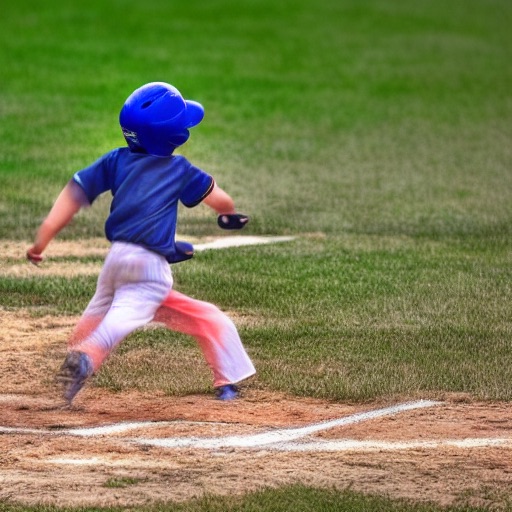}\hfill  
    \includegraphics[width=0.24\linewidth]{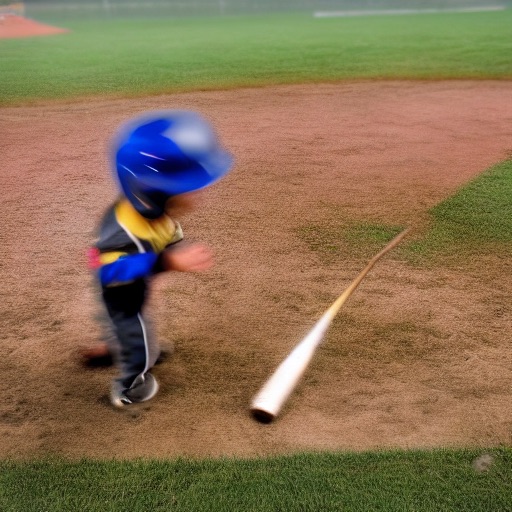}\hfill 
    \includegraphics[width=0.24\linewidth]{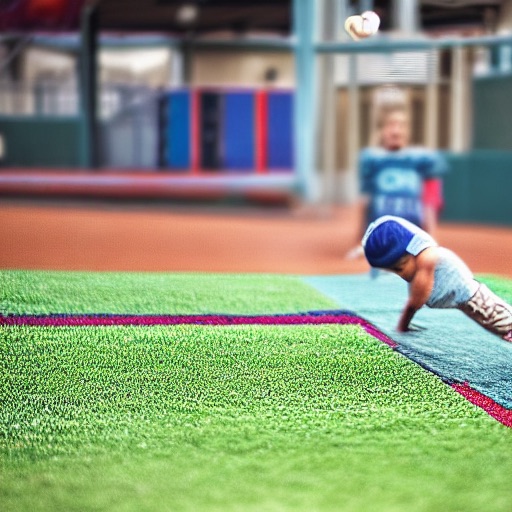}\hfill
    \includegraphics[width=0.24\linewidth]{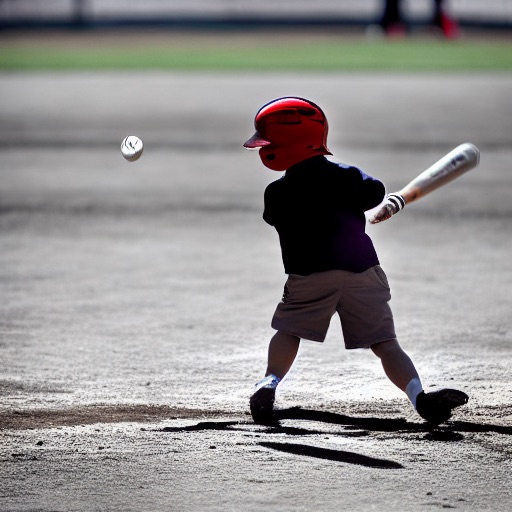}
  \end{subfigure}
  
  \vspace{1mm} 
  \begin{minipage}{0.24\linewidth}
    \centering\small DMN~\cite{xue2024accelerating}
  \end{minipage}\hfill
  \begin{minipage}{0.24\linewidth}
    \centering\small GITS~\cite{chen2024trajectory}
  \end{minipage}\hfill
  \begin{minipage}{0.24\linewidth}
    \centering\small LD3~\cite{tong2024learning}
  \end{minipage}\hfill
  \begin{minipage}{0.24\linewidth}
    \centering\small D2PO
  \end{minipage}

  \vspace{-1mm}
  \caption{Qualitative comparison of various discretization methods on Stable Diffusion v1.5 using iPNDM across different number of time steps (4 to 7). D2PO consistently produces sharper details and fewer artifacts compared to baselines.}
  \label{fig:model_comparison_all_nfe}
\end{figure}

\vspace{-1mm}
\subsection{Generalization across domains and architectures}
To evaluate the robustness of our learned policy, we tested D2PO beyond standard text-to-image tasks. 
As shown in \cref{tab:imagenet_ipndm}, D2PO matches or outperforms baseline methods on ImageNet-256 (latent space) using the 3rd-order iPNDM solver.
Furthermore, \cref{tab:instaflow_comparison} illustrates D2PO's successful application to InstaFlow, a flow-matching model on the COCO dataset.

\vspace{-1mm}
\subsection{Qualitative results}
\cref{fig:model_comparison_all_nfe} visualizes the qualitative results on Stable Diffusion v1.5 (iPNDM) for all steps reported in \cref{tab:sd15_comparison}.
Across all cases, D2PO consistently produces images with higher perceptual quality, sharper details, and fewer artifacts compared to baselines.
This visual evidence directly supports our quantitative findings.
For instance, baseline methods like LD3 and DMN frequently exhibit significant blurriness, water-color artifacts, or loss of fine-grained detail. 
In more complex scenes (Steps=7), D2PO generates a coherent and detailed image, while other methods suffer from severe structural distortion or artifacts. 
More qualitative results on different settings are provided in the supplementary material.

\subsection{Ablation study}
We ablate D2PO's key components in \cref{tab:ablation_study_no_hps}. 
First, the {w/o dynamic preference} variant replaces our dynamic $2N$-step teacher with a fixed $N+1$-step teacher (as in LD3~\cite{tong2024learning}). This degrades performance, validating that our dynamic mechanism provides a stronger, more consistent learning signal than regression to a static teacher. 
Second, the {w/o score-based energy} variant, which instead uses LPIPS, causes the most significant drop in fidelity, confirming our metric's necessity for capturing fine-grained discretization errors. 
Finally, the {w/ EMA reference timestep} variant replaces our epoch-wise copy strategy for the reference time step schedule with an EMA update. Its sub-optimal performance highlights that while EMA suits continuous parameters (like CFG weights), direct epoch-wise copying provides a more stable anchor for optimizing discrete time steps.

\section{Conclusion}
\label{sec:conclusion}
We identified a critical performance bottleneck in dominant student-teacher regression frameworks for optimizing diffusion samplers.
We demonstrated that as the teacher-student NFE gap increases, standard regression losses force the low-NFE student to sacrifice high-frequency texture fidelity, leading to degraded perceptual quality.
To address this, we proposed {D2PO}, a novel framework that reframes sampler optimization as a preference-based alignment task.
We introduced a novel {score-based energy function} that leverages the score model itself to capture the fine-grained textural and structural errors that standard metrics miss, and a dynamic preference mechanism that creates a self-improving loop, where the student policy is aligned with a dynamically refined, higher-fidelity version of itself.
This dynamic teacher provides a stronger, theoretically-grounded learning signal that forces the student to minimize its own discretization error, rather than converging to a suboptimal fixed teacher.
Our extensive experiments across multiple solvers demonstrated that D2PO successfully aligns diffusion samplers with true perceptual quality, effectively solving the existing bottleneck of static teacher regression.

\subsubsection*{Acknowledgements.}
This work was partly supported by the Samsung Electronics Co., Ltd. (IO250418-12669-01).
It was also partly supported by the NRF grant [RS-2022-NR070855] and the IITP grants [RS-2025-25442338; RS-2026-25526850; No.RS-2021-II211343; No.RS-2020-II201336] funded by the Korea government (MSIT).

\newpage

\bibliographystyle{splncs04}
\bibliography{main}

\newpage
\title{D2PO: Optimizing Diffusion Samplers via Dynamic Preference\\ \textit{Supplementary Document}}

\renewcommand\thesection{\Alph{section}}
\renewcommand\thetable{\Alph{table}}
\renewcommand\thefigure{\Alph{figure}}

\setcounter{table}{0}
\setcounter{figure}{0}
\setcounter{theorem}{0}
\setcounter{section}{0}
\section{Experimental Setup}
\label{sec:experimental_setup}

\subsection{Implementation details}

We implement our method using PyTorch~\cite{paszke2019pytorch} and adopt the pretrained, frozen Stable Diffusion v1.5~\cite{rombach2022high} as the base denoiser~$\epsilon_\theta$, following prior work~\cite{frankels4s,tong2024learning}.  
All experiments are performed on a single NVIDIA A6000 GPU.

Our learnable policy is $\phi = \{\mathcal{S}, \boldsymbol{\omega}\}$. 
Following LD3~\cite{tong2024learning}, the timestep schedule $\mathcal{S}$ is parameterized by two coupled sequences: $\mathcal{S}_1$ defines the ODE solver discretization grid, while $\mathcal{S}_2$ provides the time-conditioning inputs to the diffusion model. 
We optimize each parameter group with a separate optimizer: $\mathcal{S}_1$ uses RMSprop (momentum $0.9$), while $\mathcal{S}_2$ and the CFG weights $\boldsymbol{\omega}$ use SGD. 
For the D2PO objective, we set the temperature $\beta \in \{10, 50, 100\}$.
To keep the effective step size stable, the base learning rates are scaled as $\frac{1\times10^{-3}}{\beta}$ for $\mathcal{S}_1$ and $\frac{2\times10^{-4}}{\beta N}$ for $\mathcal{S}_2$ and $\boldsymbol{\omega}$, where $N$ is the number of sampling steps.
We employ gradient accumulation over four iterations, yielding an effective batch size of~4.

To construct the winning policy~$\phi'$, we linearly interpolate the current student timestep schedule $\mathcal{S}$ to generate a denser schedule with $2N$ function evaluations.  
The linear interpolation is performed in continuous time $t \in [0, T]$.

\subsection{Evaluation protocol}

Following standard practice in text-to-image evaluation~\cite{frankels4s,tong2024learning}, we conduct experiments on the COCO 2014 validation set~\cite{lin2014microsoft} under the zero-shot generation setting.  
For quantitative evaluation, we use a random subset of 30,000 captions sampled from the validation set.
Training and evaluation prompts remain fully disjoint: D2PO is trained using 400 prompts from the training split, whereas evaluation is performed on unseen captions from the validation split.
On the generated samples, we compute standard metrics assessing distributional fidelity and perceptual quality: HPSv2~\cite{wu2023human}, the Aesthetic score~\cite{Schuhmann2022LaionAesthetics}, and FID~\cite{heusel2017gans}.

\subsection{Gradient rematerialization}

Optimizing sampling parameters requires backpropagating gradients through the entire reverse-diffusion trajectory, which involves repeated evaluations of the heavy diffusion backbone~$\epsilon_\theta$.  
A naive implementation would store all intermediate activations, resulting in memory consumption that scales linearly with the number of function evaluations, which is infeasible for large diffusion models.

We therefore apply gradient rematerialization~\cite{chen2016training}, following the protocol established in~\cite{tong2024learning}.  
Instead of storing intermediate activations during the forward pass, rematerialization recomputes necessary activations on demand during backpropagation.  
This trades additional compute for a substantial reduction in memory, enabling efficient training on a single GPU.

\vspace{-3mm}

\section{Additional Quantitative Results}
\label{sec:additional_quantitative}

\begin{table}[t]
\centering
\caption{
Comparison of D2PO with standard LD3 and an enhanced variant (LD3$^\dagger$) on the COCO dataset~\cite{lin2014microsoft}.
{LD3$^\dagger$} is trained with a significantly larger computational budget ($16\times$ training time), more training data ($16\times$), and a stronger teacher ($\Delta = 4$) to match the resource allocation of D2PO.
 Higher HPS and Aesthetic scores (\(\uparrow\)) are better, while lower FID scores (\(\downarrow\)) are better.
}

\label{tab:computational_cost}
\setlength{\tabcolsep}{11pt}
\scalebox{0.85}{
\begin{tabular}{c l ccc}
\toprule
{Steps} & {Method} & {HPS \(\uparrow\)} & {Aesthetic \(\uparrow\)} & {FID \(\downarrow\)} \\
\midrule
\multirow{3}{*}{4} 
&  LD3~\cite{tong2024learning}  & 0.2191 & 5.1756 & 17.60 \\
&  LD3$^\dagger$  & 0.2146 & 5.1678 & 21.34 \\
&  D2PO & \textbf{0.2237} & \textbf{5.2024} & \textbf{15.69} \\
\midrule
\multirow{3}{*}{5} 
&  LD3~\cite{tong2024learning}  & 0.2346 & 5.2463 & 13.59 \\
&  LD3$^\dagger$  & 0.2361 & 5.2644 & 14.89 \\
&  D2PO & \textbf{0.2385} & \textbf{5.2701} & \textbf{13.38} \\
\midrule
\multirow{3}{*}{6} 
&  LD3~\cite{tong2024learning}  & 0.2375 & 5.2565 & \textbf{13.10} \\
&  LD3$^\dagger$  & 0.2463 & 5.3196 & 13.46 \\
&  D2PO & \textbf{0.2482} & \textbf{5.3309} & 13.54 \\
\midrule
\multirow{3}{*}{7} 
&  LD3~\cite{tong2024learning}  & 0.2434 & 5.2763 & \textbf{12.41} \\
&  LD3$^\dagger$  & 0.2502 & \textbf{5.3264} & 13.15 \\
&  D2PO & \textbf{0.2513} & 5.3257 & 12.71 \\
\bottomrule
\end{tabular}}
\end{table}

\begin{table}[t]
\centering
\caption{
Quantitative comparison of text-image alignment using CLIP score on the COCO dataset~\cite{lin2014microsoft} with the iPNDM solver.
Higher CLIP scores ($\uparrow$) indicate better alignment.
}

\label{tab:clip_score}
\setlength{\tabcolsep}{8pt}
\scalebox{0.85}{
\begin{tabular}{ccccc}
\toprule
{Method} & {Steps = 4} & {Steps = 5} & {Steps = 6} & {Steps = 7}\\
\midrule
LD3~\cite{tong2024learning}  & 25.87 & 26.21 & 26.22 &26.37 \\
D2PO & \bf{26.01} & \bf{26.30} & \bf{26.52} & \bf{26.57} \\
\bottomrule
\end{tabular}}
\end{table}

\subsection{Decoupling methodological gains from computational budget}
We analyze whether D2PO's gains stem from its design rather than its larger training budget.
LD3~\cite{tong2024learning} training is typically lightweight, utilizing a small subset of 25 prompts for 5 epochs.
D2PO trains on a significantly larger scale using 400 prompts.
To verify that D2PO's superiority stems from its methodological design rather than resource scaling, we compare against an enhanced baseline, LD3$^\dagger$, designed to match D2PO's resource budget.
LD3$^\dagger$'s budget has $16\times$ training duration, $16\times$ more data, and a stronger teacher with an increased step gap ($\Delta = T - S = 4$).
As reported in \cref{tab:computational_cost}, granting LD3 the same budget as D2PO fails to close the gap on the perceptual metrics D2PO is designed to optimize: D2PO retains the highest HPS at every step count and matches or exceeds LD3$^\dagger$ on Aesthetic, confirming that its perceptual advantage is methodological rather than a by-product of resource scaling. 

\subsection{Evaluation of text-image semantic alignment}
Beyond visual aesthetics and distributional fidelity, faithfully reflecting the conditioning prompt is a central requirement for text-to-image synthesis. We therefore measure CLIP score on COCO to assess semantic alignment.

As shown in \cref{tab:clip_score}, D2PO achieves higher CLIP scores than LD3 at every step count. 
We attribute this to a difference in objective. Distillation methods such as LD3 enforce pointwise $\ell^2$ or LPIPS matching to a fixed teacher; in the extreme few-step regime, where the solver already incurs large truncation error, this rigid structural constraint leaves little freedom to preserve prompt-relevant content, diluting semantic alignment. D2PO instead optimizes an ordinal preference toward its own refined, higher-fidelity output, which does not tie the student to a single pointwise target and thus retains more flexibility to keep the generation consistent with the prompt. As a result, D2PO improves faithfulness to the input text alongside its perceptual gains.

\begin{table}[t]
\centering
\caption{ {Generalization to the modern Stable Diffusion~3.5-Medium~\cite{esser2024scalingrectifiedflowtransformers} backbone (MM-DiT, flow matching, Euler solver) on the COCO dataset. D2PO improves all metrics, demonstrating that the gains are not specific to SD~v1.5.}}
\label{tab:sd35_comparison}
\vspace{-1mm}
\setlength{\tabcolsep}{12pt}
\scalebox{0.85}{
\begin{tabular}{c l ccc}
\toprule
{Steps} & {Method} & {HPS \(\uparrow\)} & {Aesthetic \(\uparrow\)} & {FID \(\downarrow\)} \\
\midrule
\multirow{3}{*}{4}
& Uniform & 0.1289 & 4.6185 & 92.09 \\
& LD3~\cite{tong2024learning} & 0.1336 & 4.6476 & 86.34 \\
& \textbf{D2PO} & \bfseries 0.1468 & \bfseries 4.7812 & \bfseries 73.89 \\
\midrule
\multirow{3}{*}{6}
& Uniform & 0.1675 & 4.8922 & 50.94 \\
& LD3~\cite{tong2024learning} & 0.1698 & 4.9184 & 48.95 \\
& \textbf{D2PO} & \bfseries 0.1902 & \bfseries 5.0190 & \bfseries 33.64 \\
\bottomrule
\end{tabular}}
\vspace{-1mm}
\end{table}

\begin{table}[t]
\centering
\caption{ { Blind user study: vote shares on 30 COCO prompts (iPNDM, 7 steps, same initial noise, hidden method names).}}
\label{tab:user_study}
\vspace{-1mm}
\setlength{\tabcolsep}{12pt}
\scalebox{0.85}{
\begin{tabular}{l cccc}
\toprule
{Method} & {Align.} & {Qual.} & {Overall} & {Avg.} \\
\midrule
GITS~\cite{chen2024trajectory} & 15.8\% & 9.4\% & 11.2\% & 12.1\% \\
LD3~\cite{tong2024learning} & 29.4\% & 26.1\% & 27.3\% & 27.6\% \\
\textbf{D2PO} & \bfseries 54.8\% & \bfseries 64.5\% & \bfseries 61.5\% & \bfseries 60.3\% \\
\bottomrule
\end{tabular}}
\vspace{-1mm}
\end{table}

\begin{table}[t]
\centering
\caption{ {FID comparison on the pixel-space AFHQv2 dataset with the iPNDM solver.}}
\label{tab:additional_fid_afhq}
\setlength{\tabcolsep}{8pt}
\scalebox{0.85}{
\begin{tabular}{lcccc}
\toprule
Method & 4 & 6 & 8 & 10 \\
\midrule
Uniform & 23.20 & 9.55 & 4.49 & 3.19 \\
GITS~\cite{chen2024trajectory} & 12.89 & 6.10 & 4.03 & 3.26 \\
LD3~\cite{tong2024learning} & \ \ 9.96 & 3.63 & 2.63 & 2.27 \\
\textbf{D2PO} & \ \ \textbf{9.94} & \textbf{3.60} & \textbf{2.61} & \textbf{2.20} \\
\bottomrule
\end{tabular}}
\end{table}

\begin{table}[t]
\centering
\caption{ {FID comparison on ImageNet-256 (latent space) against search-based optimization (CMA-ES and the GITS search baseline) with the iPNDM solver.}}
\label{tab:additional_fid_cmaes}
\setlength{\tabcolsep}{8pt}
\scalebox{0.85}{
\begin{tabular}{lcccc}
\toprule
Method & 4 & 5 & 6 & 7 \\
\midrule
GITS~\cite{chen2024trajectory} & 56.00 & 43.56 & 19.33 & 10.33 \\
CMA-ES~\cite{hansen2016cma} & 20.01 & 17.83 & \ \ 7.20 & \ \ 6.99 \\
\textbf{D2PO} & \ \ \textbf{7.28} & \ \ \textbf{5.48} & \ \ \textbf{4.80} & \ \ \textbf{4.70} \\
\bottomrule
\end{tabular}}
\end{table}

\subsection{Generalization to modern backbones}
 {To confirm that the gains are not specific to the SD~v1.5 backbone, we apply D2PO to the modern Stable Diffusion~3.5-Medium~\cite{esser2024scalingrectifiedflowtransformers}, a multimodal diffusion transformer (MM-DiT) trained with flow matching and sampled with the Euler solver. As reported in \cref{tab:sd35_comparison}, D2PO outperforms both the Time-Uniform baseline and LD3 on all metrics, including a large FID improvement, demonstrating that the benefits of our preference-based optimization transfer to state-of-the-art backbones.}

\subsection{ {Human evaluation}}
 {To verify that our gains on proxy metrics reflect genuine perceptual improvements, we conduct a blind, randomized user study on $30$ COCO prompts (iPNDM, $7$ steps), in which all methods share the same initial noise and the method names are hidden. Participants select the best result among GITS, LD3, and D2PO under three criteria---prompt alignment, visual quality, and overall preference ($90$ votes per participant). As shown in \cref{tab:user_study}, D2PO receives $60.3\%$ of the votes on average, far ahead of LD3 ($27.6\%$) and GITS ($12.1\%$), confirming that the improvements in HPSv2 and Aesthetic scores correspond to human-perceived quality.}

\subsection{ {Generalization to pixel-space generation}}
 {We assess D2PO on the pixel-space AFHQv2 dataset with the iPNDM solver (\cref{tab:additional_fid_afhq}), where it matches or slightly outperforms LD3 across $4$--$10$ steps. 
The marginal gap is partly because AFHQv2 uses unconditional generation, so the per-step CFG weights $\boldsymbol{\omega}$ carry no optimization signal and only the timestep schedule $\mathcal{S}$ is effectively optimized. 
}

\subsection{ {Comparison with search-based optimization}}
 {On ImageNet-256 (latent space), we compare against search-based optimization (\cref{tab:additional_fid_cmaes}). 
Although the sampler parameters are low-dimensional, black-box search such as CMA-ES~\cite{hansen2016cma} requires many sample evaluations per candidate. 
Under the same parameterization and score-based distance, D2PO substantially outperforms both CMA-ES and the GITS~\cite{chen2024trajectory} search baseline, as it backpropagates a preference signal through the solver rather than relying on scalar function evaluations alone.}

\begin{figure*}[t] 
  \centering
  
  \captionsetup[subfigure]{font=small, labelfont=small}
  
  \begin{subfigure}{\linewidth}
    \centering
    
    \begin{minipage}{0.24\linewidth}\centering\small DMN\end{minipage} \hfill
    \begin{minipage}{0.24\linewidth}\centering\small GITS\end{minipage} \hfill
    \begin{minipage}{0.24\linewidth}\centering\small LD3\end{minipage} \hfill
    \begin{minipage}{0.24\linewidth}\centering\small D2PO\end{minipage}
    \vspace{1mm} 
    
    \includegraphics[width=0.24\linewidth]{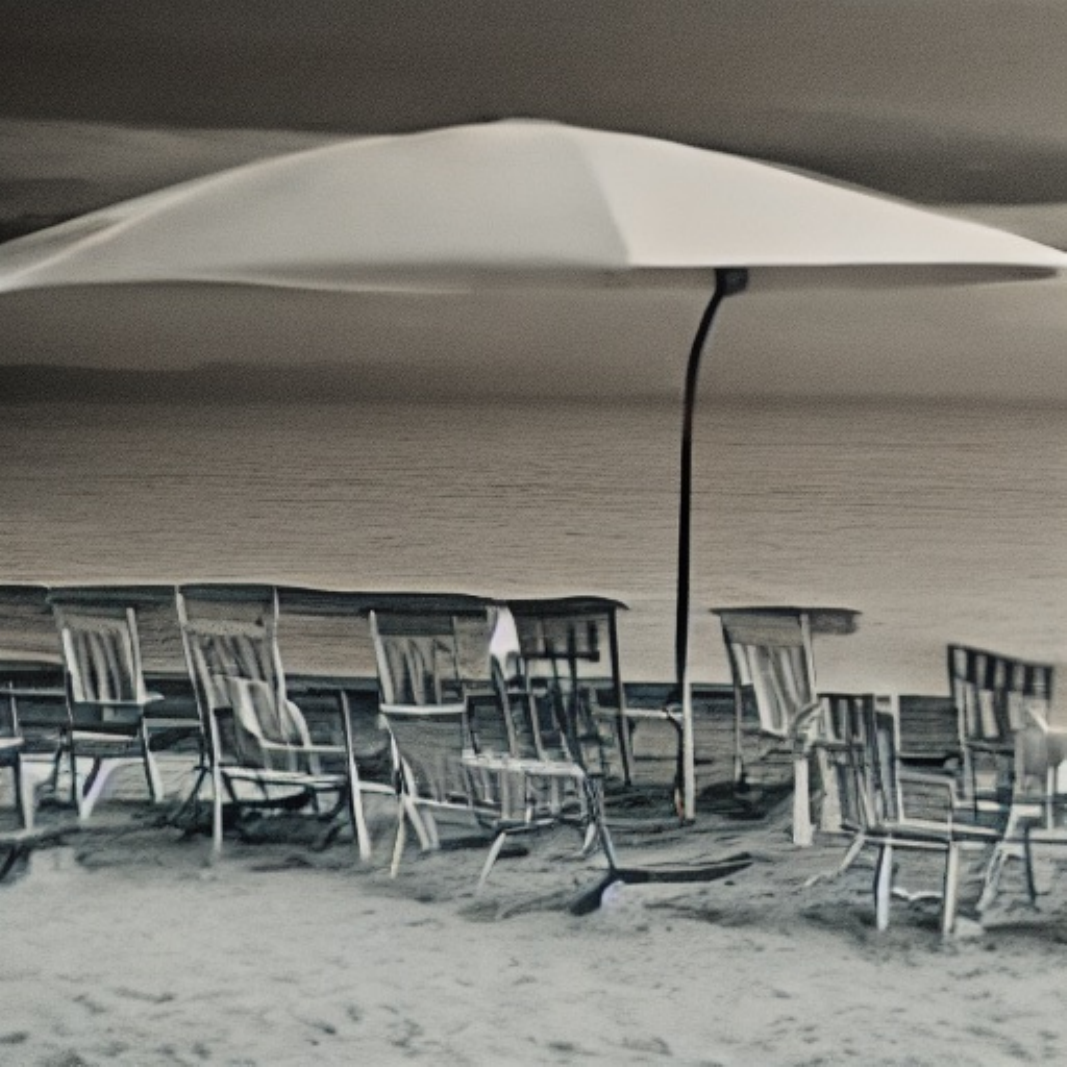}\hfill
    \includegraphics[width=0.24\linewidth]{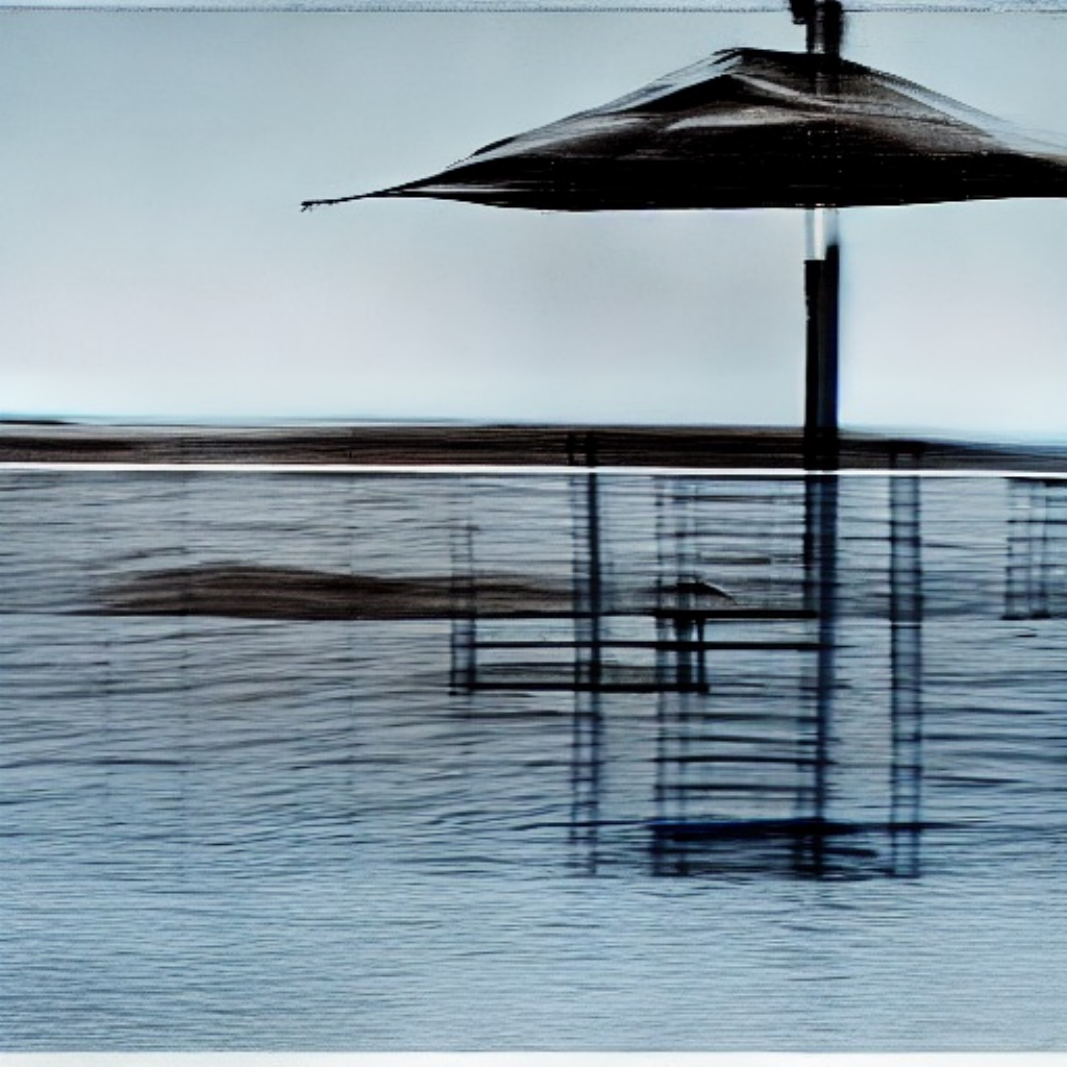}\hfill
    \includegraphics[width=0.24\linewidth]{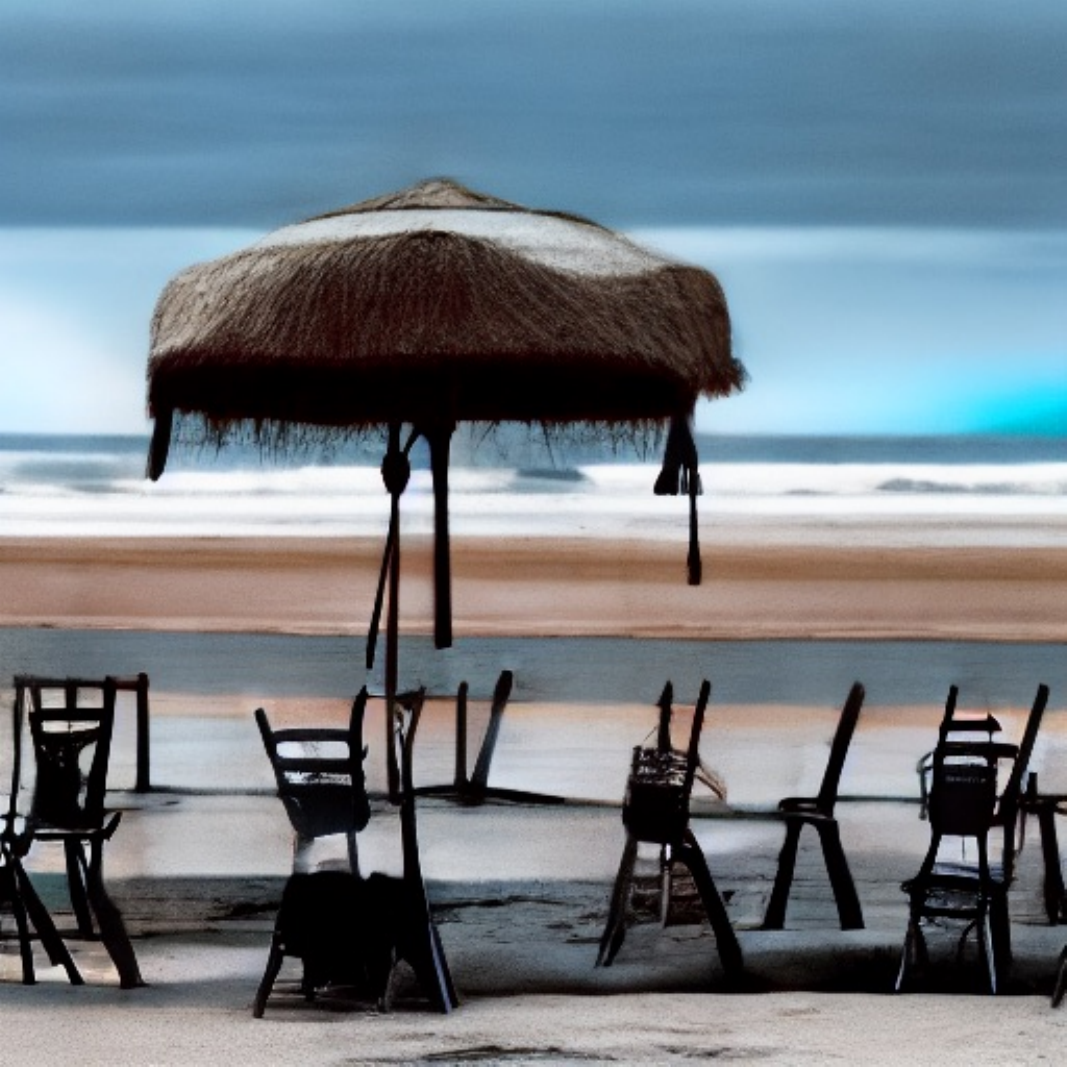}\hfill
    \includegraphics[width=0.24\linewidth]{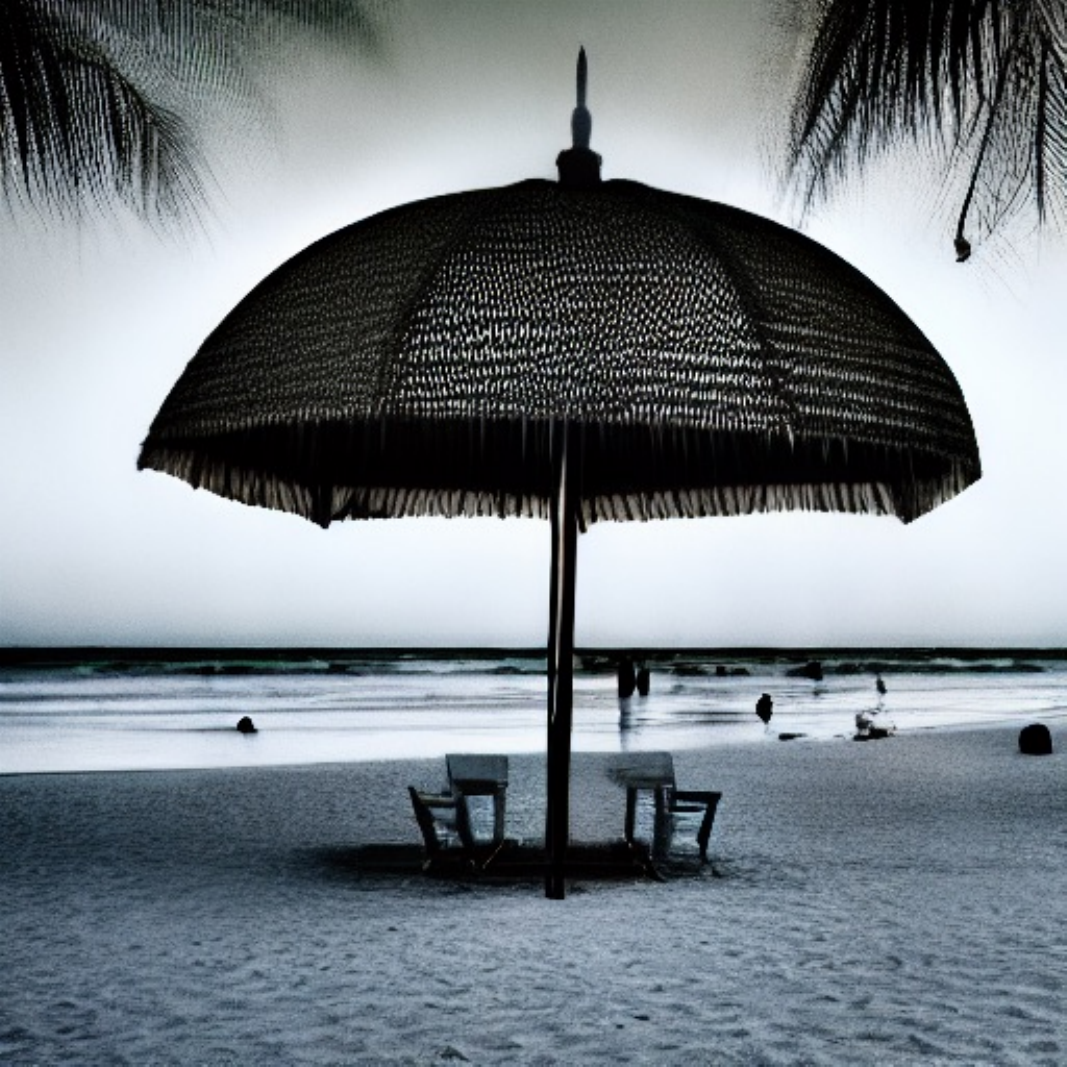}
    
    \vspace{0.5mm} 
    \caption*{ ``An old photo of an umbrella and chairs at the beach."}
    \vspace{1mm} %

    \includegraphics[width=0.24\linewidth]{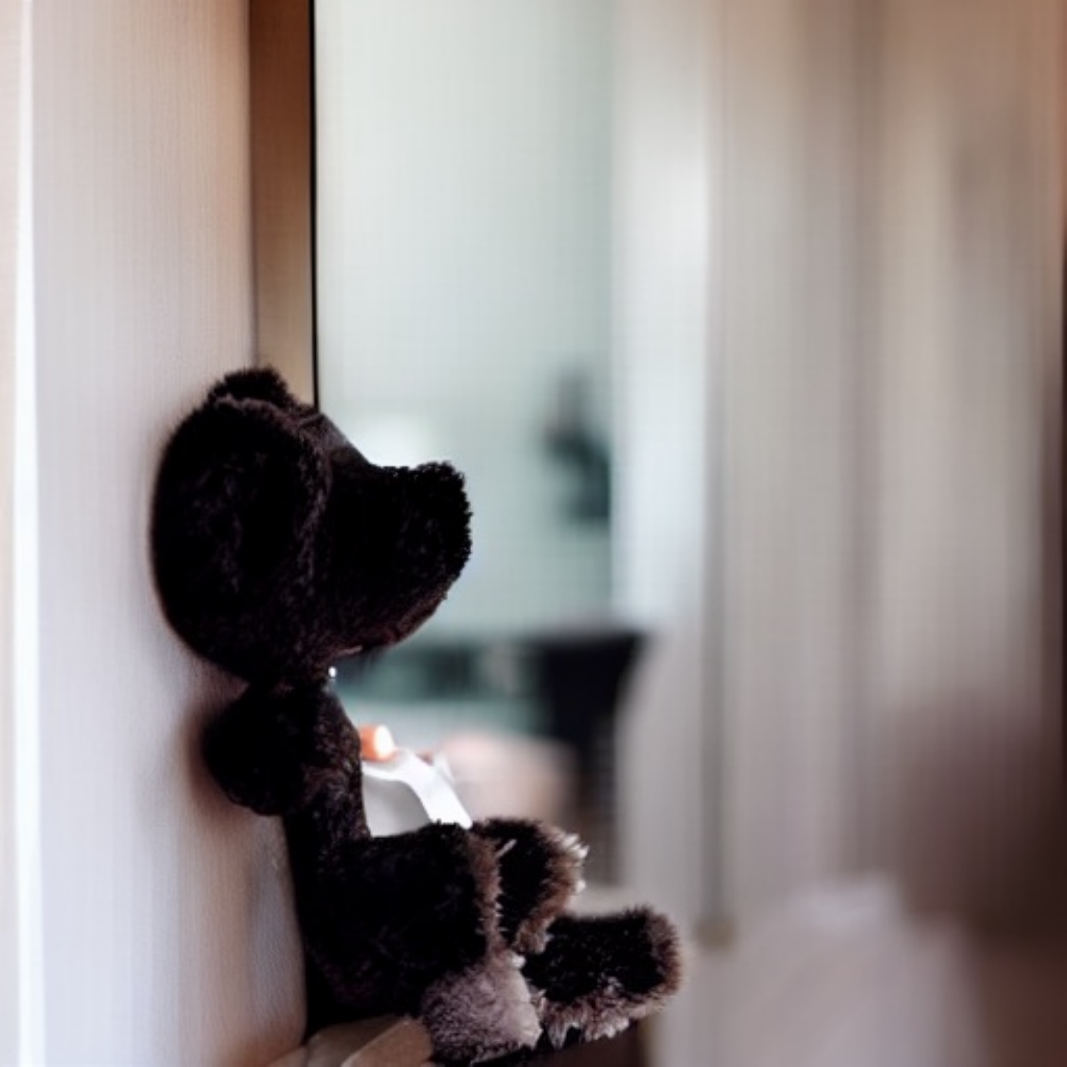}\hfill
    \includegraphics[width=0.24\linewidth]{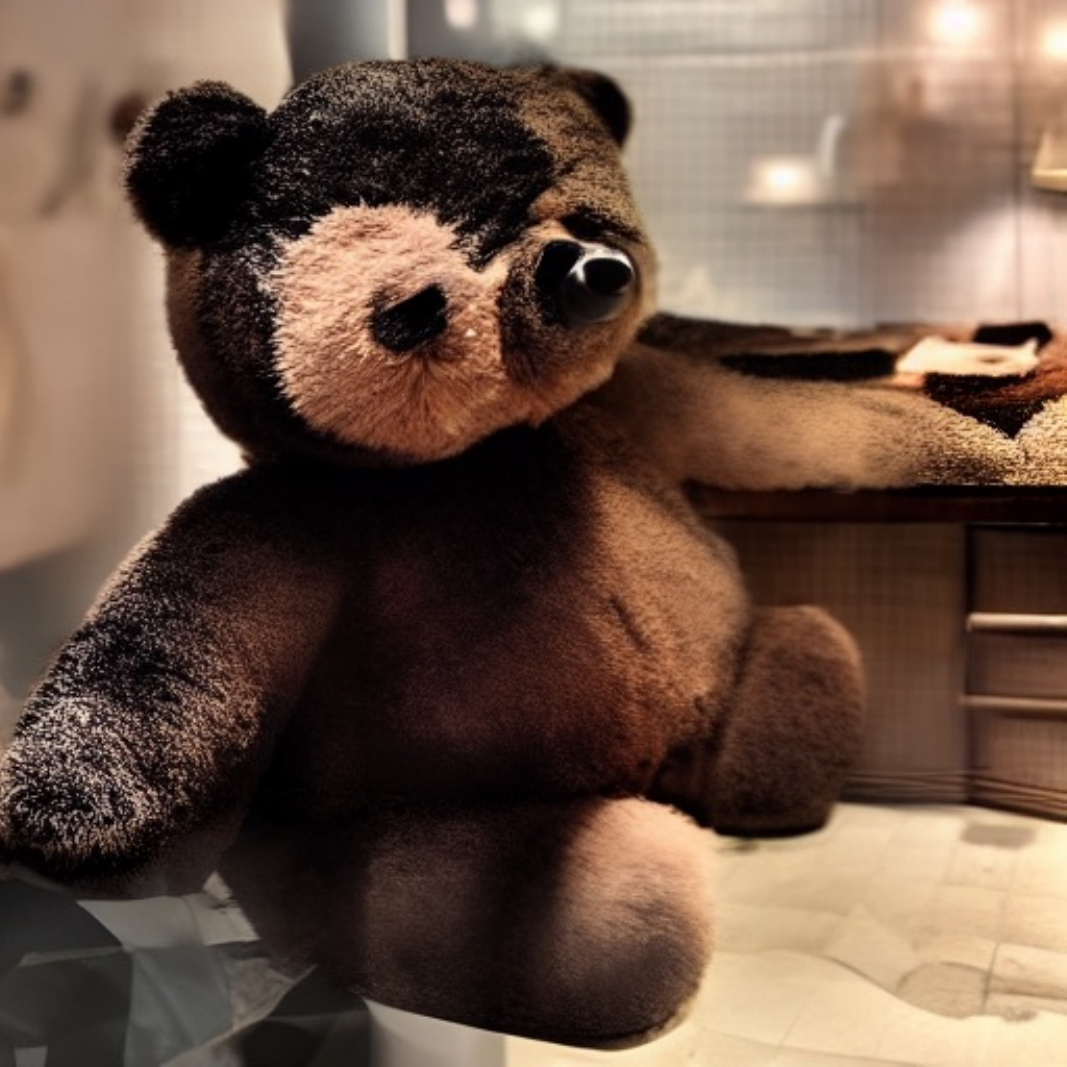}\hfill
    \includegraphics[width=0.24\linewidth]{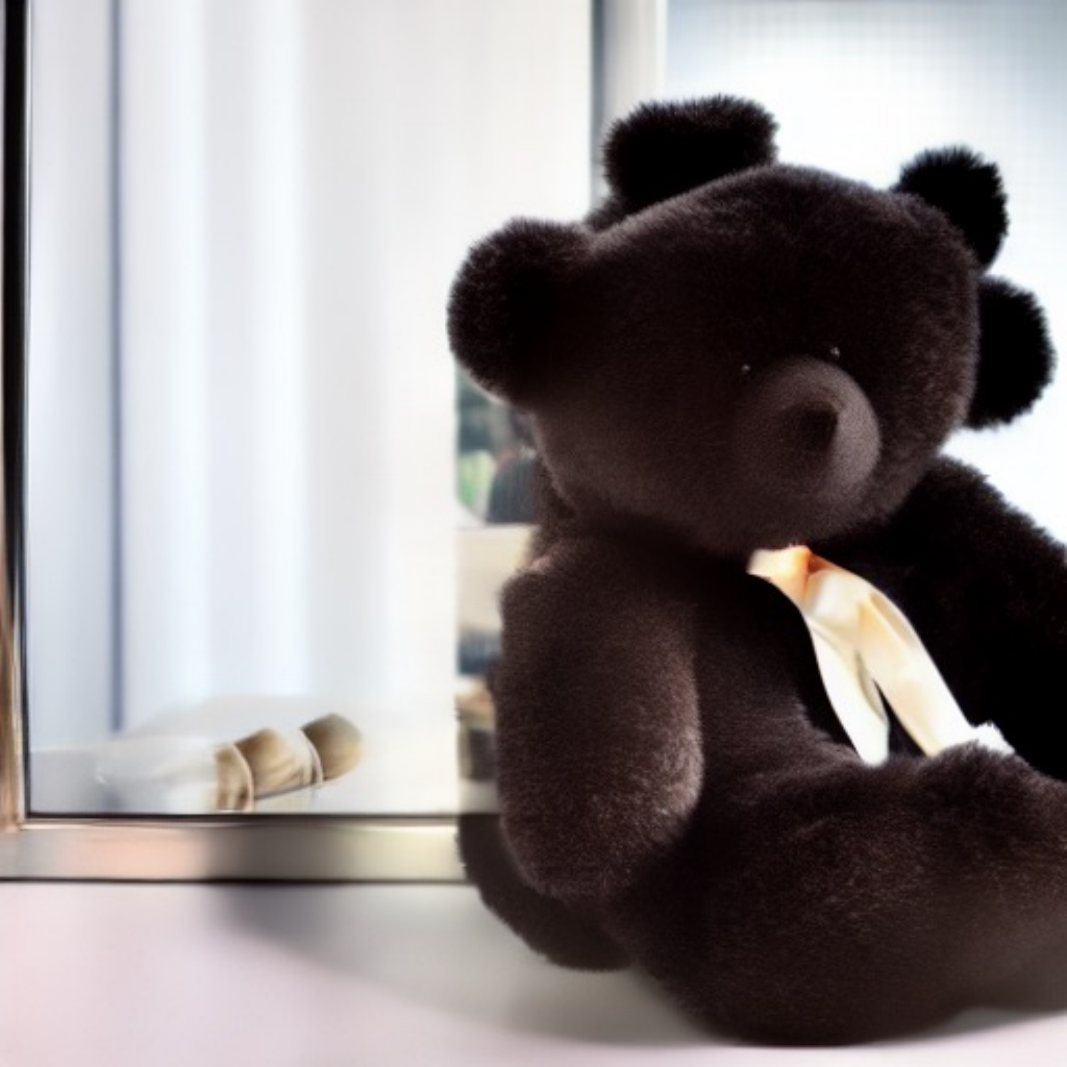}\hfill
    \includegraphics[width=0.24\linewidth]{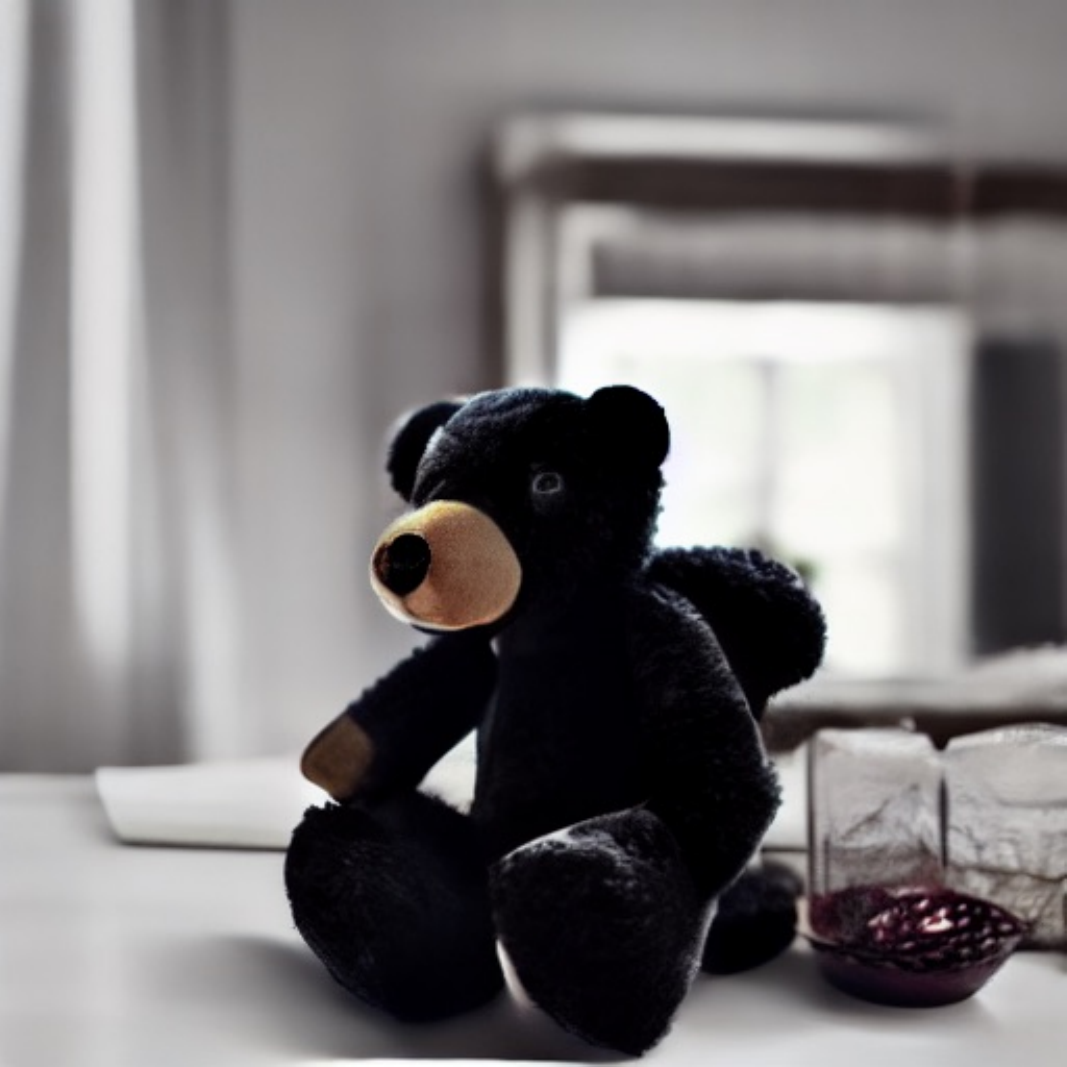}
    
    \vspace{0.5mm}
    \caption*{ ``black teddy bear placed in front of mirror."}
    
    \caption{ Steps = 4} 
    \label{subfig:nfe4}
  \end{subfigure}
  
  \vspace{1em} %

  \begin{subfigure}{\linewidth}
    \centering
    
    \includegraphics[width=0.24\linewidth]{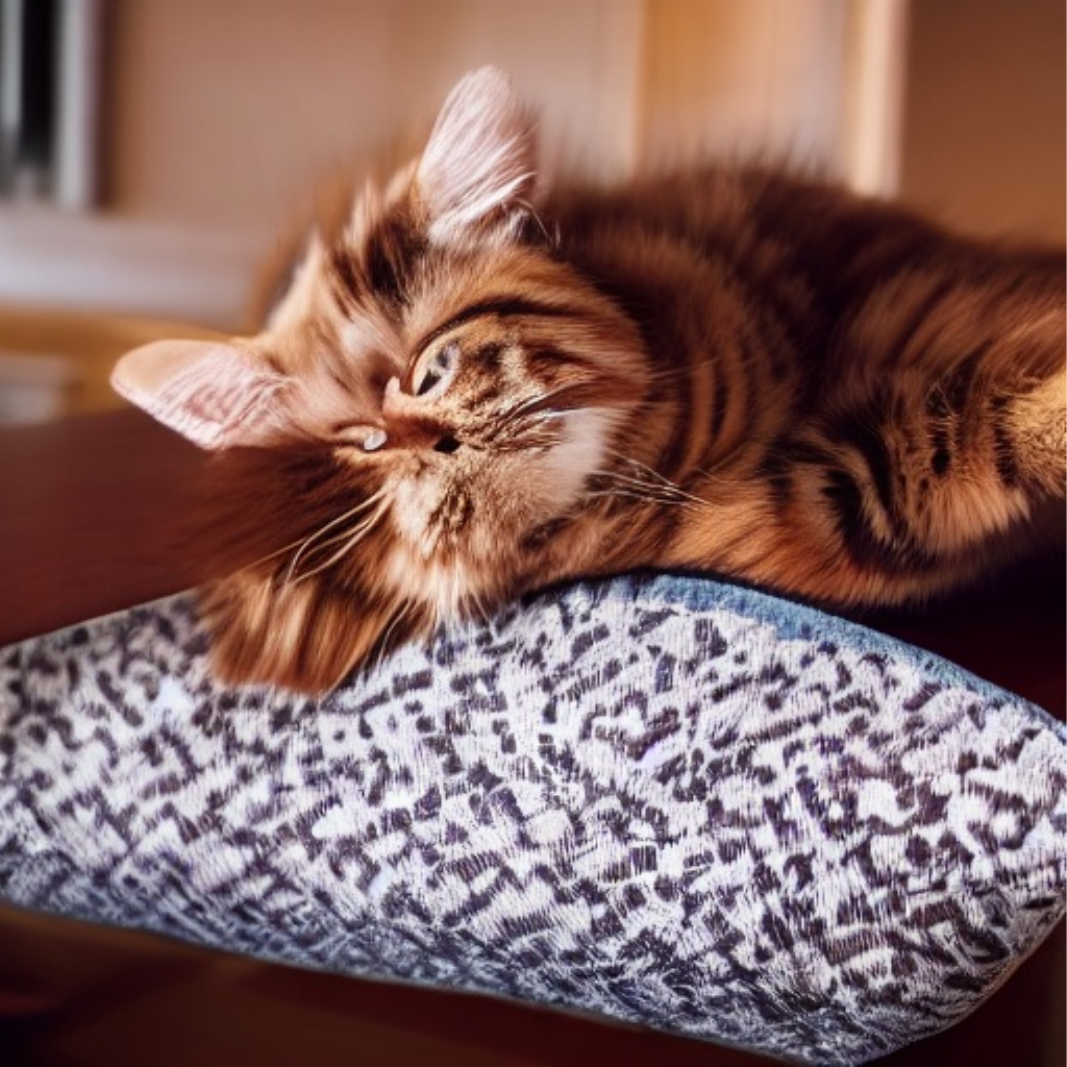}\hfill
    \includegraphics[width=0.24\linewidth]{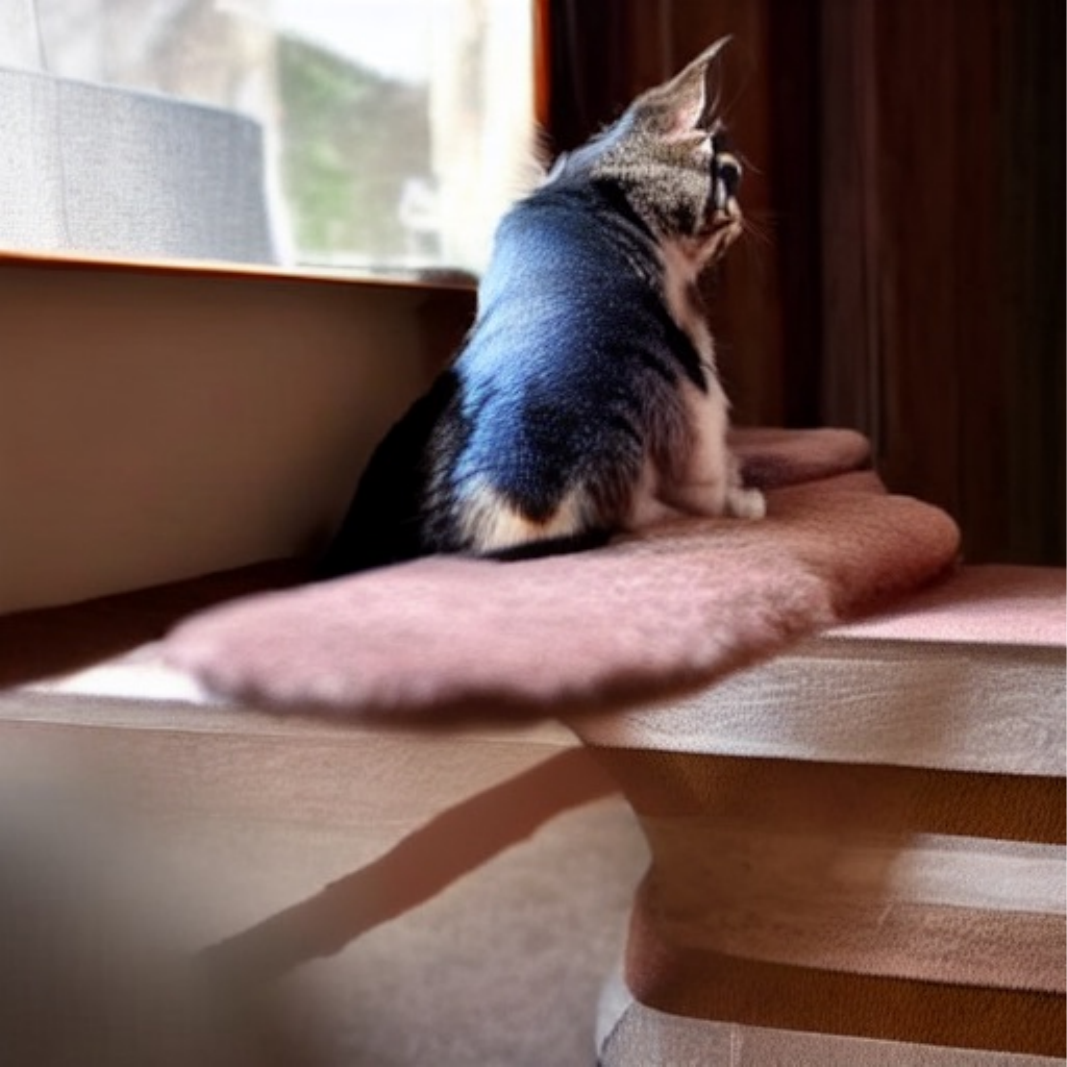}\hfill
    \includegraphics[width=0.24\linewidth]{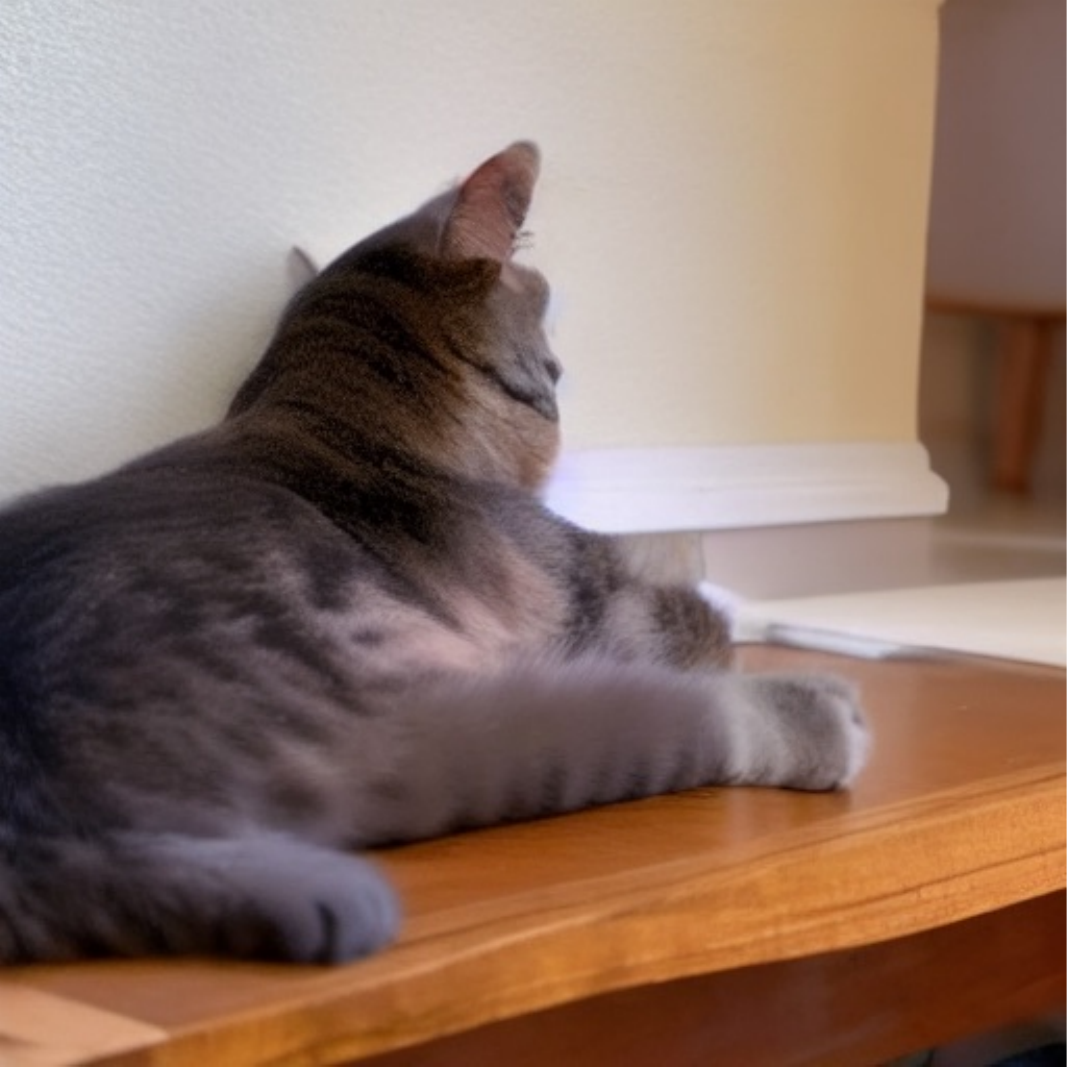}\hfill
    \includegraphics[width=0.24\linewidth]{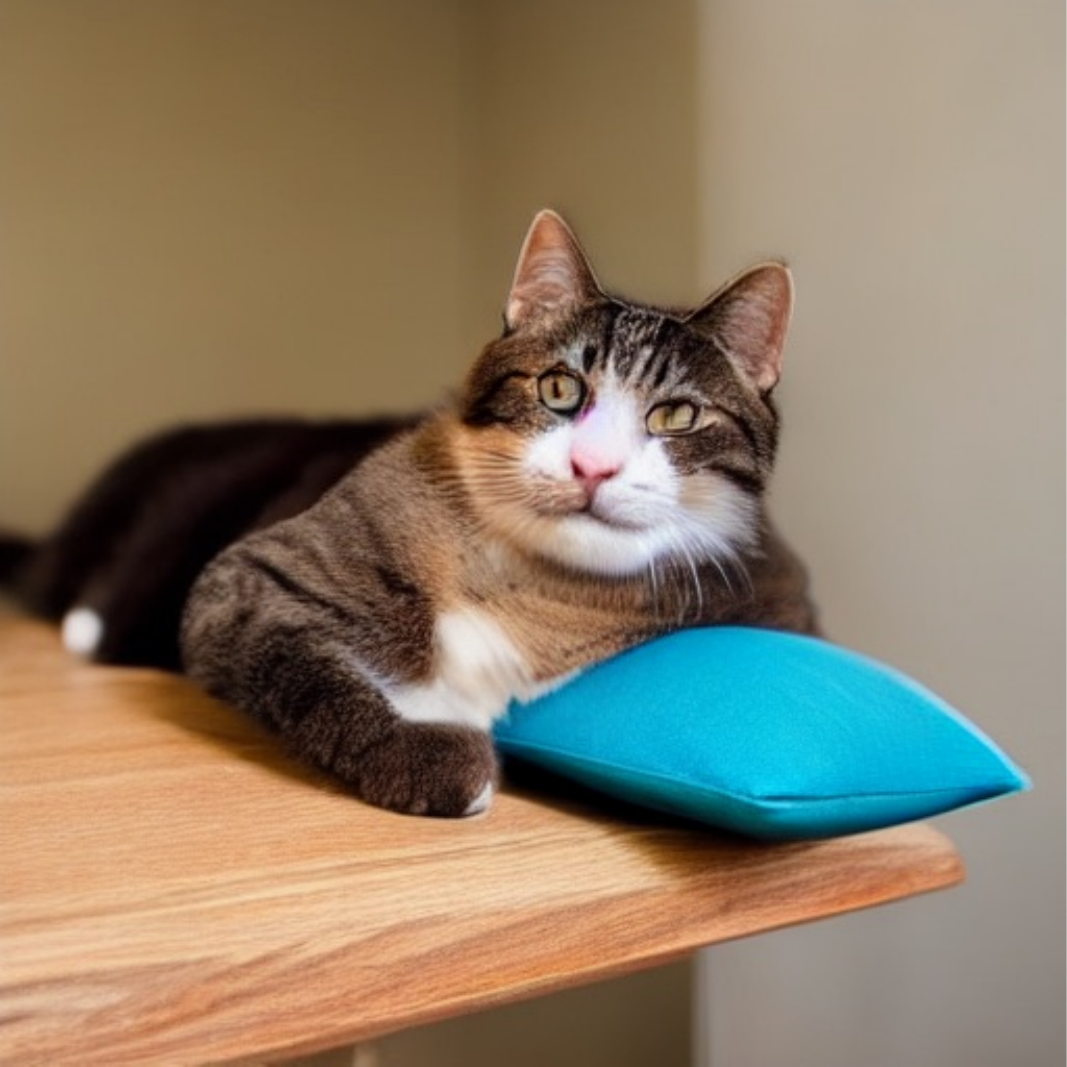}
    
    \vspace{0.5mm}
    \caption*{``A cat laying on a cushion on top of a table."}
    \vspace{1mm}

    \includegraphics[width=0.24\linewidth]{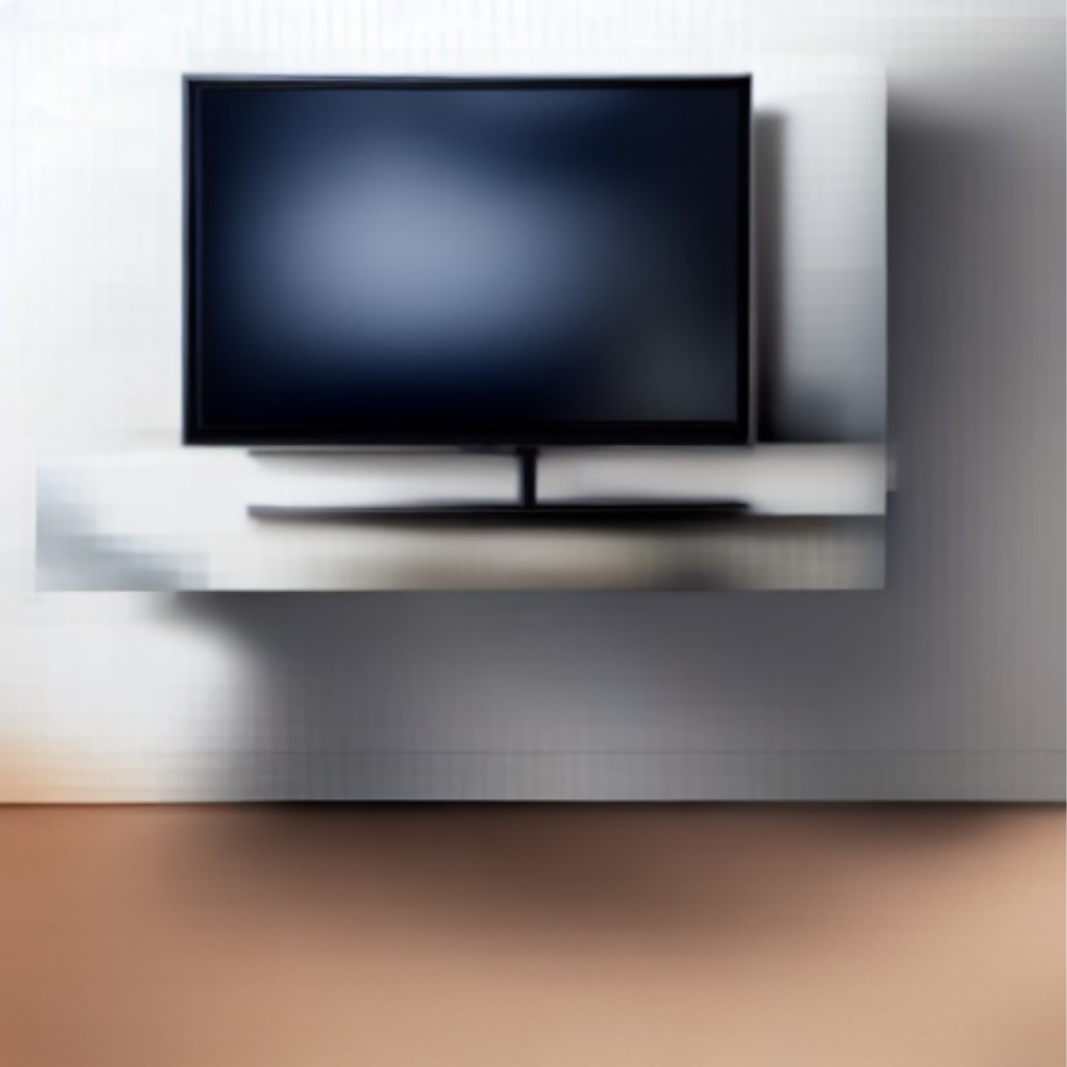}\hfill
    \includegraphics[width=0.24\linewidth]{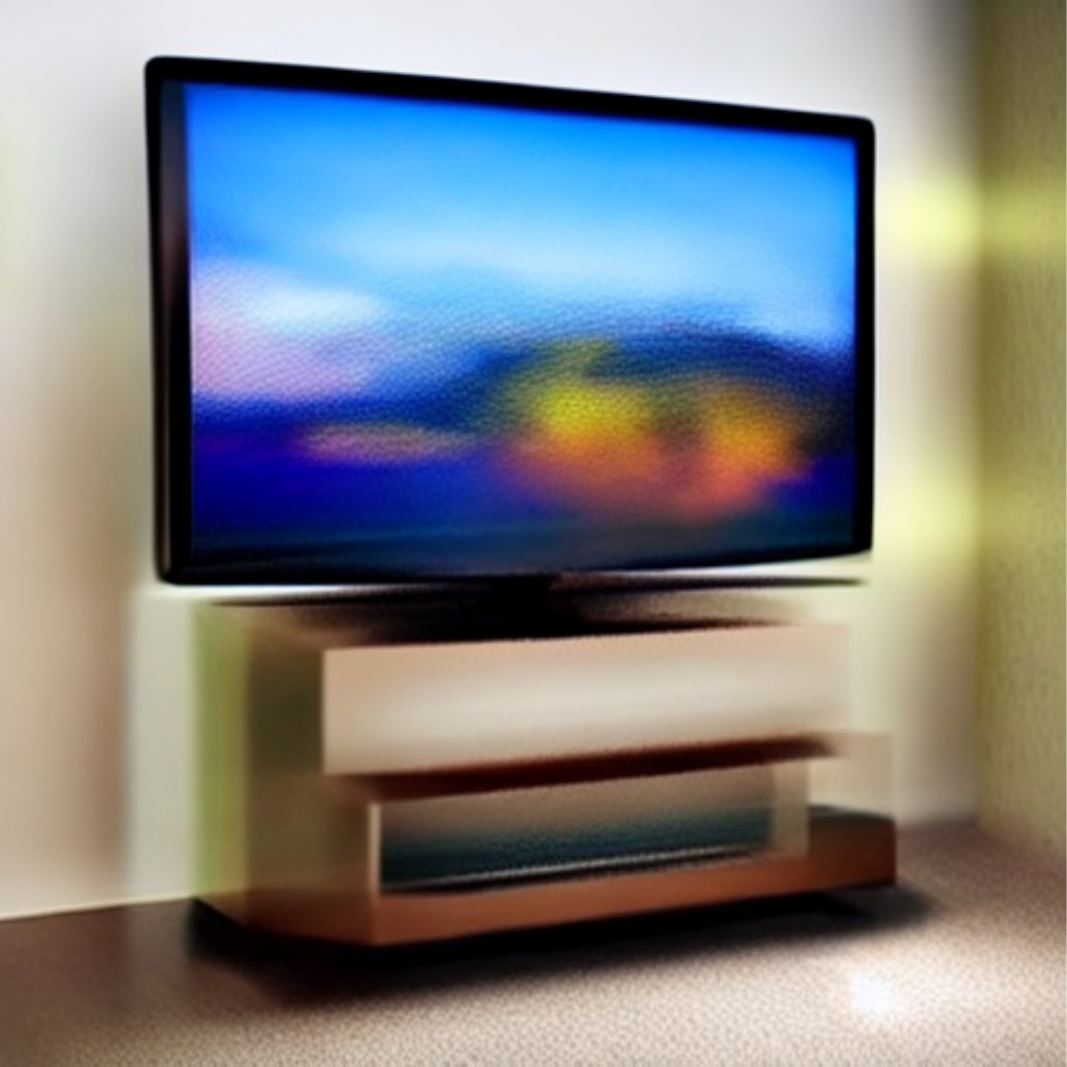}\hfill
    \includegraphics[width=0.24\linewidth]{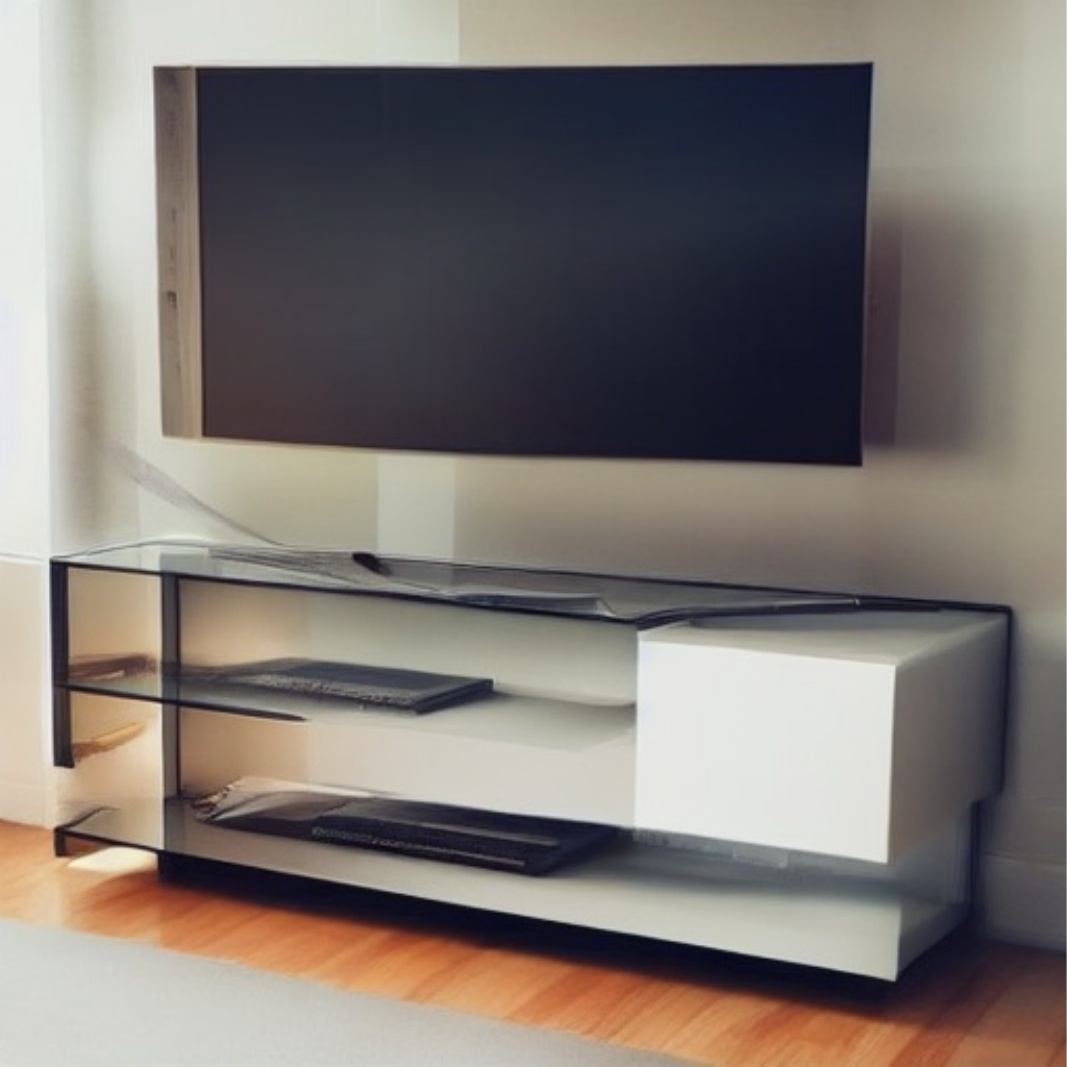}\hfill
    \includegraphics[width=0.24\linewidth]{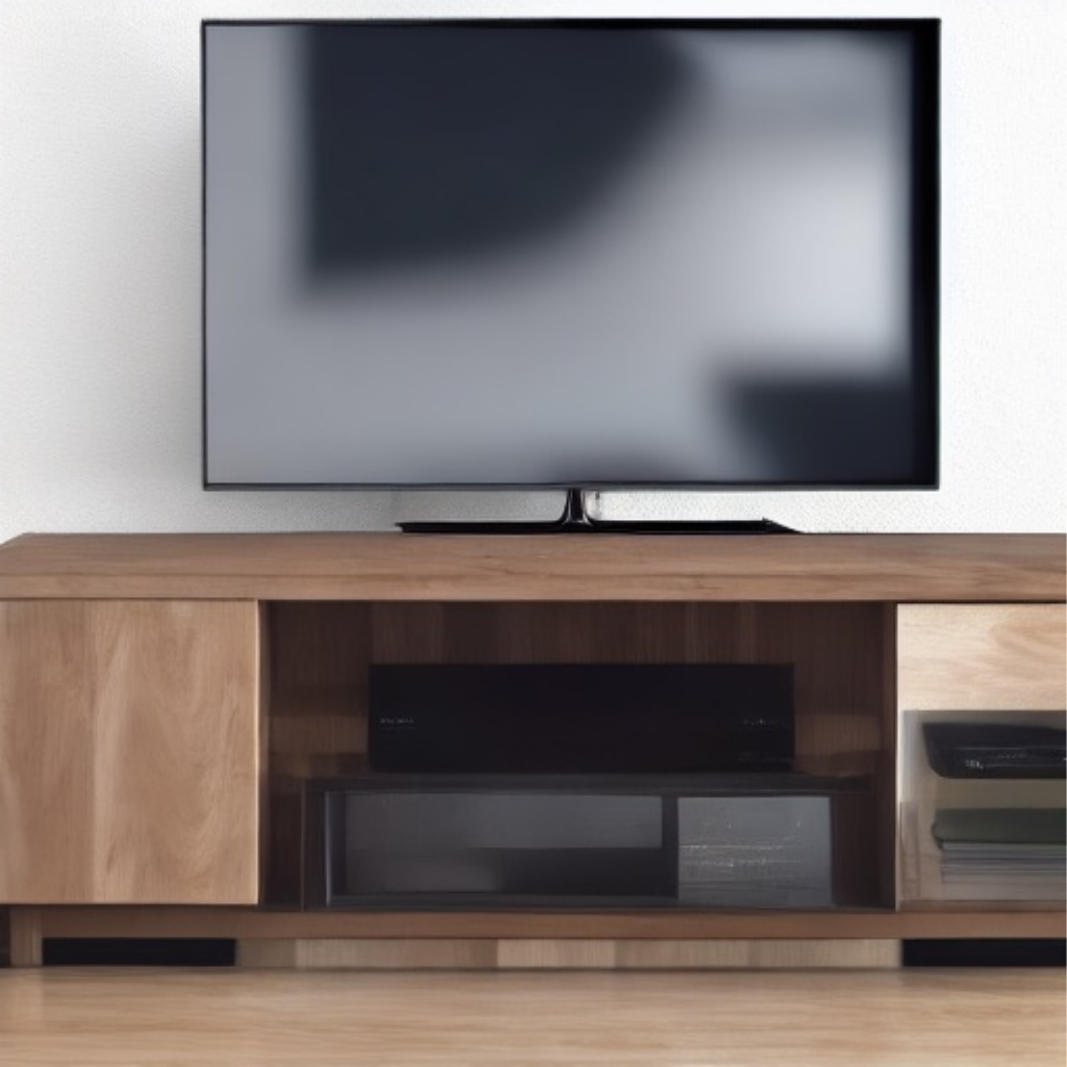}
    
    \vspace{0.5mm}
    \caption*{``A television that is sitting on a stand."}
    
    \caption{Steps = 5}
    \label{subfig:nfe5}
  \end{subfigure}
  \caption{Side-by-side comparison of selected images generated with Stable Diffusion and iPNDM solver in the low NFE regime (Steps $\in \{4, 5\}$).
    Methods (from left to right): DMN, GITS, LD3, and D2PO.}
  \label{fig:sd_ipndm_nfelow}
\end{figure*}

\begin{figure*}[t] 
  \centering
  
  \captionsetup[subfigure]{font=small, labelfont=small}

  \begin{subfigure}{\linewidth}
    \centering
    
    \begin{minipage}{0.24\linewidth}\centering\small DMN\end{minipage} \hfill
    \begin{minipage}{0.24\linewidth}\centering\small GITS\end{minipage} \hfill
    \begin{minipage}{0.24\linewidth}\centering\small LD3\end{minipage} \hfill
    \begin{minipage}{0.24\linewidth}\centering\small D2PO\end{minipage}
    \vspace{1mm} 
    
    \includegraphics[width=0.24\linewidth]{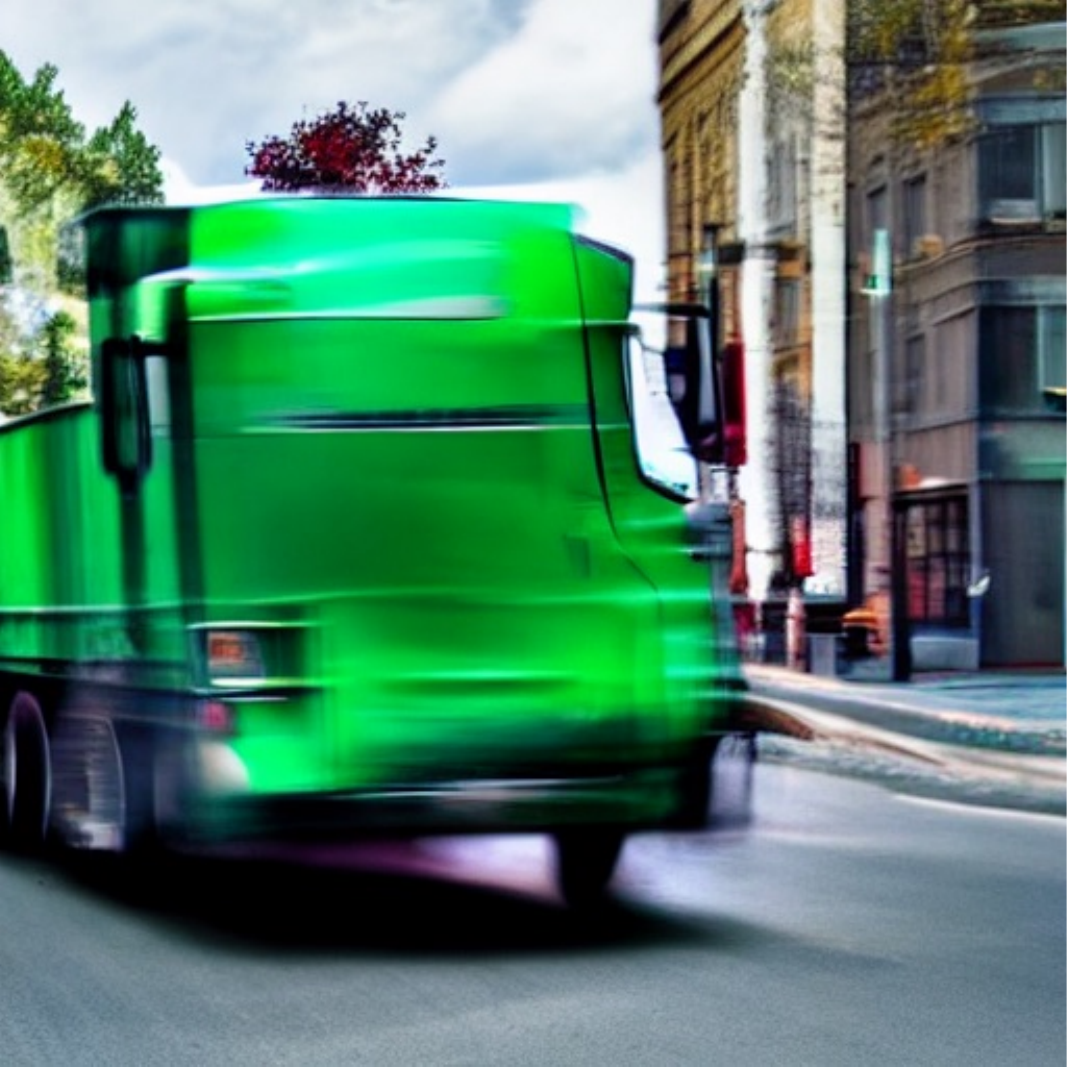}\hfill
    \includegraphics[width=0.24\linewidth]{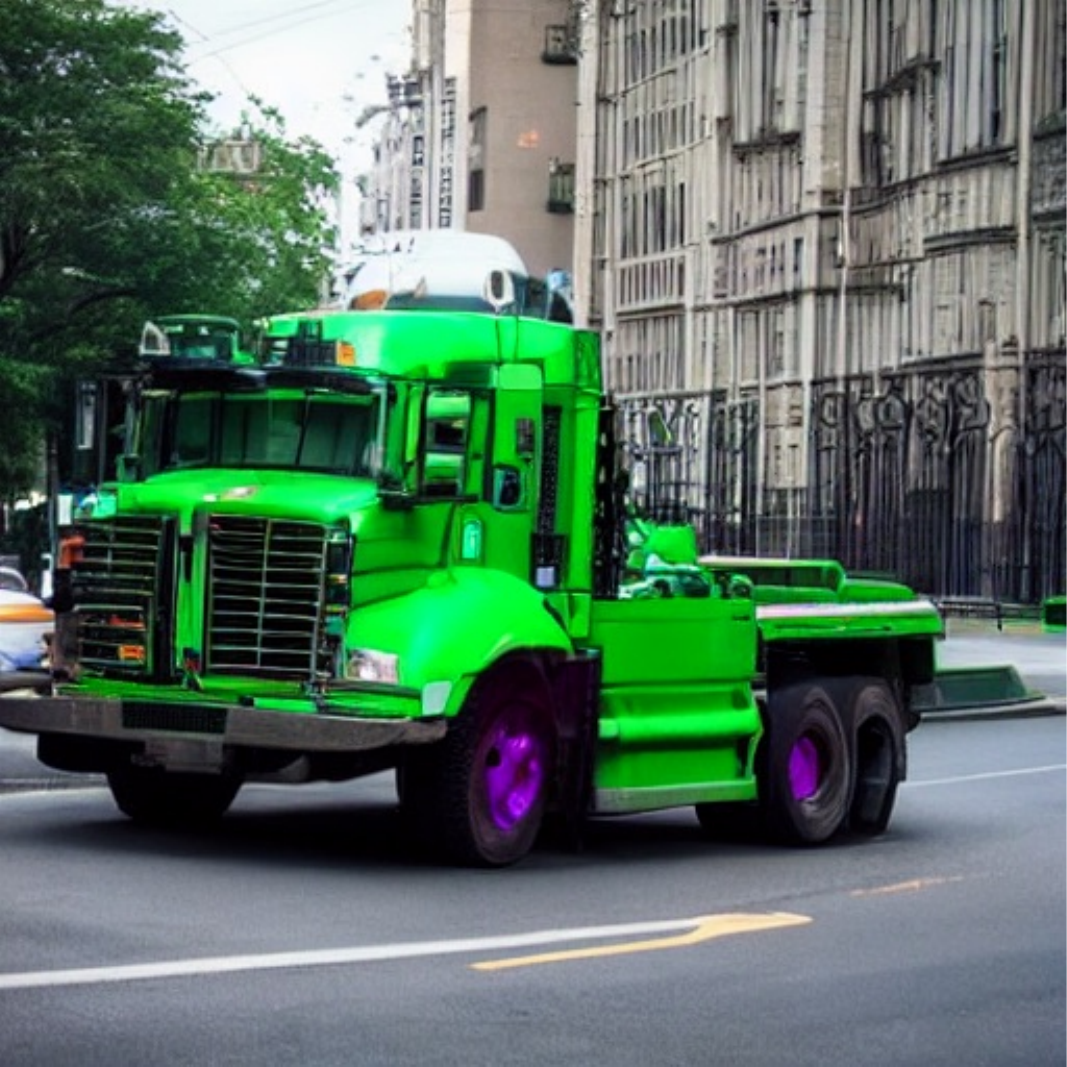}\hfill
    \includegraphics[width=0.24\linewidth]{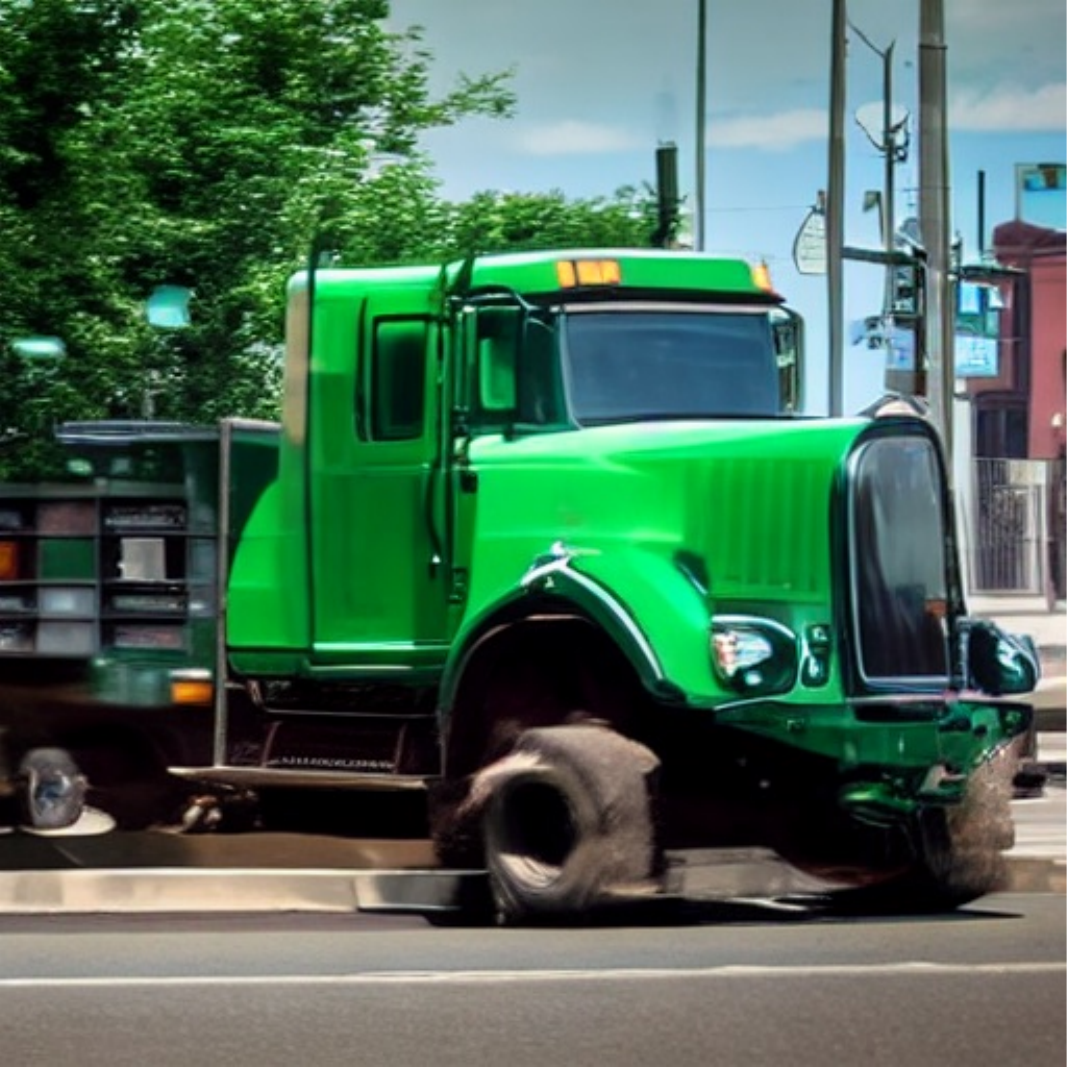}\hfill
    \includegraphics[width=0.24\linewidth]{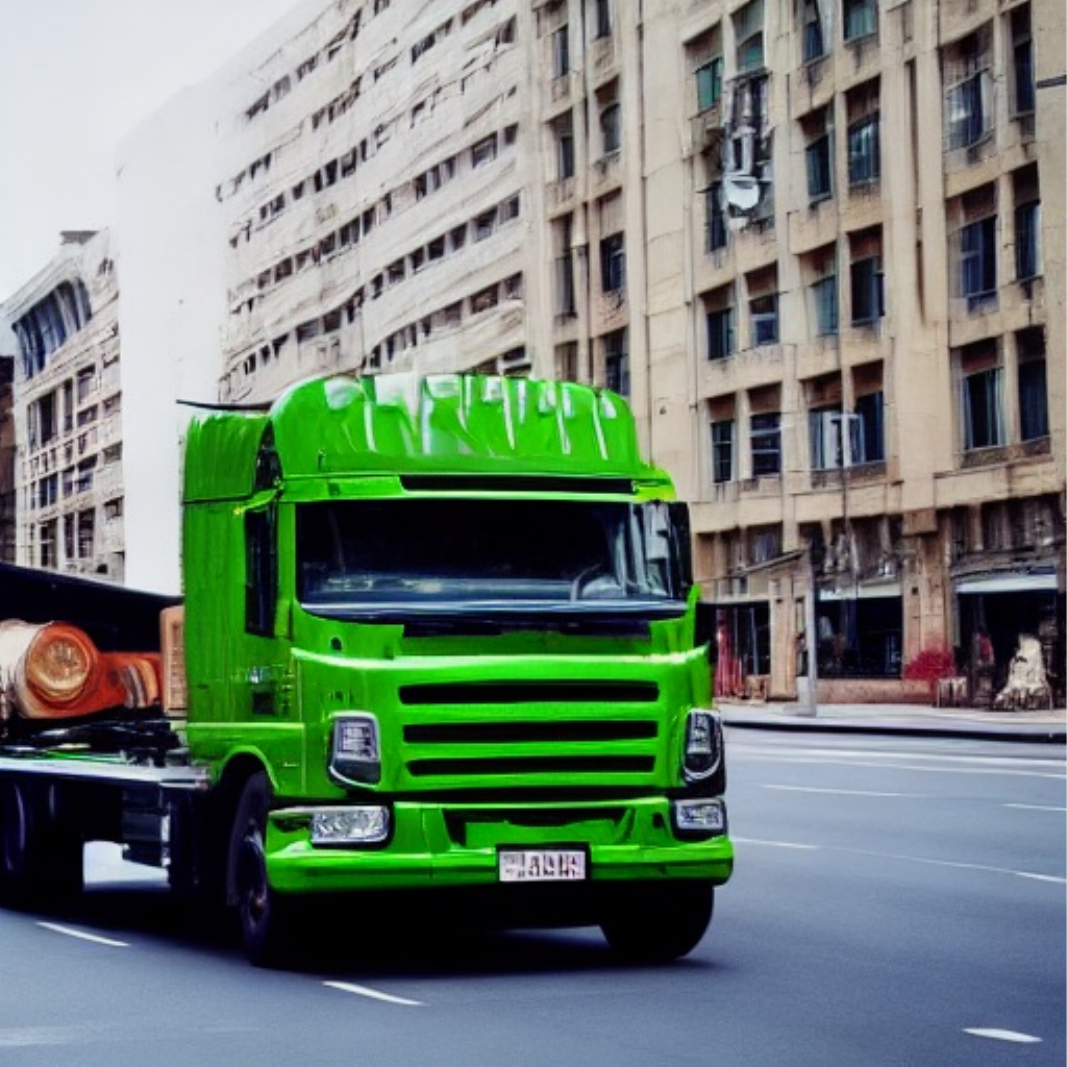}
    
    \vspace{0.5mm} 
    \caption*{\small ``A large green truck on a city street."}
    \vspace{1mm} %

    \includegraphics[width=0.24\linewidth]{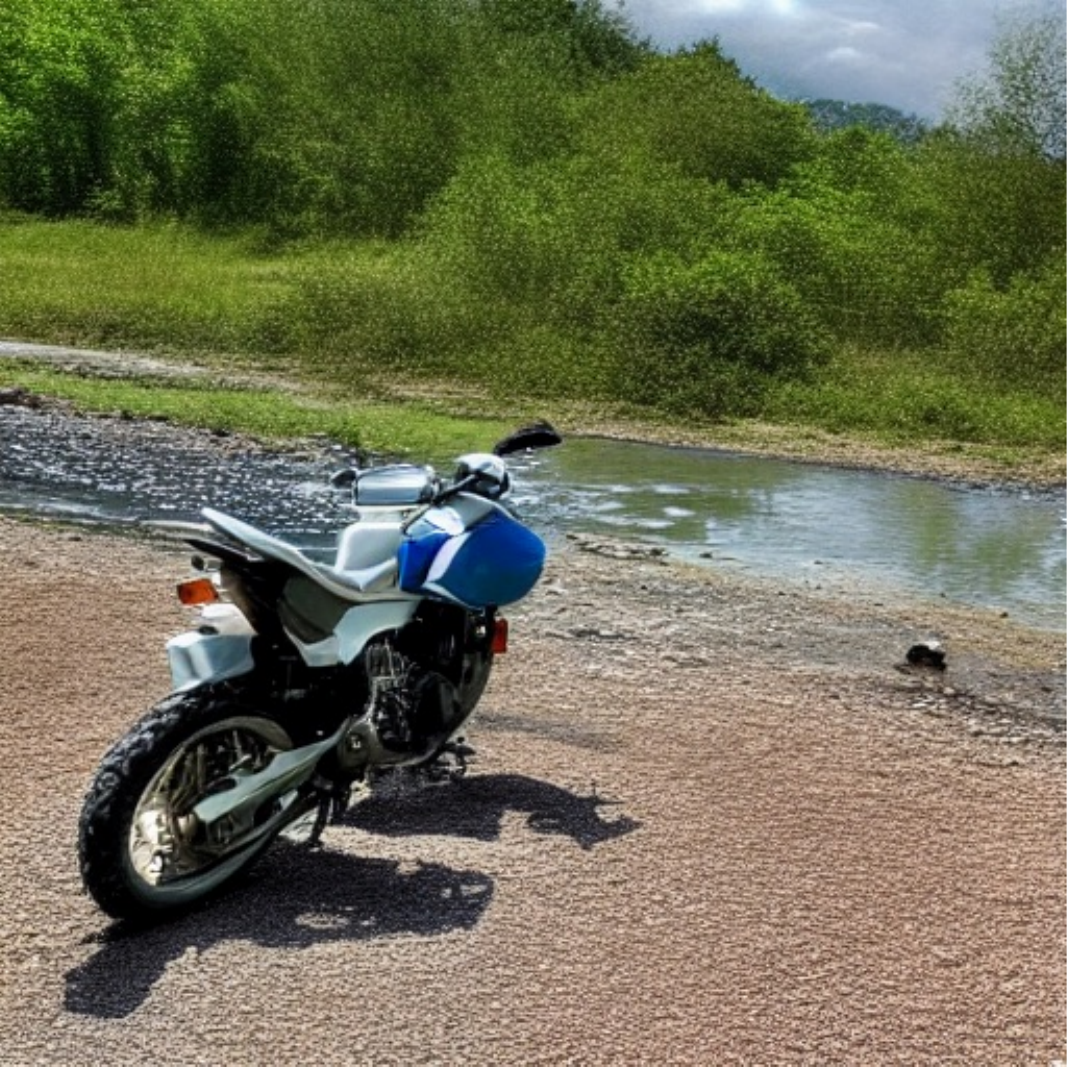}\hfill
    \includegraphics[width=0.24\linewidth]{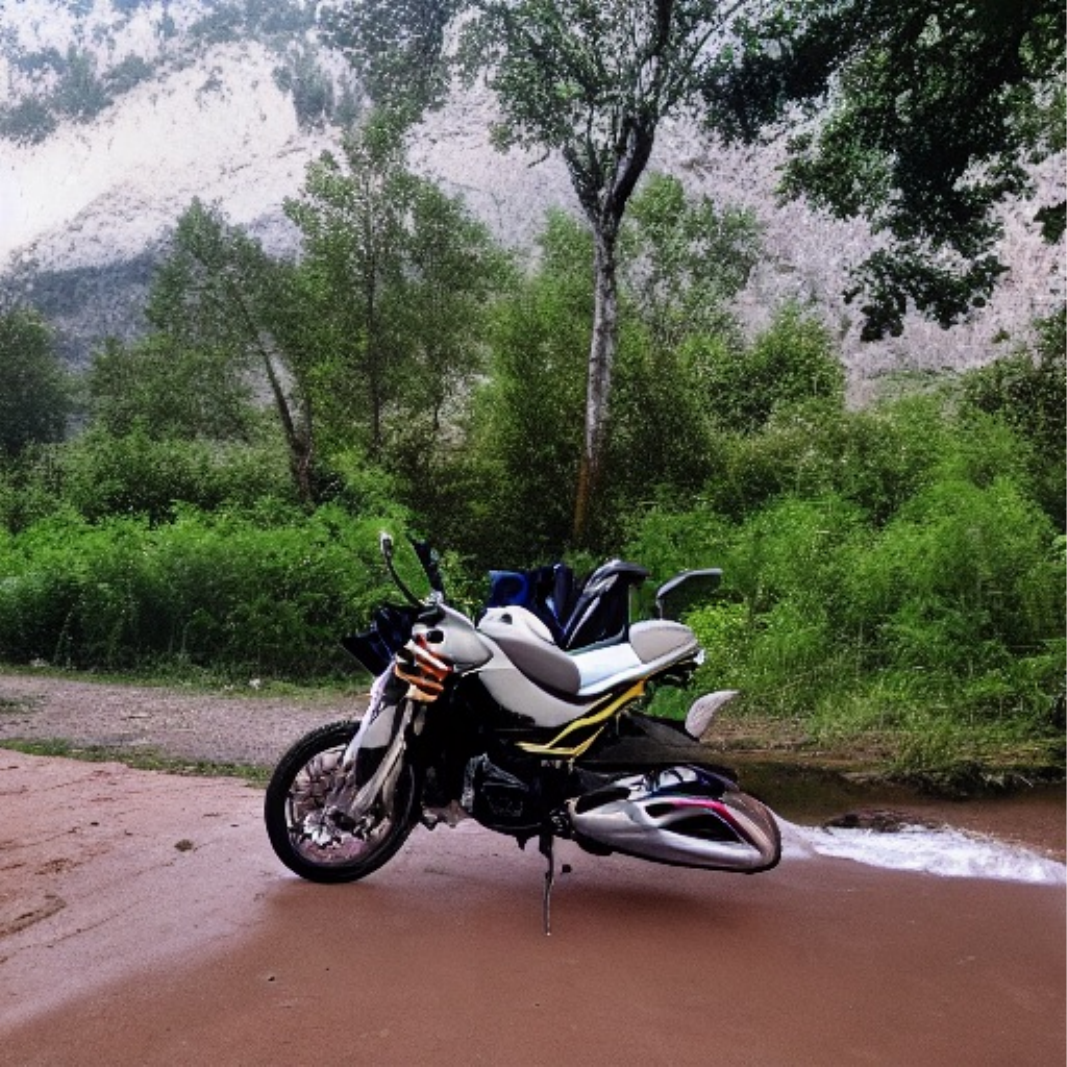}\hfill
    \includegraphics[width=0.24\linewidth]{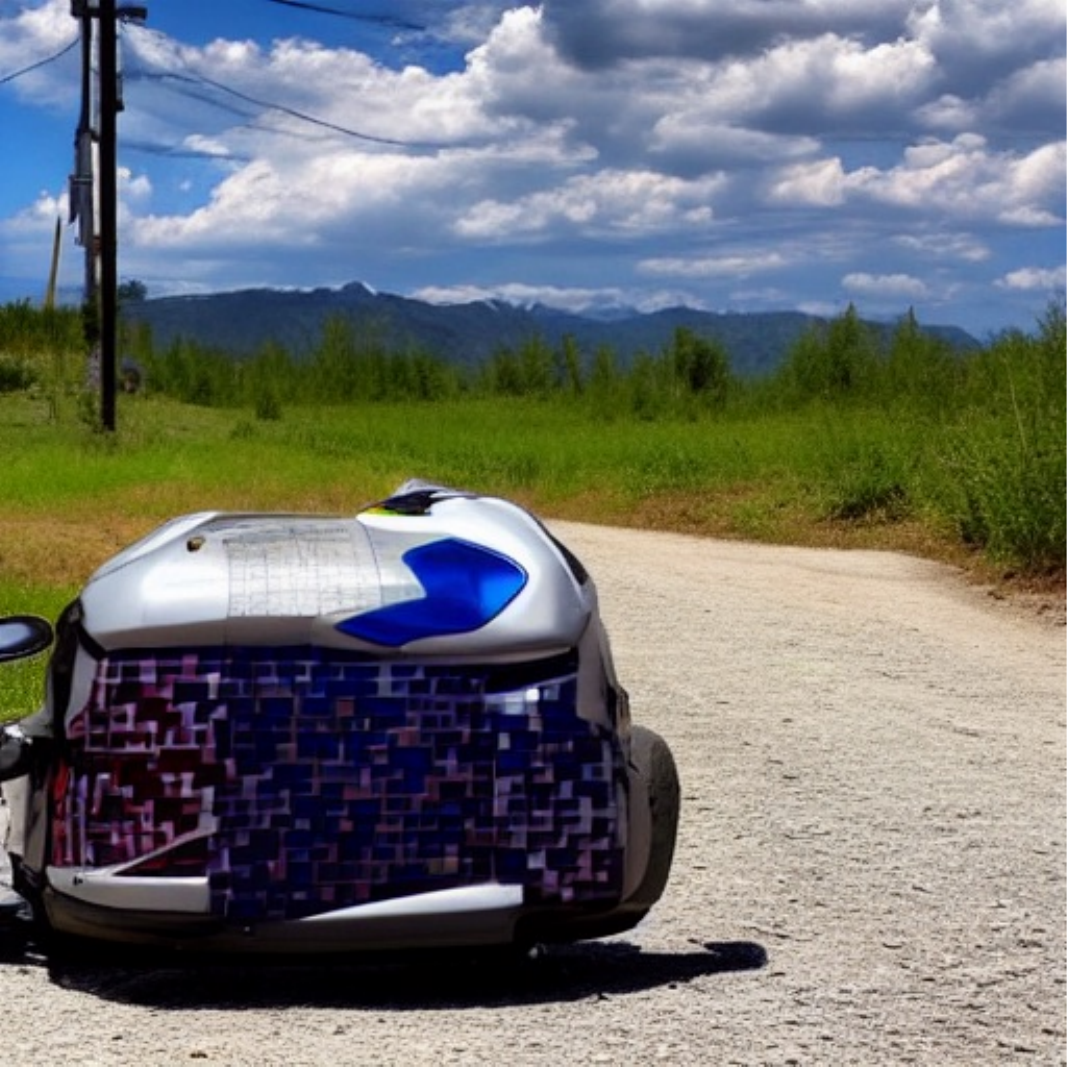}\hfill
    \includegraphics[width=0.24\linewidth]{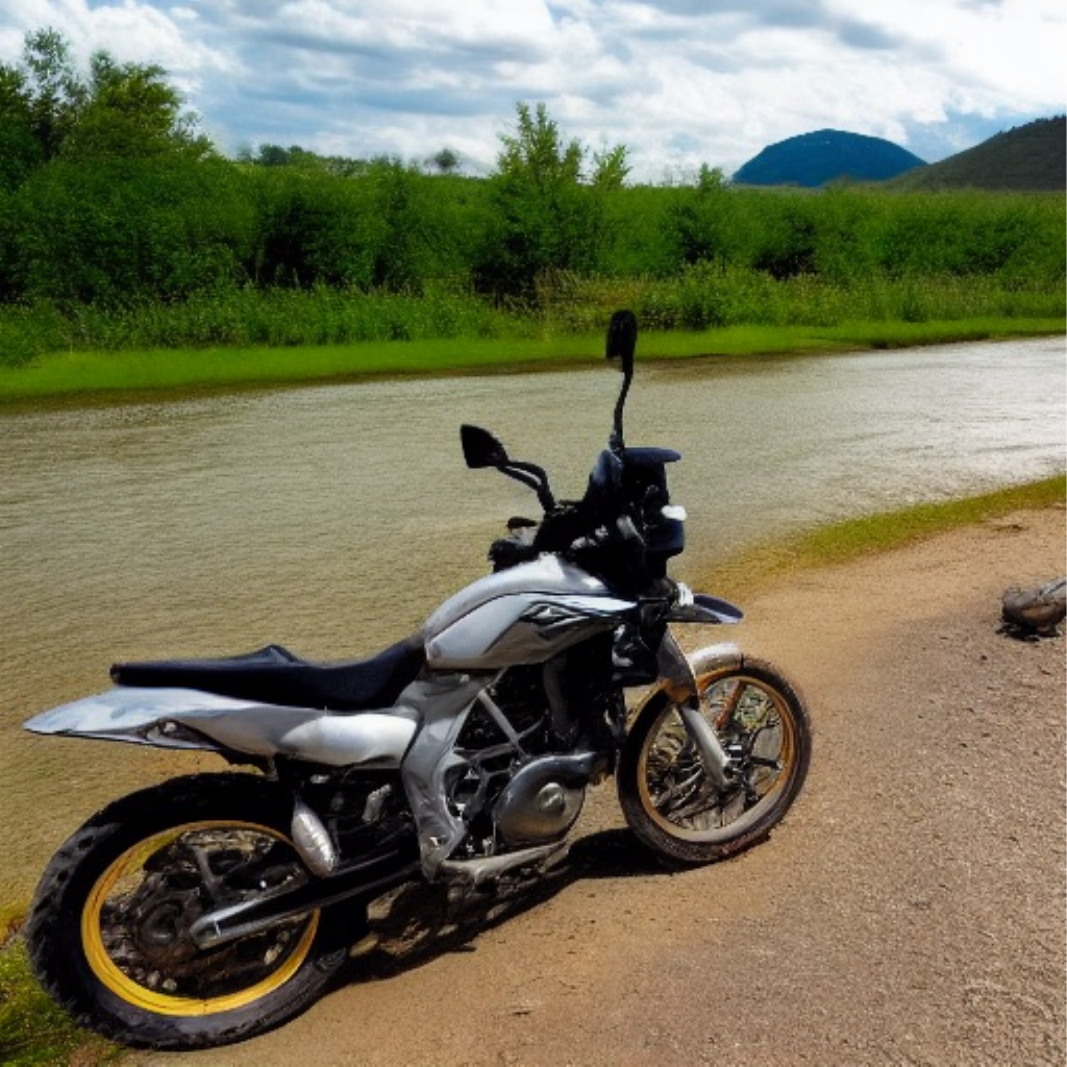}
    
    \vspace{0.5mm}
    \caption*{\small ``A sports motorcycle is parked on a gravel road by a river."}
    
    \caption{Steps = 6} 
    \label{subfig:nfe6}
  \end{subfigure}
  
  \vspace{1em} %

  \begin{subfigure}{\linewidth}
    \centering
    
    \includegraphics[width=0.24\linewidth]{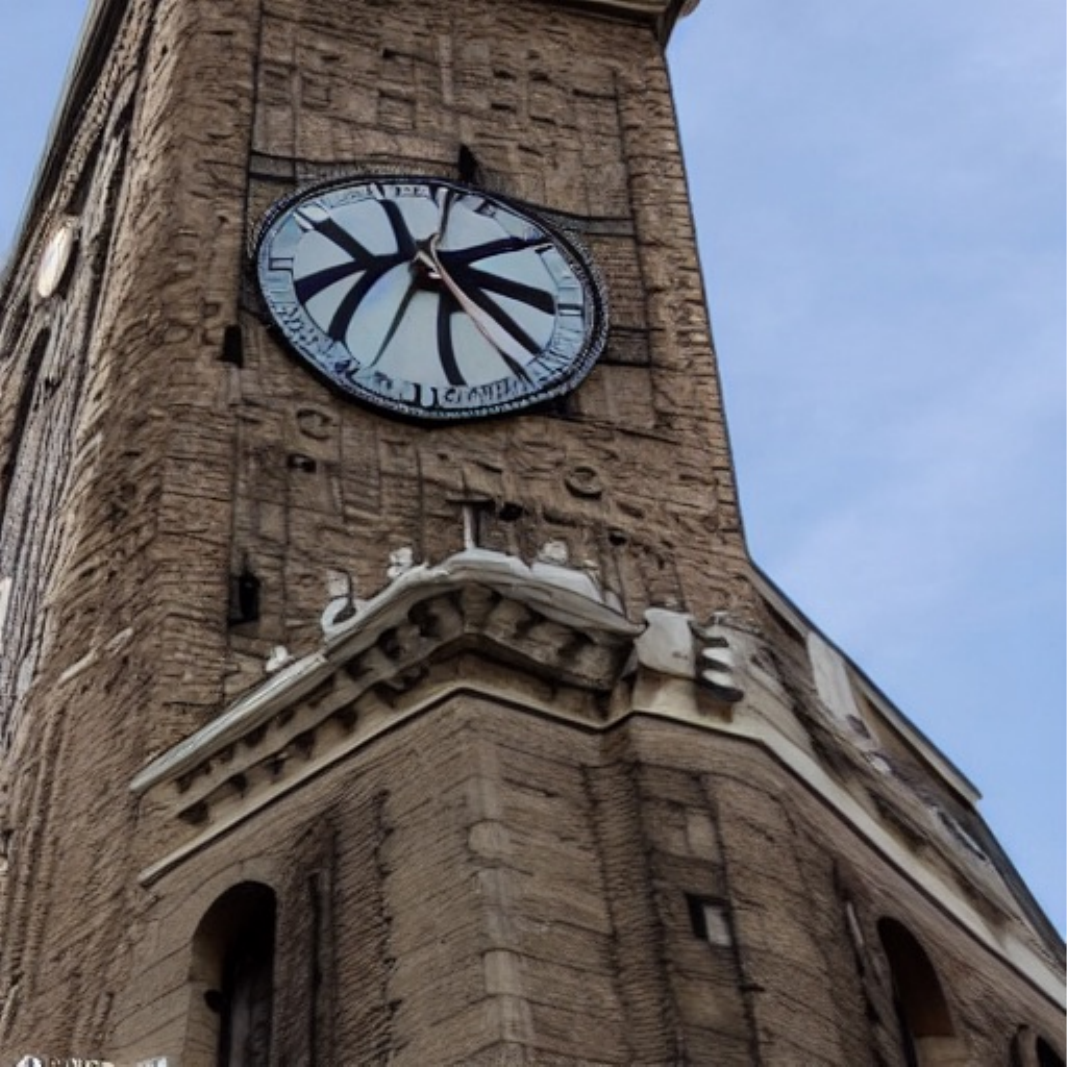}\hfill
    \includegraphics[width=0.24\linewidth]{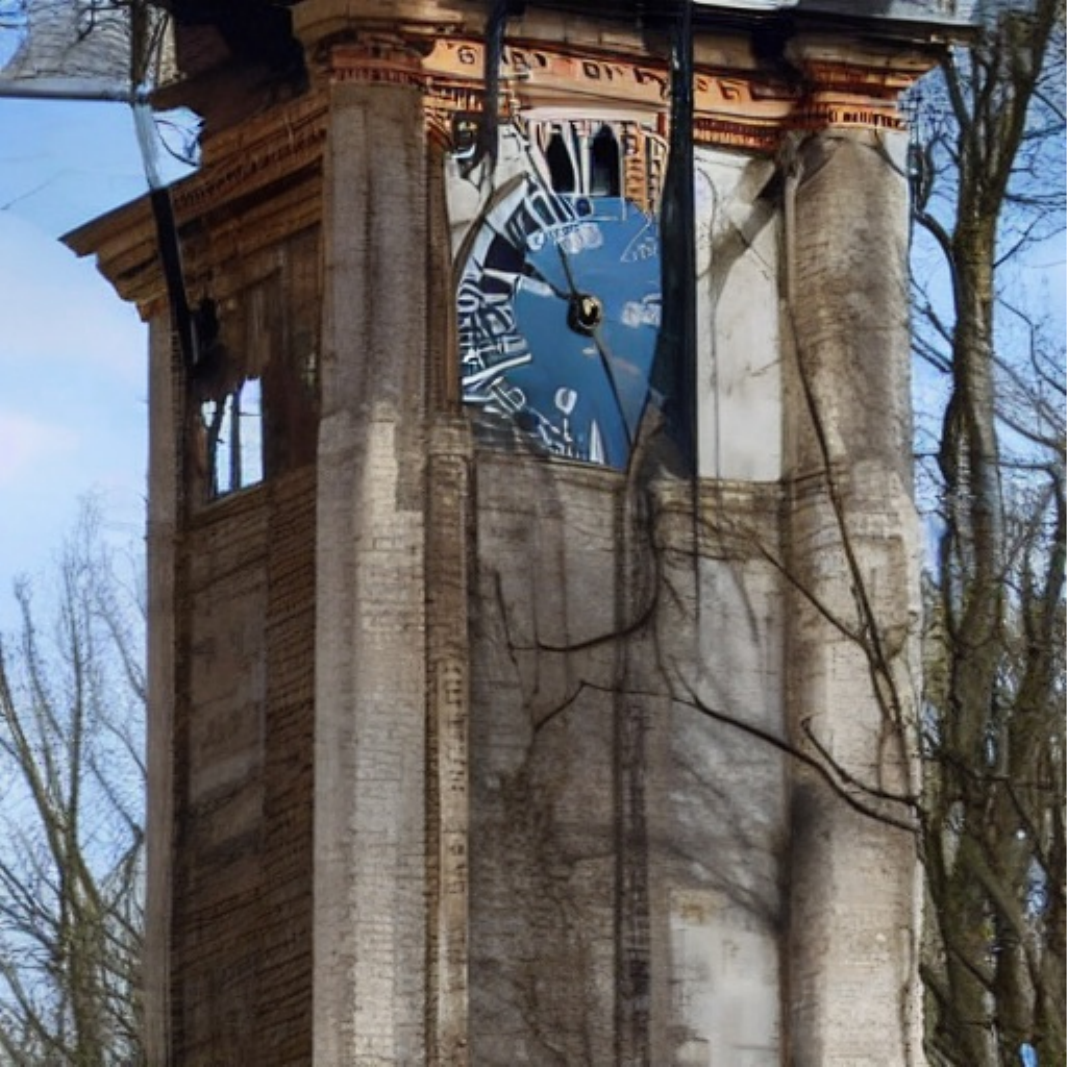}\hfill
    \includegraphics[width=0.24\linewidth]{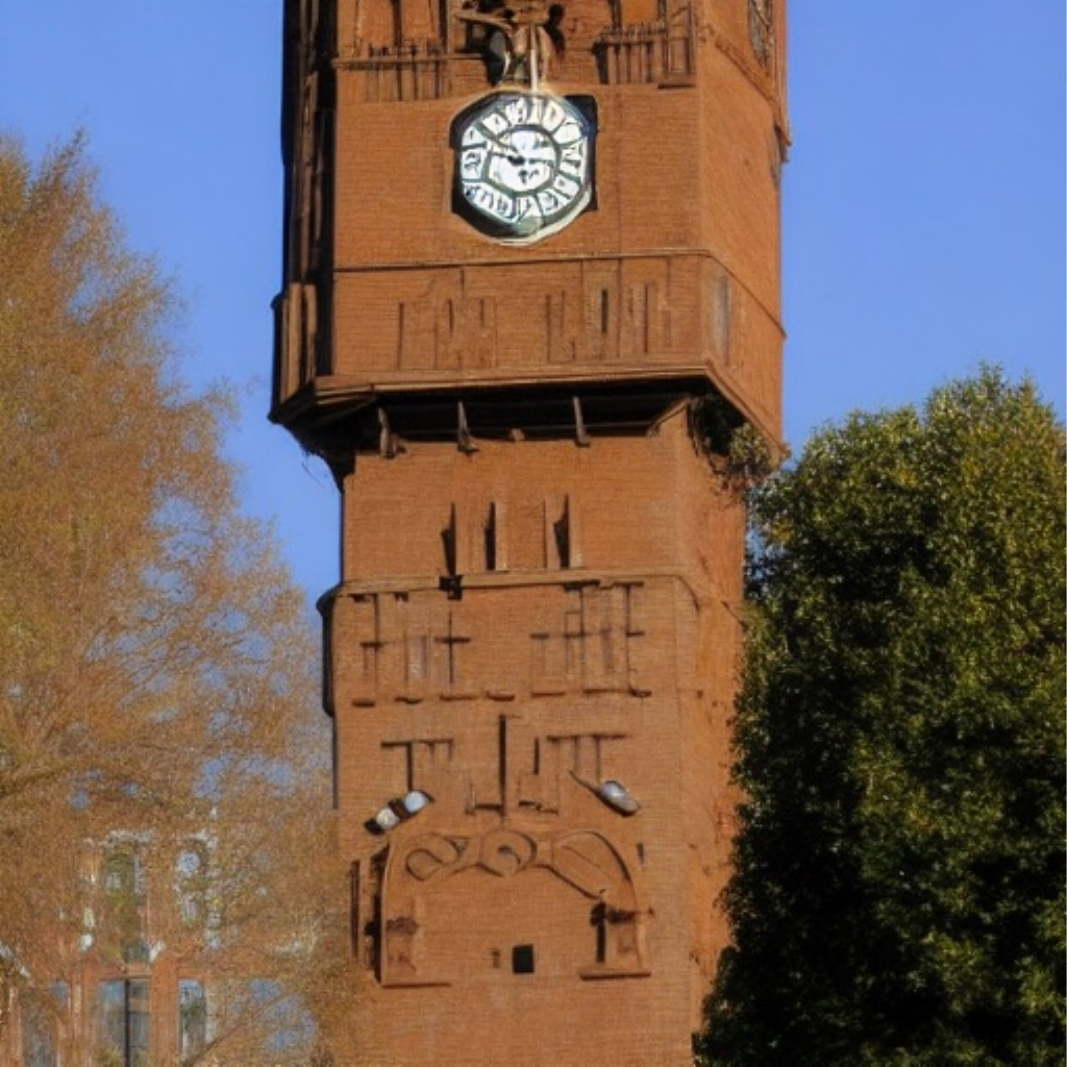}\hfill
    \includegraphics[width=0.24\linewidth]{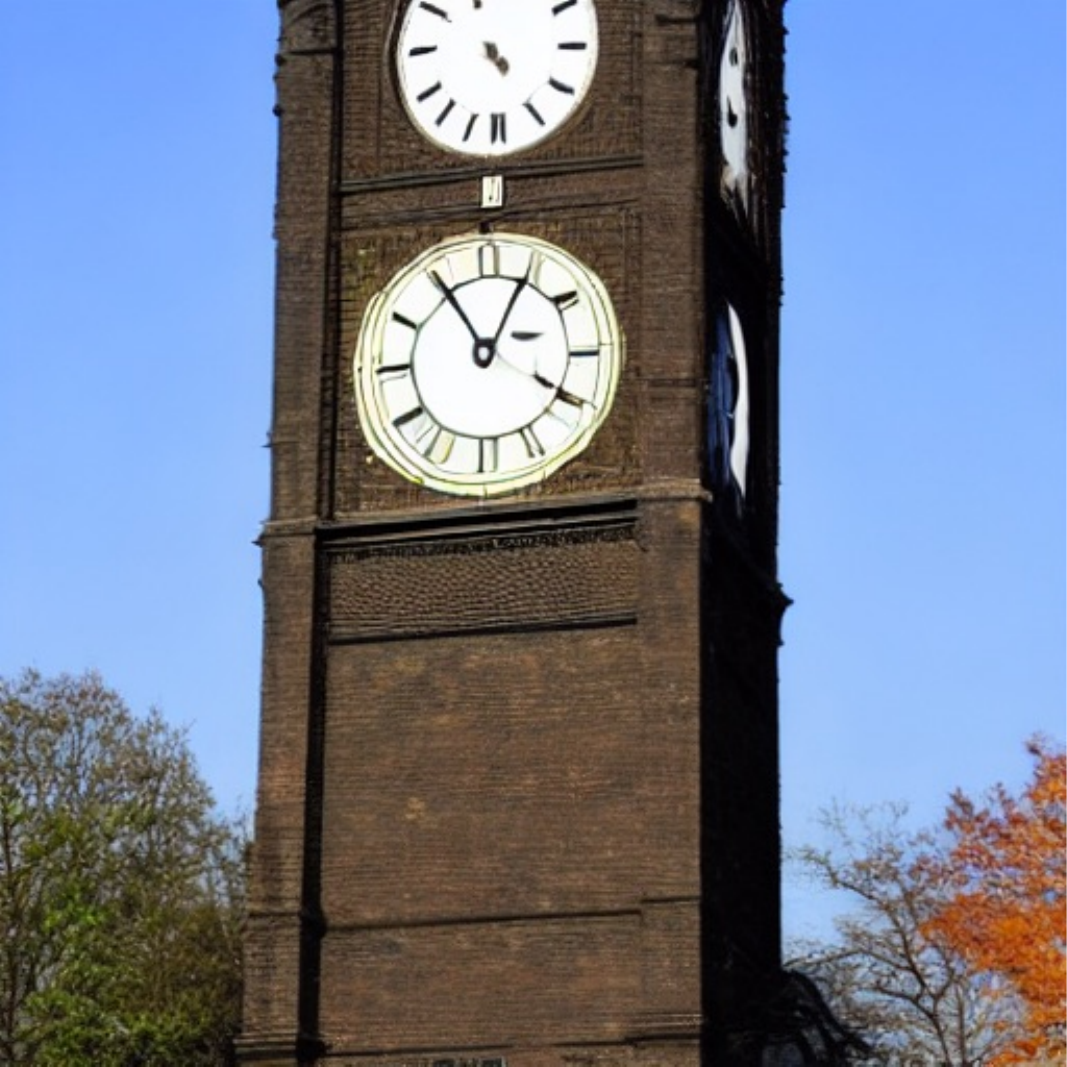}
    
    \vspace{0.5mm}
    \caption*{\small ``A large clock tower with a clock on it's face."}
    \vspace{1mm}

    \includegraphics[width=0.24\linewidth]{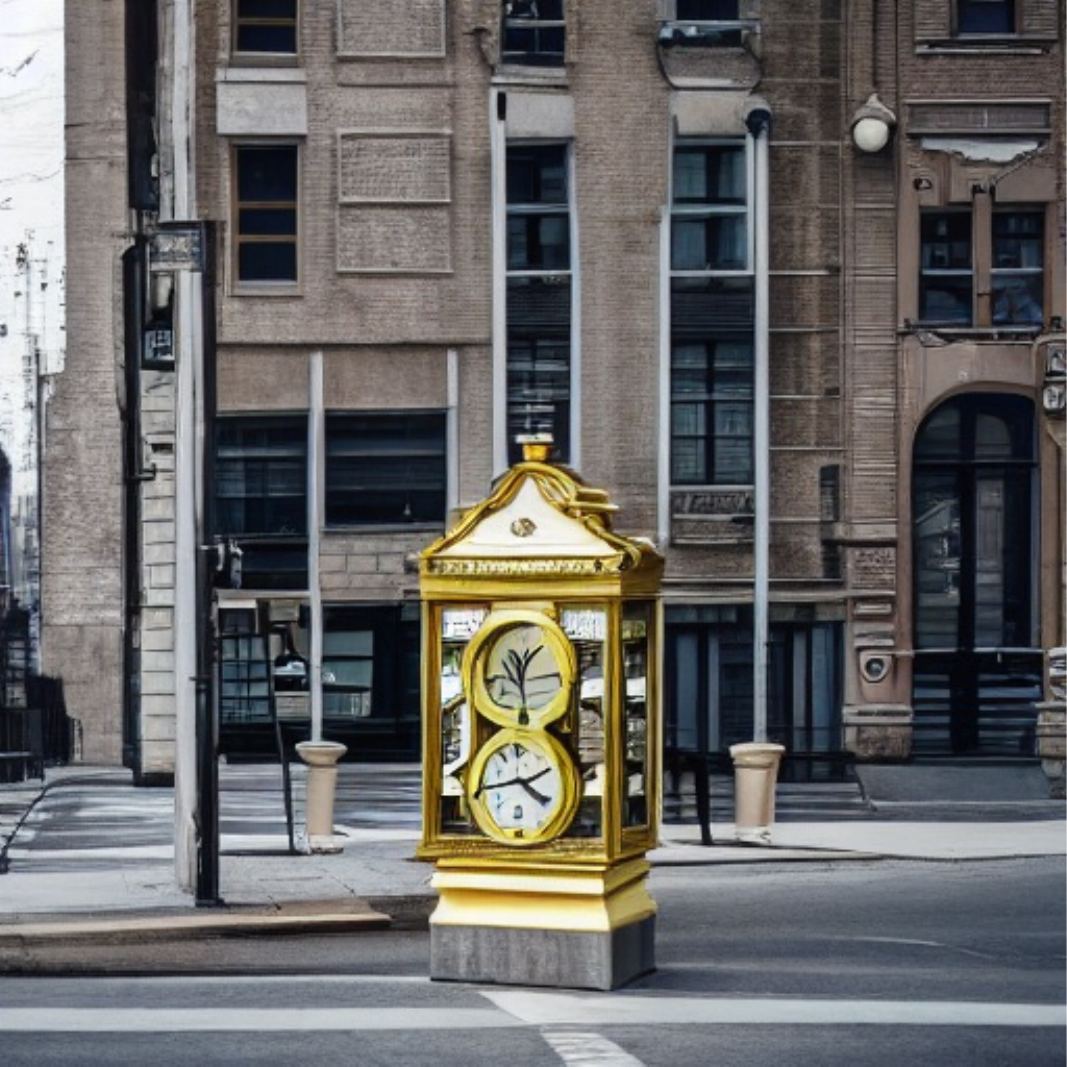}\hfill
    \includegraphics[width=0.24\linewidth]{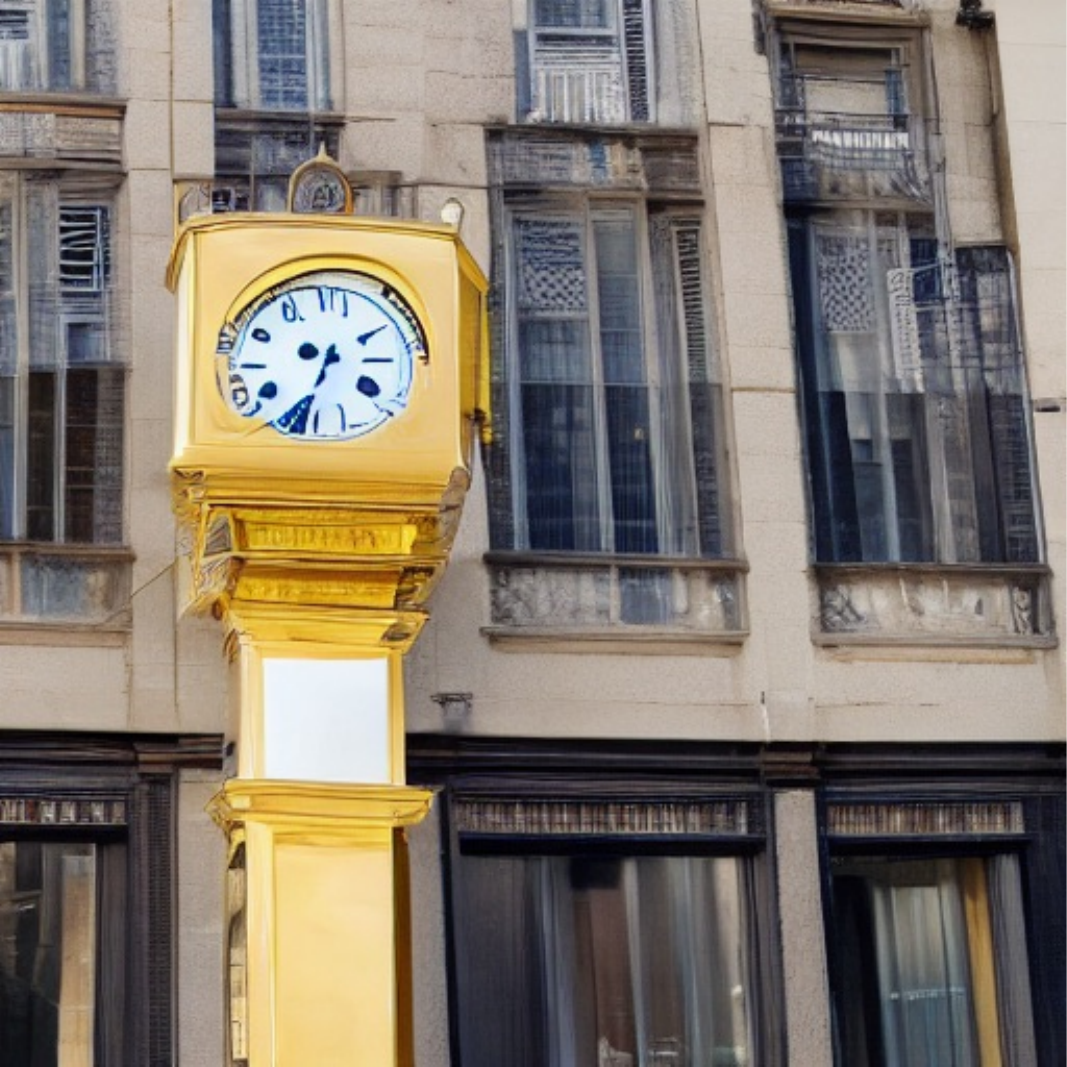}\hfill
    \includegraphics[width=0.24\linewidth]{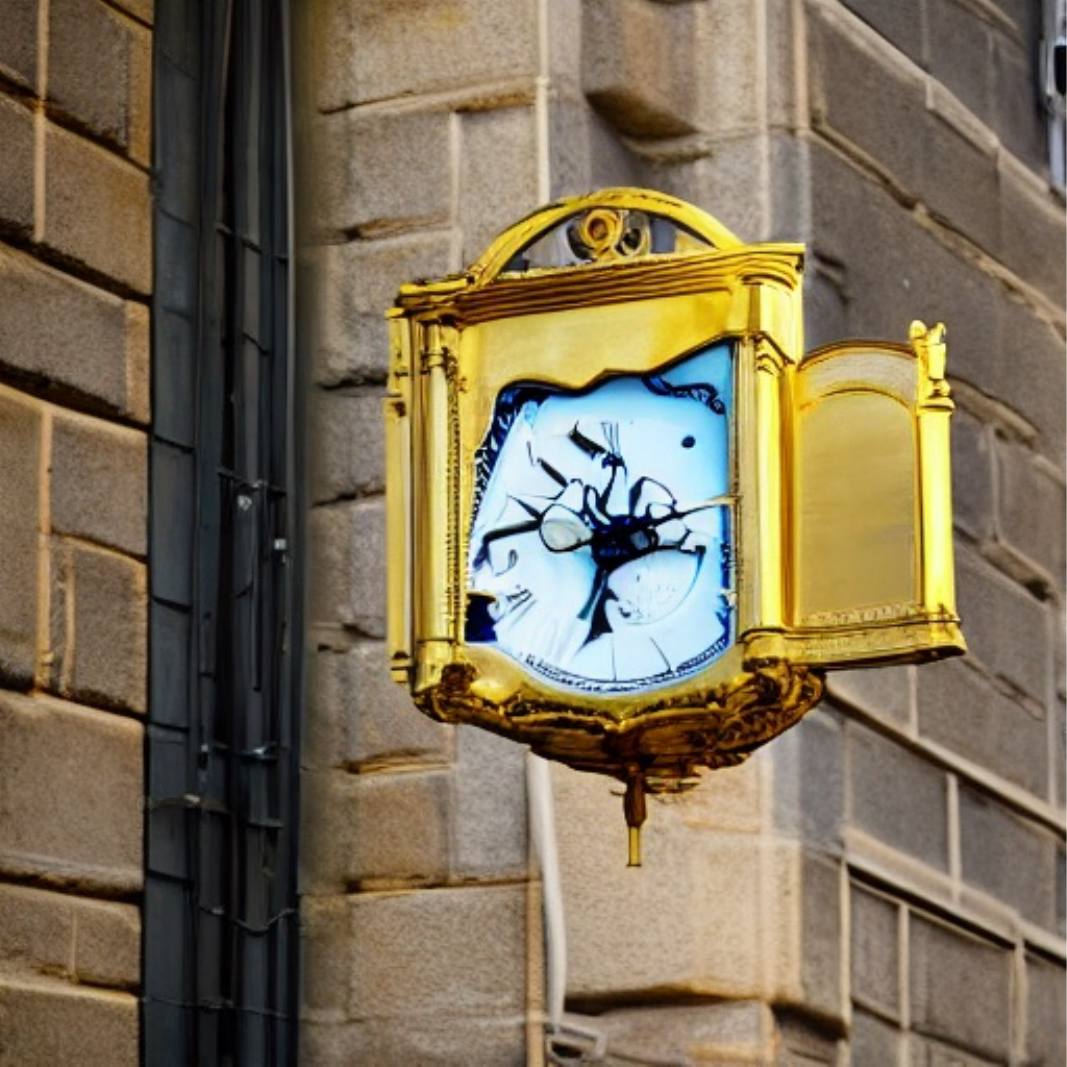}\hfill
    \includegraphics[width=0.24\linewidth]{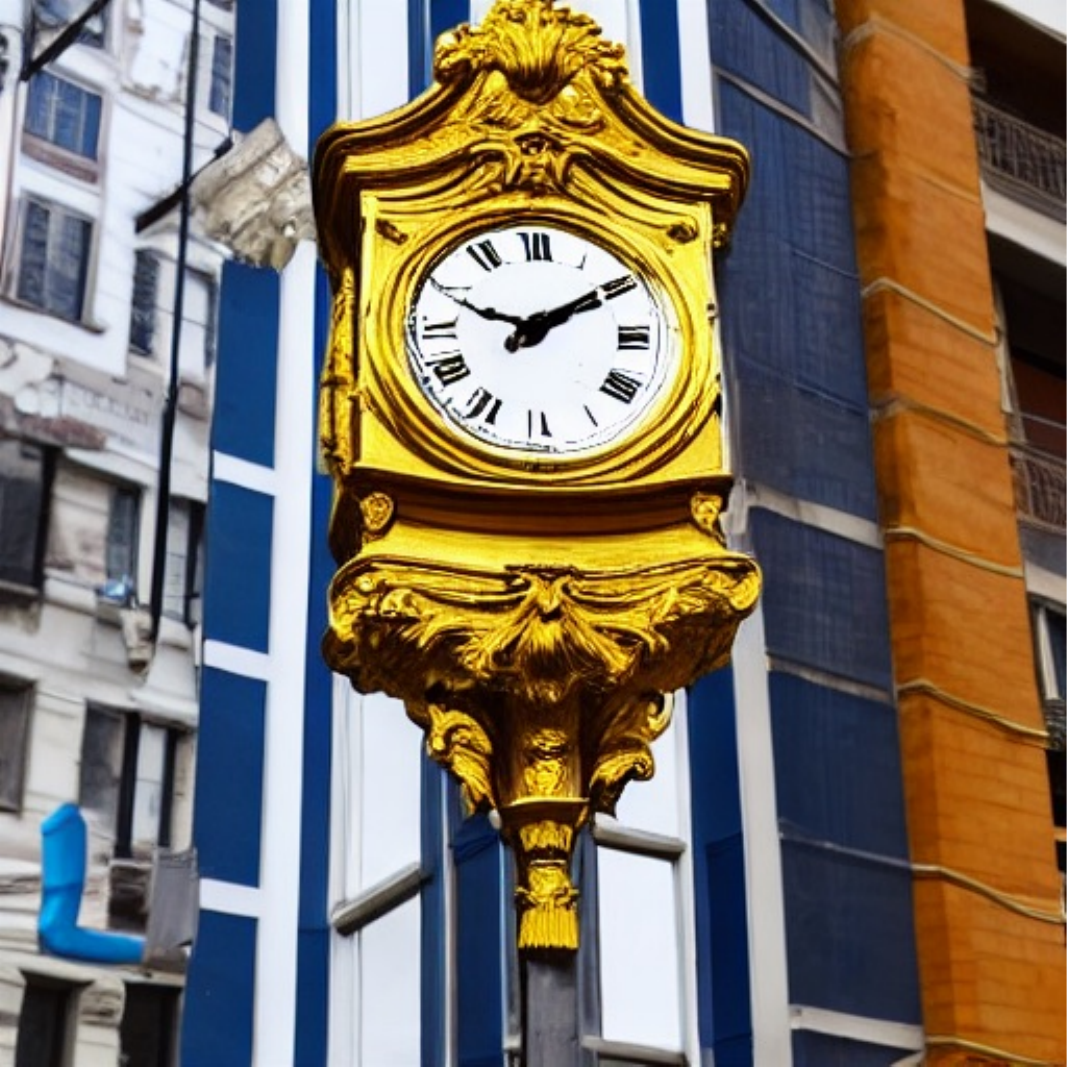}
    
    \vspace{0.5mm}
    \caption*{\small ``A gold and white clock on street next to a building."}
    
    \caption{Steps = 7}
    \label{subfig:nfe7}
  \end{subfigure}
  
  \caption{Side-by-side comparison of selected images generated with Stable Diffusion and iPNDM solver in the high NFE regime (Steps $\in \{6, 7\}$).
    Methods (from left to right): DMN, GITS, LD3, and D2PO.}
  \label{fig:sd_ipndm_nfehigh}
\end{figure*}

\begin{figure*}[t] 
  \centering
  
  \captionsetup[subfigure]{font=small, labelfont=small}
  
  \begin{subfigure}{\linewidth}
    \centering
    
    \begin{minipage}{0.24\linewidth}\centering\small DMN\end{minipage} \hfill
    \begin{minipage}{0.24\linewidth}\centering\small GITS\end{minipage} \hfill
    \begin{minipage}{0.24\linewidth}\centering\small LD3\end{minipage} \hfill
    \begin{minipage}{0.24\linewidth}\centering\small D2PO\end{minipage}
    \vspace{1mm} 
    
    \includegraphics[width=0.24\linewidth]{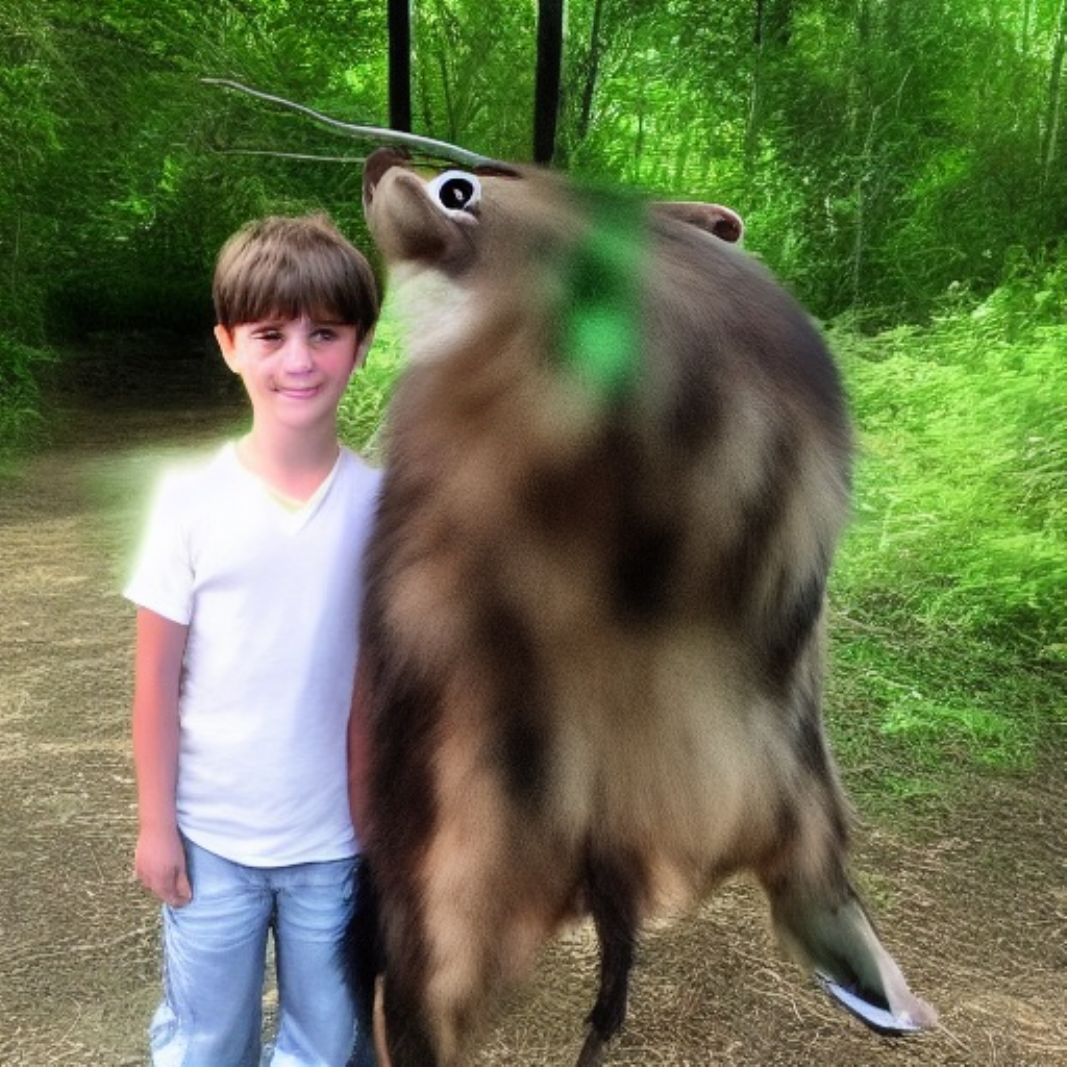}\hfill
    \includegraphics[width=0.24\linewidth]{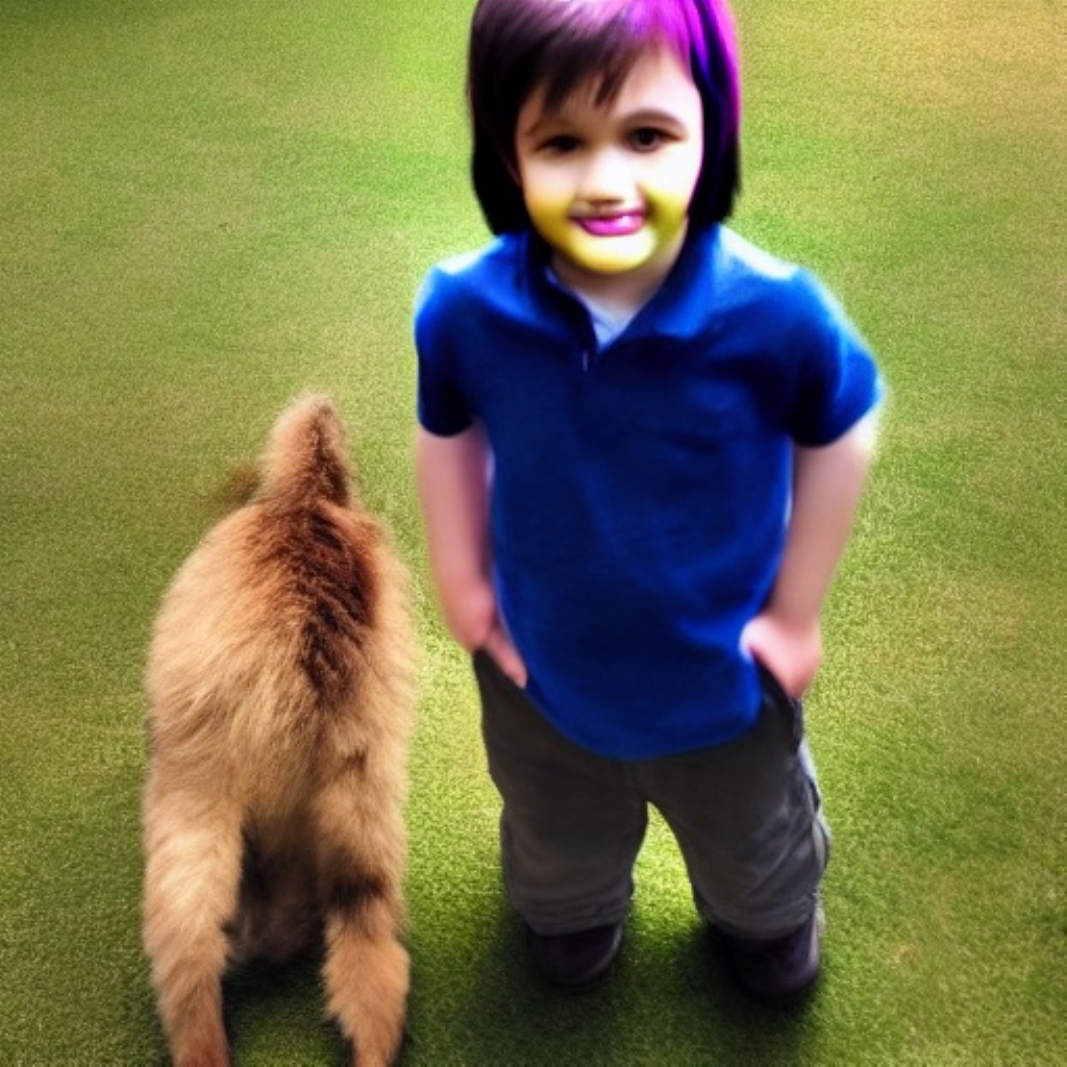}\hfill
    \includegraphics[width=0.24\linewidth]{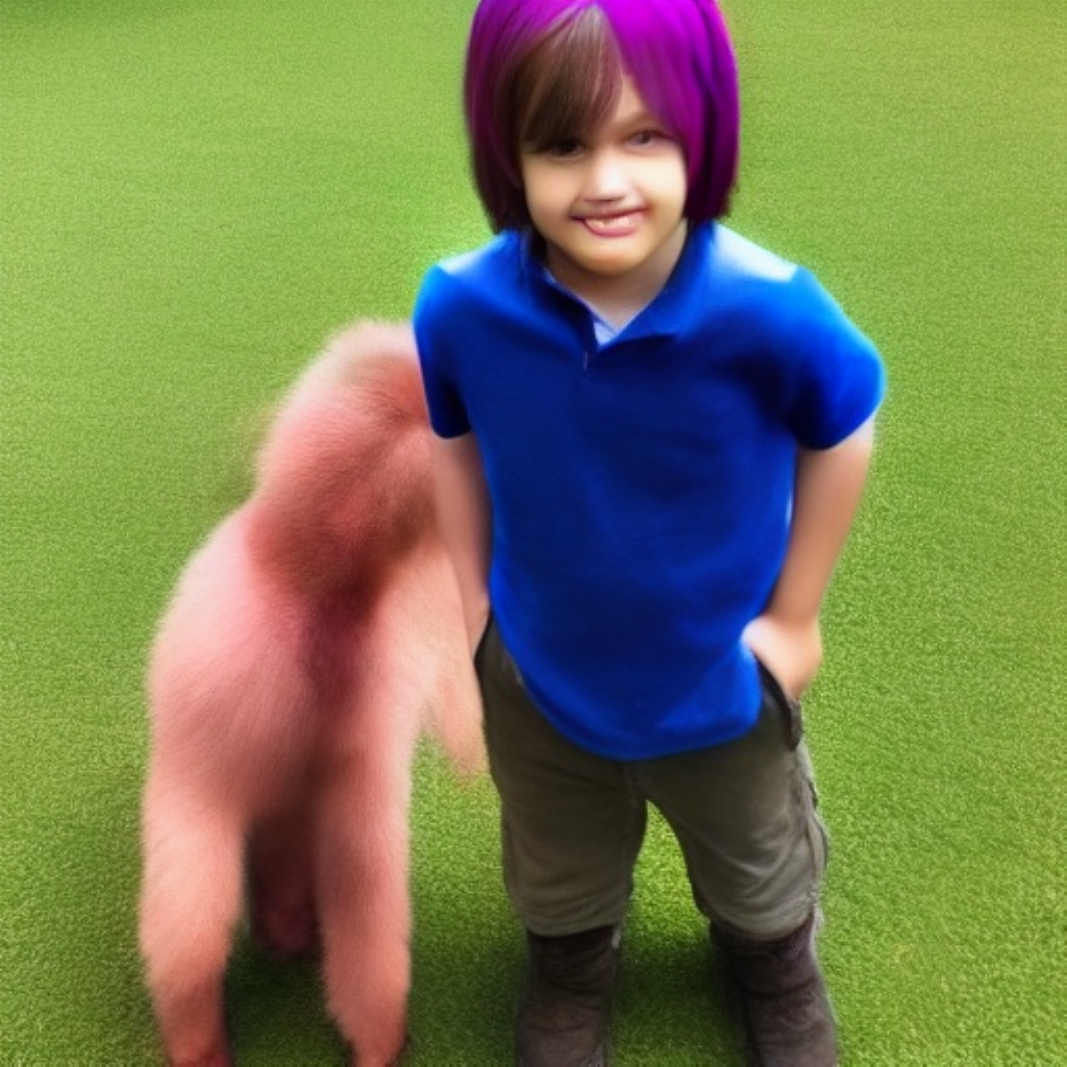}\hfill
    \includegraphics[width=0.24\linewidth]{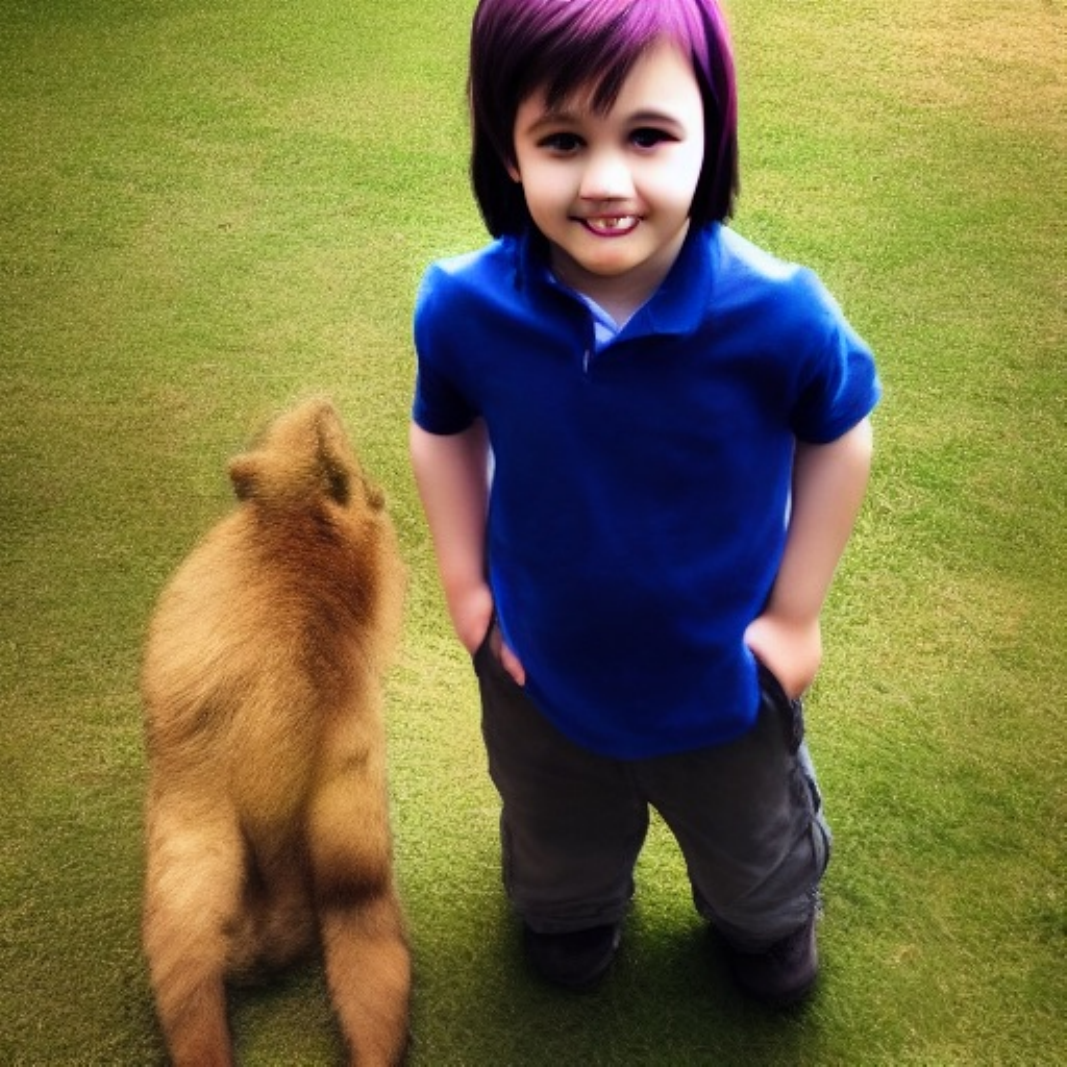}
    
    \vspace{0.5mm} 
    \caption*{ ``A boy that is standing next to an animal."}
    \vspace{1mm} %

    \includegraphics[width=0.24\linewidth]{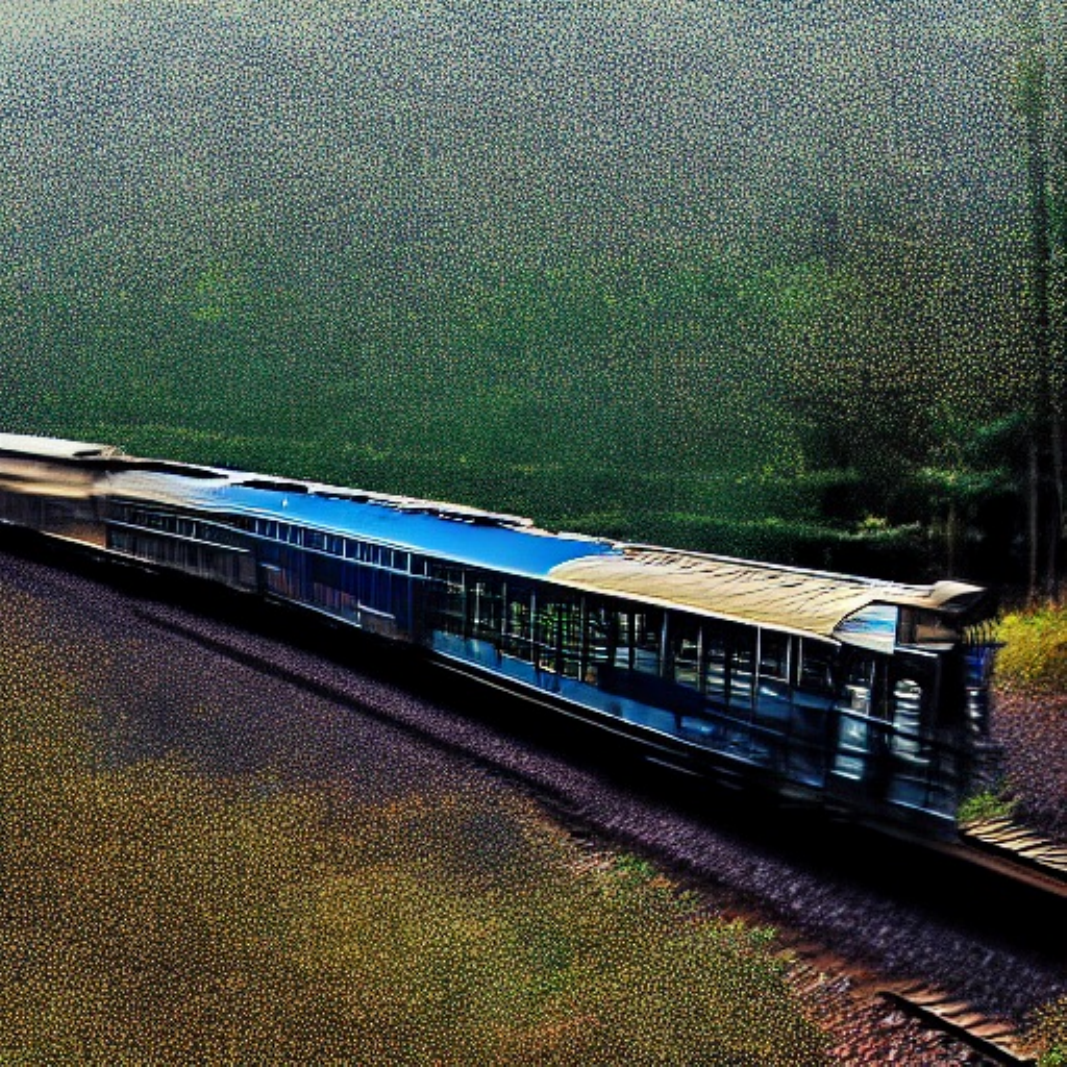}\hfill
    \includegraphics[width=0.24\linewidth]{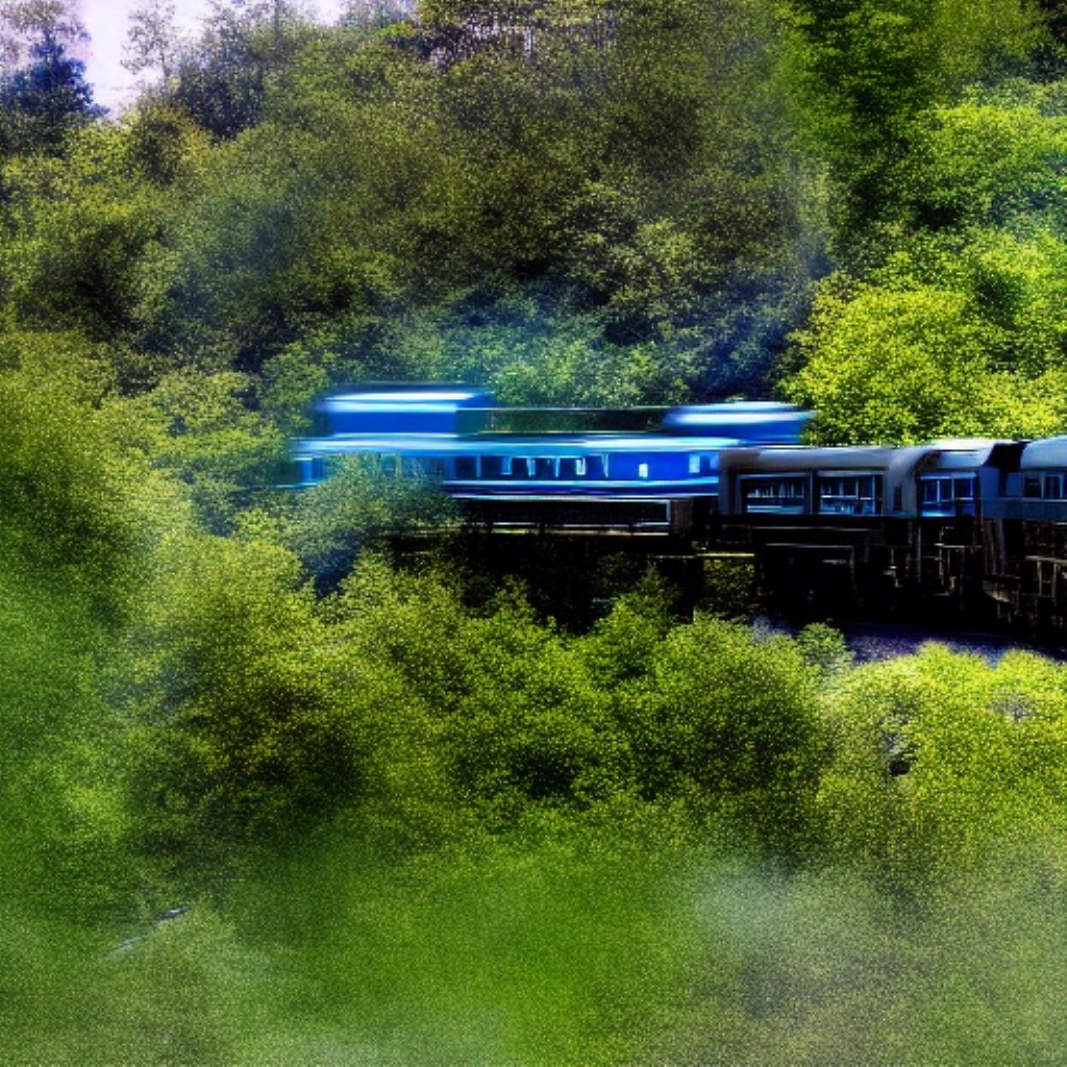}\hfill
    \includegraphics[width=0.24\linewidth]{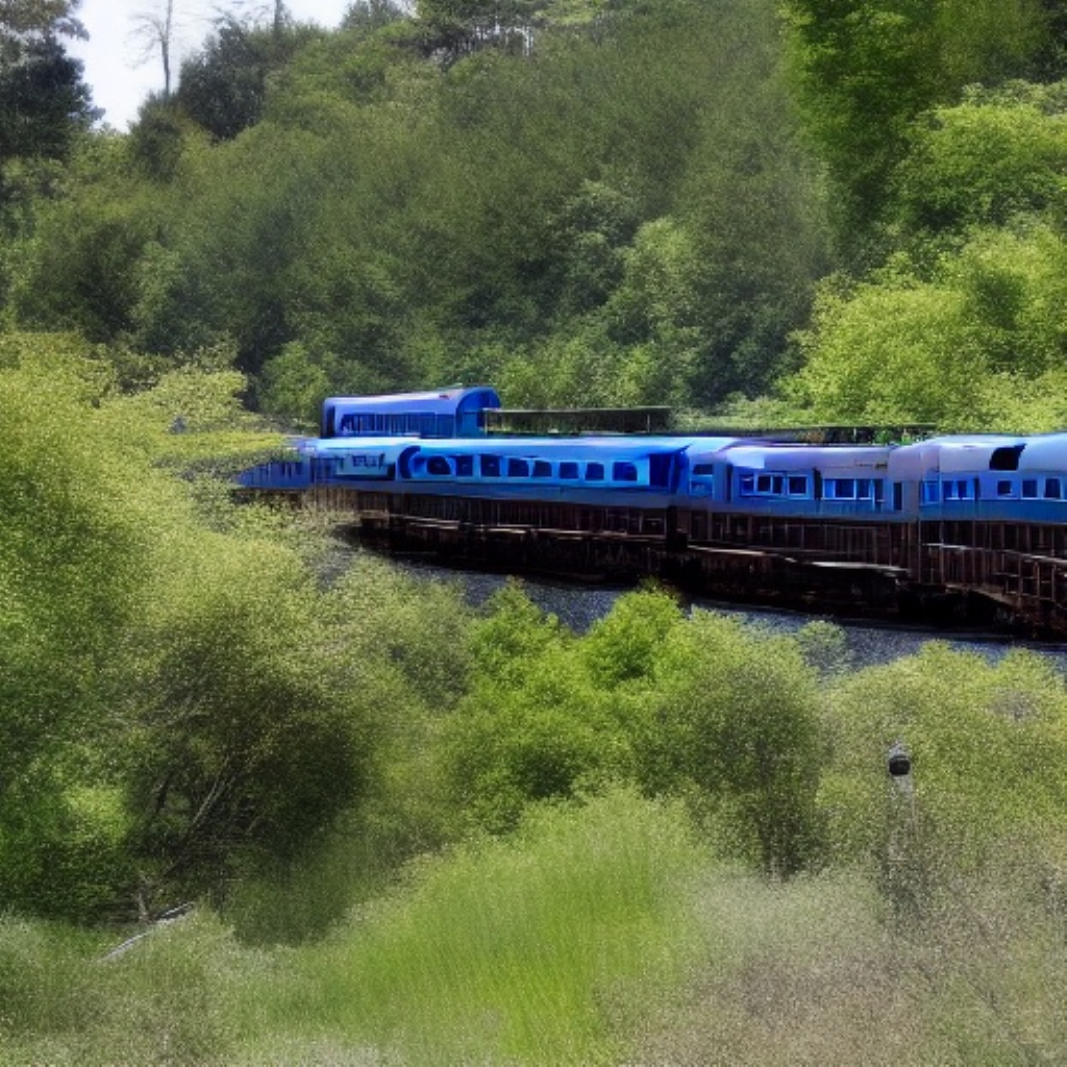}\hfill
    \includegraphics[width=0.24\linewidth]{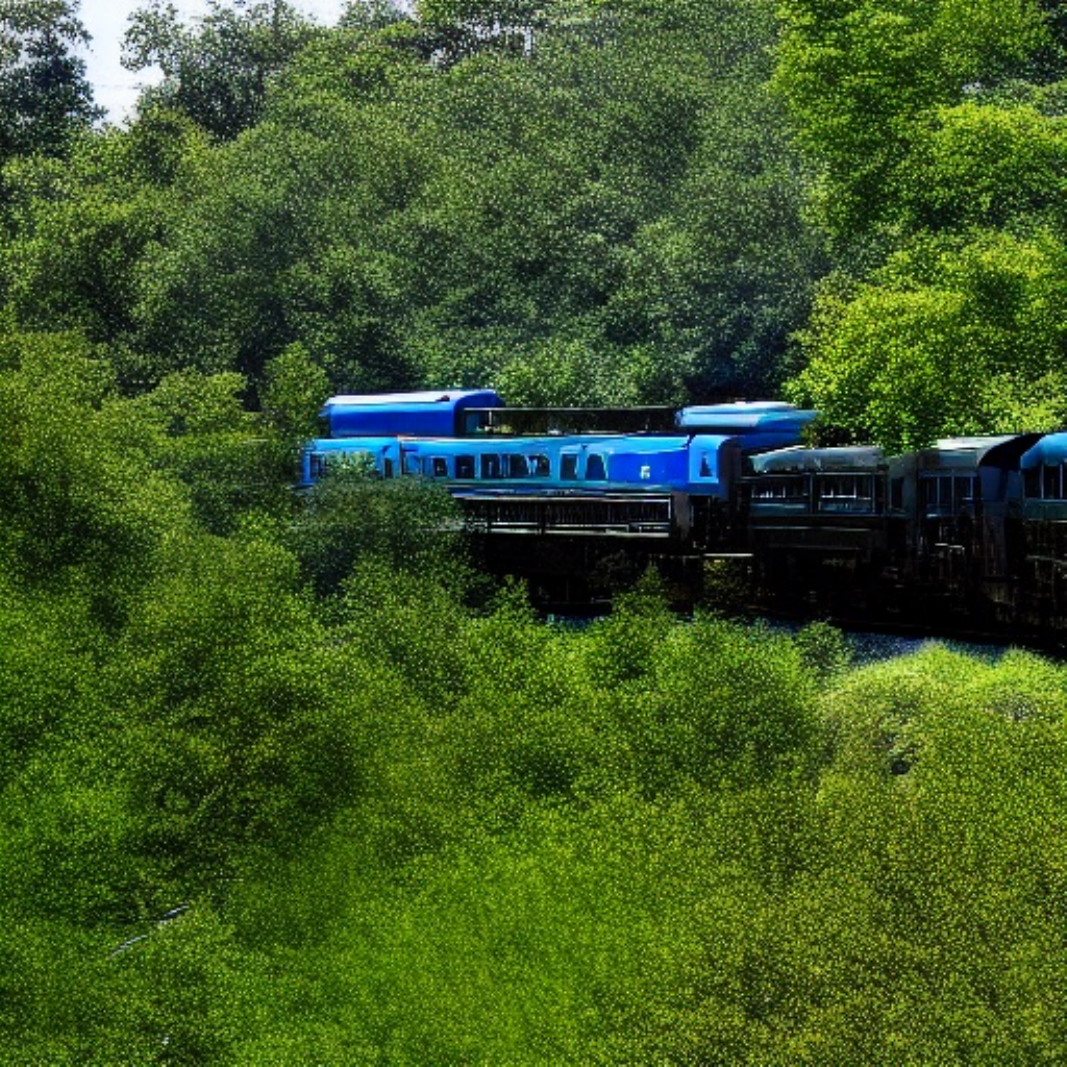}
    
    \vspace{0.5mm}
    \caption*{ ``The blue train is passing through a wooded area."}
    
    \caption{ Steps = 4} 
    \label{subfig:nfe4}
  \end{subfigure}
  
  \vspace{1em} %

  \begin{subfigure}{\linewidth}
    \centering
    
    \includegraphics[width=0.24\linewidth]{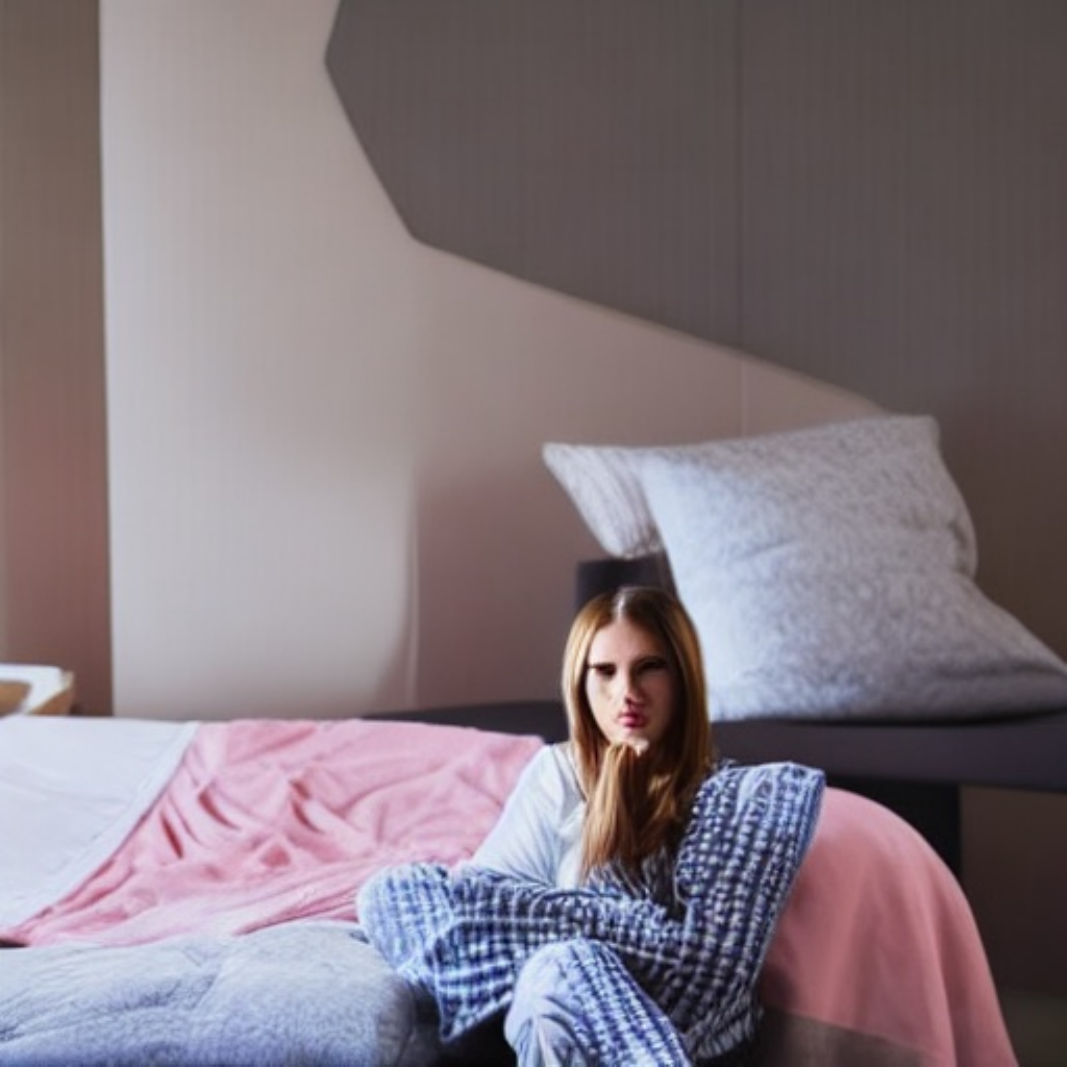}\hfill
    \includegraphics[width=0.24\linewidth]{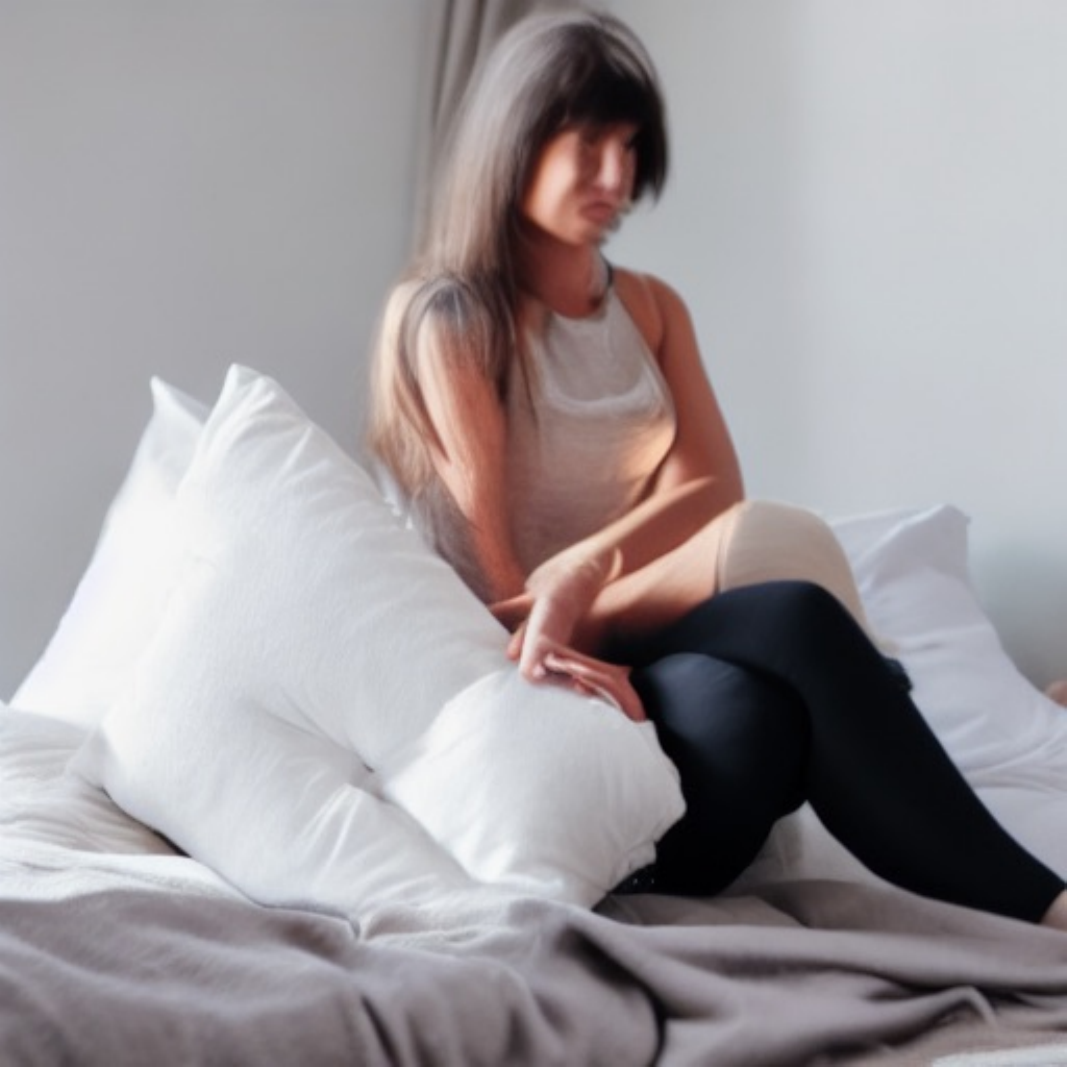}\hfill
    \includegraphics[width=0.24\linewidth]{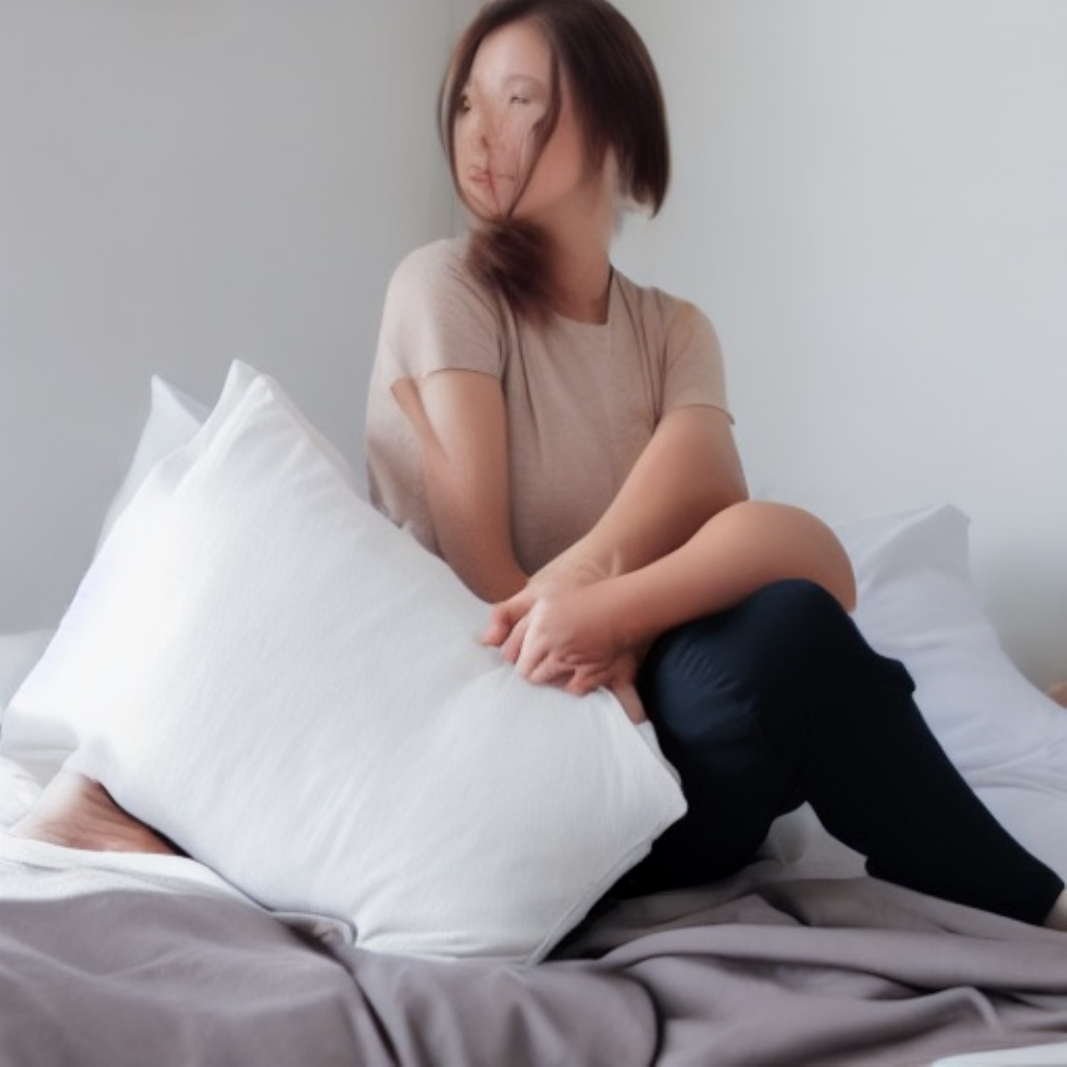}\hfill
    \includegraphics[width=0.24\linewidth]{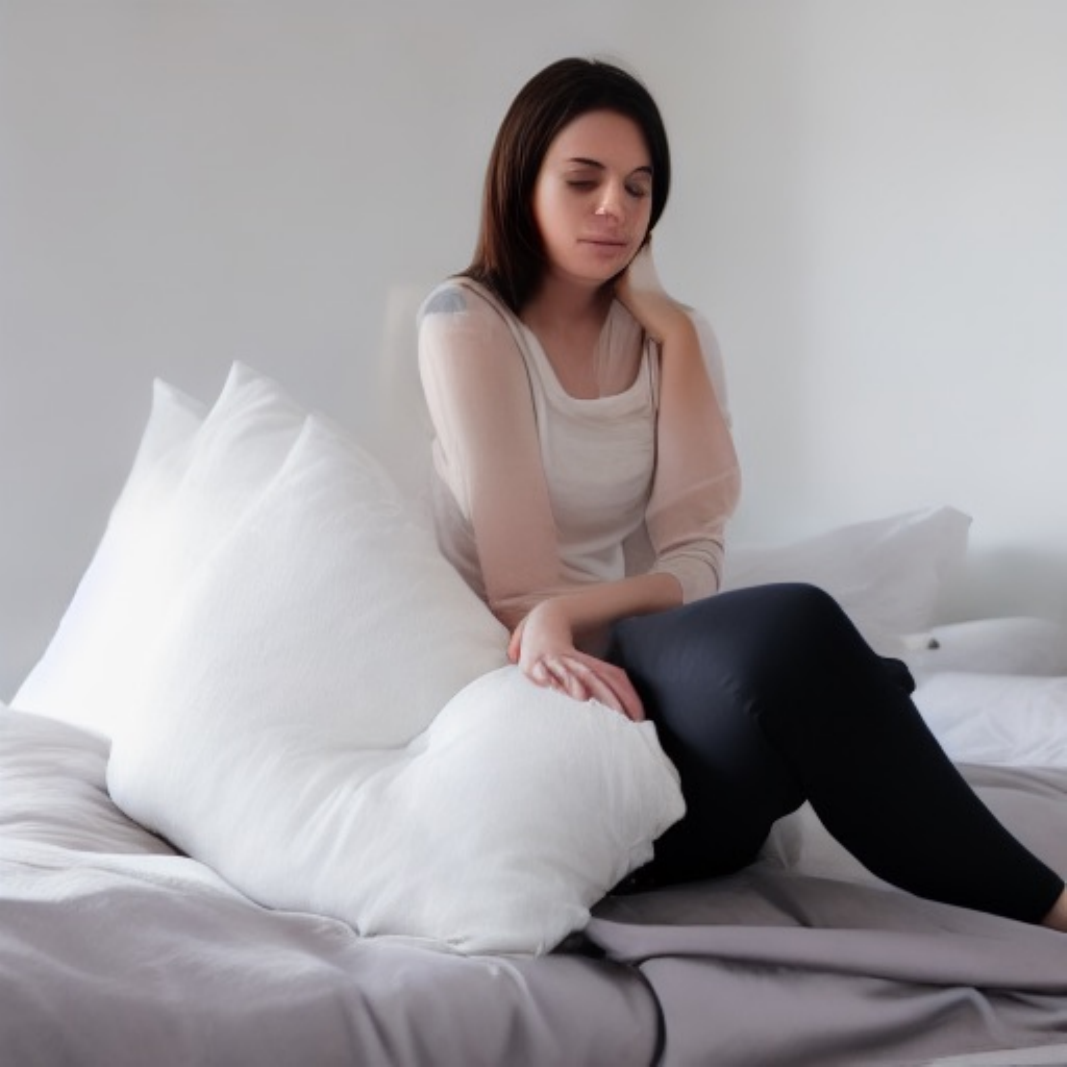}
    
    \vspace{0.5mm}
    \caption*{``A woman sits on a bed with pillows."}
    \vspace{1mm}

    \includegraphics[width=0.24\linewidth]{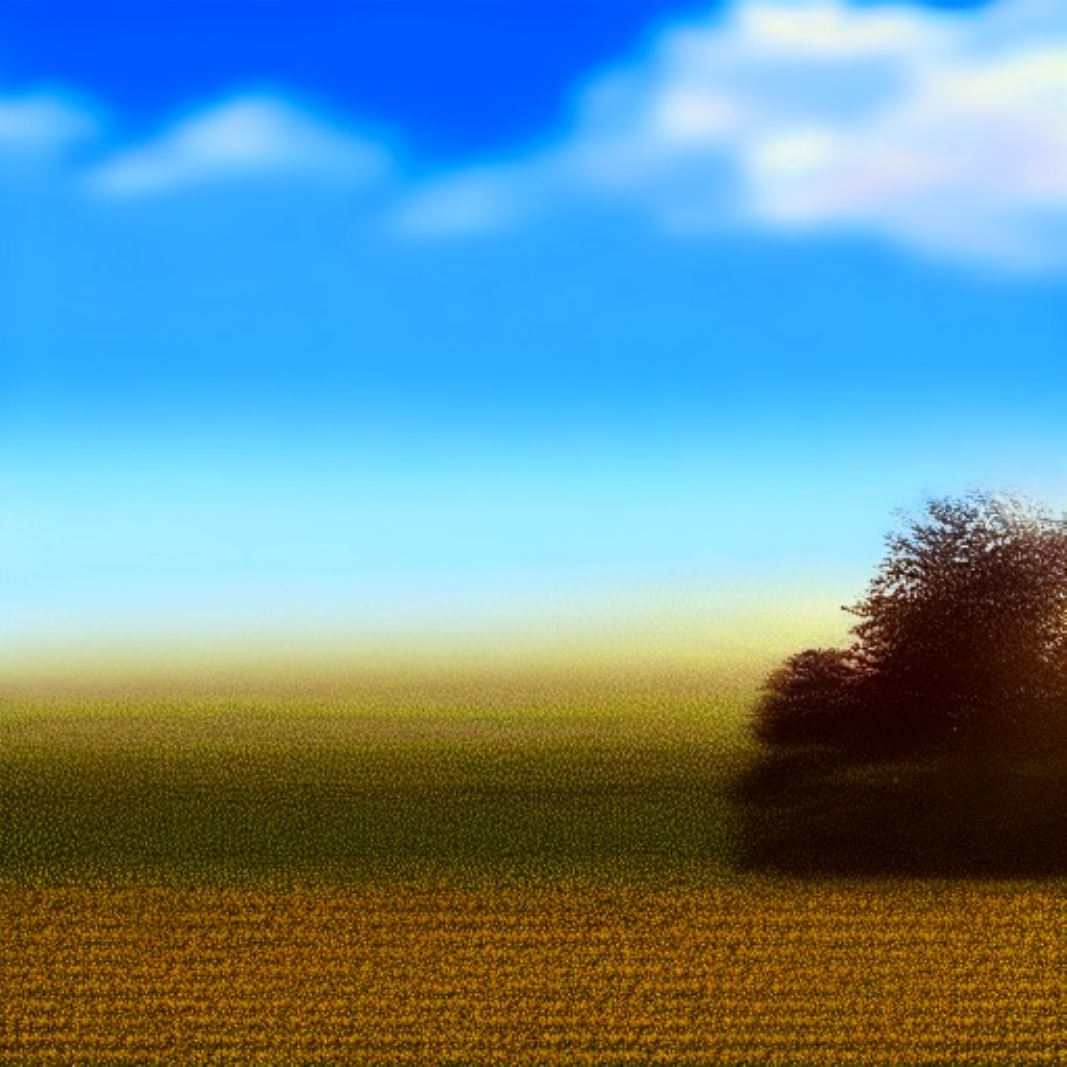}\hfill
    \includegraphics[width=0.24\linewidth]{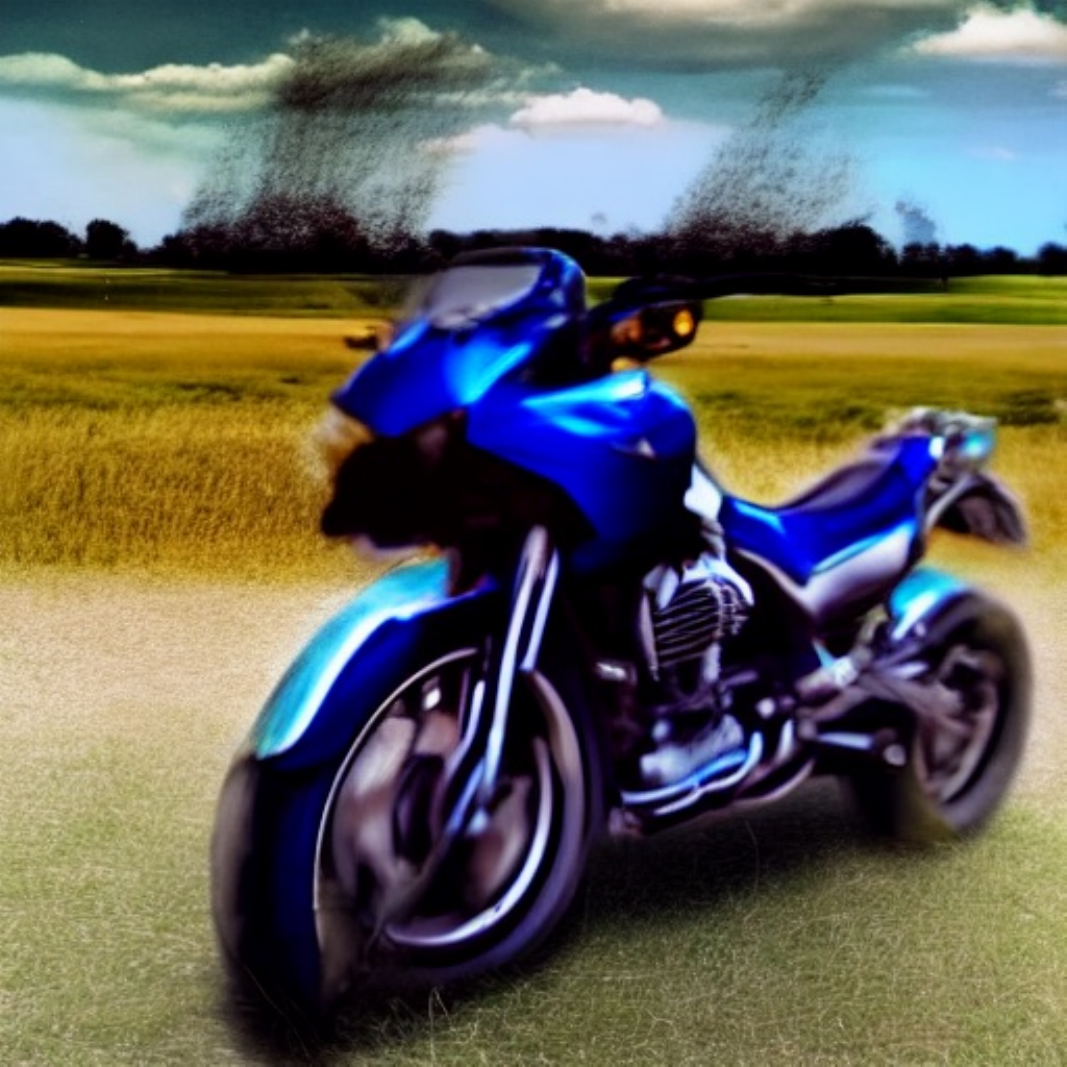}\hfill
    \includegraphics[width=0.24\linewidth]{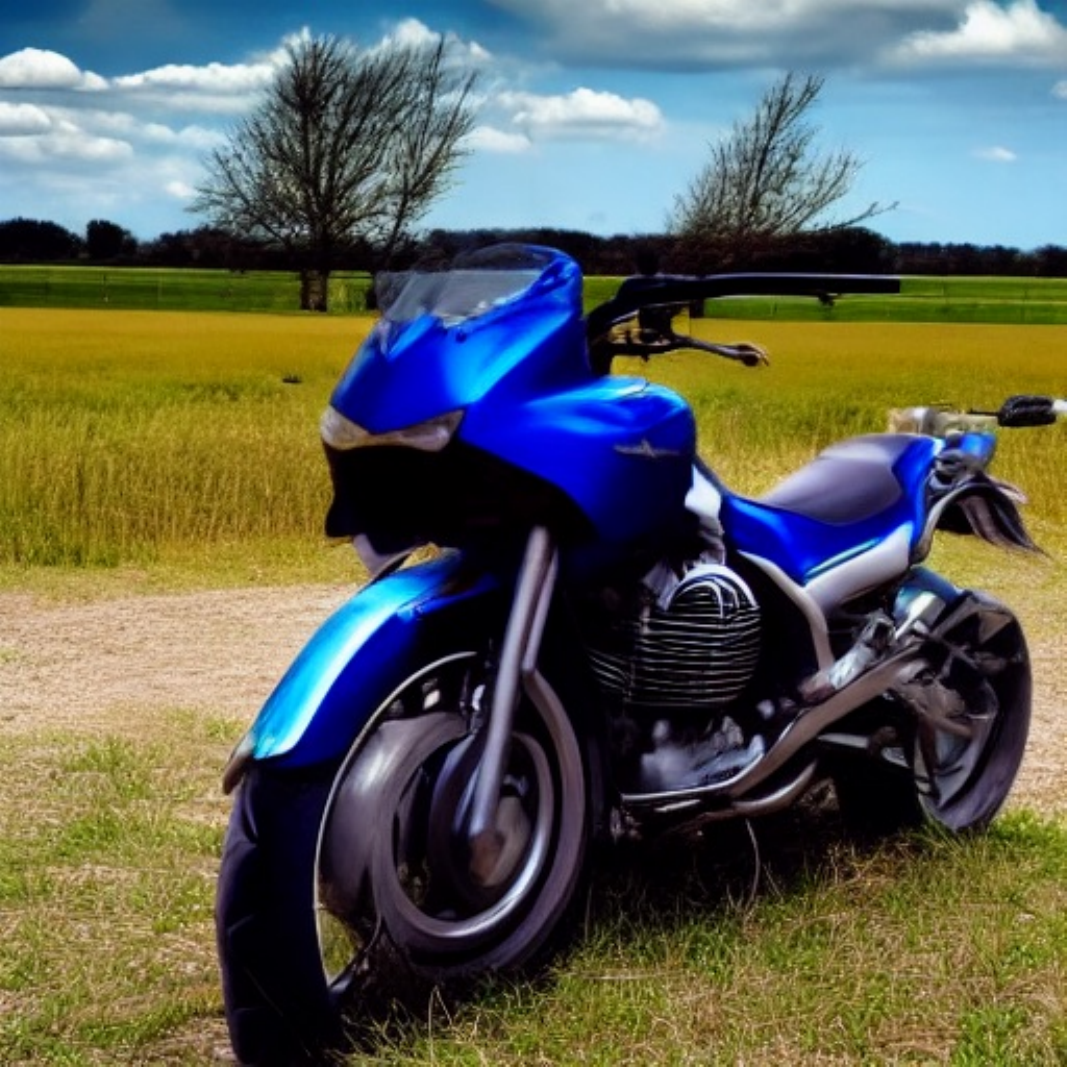}\hfill
    \includegraphics[width=0.24\linewidth]{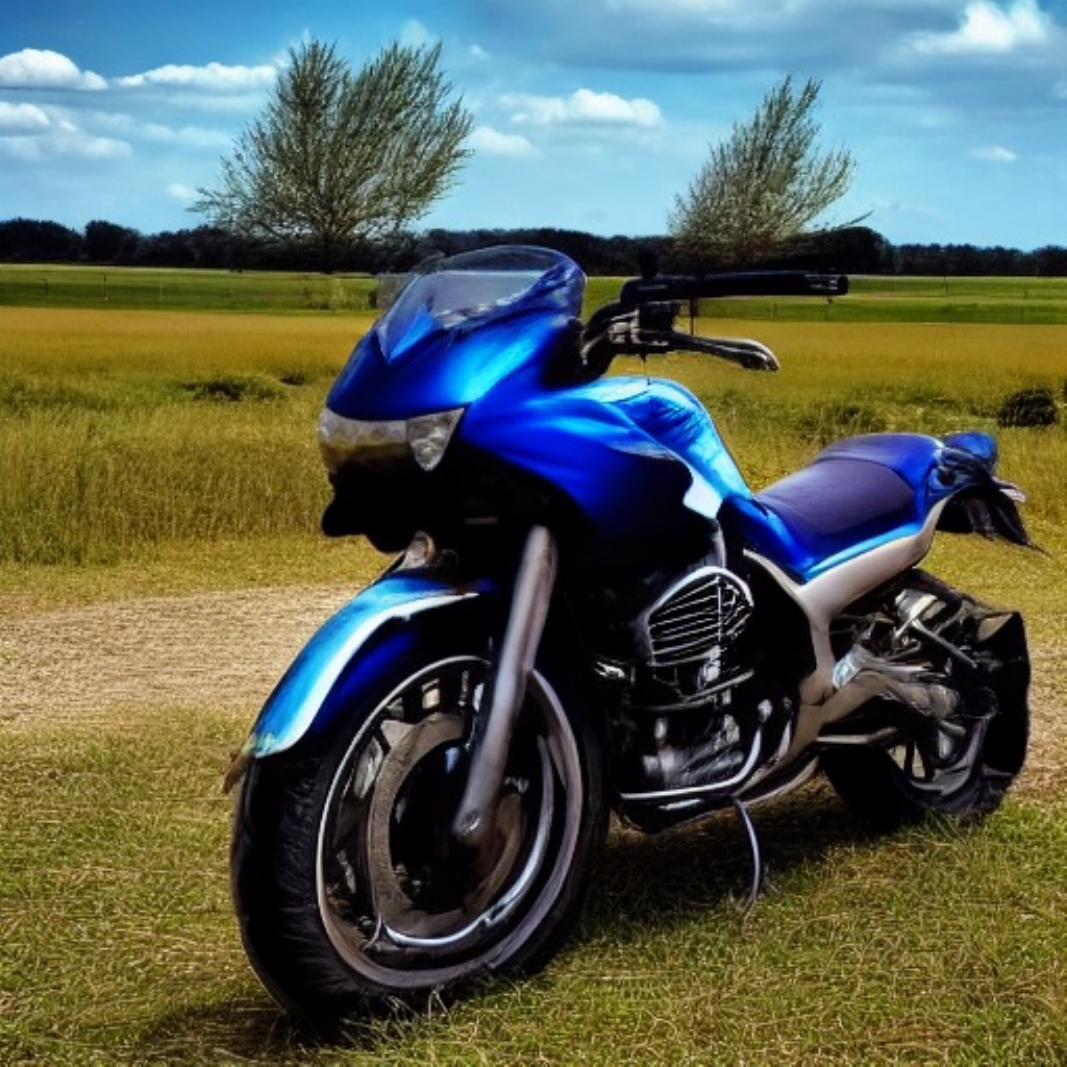}
    
    \vspace{0.5mm}
    \caption*{``A motorcycle parked in a field with a sky background."}
    
    \caption{Steps = 5}
    \label{subfig:nfe5}
  \end{subfigure}
  \caption{Side-by-side comparison of selected images generated with Stable Diffusion and UniPC solver in the low NFE regime (Steps $\in \{4, 5\}$).
    Methods (from left to right): DMN, GITS, LD3, and D2PO.}
  \label{fig:sd_unipc_nfelow}
\end{figure*}

\begin{figure*}[t] 
  \centering
  
  \captionsetup[subfigure]{font=small, labelfont=small}

  \begin{subfigure}{\linewidth}
    \centering
    
    \begin{minipage}{0.24\linewidth}\centering\small DMN\end{minipage} \hfill
    \begin{minipage}{0.24\linewidth}\centering\small GITS\end{minipage} \hfill
    \begin{minipage}{0.24\linewidth}\centering\small LD3\end{minipage} \hfill
    \begin{minipage}{0.24\linewidth}\centering\small D2PO\end{minipage}
    \vspace{1mm} 
    
    \includegraphics[width=0.24\linewidth]{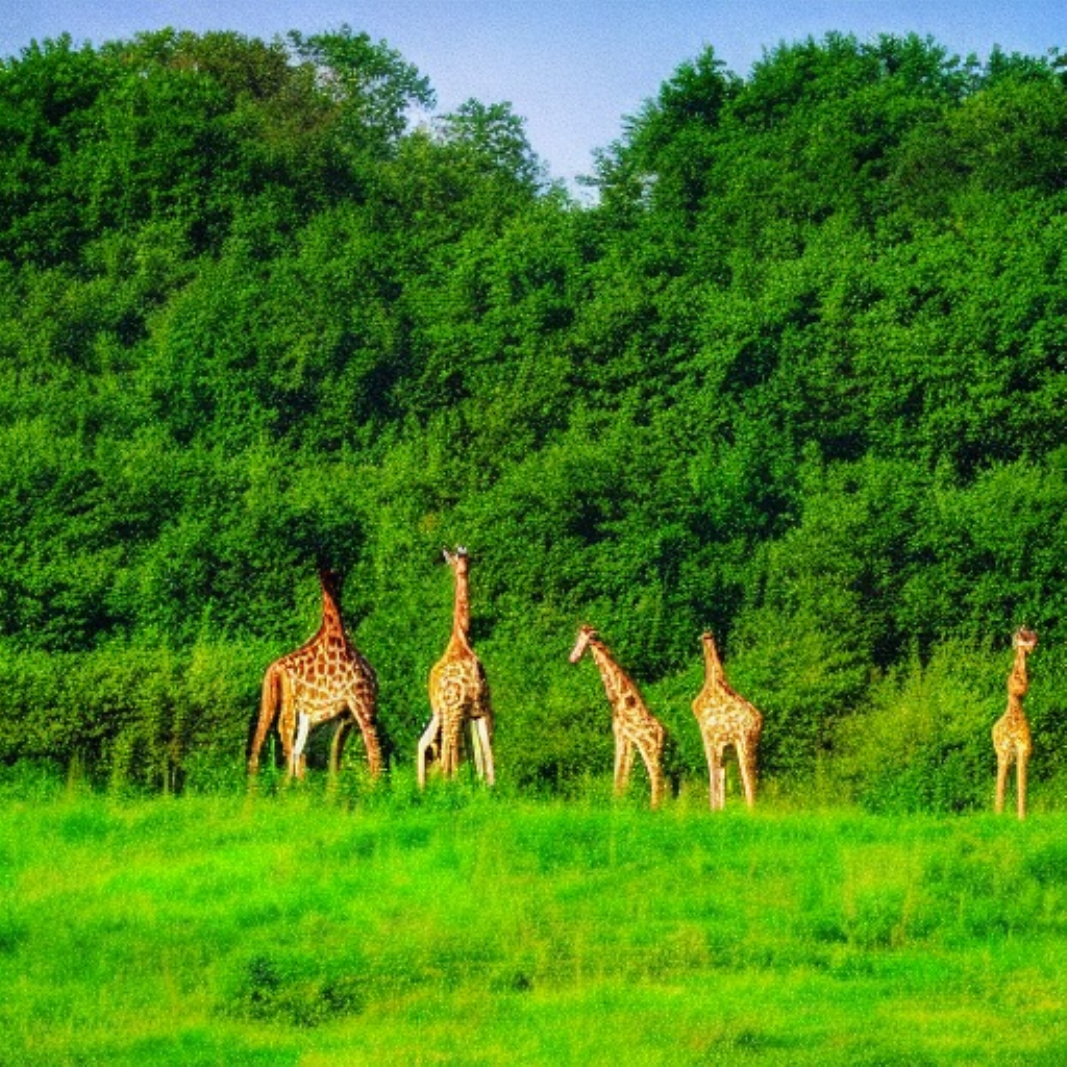}\hfill
    \includegraphics[width=0.24\linewidth]{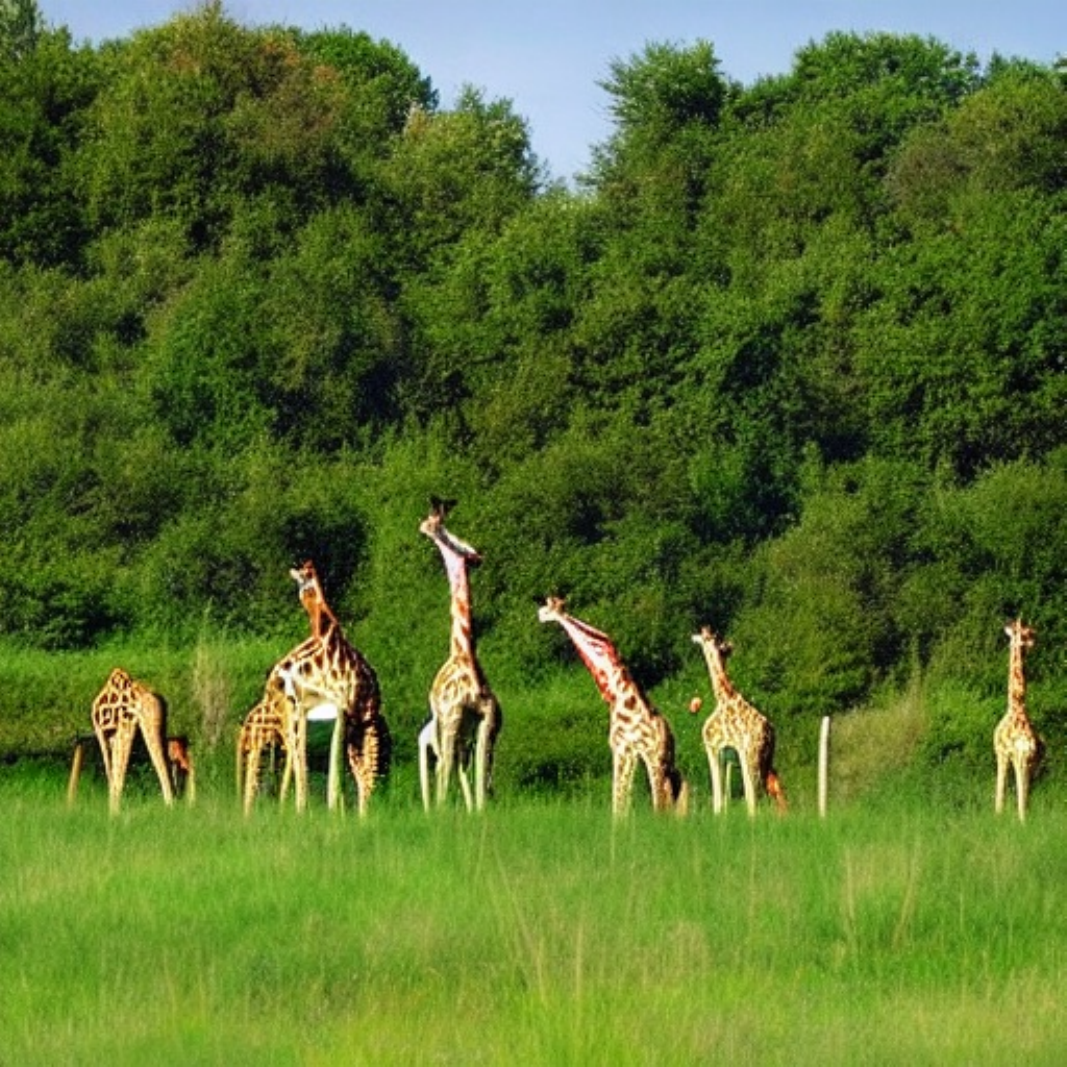}\hfill
    \includegraphics[width=0.24\linewidth]{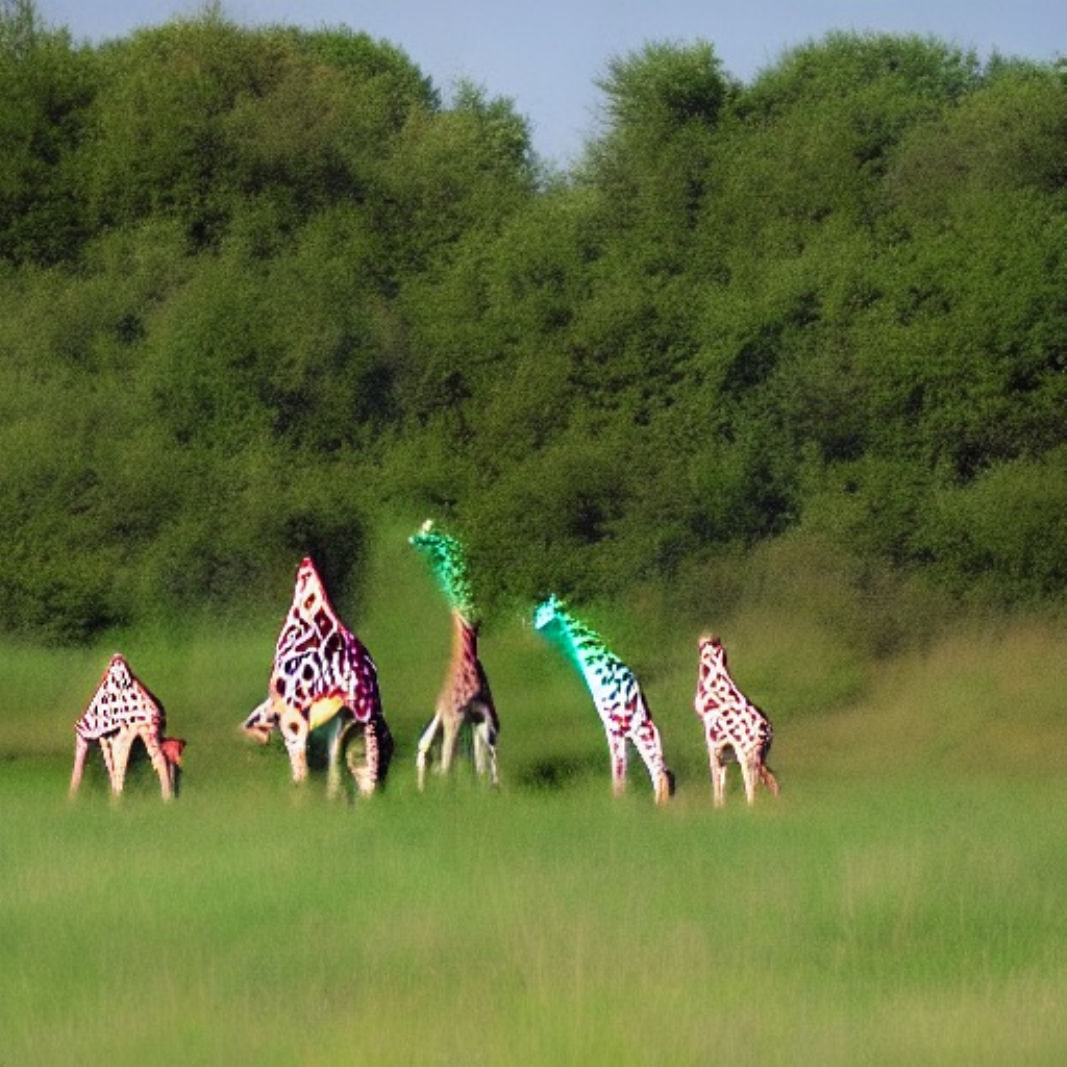}\hfill
    \includegraphics[width=0.24\linewidth]{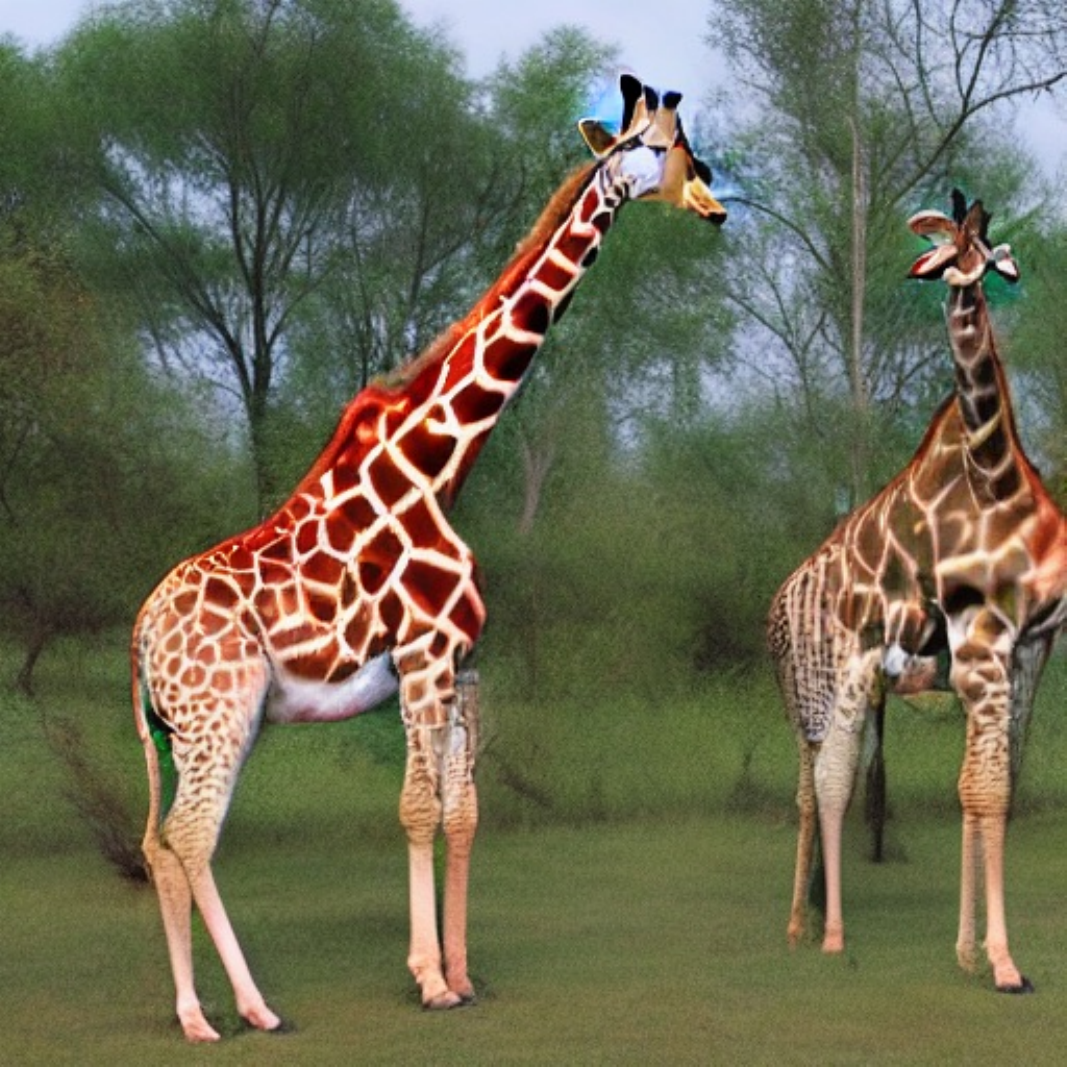}
    
    \vspace{0.5mm} 
    \caption*{\small ``Giraffes in their wood and grass zoo enclosure."}
    \vspace{1mm} %

    \includegraphics[width=0.24\linewidth]{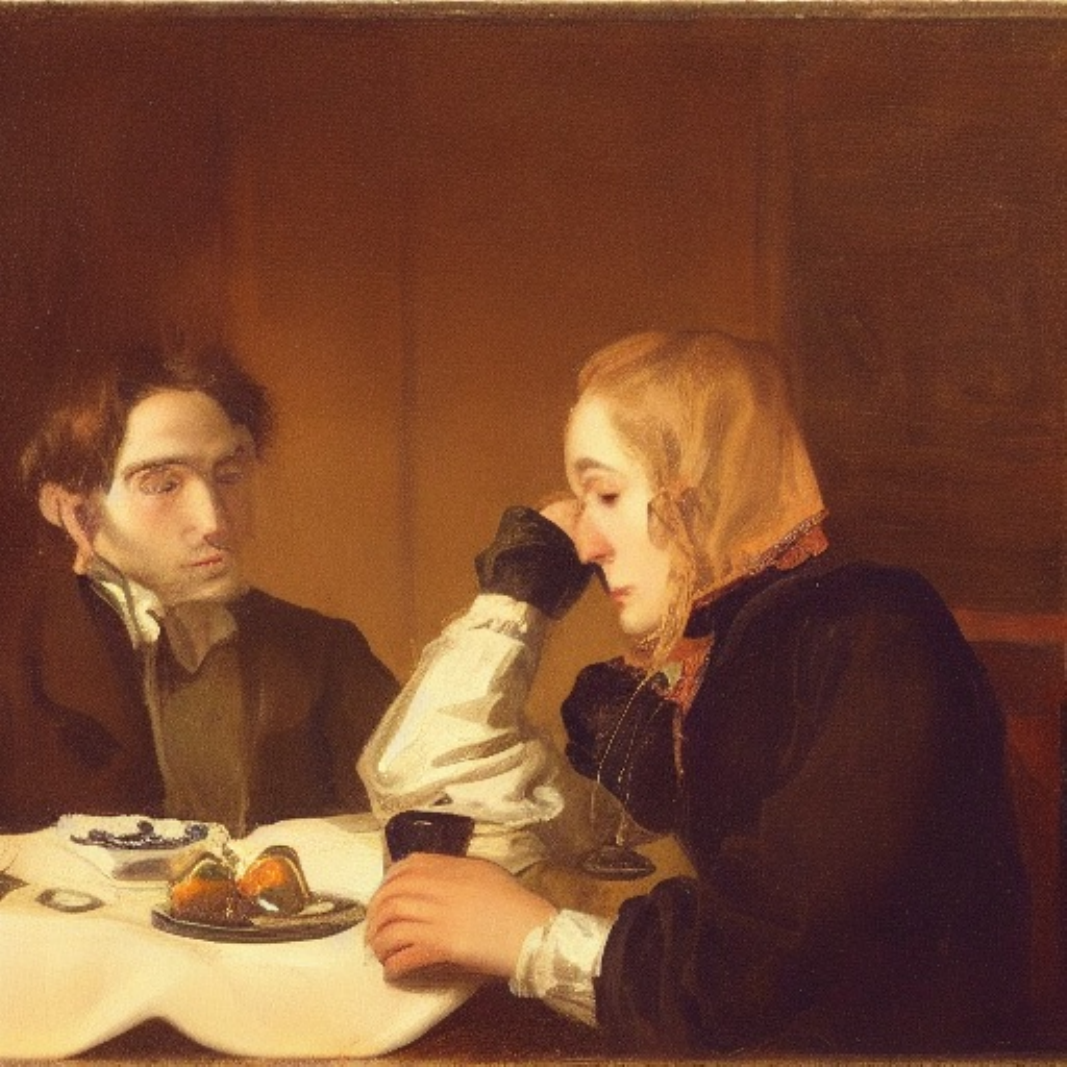}\hfill
    \includegraphics[width=0.24\linewidth]{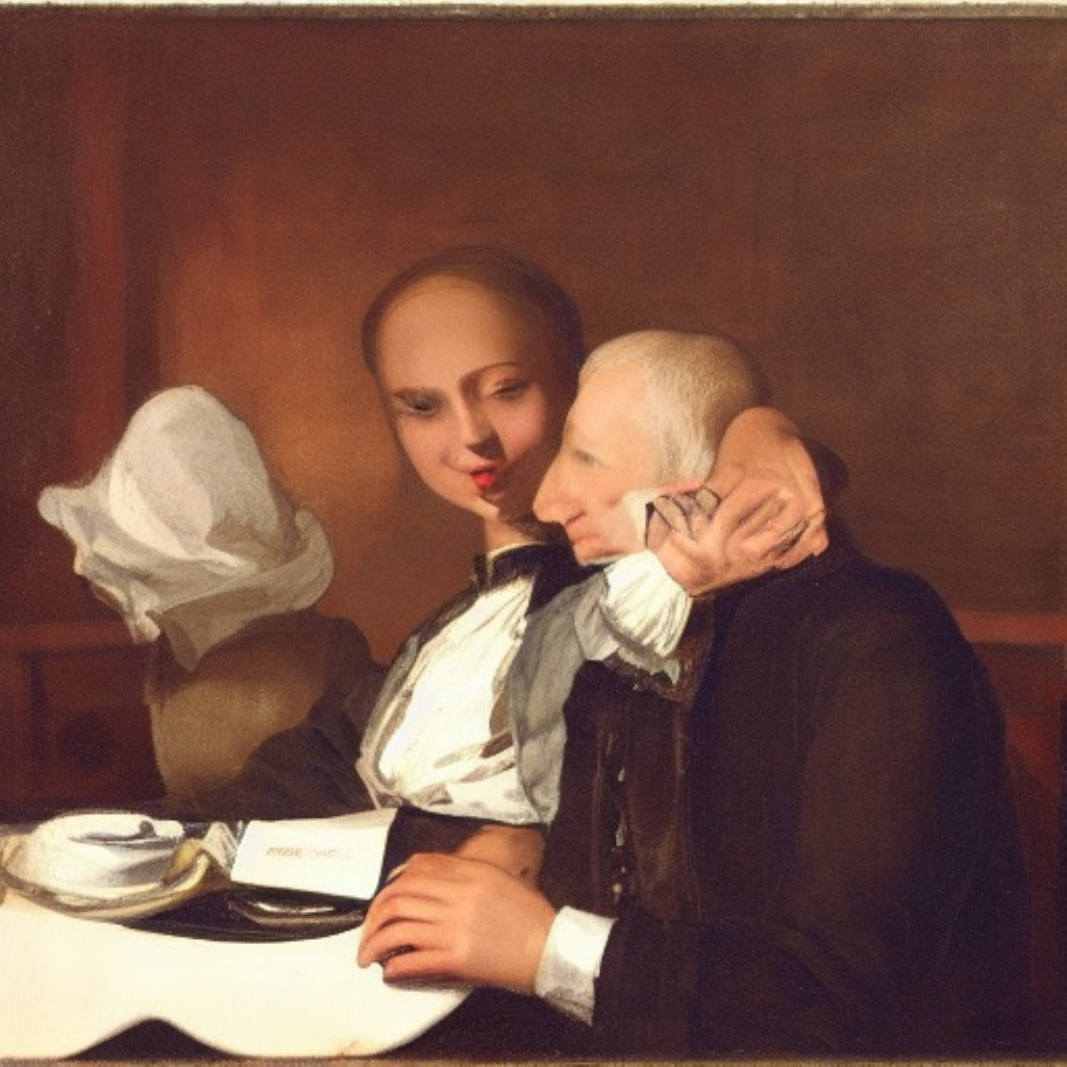}\hfill
    \includegraphics[width=0.24\linewidth]{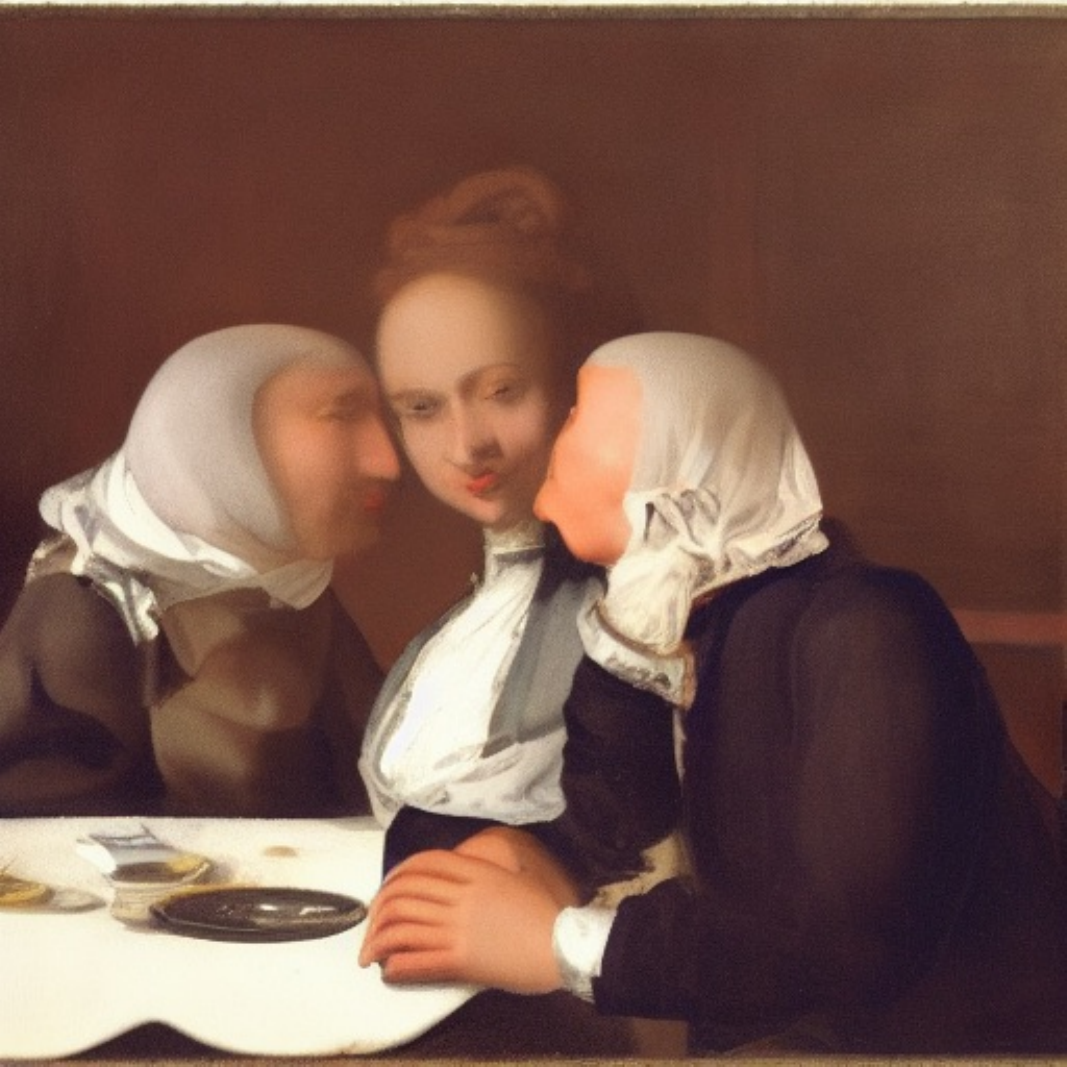}\hfill
    \includegraphics[width=0.24\linewidth]{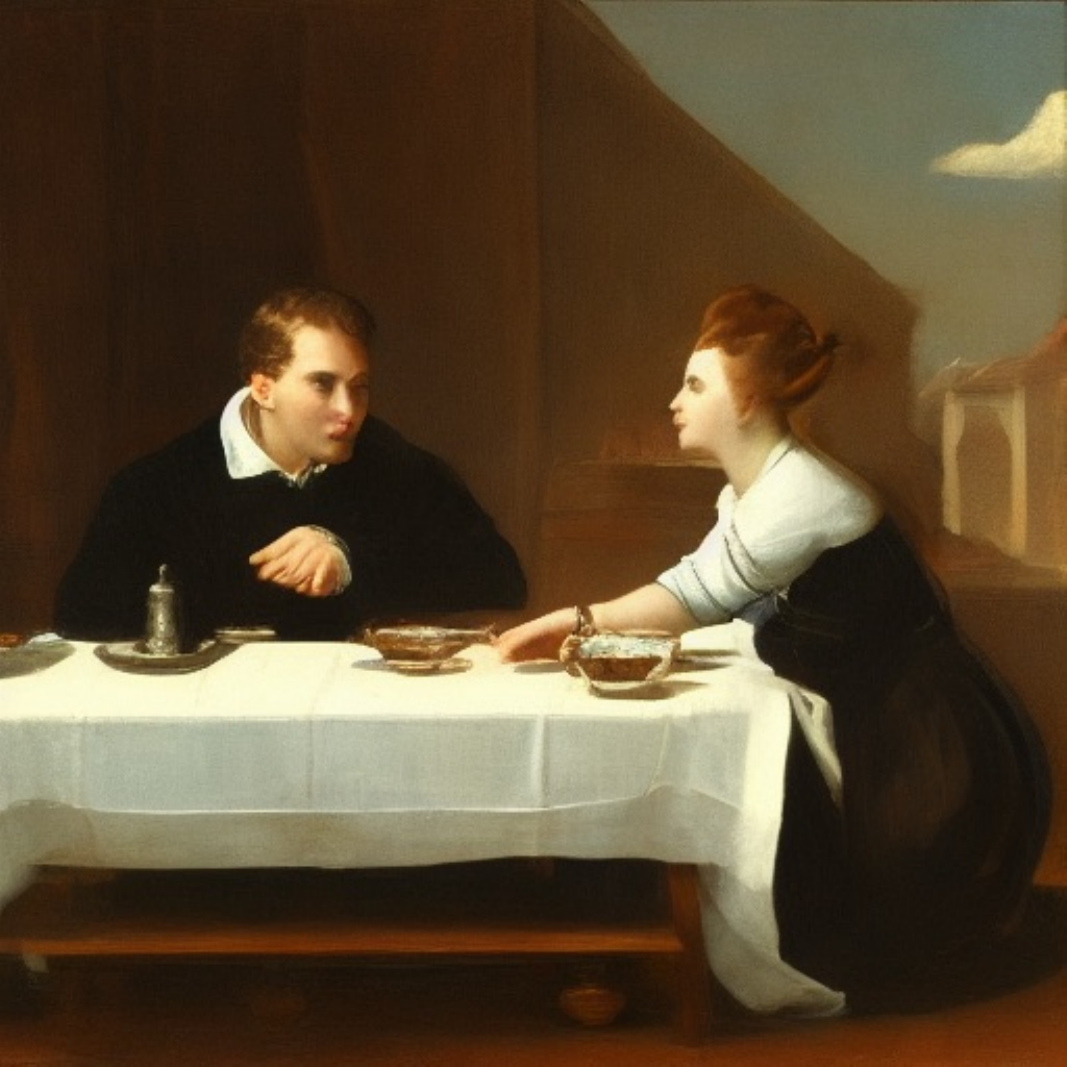}
    
    \vspace{0.5mm}
    \caption*{\small ``A young man and woman sitting at a table."}
    
    \caption{Steps = 6} 
    \label{subfig:nfe6}
  \end{subfigure}
  
  \vspace{1em} %

  \begin{subfigure}{\linewidth}
    \centering
    
    \includegraphics[width=0.24\linewidth]{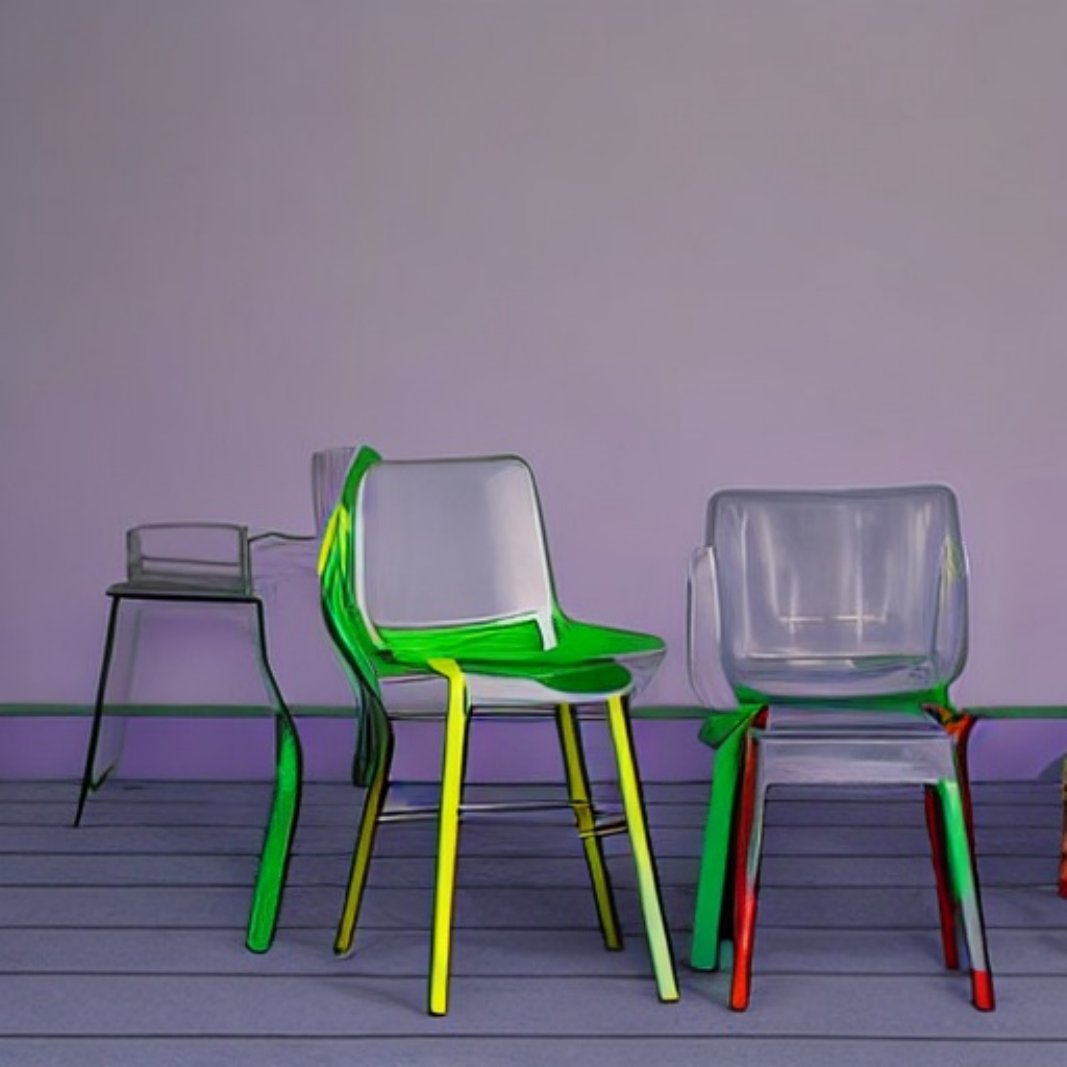}\hfill
    \includegraphics[width=0.24\linewidth]{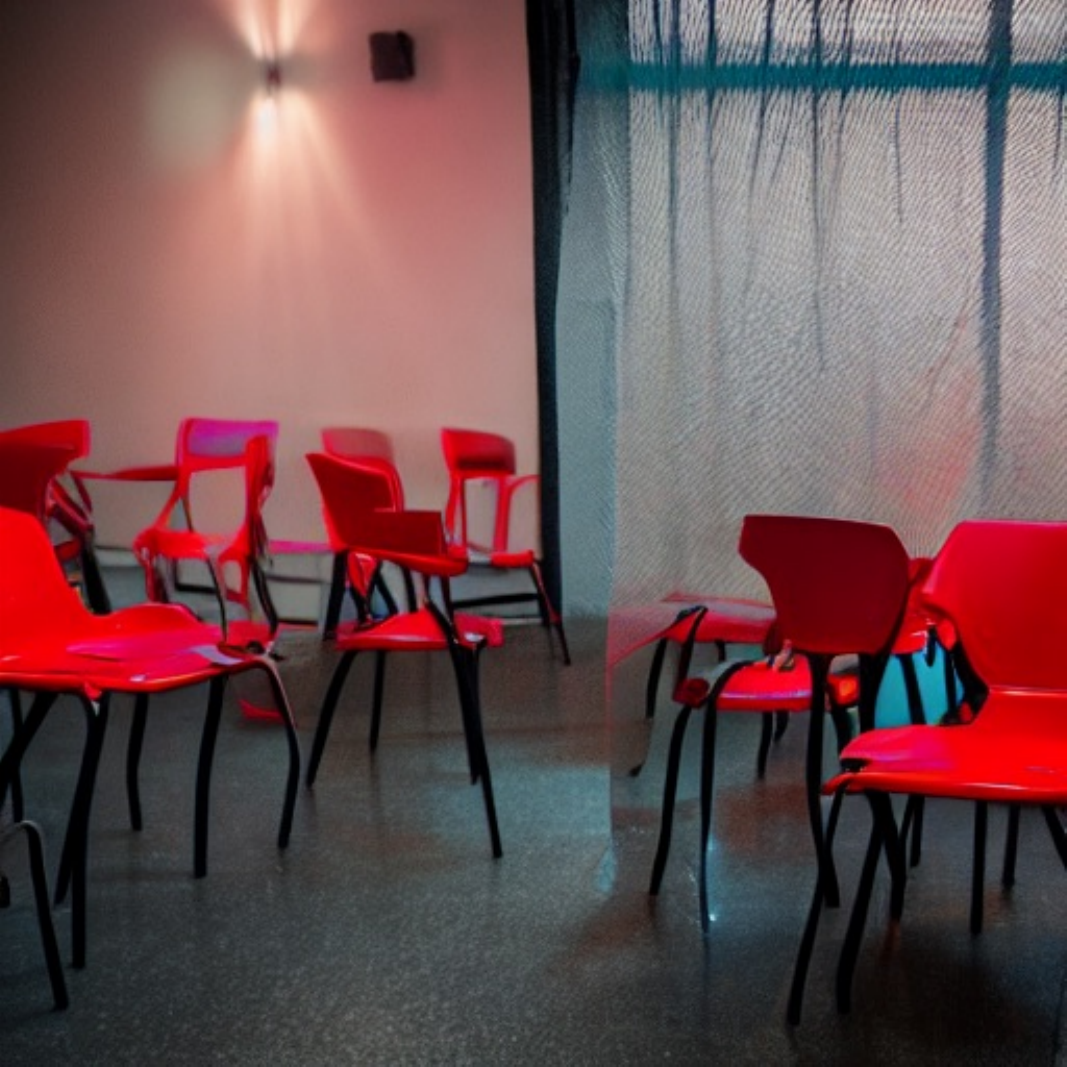}\hfill
    \includegraphics[width=0.24\linewidth]{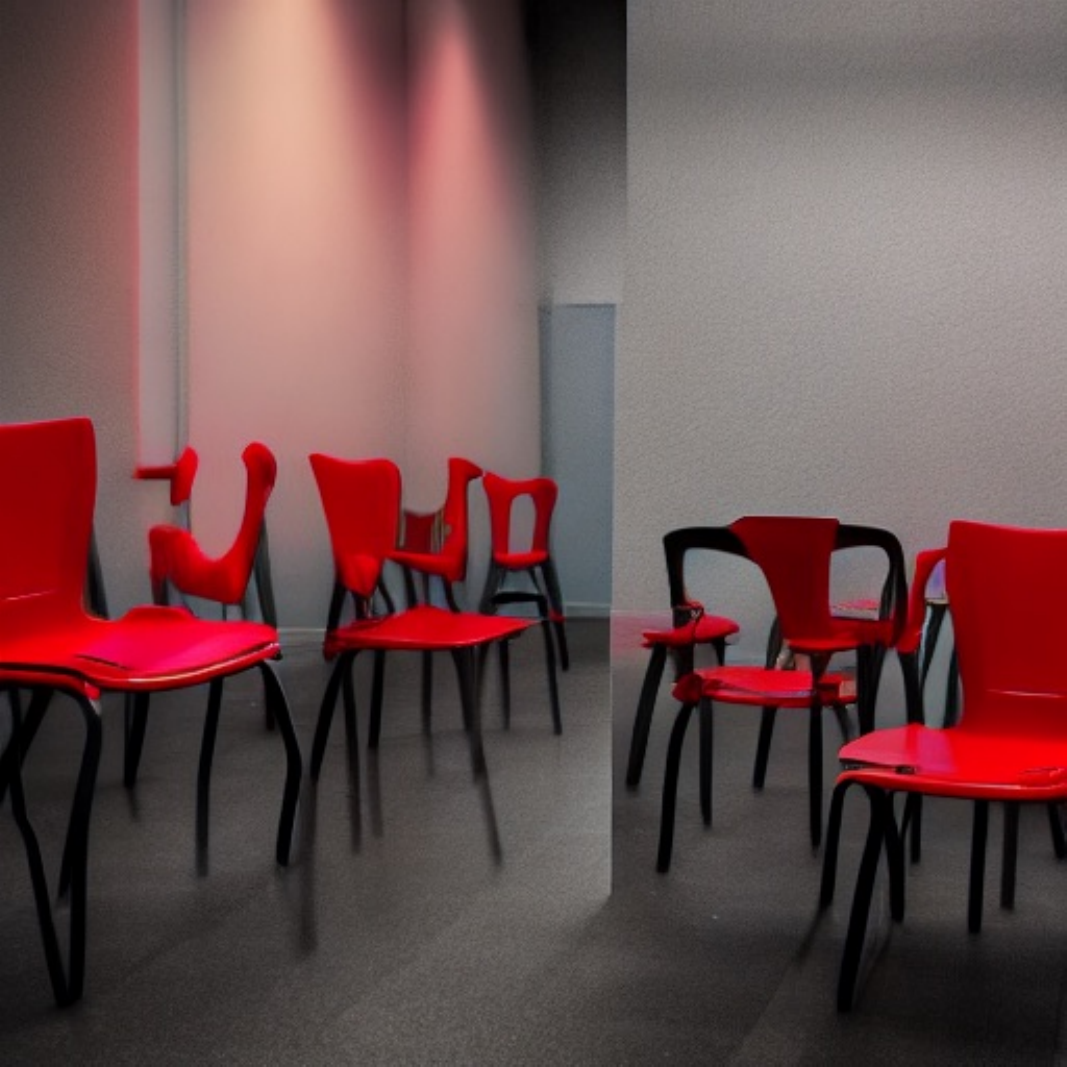}\hfill
    \includegraphics[width=0.24\linewidth]{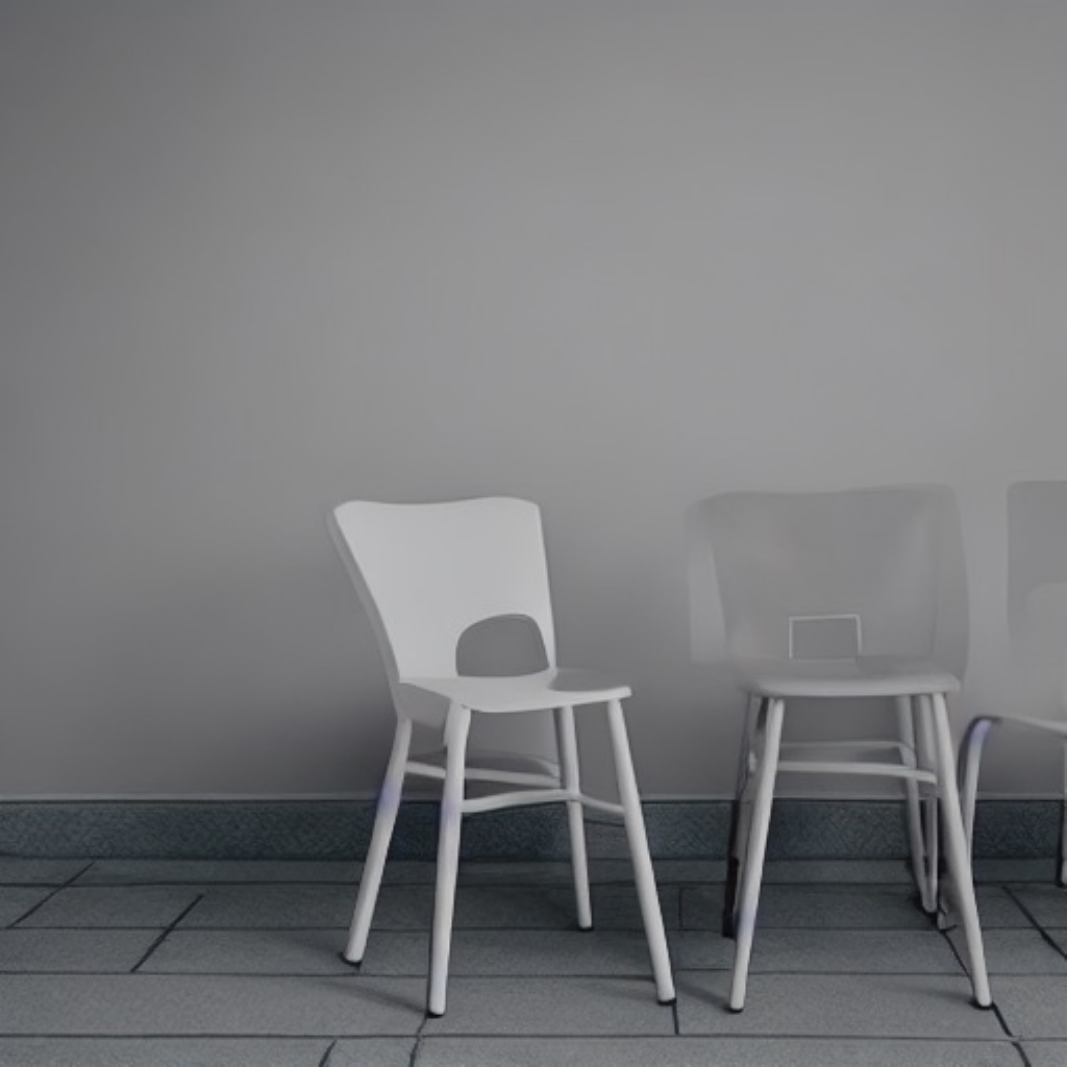}
    
    \vspace{0.5mm}
    \caption*{\small ``Hard plastic chairs in a dimly lit room."}
    \vspace{1mm}

    \includegraphics[width=0.24\linewidth]{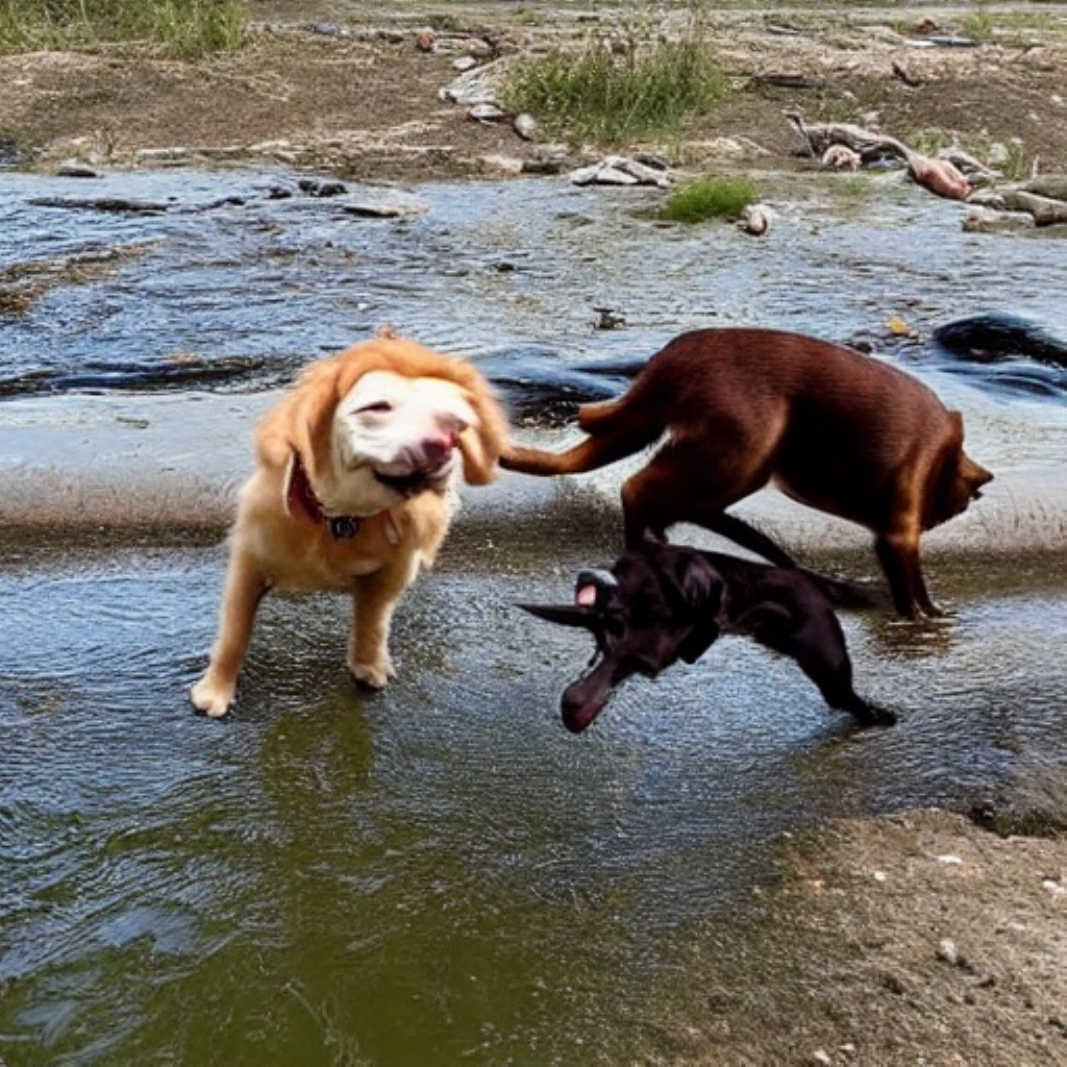}\hfill
    \includegraphics[width=0.24\linewidth]{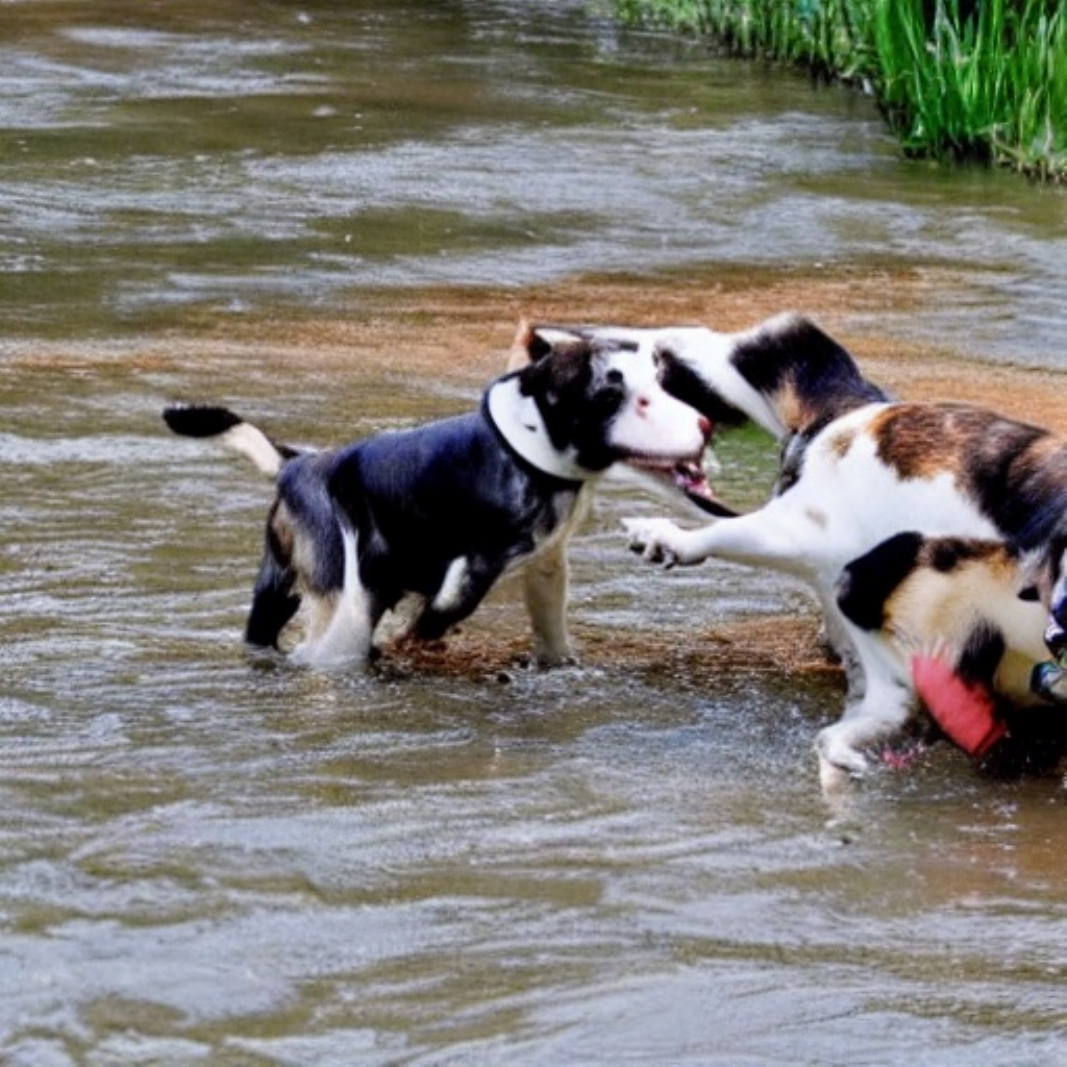}\hfill
    \includegraphics[width=0.24\linewidth]{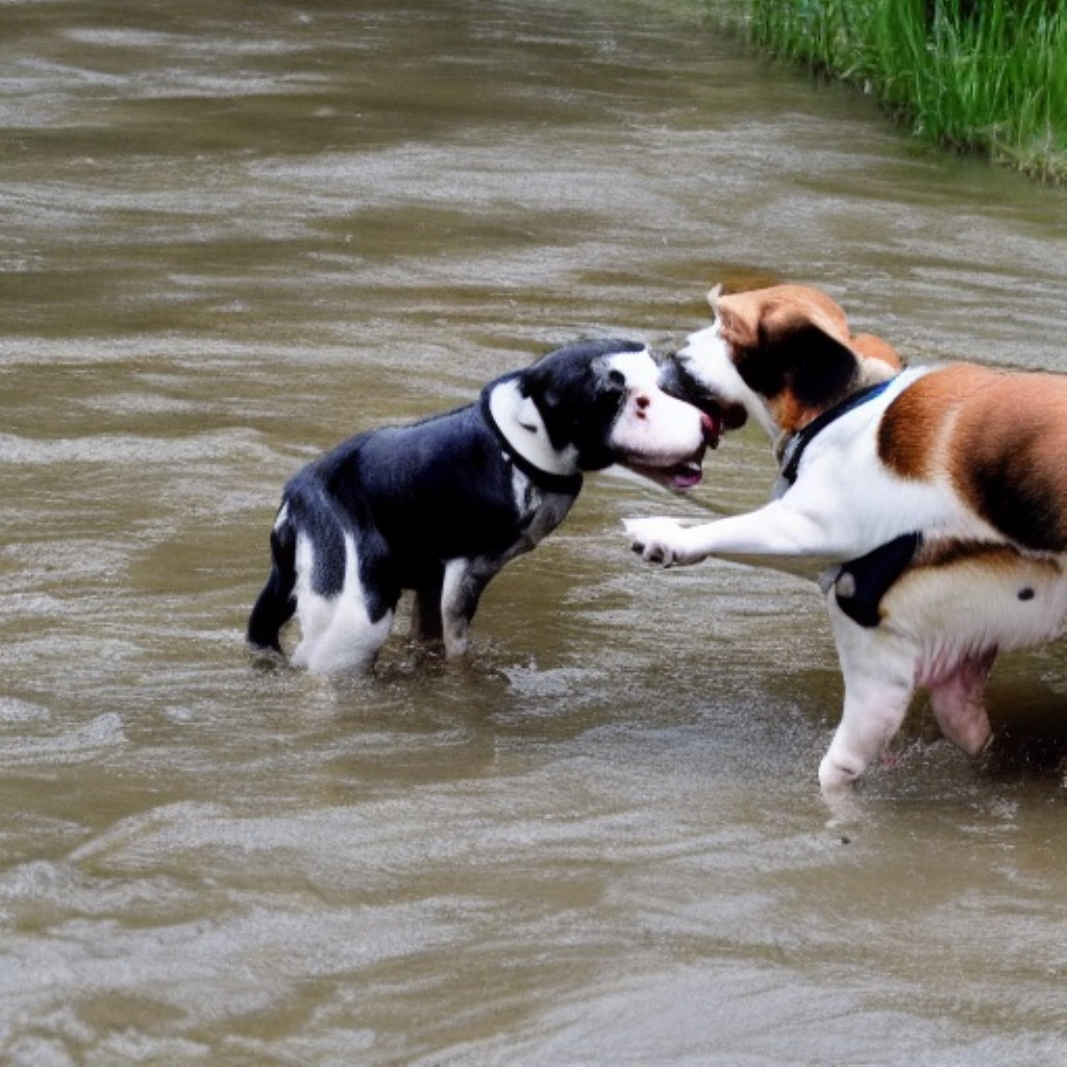}\hfill
    \includegraphics[width=0.24\linewidth]{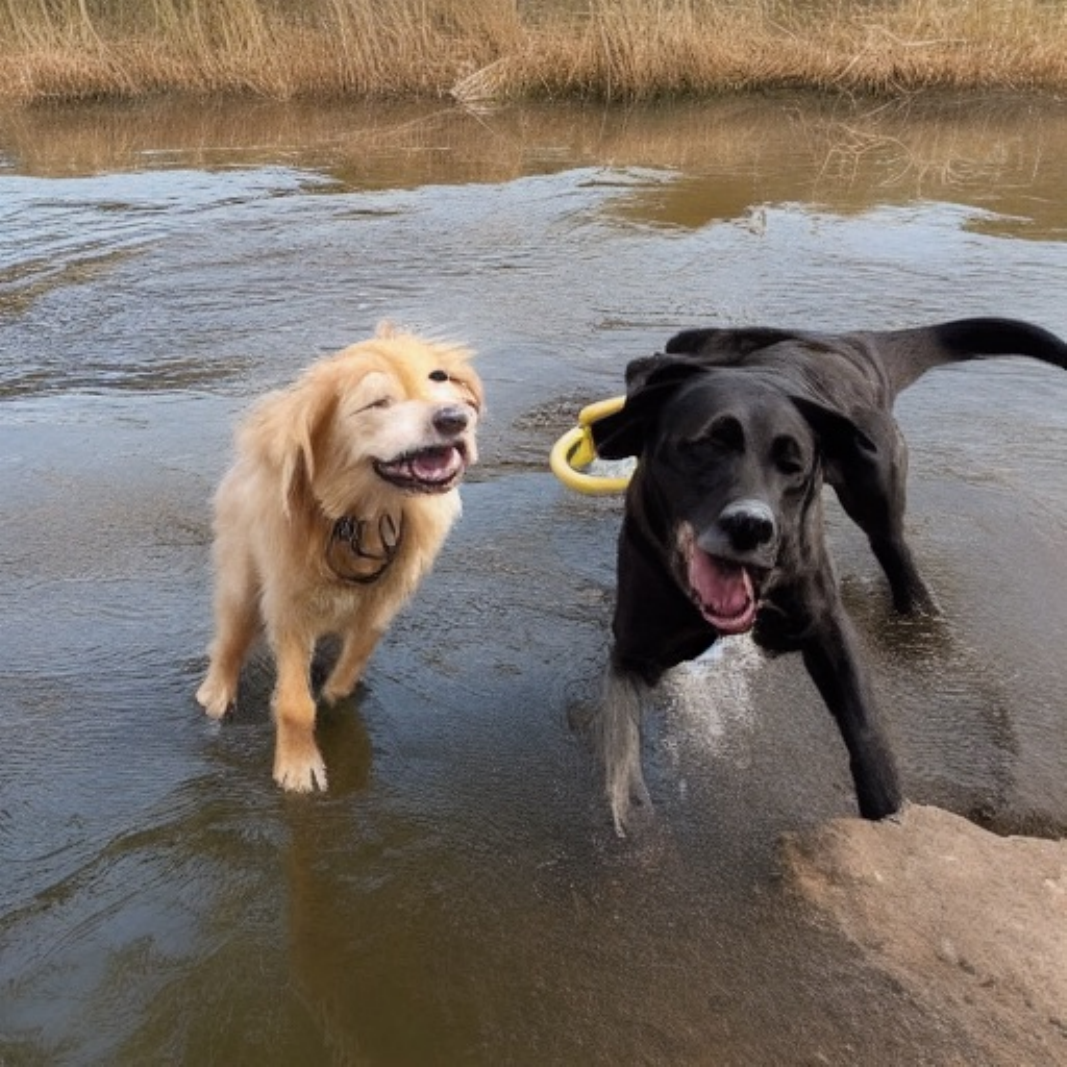}
    
    \vspace{0.5mm}
    \caption*{\small ``Two dogs play in a dammed up river."}
    
    \caption{Steps = 7}
    \label{subfig:nfe7}
  \end{subfigure}
  
  \caption{Side-by-side comparison of selected images generated with Stable Diffusion and UniPC solver in the high NFE regime (Steps $\in \{6, 7\}$).
    Methods (from left to right): DMN, GITS, LD3, and D2PO.}
  \label{fig:sd_unipc_nfehigh}
\end{figure*}

\begin{figure*}[t] 
  \centering
  
  \captionsetup[subfigure]{font=small, labelfont=small}
  
  \begin{subfigure}{\linewidth}
    \centering
    
    \begin{minipage}{0.24\linewidth}\centering\small DMN\end{minipage} \hfill
    \begin{minipage}{0.24\linewidth}\centering\small GITS\end{minipage} \hfill
    \begin{minipage}{0.24\linewidth}\centering\small LD3\end{minipage} \hfill
    \begin{minipage}{0.24\linewidth}\centering\small D2PO\end{minipage}
    \vspace{1mm} 
    
    \includegraphics[width=0.24\linewidth]{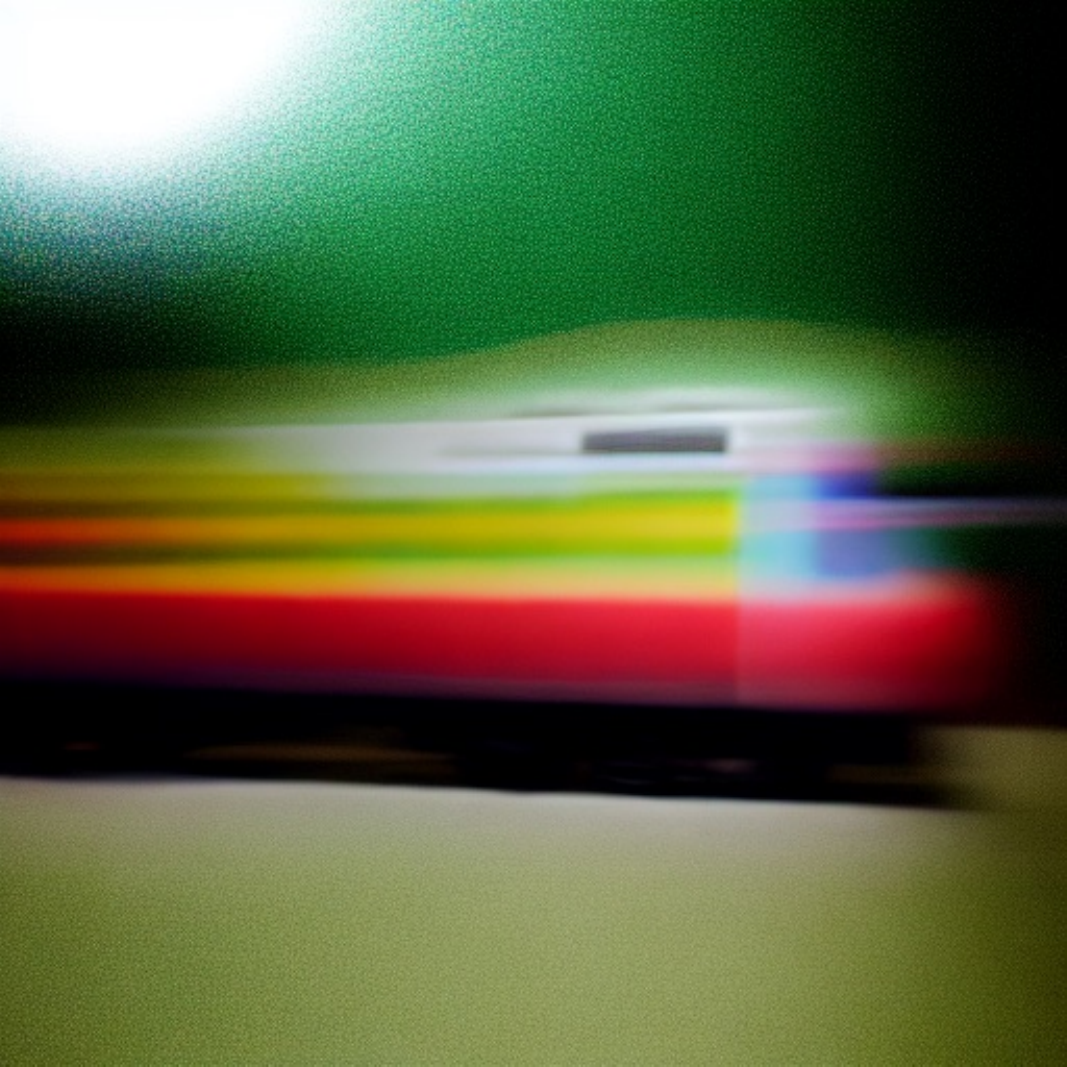}\hfill
    \includegraphics[width=0.24\linewidth]{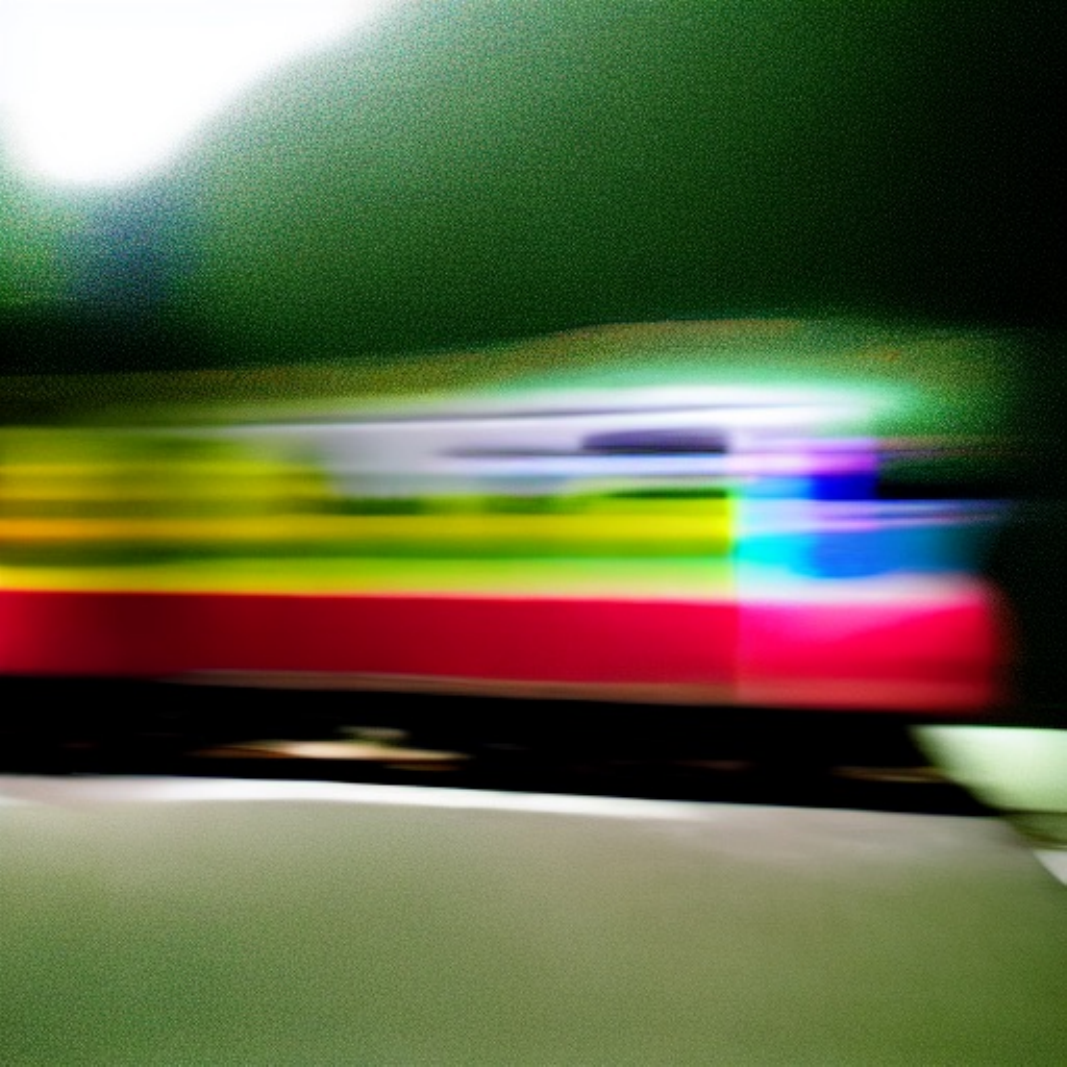}\hfill
    \includegraphics[width=0.24\linewidth]{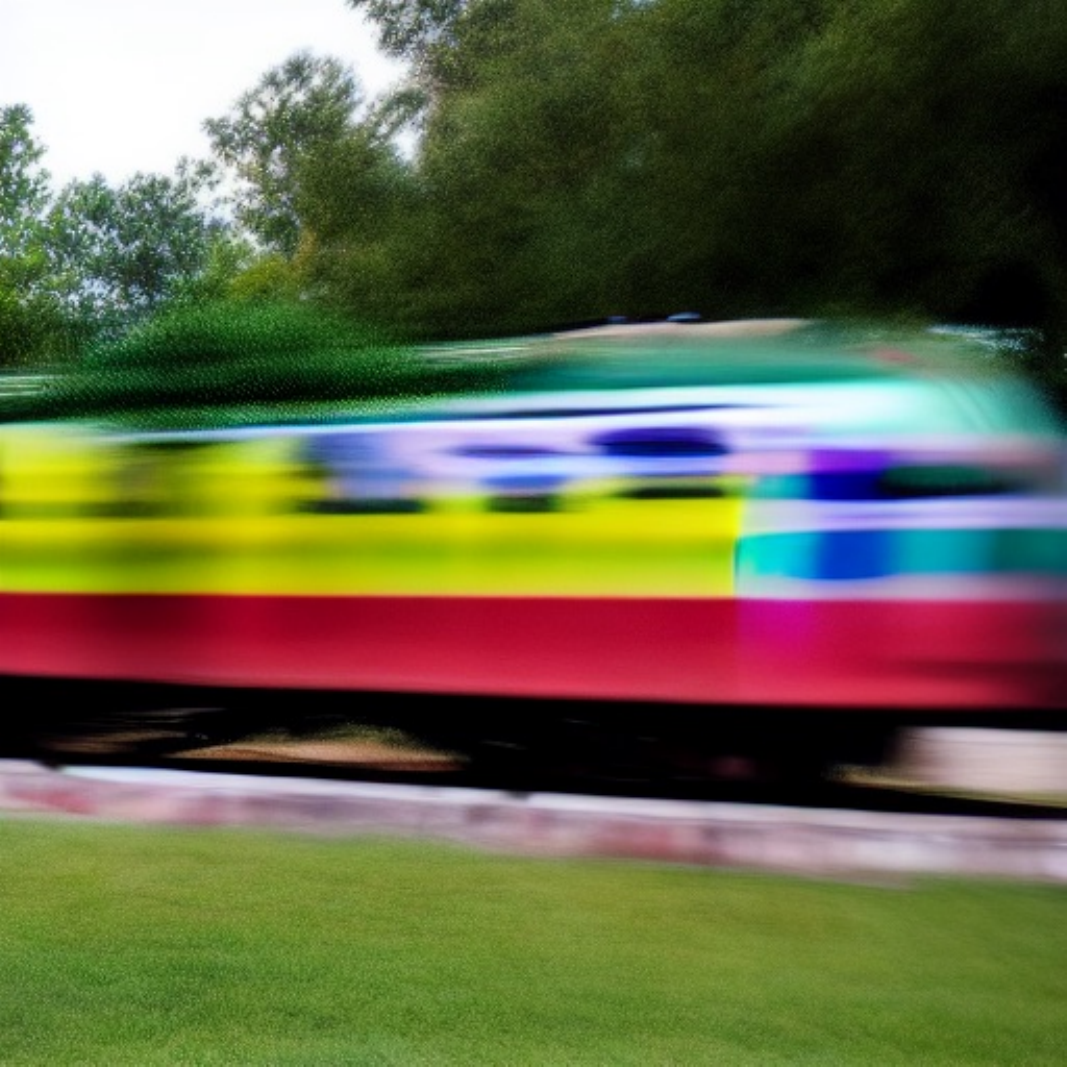}\hfill
    \includegraphics[width=0.24\linewidth]{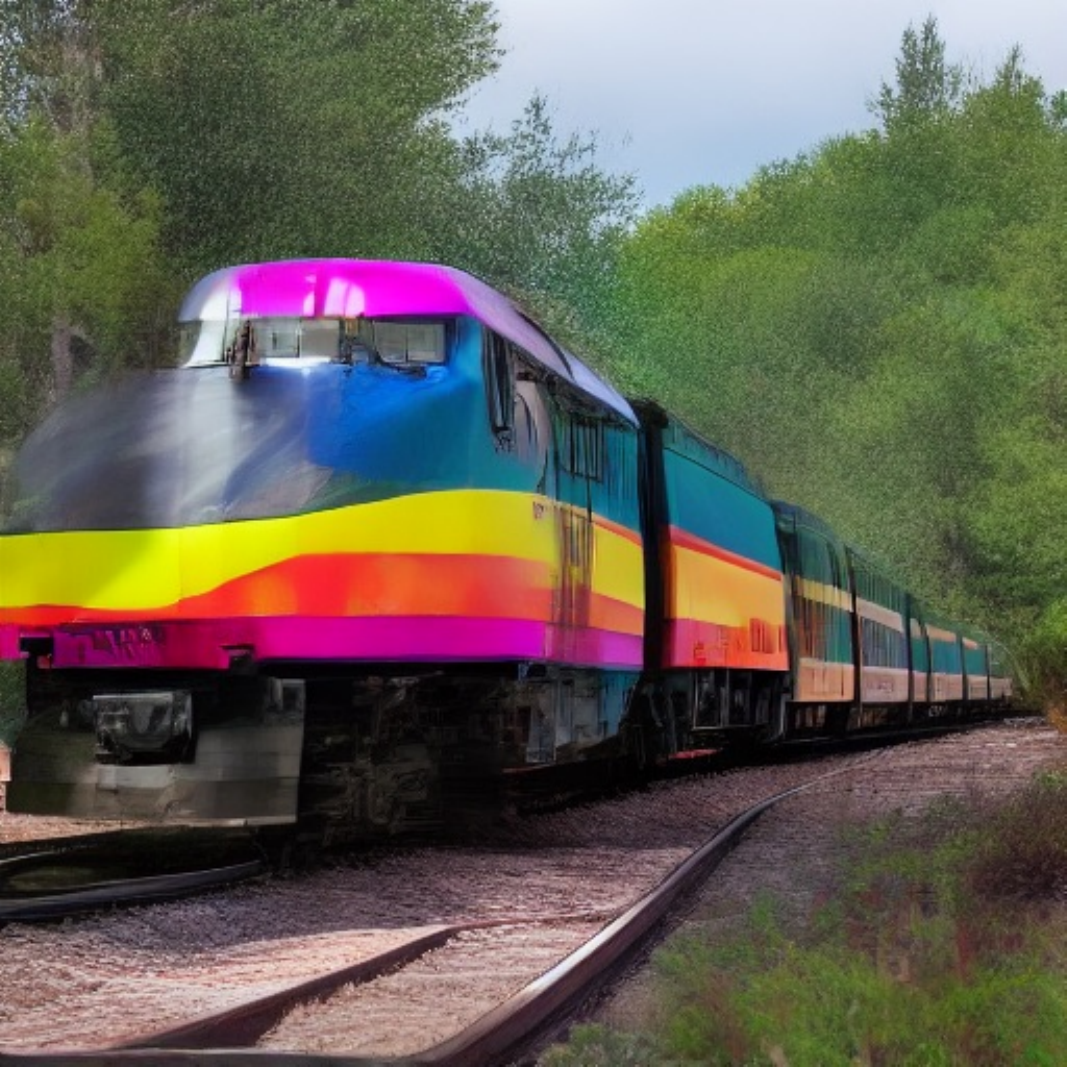}
    
    \vspace{0.5mm} 
    \caption*{ ``A multicolored train passing another set of tracks."}
    \vspace{1mm} %

    \includegraphics[width=0.24\linewidth]{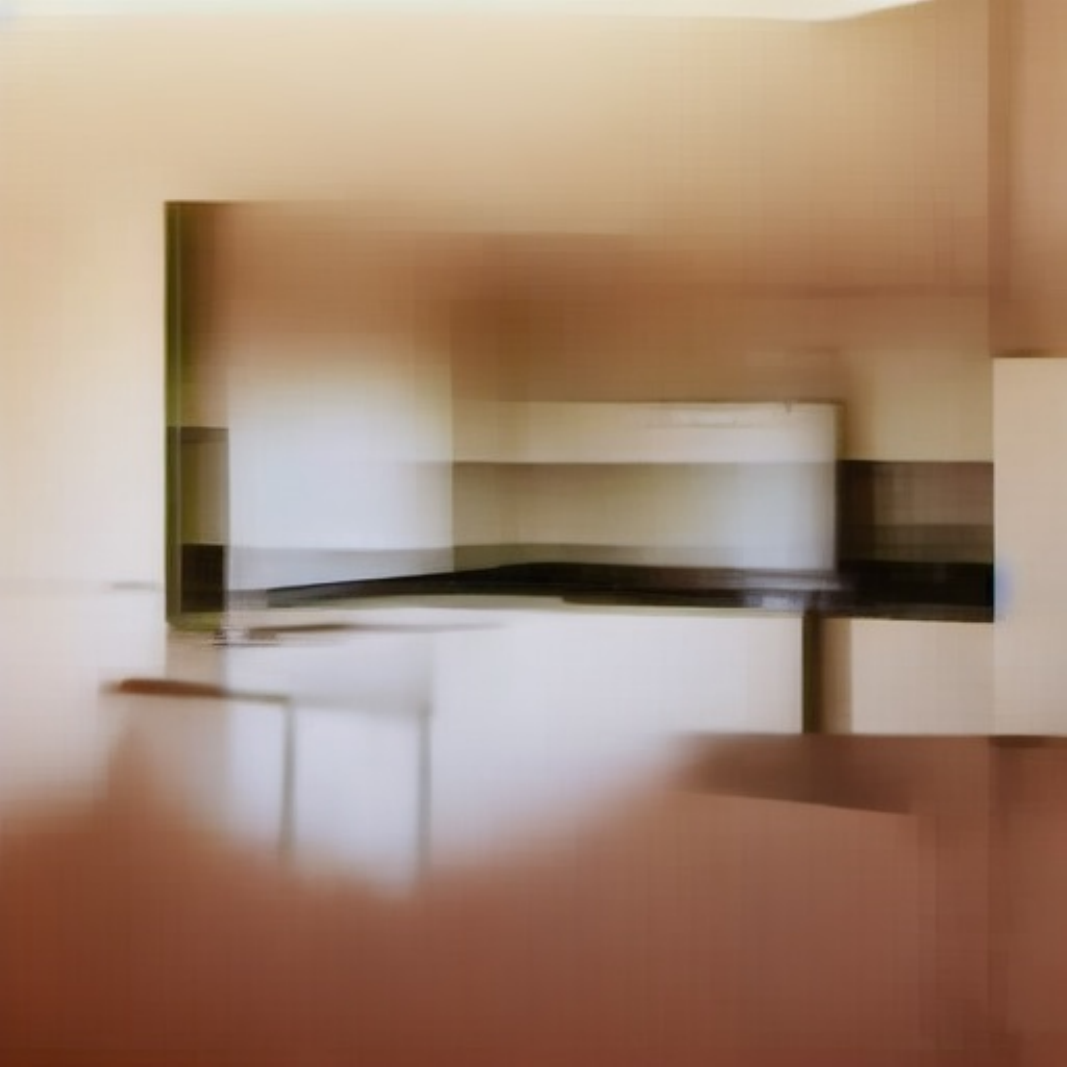}\hfill
    \includegraphics[width=0.24\linewidth]{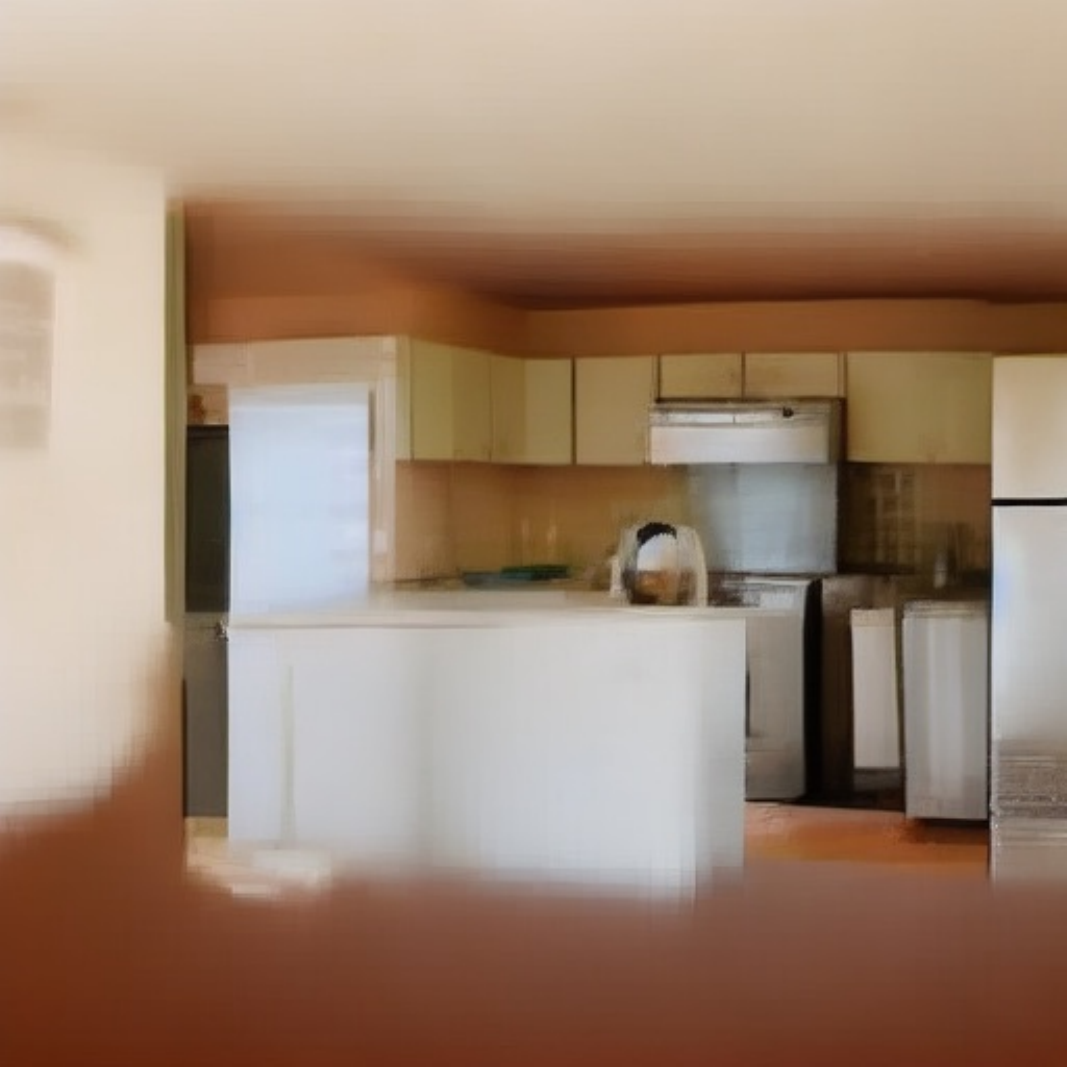}\hfill
    \includegraphics[width=0.24\linewidth]{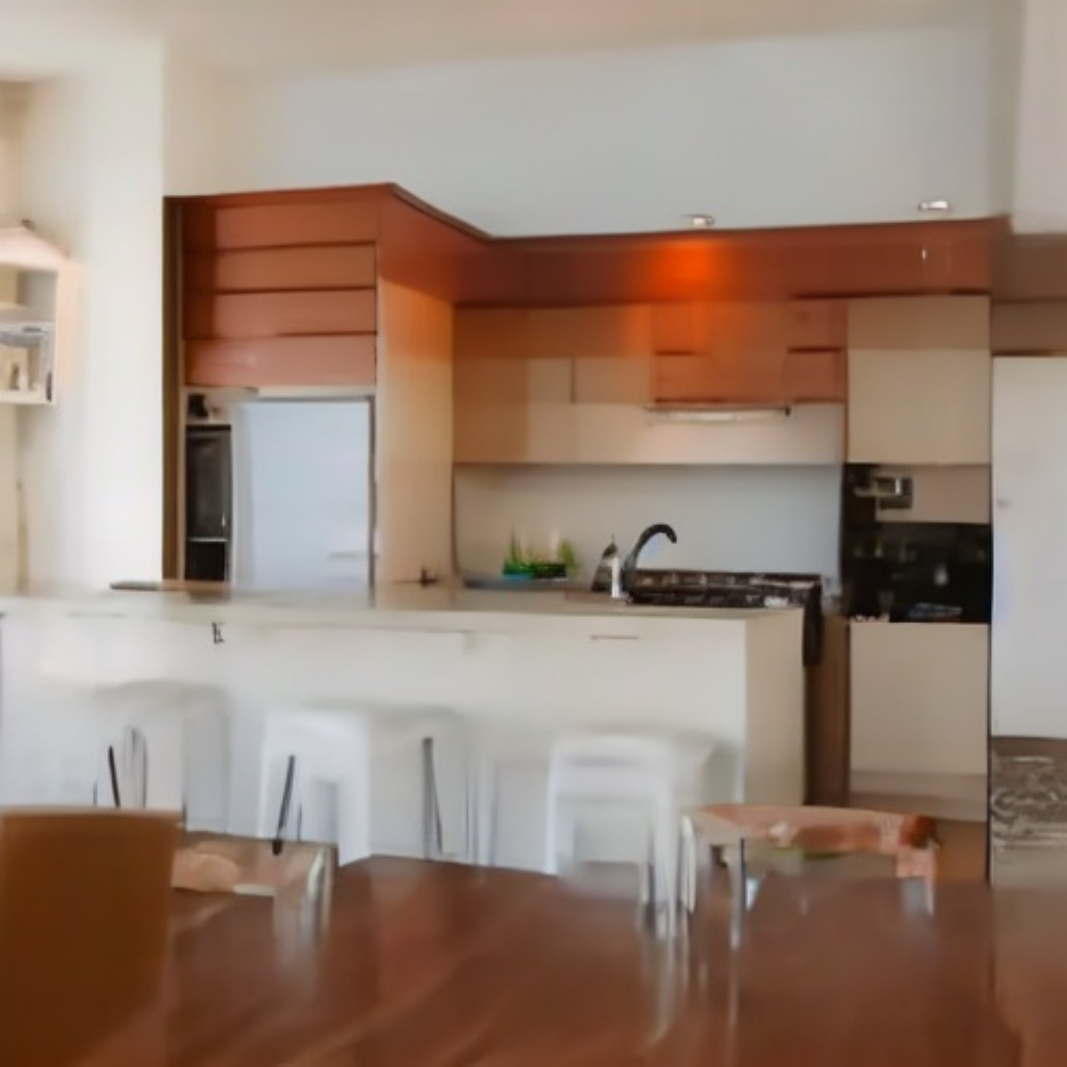}\hfill
    \includegraphics[width=0.24\linewidth]{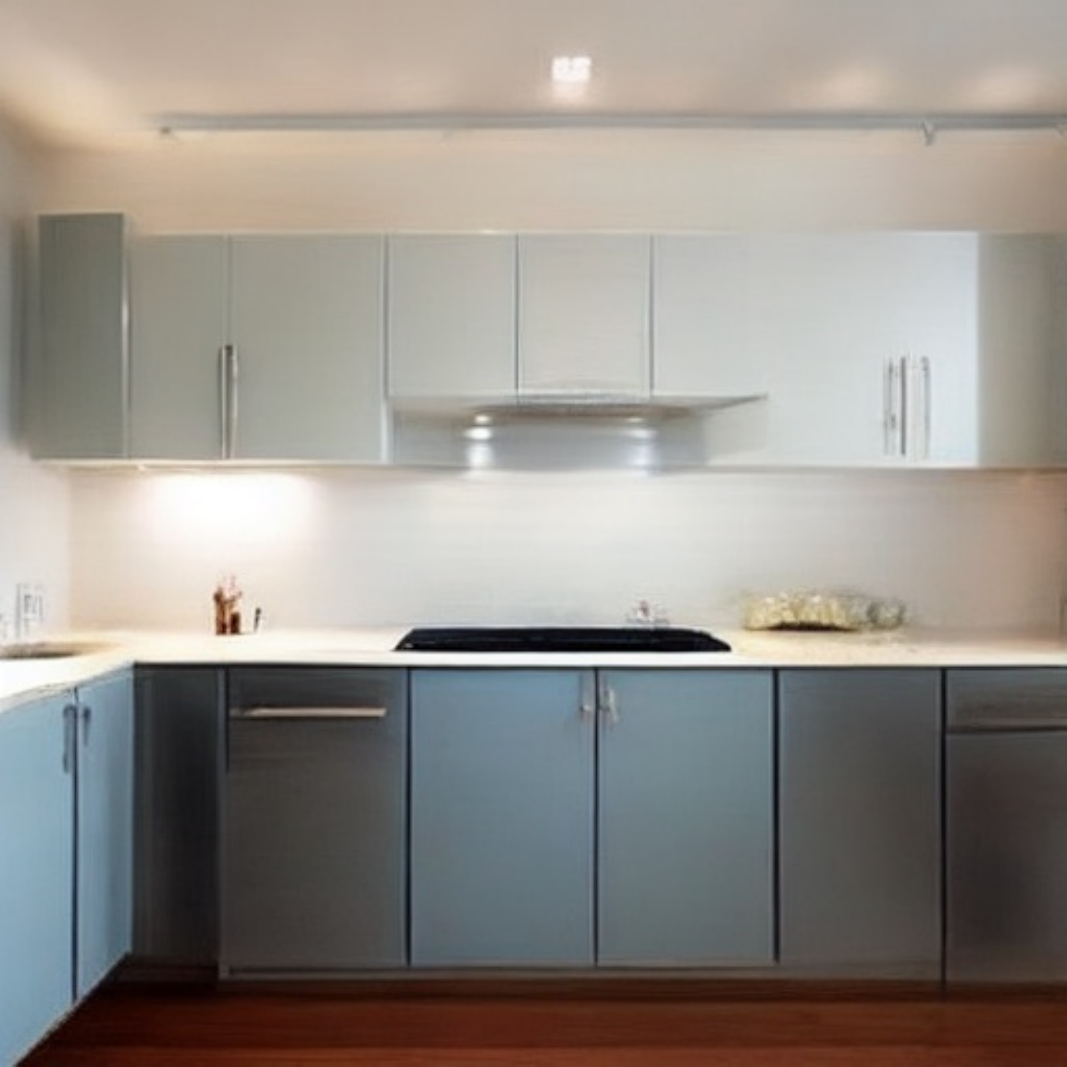}
    
    \vspace{0.5mm}
    \caption*{ ``A kitchen that is very clean in a house."}
    
    \caption{ Steps = 4} 
    \label{subfig:nfe4}
  \end{subfigure}
  
  \vspace{1em} %

  \begin{subfigure}{\linewidth}
    \centering
    
    \includegraphics[width=0.24\linewidth]{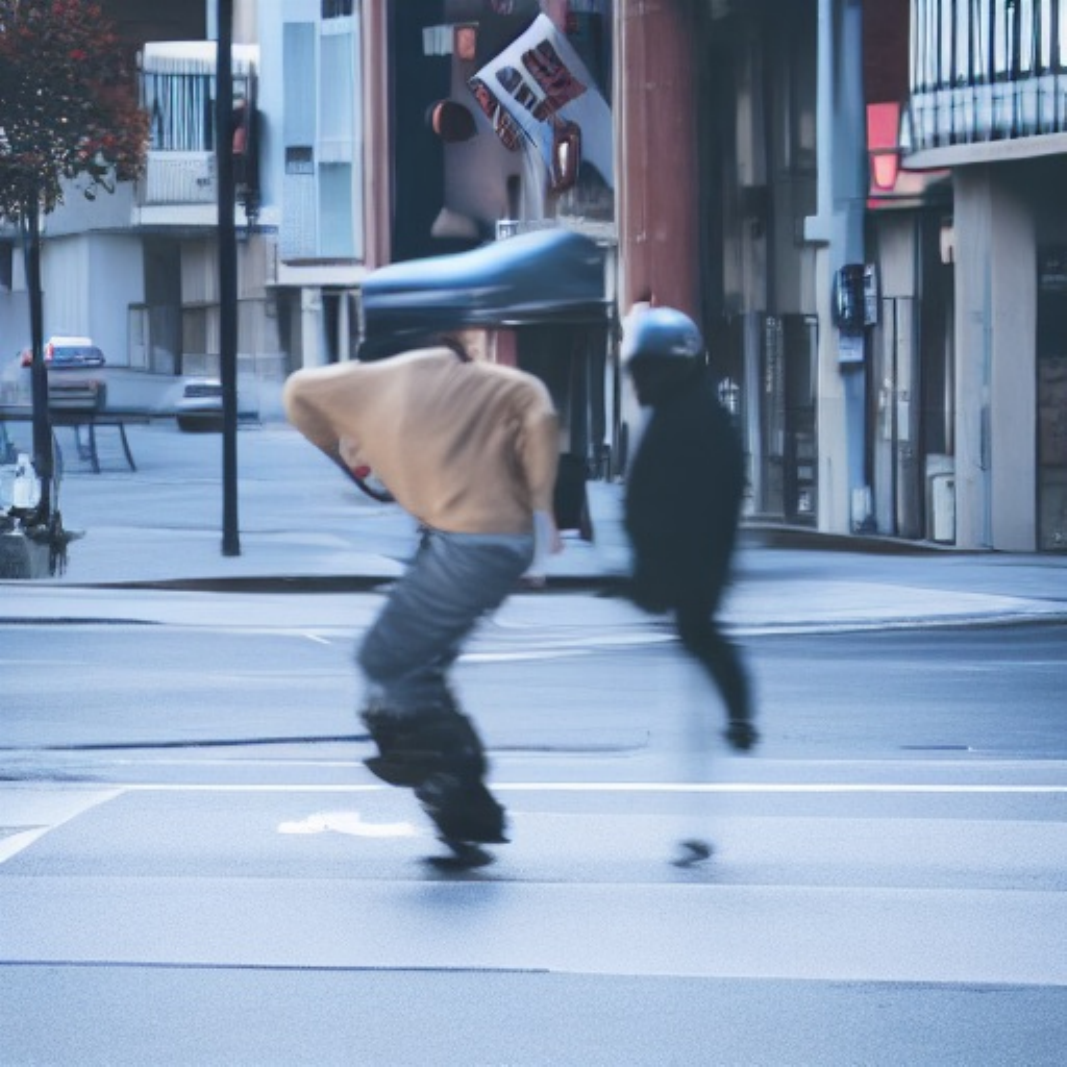}\hfill
    \includegraphics[width=0.24\linewidth]{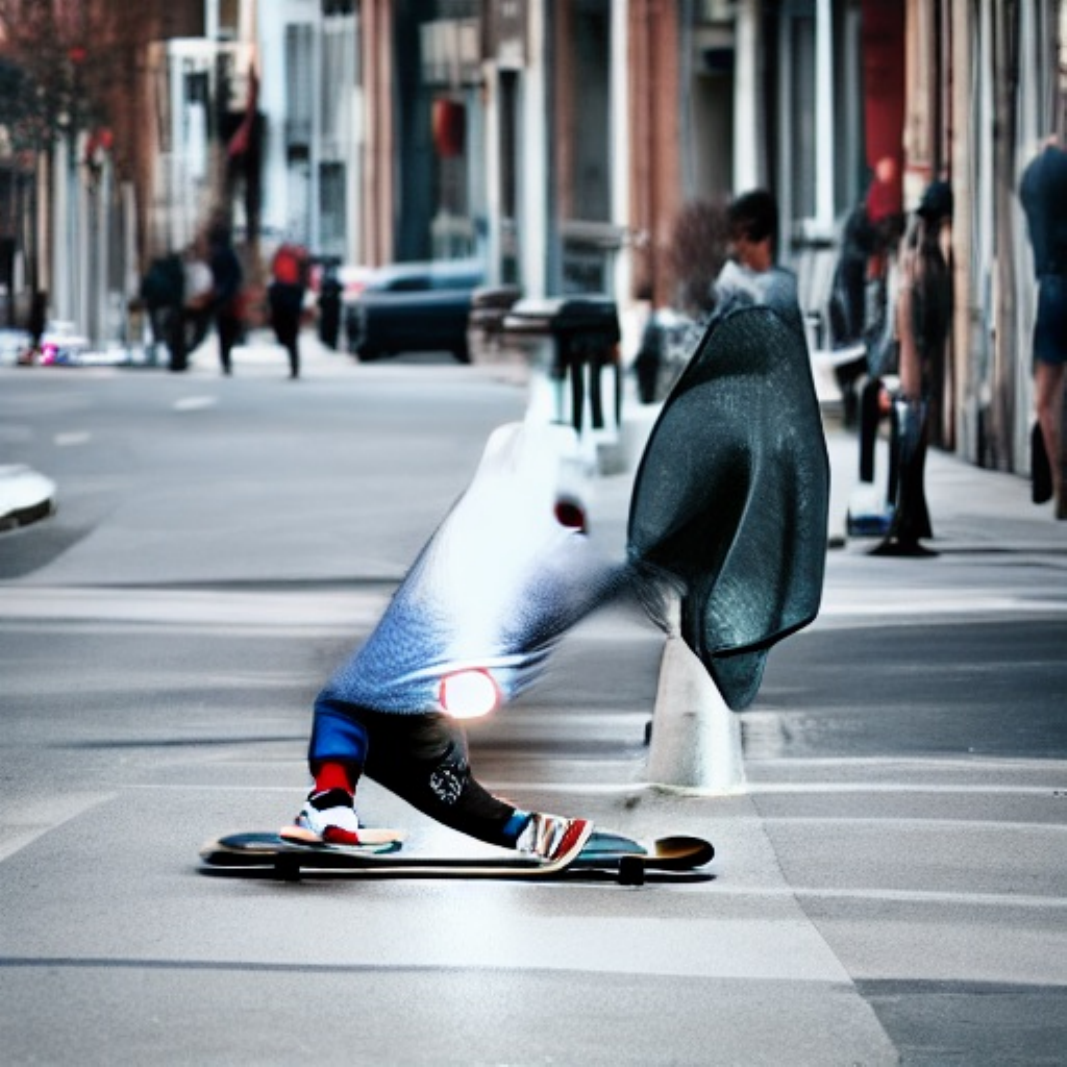}\hfill
    \includegraphics[width=0.24\linewidth]{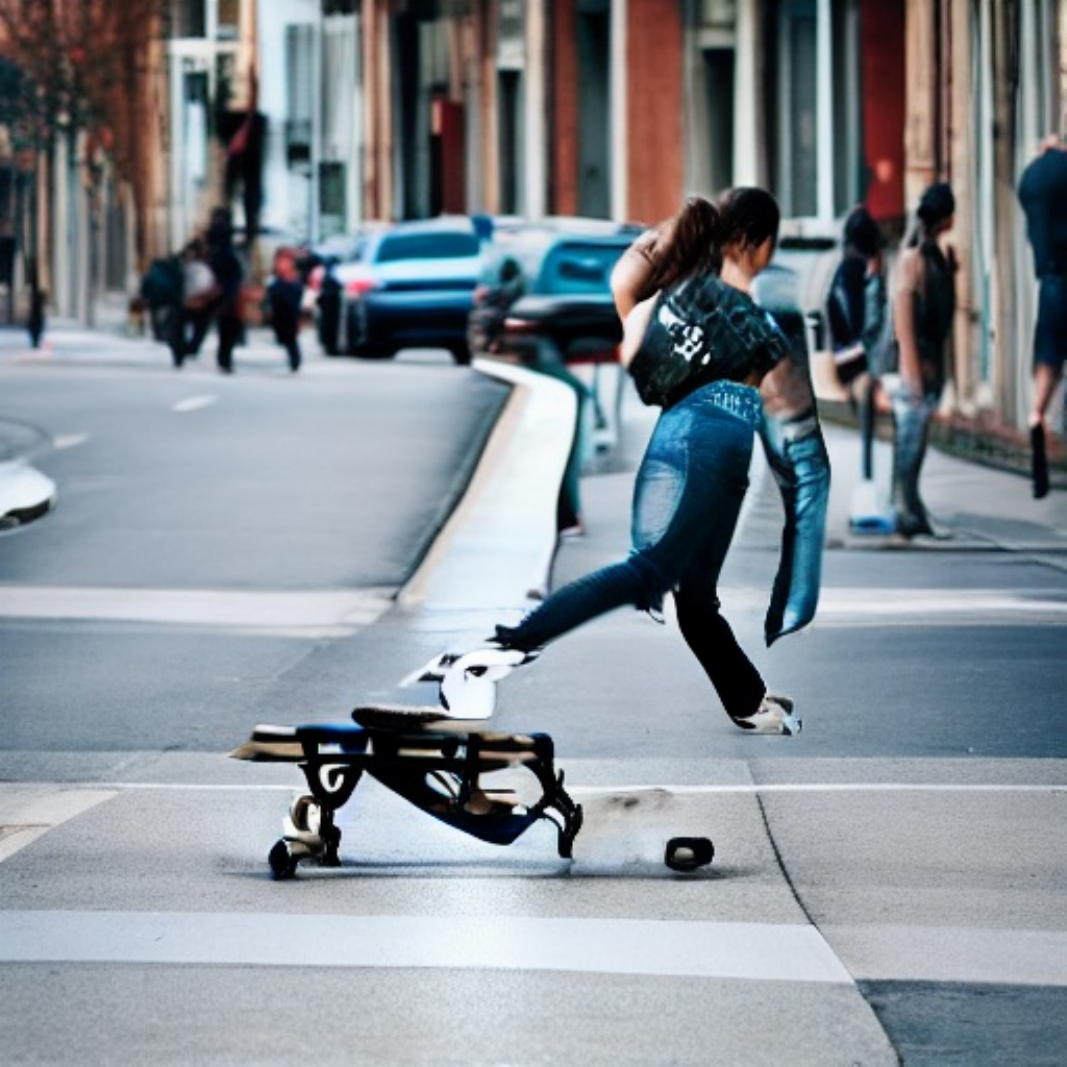}\hfill
    \includegraphics[width=0.24\linewidth]{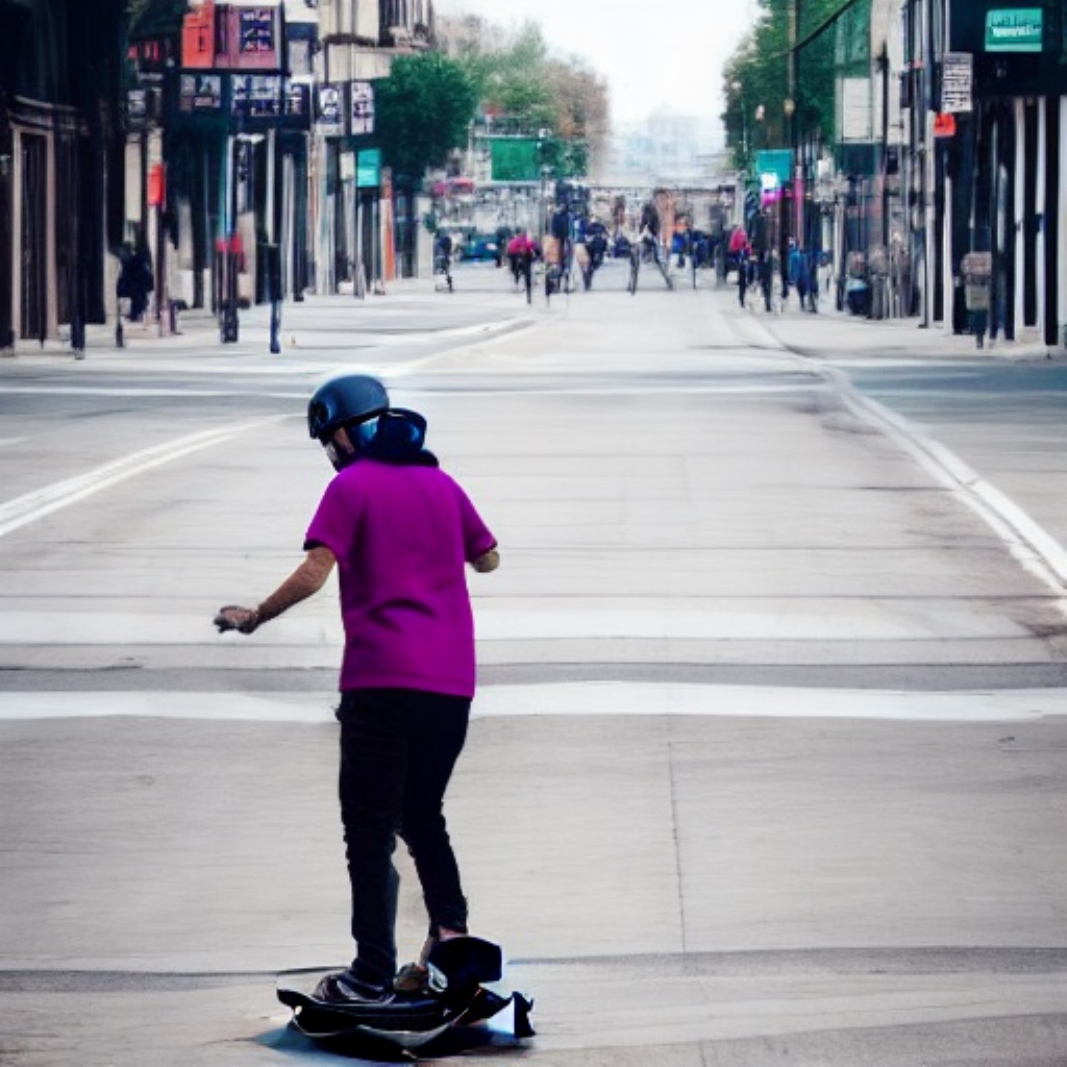}
    
    \vspace{0.5mm}
    \caption*{``A person with a skateboard on a street."}
    \vspace{1mm}

    \includegraphics[width=0.24\linewidth]{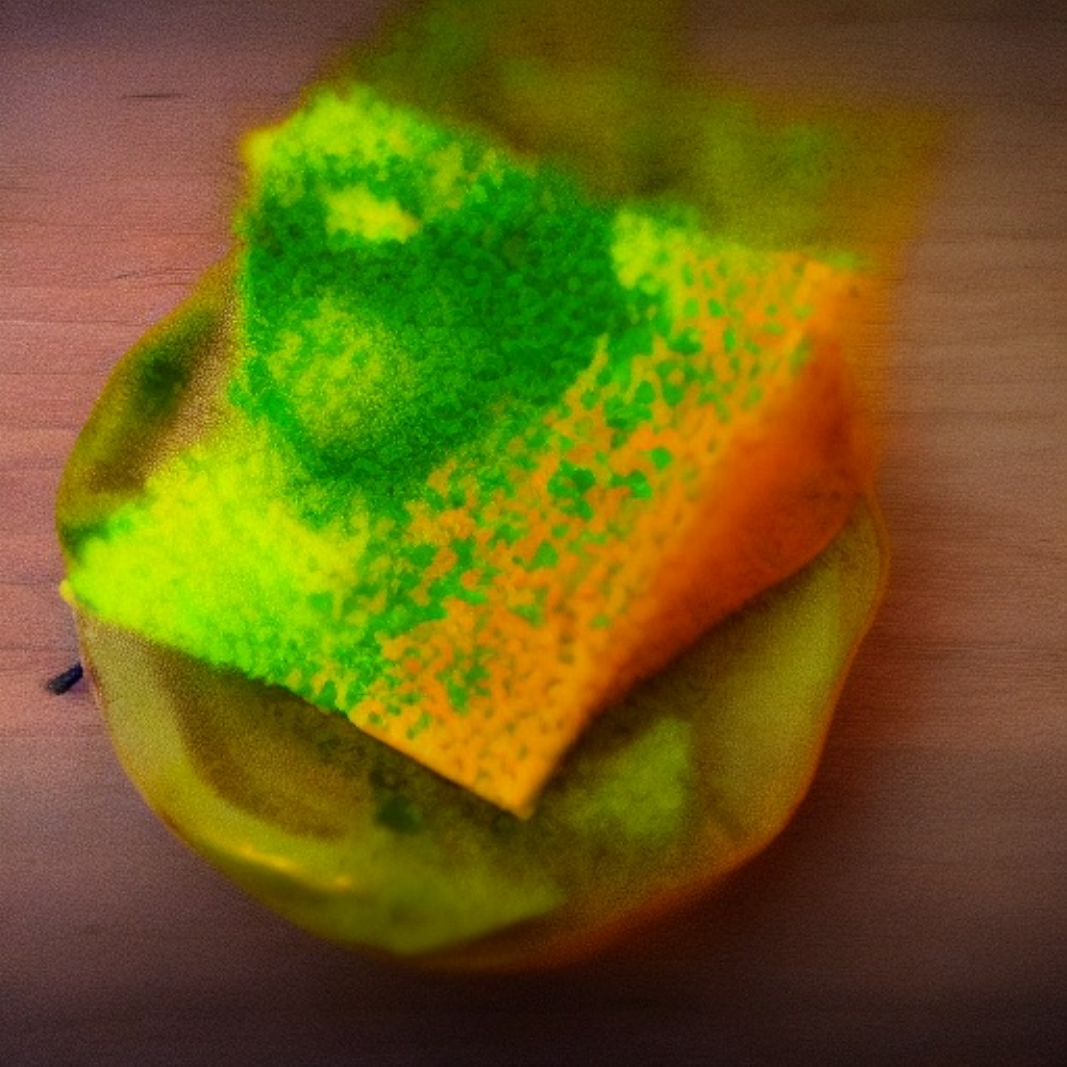}\hfill
    \includegraphics[width=0.24\linewidth]{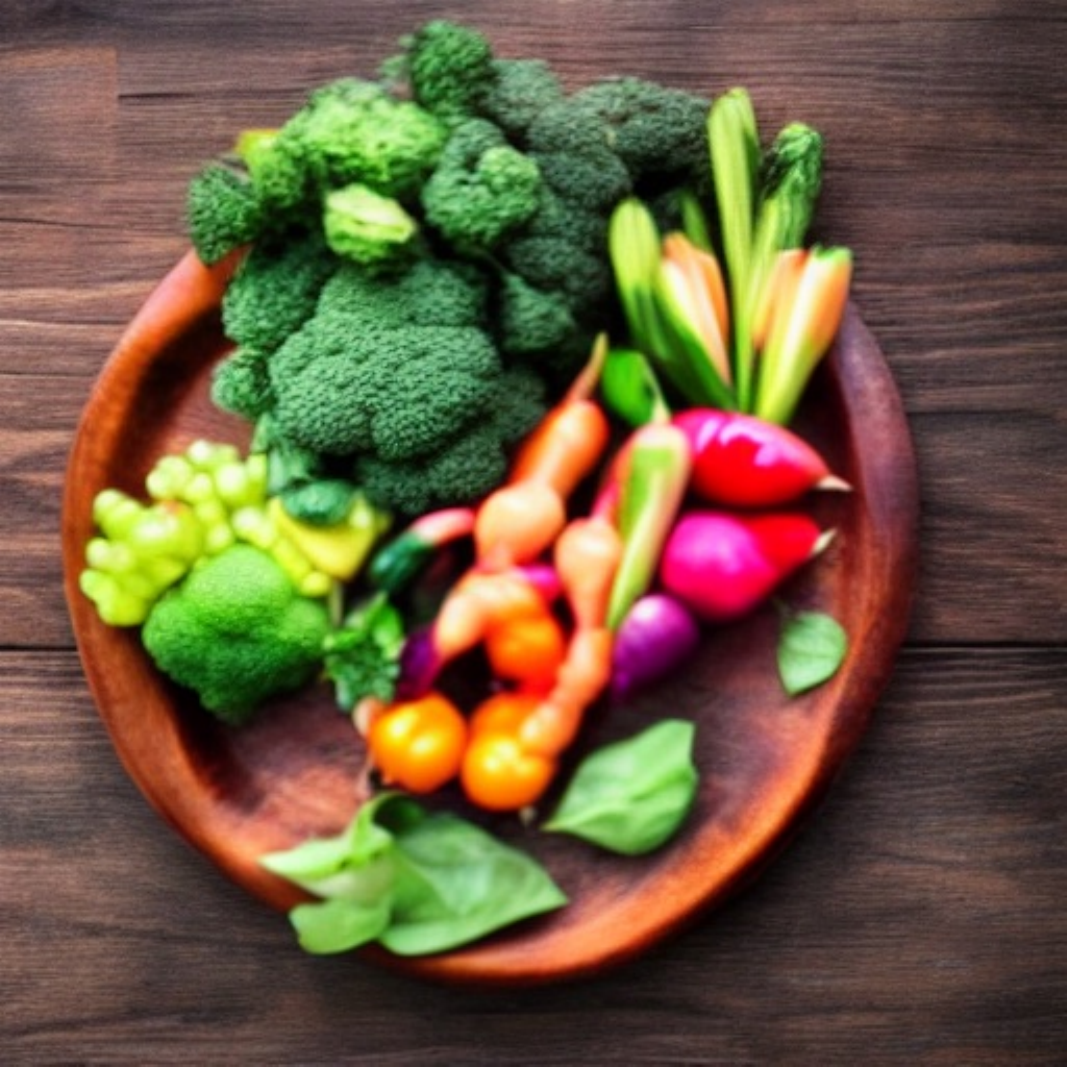}\hfill
    \includegraphics[width=0.24\linewidth]{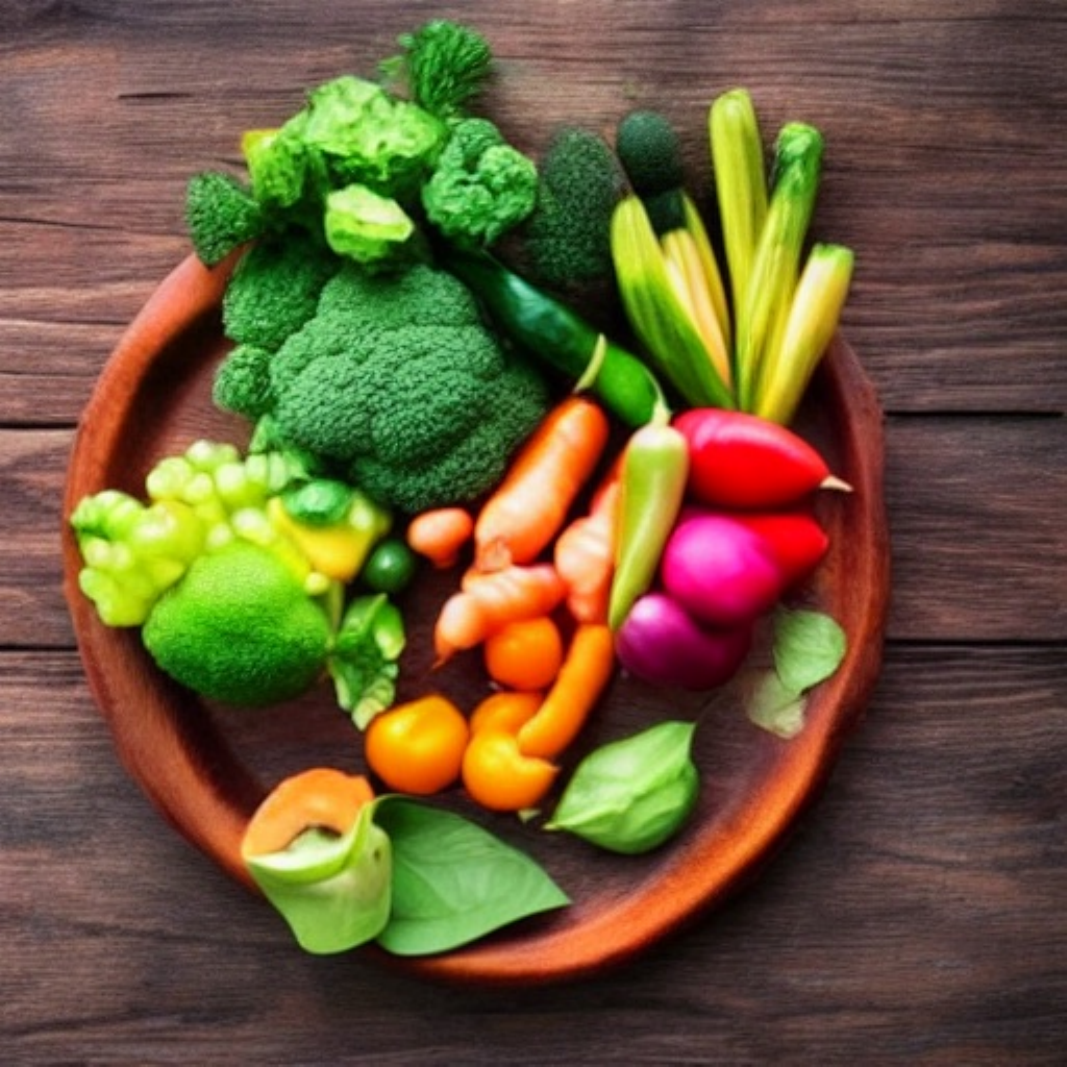}\hfill
    \includegraphics[width=0.24\linewidth]{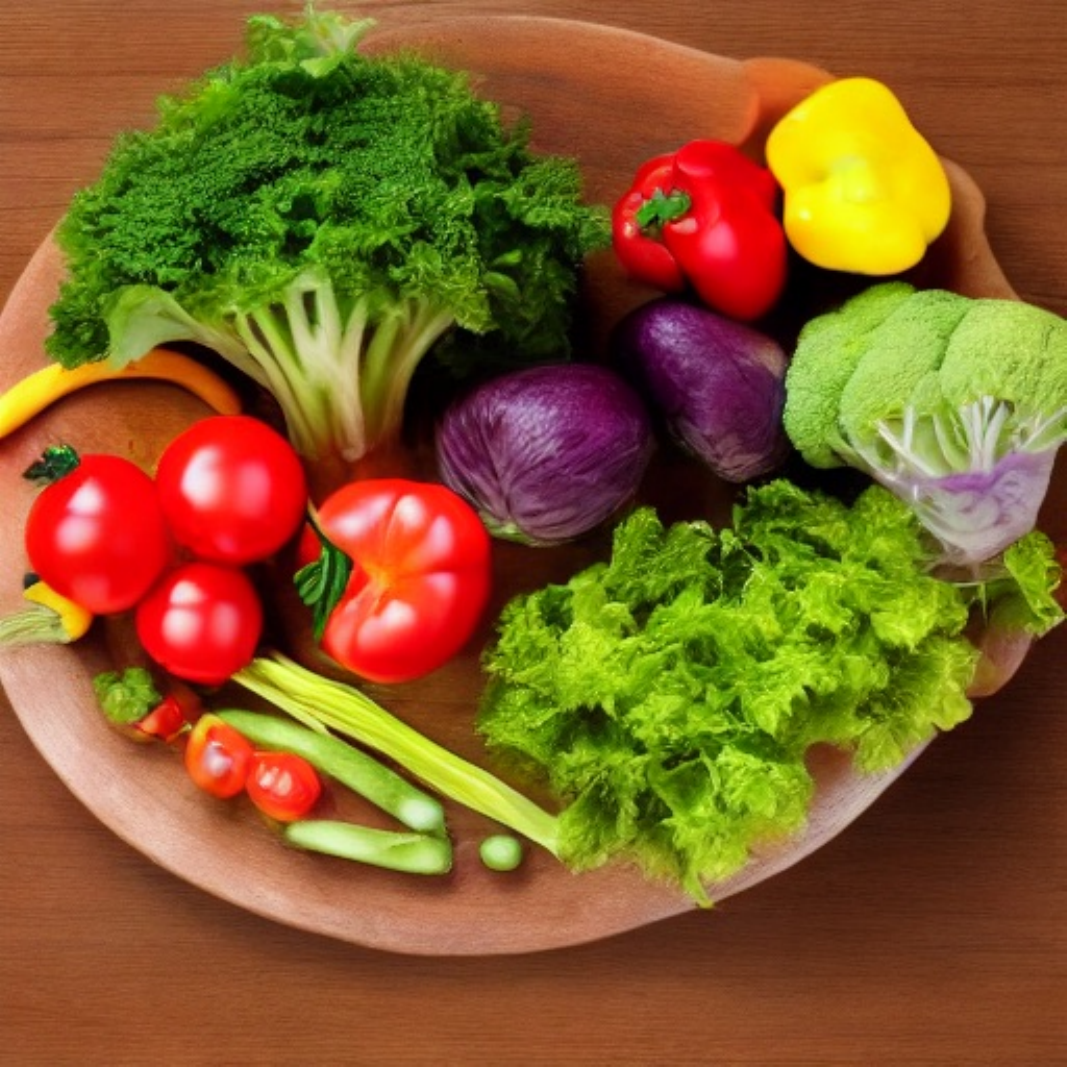}
    
    \vspace{0.5mm}
    \caption*{``A small wooden table covered with delicious vegetables."}
    
    \caption{Steps = 5}
    \label{subfig:nfe5}
  \end{subfigure}
  \caption{Side-by-side comparison of selected images generated with Stable Diffusion and DPM-Solver++ in the low NFE regime (Steps $\in \{4, 5\}$).
    Methods (from left to right): DMN, GITS, LD3, and D2PO.}
  \label{fig:sd_dpm++_nfelow}
\end{figure*}

\begin{figure*}[t] 
  \centering
  
  \captionsetup[subfigure]{font=small, labelfont=small}

  \begin{subfigure}{\linewidth}
    \centering
    
    \begin{minipage}{0.24\linewidth}\centering\small DMN\end{minipage} \hfill
    \begin{minipage}{0.24\linewidth}\centering\small GITS\end{minipage} \hfill
    \begin{minipage}{0.24\linewidth}\centering\small LD3\end{minipage} \hfill
    \begin{minipage}{0.24\linewidth}\centering\small D2PO\end{minipage}
    \vspace{1mm} 
    
    \includegraphics[width=0.24\linewidth]{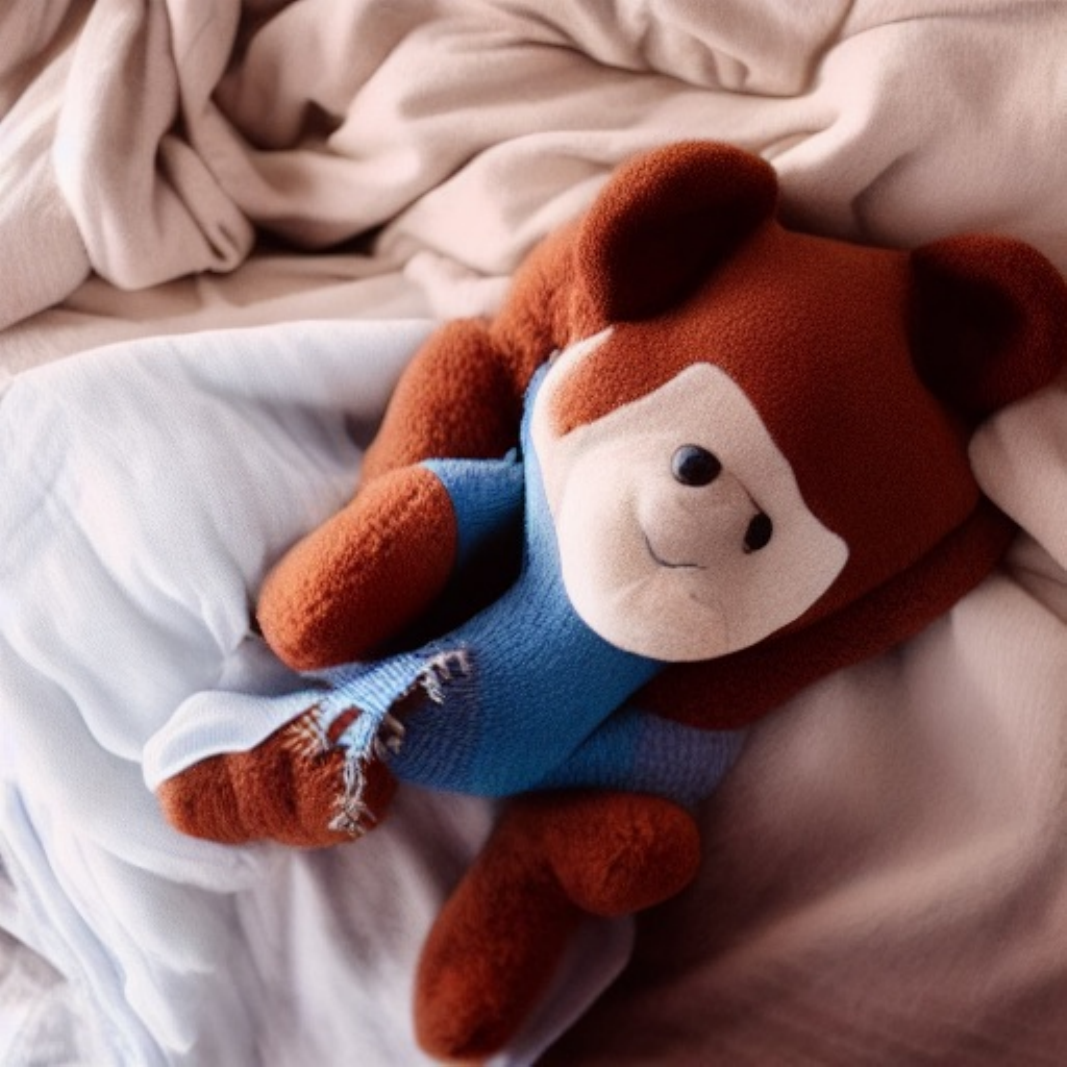}\hfill
    \includegraphics[width=0.24\linewidth]{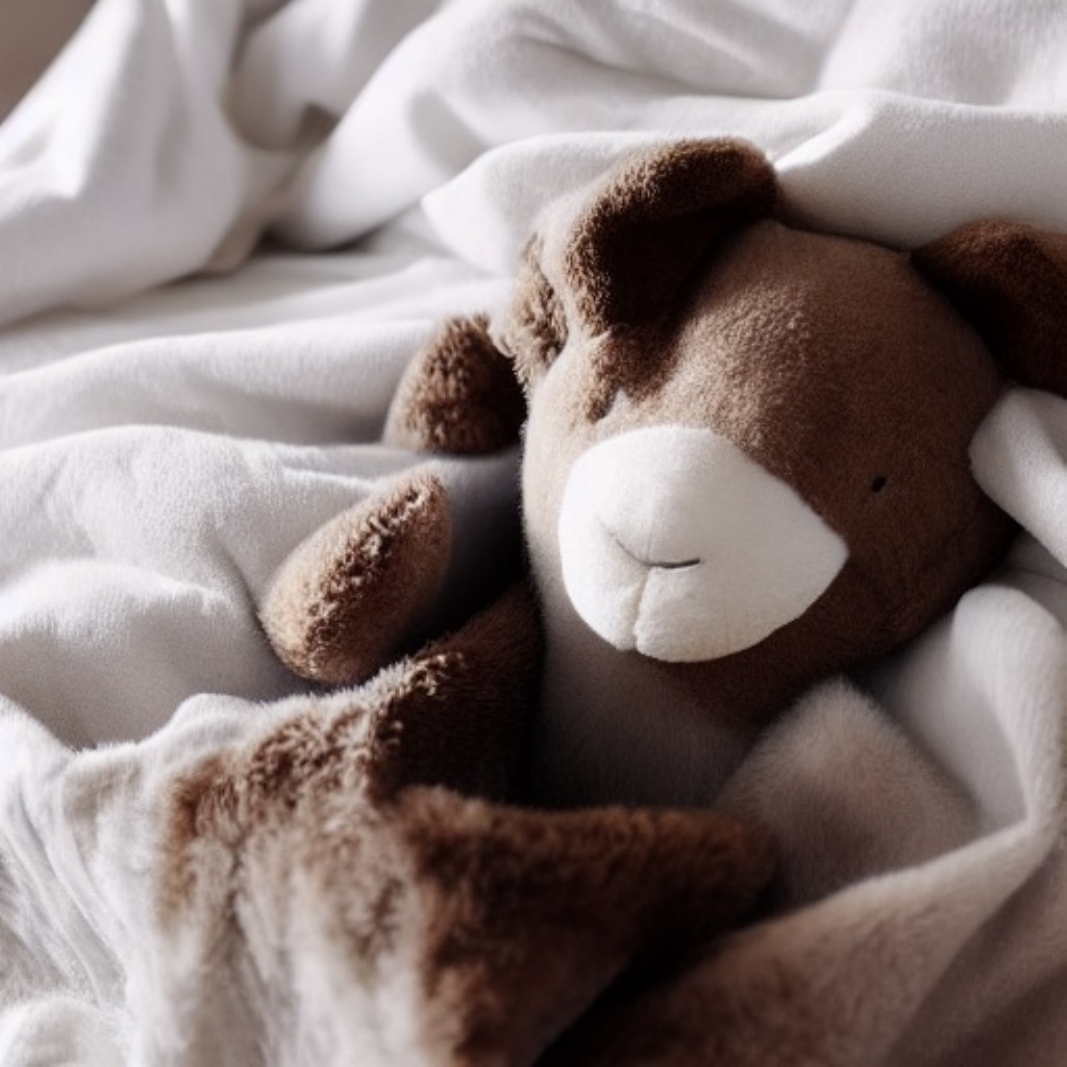}\hfill
    \includegraphics[width=0.24\linewidth]{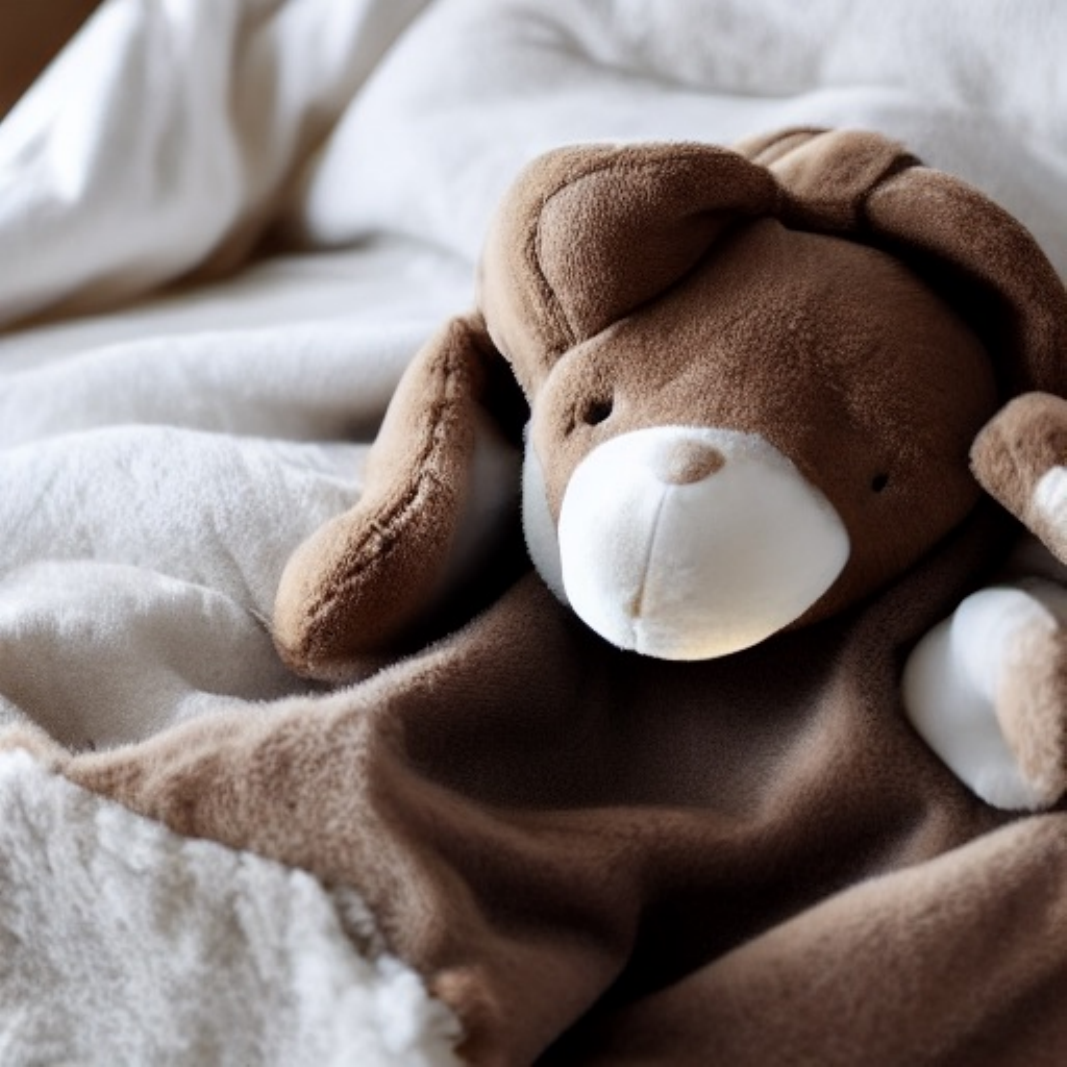}\hfill
    \includegraphics[width=0.24\linewidth]{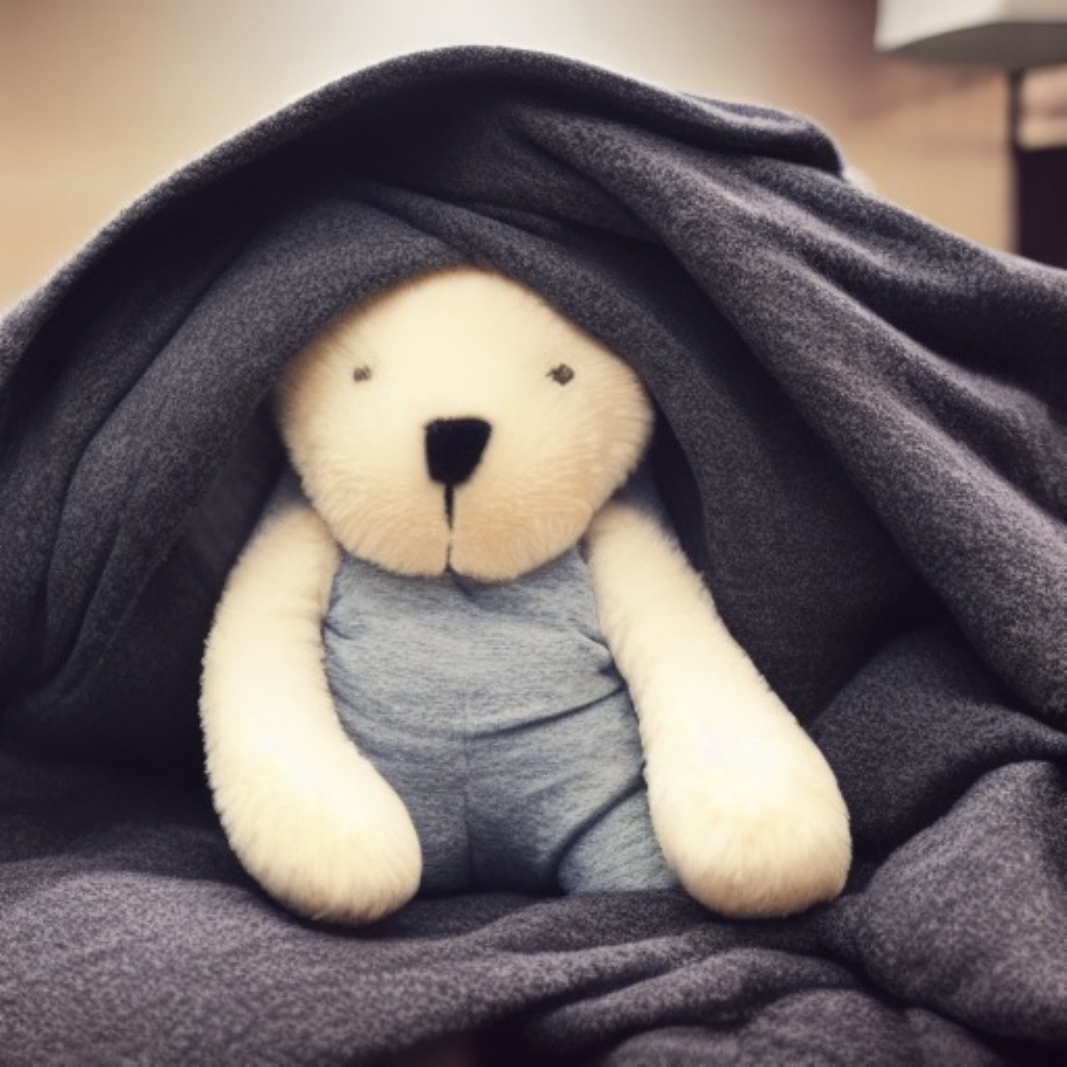}
    
    \vspace{0.5mm} 
    \caption*{\small ``A stuffed animal has been placed inside of blankets."}
    \vspace{1mm} %

    \includegraphics[width=0.24\linewidth]{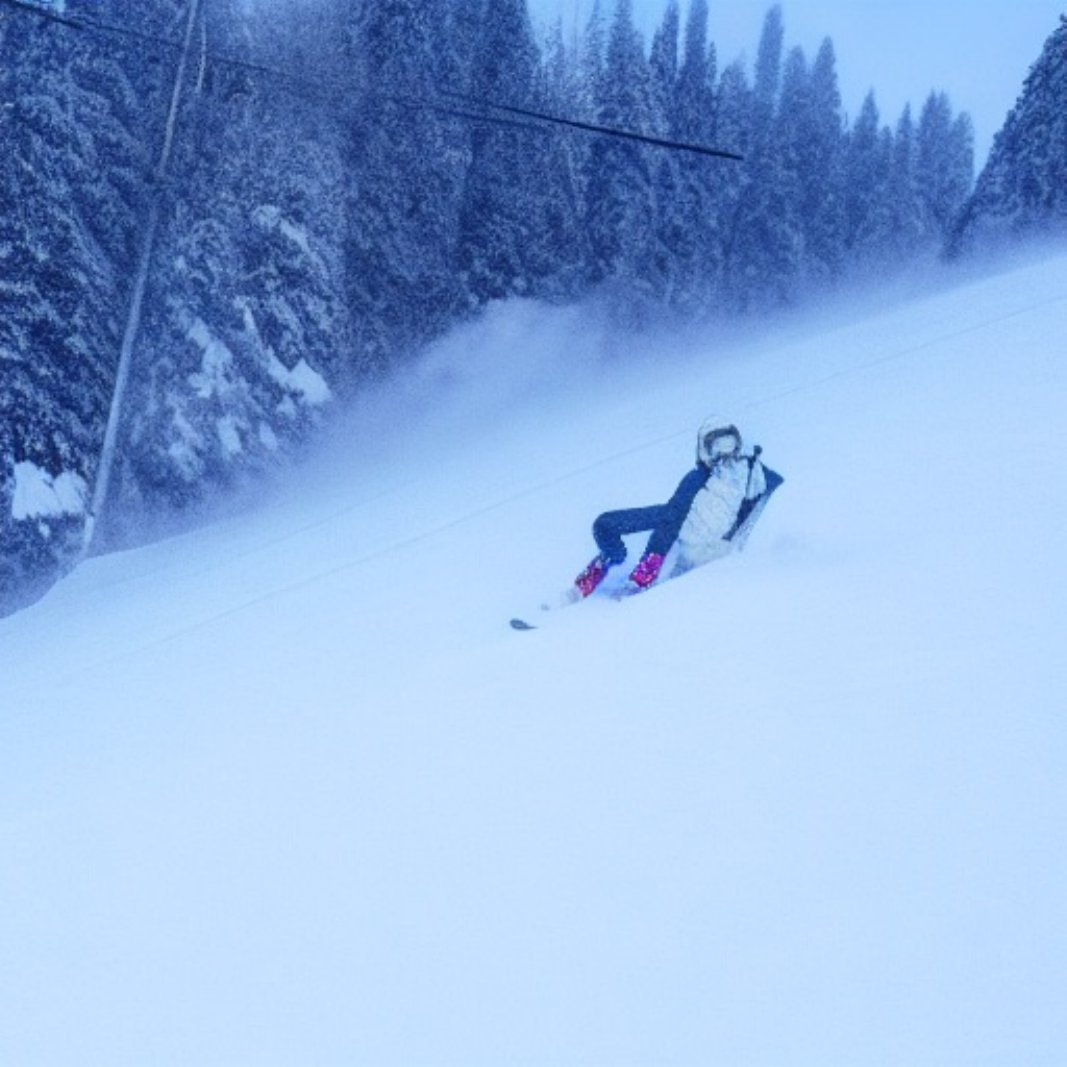}\hfill
    \includegraphics[width=0.24\linewidth]{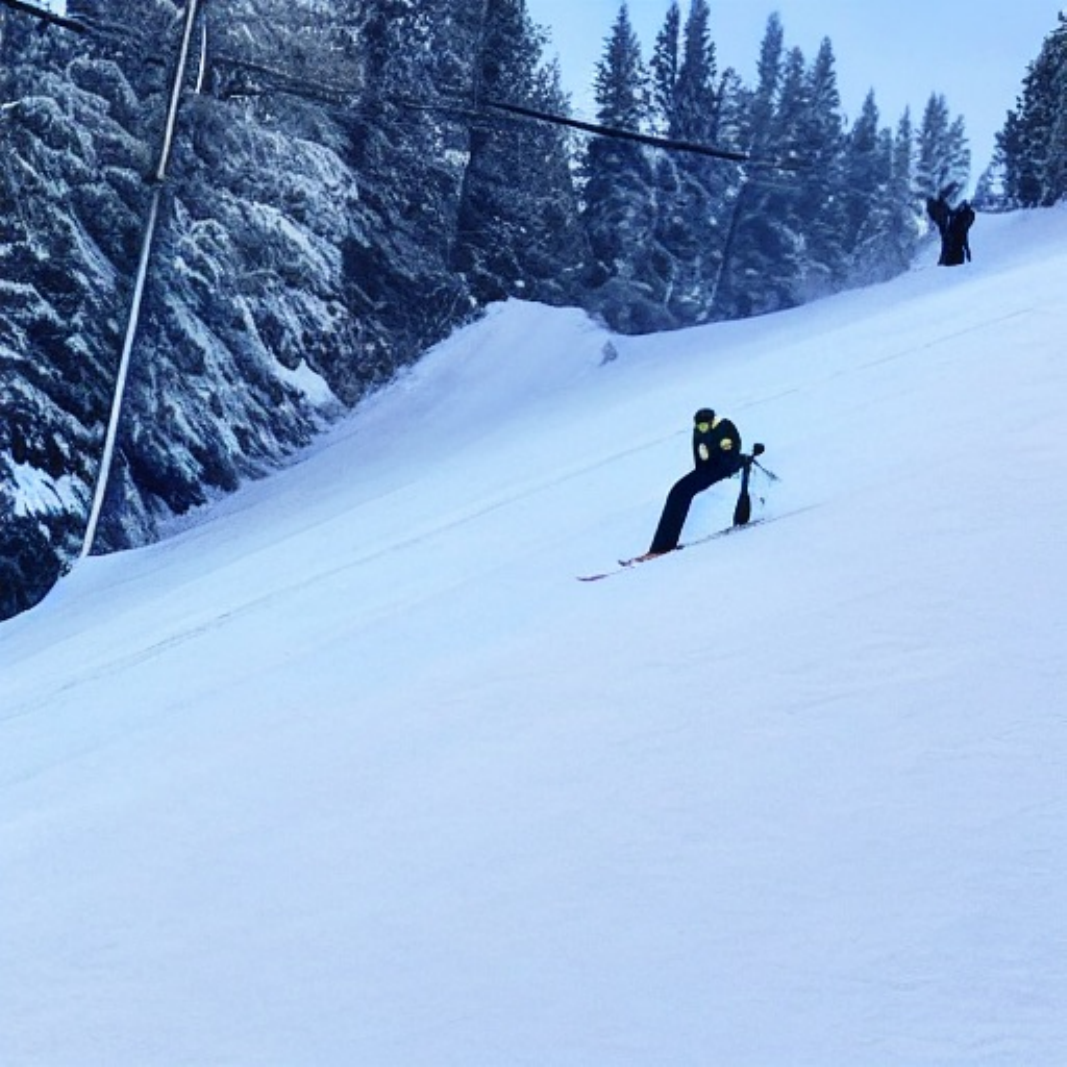}\hfill
    \includegraphics[width=0.24\linewidth]{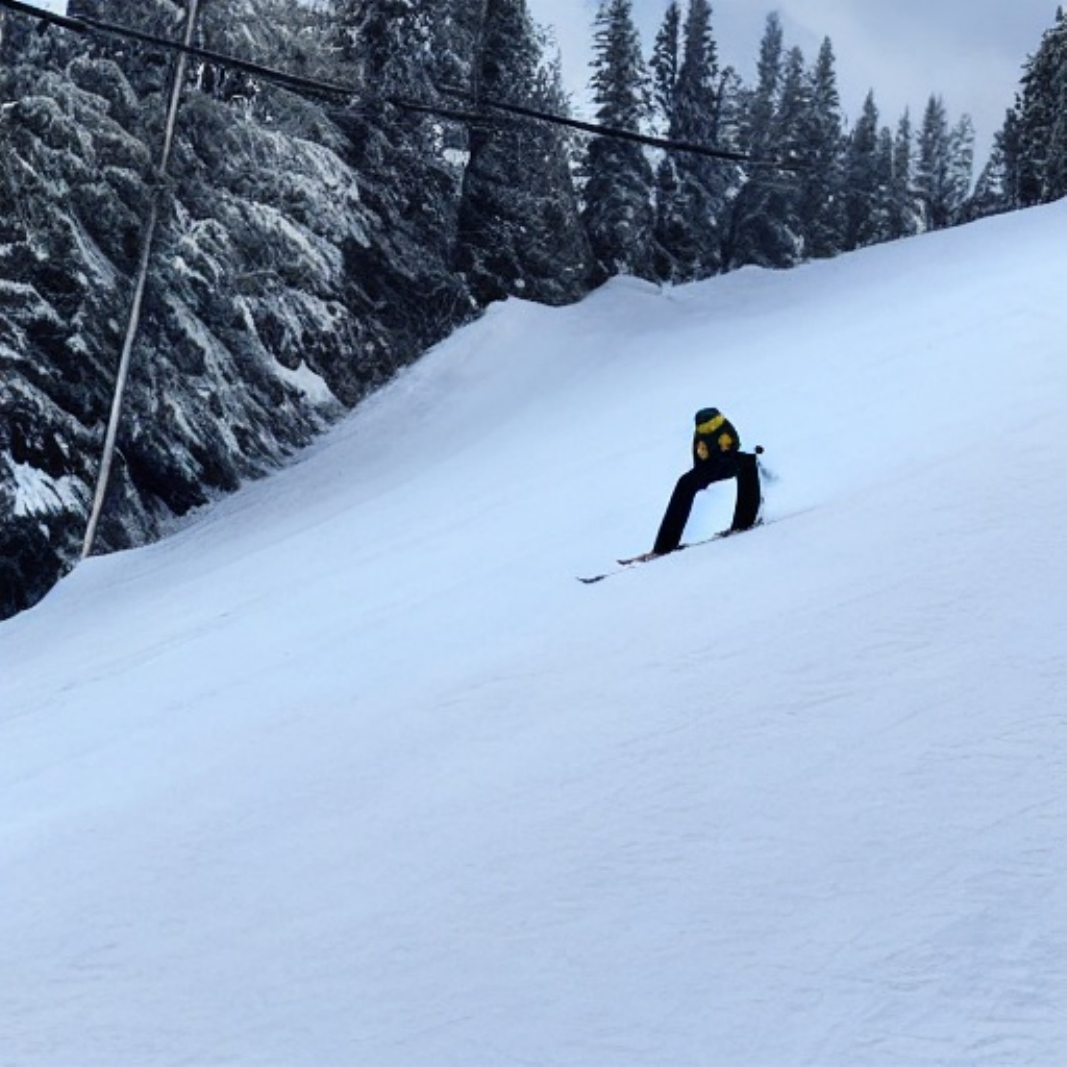}\hfill
    \includegraphics[width=0.24\linewidth]{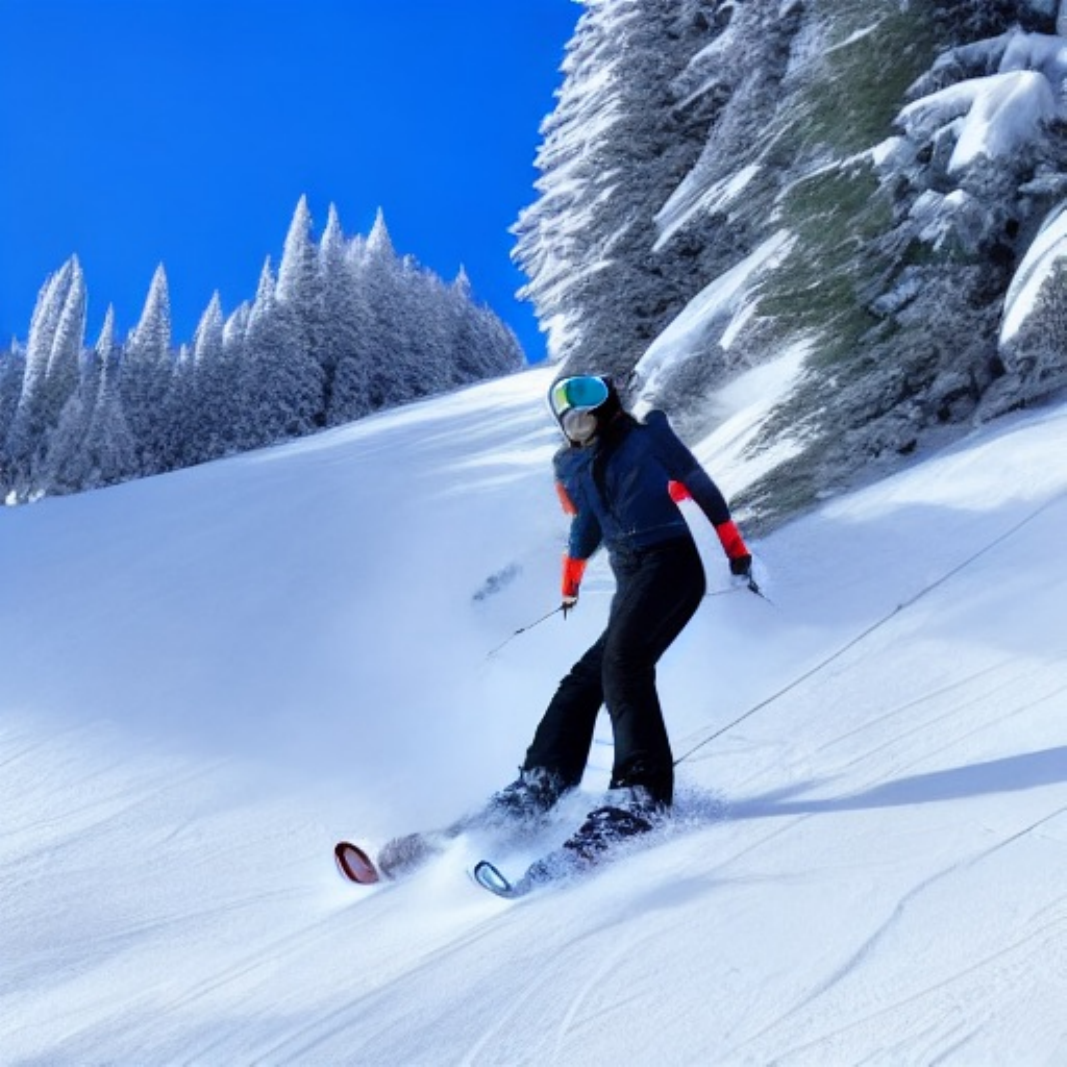}
    
    \vspace{0.5mm}
    \caption*{\small ``A woman wearing a jacket and jeans skiing down a hill."}
    
    \caption{Steps = 6} 
    \label{subfig:nfe6}
  \end{subfigure}
  
  \vspace{1em} %

  \begin{subfigure}{\linewidth}
    \centering
    
    \includegraphics[width=0.24\linewidth]{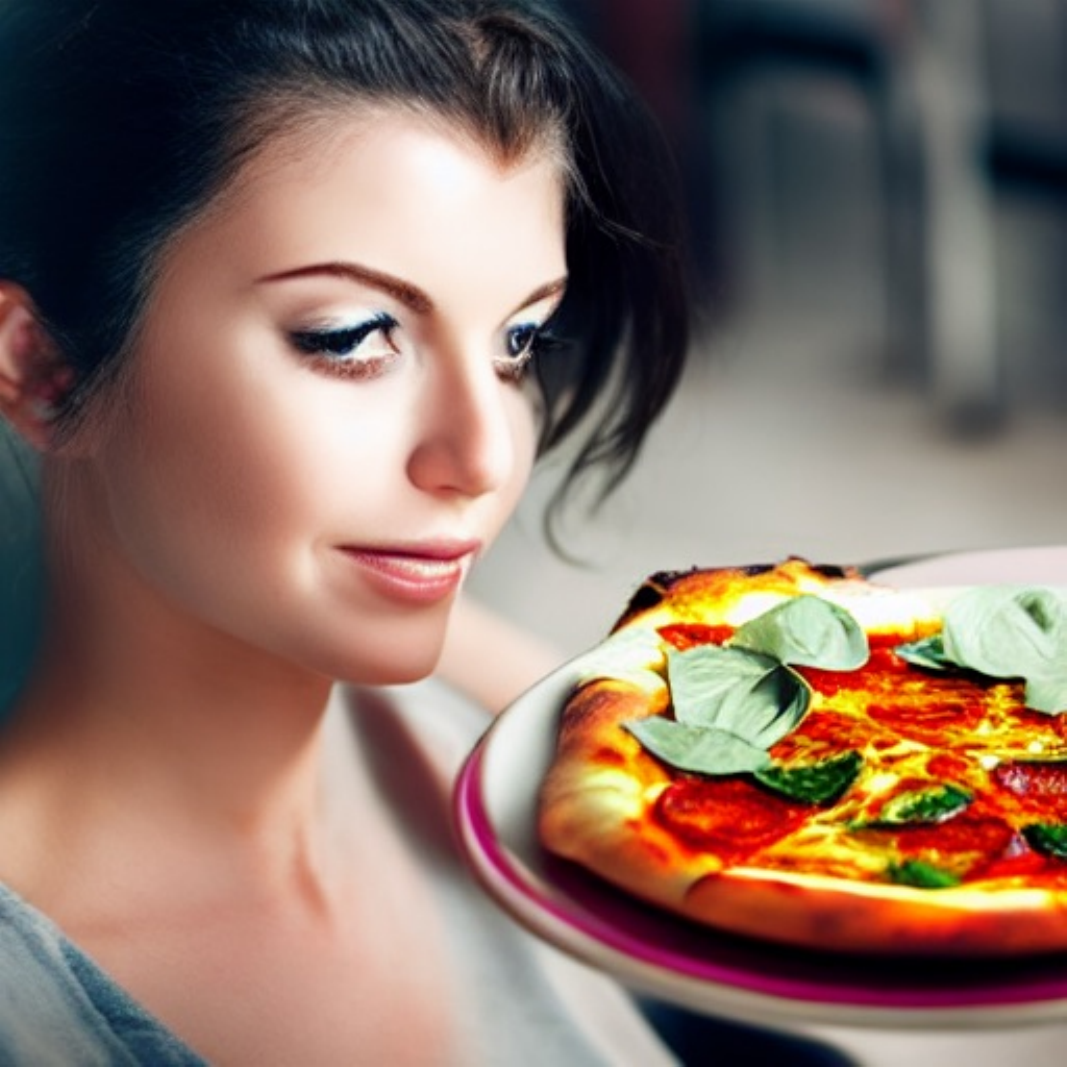}\hfill
    \includegraphics[width=0.24\linewidth]{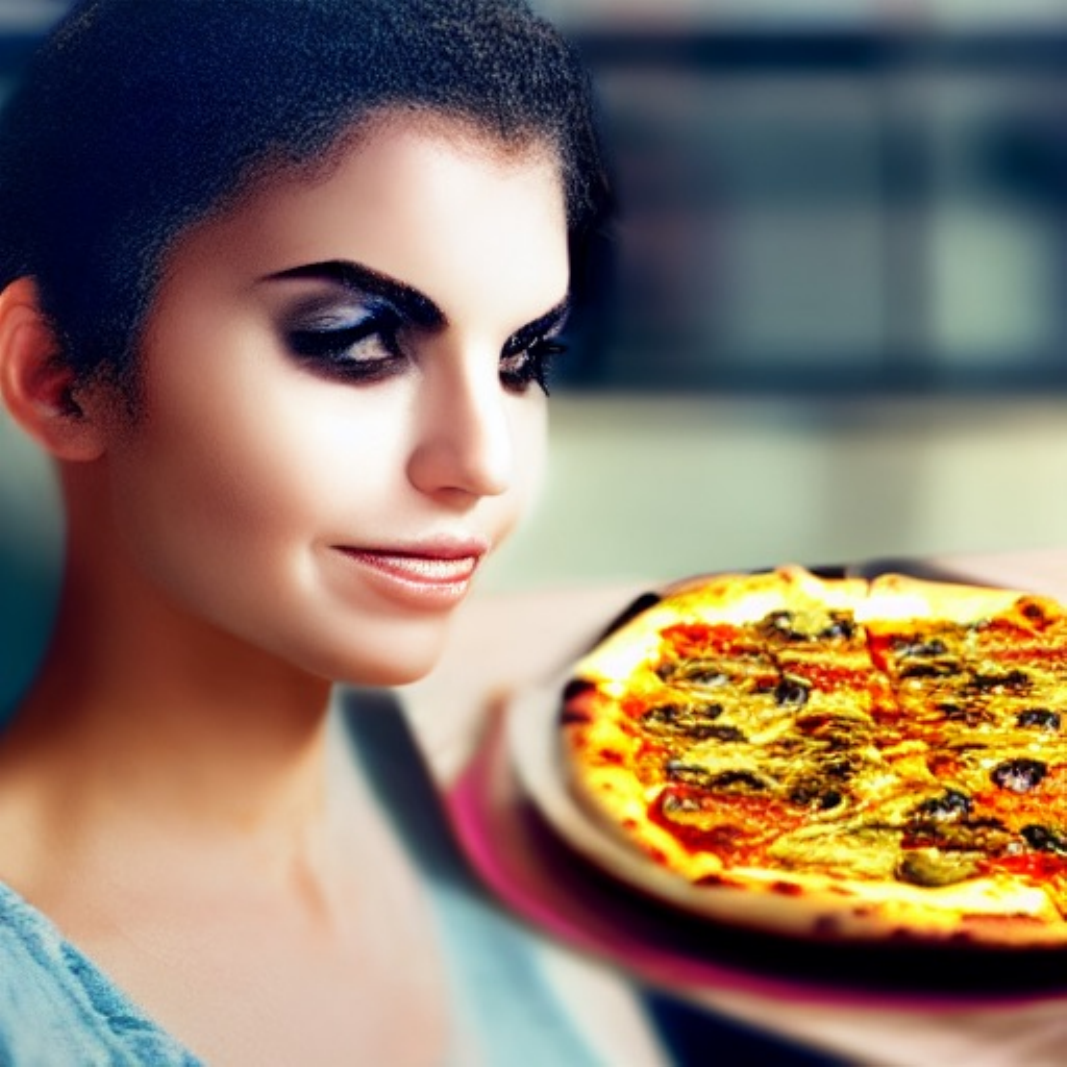}\hfill
    \includegraphics[width=0.24\linewidth]{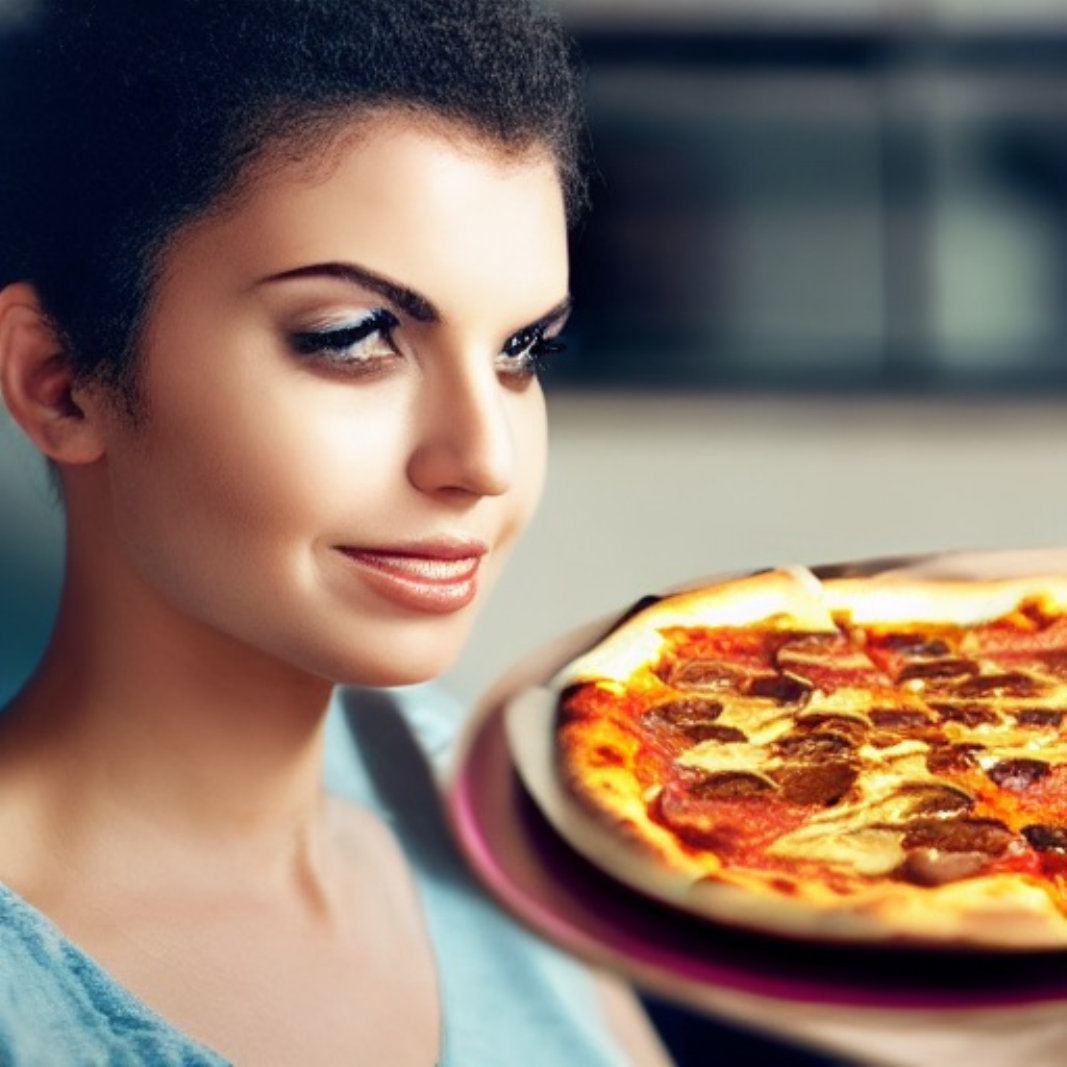}\hfill
    \includegraphics[width=0.24\linewidth]{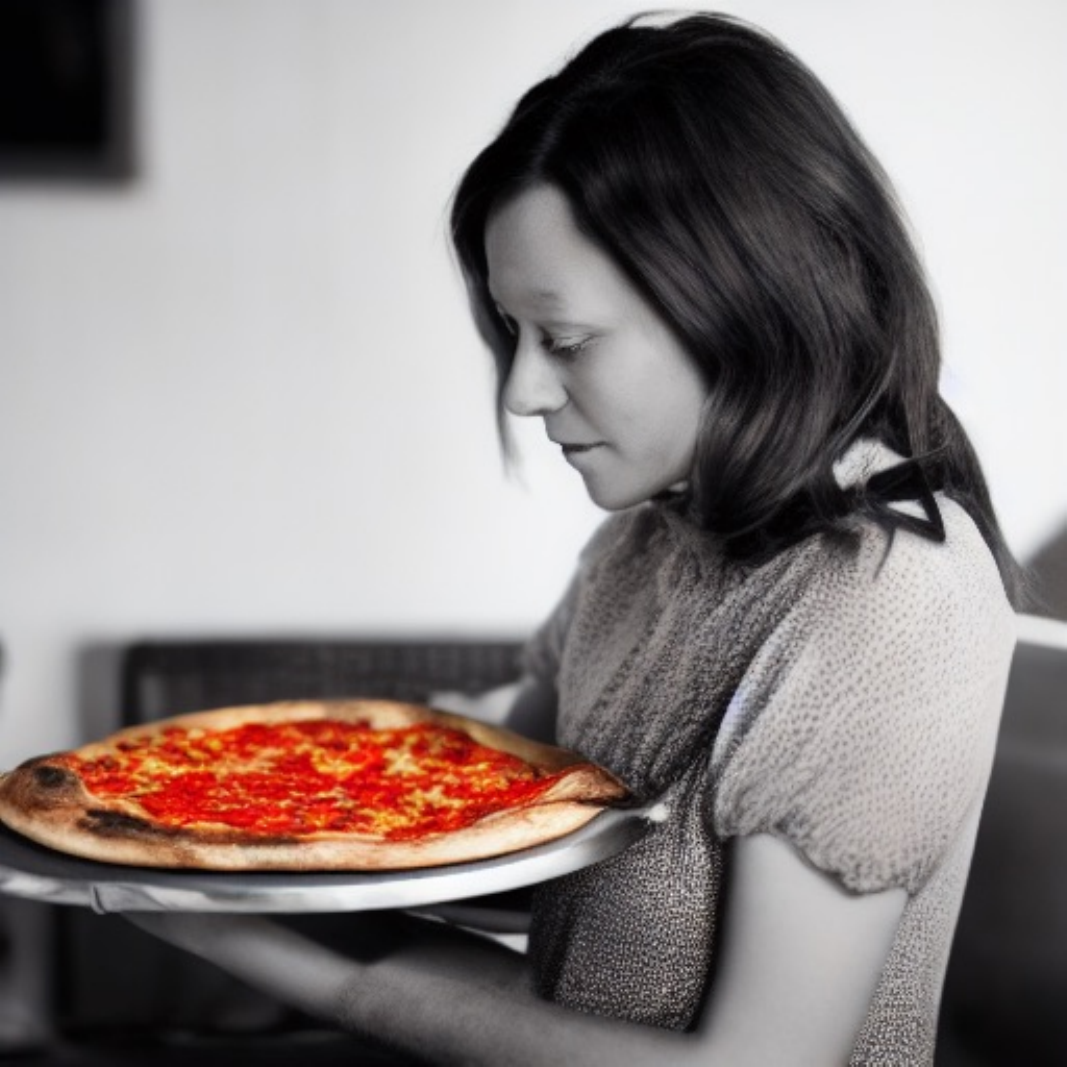}
    
    \vspace{0.5mm}
    \caption*{\small ``A lady staring lovingly into her pizza."}
    \vspace{1mm}

    \includegraphics[width=0.24\linewidth]{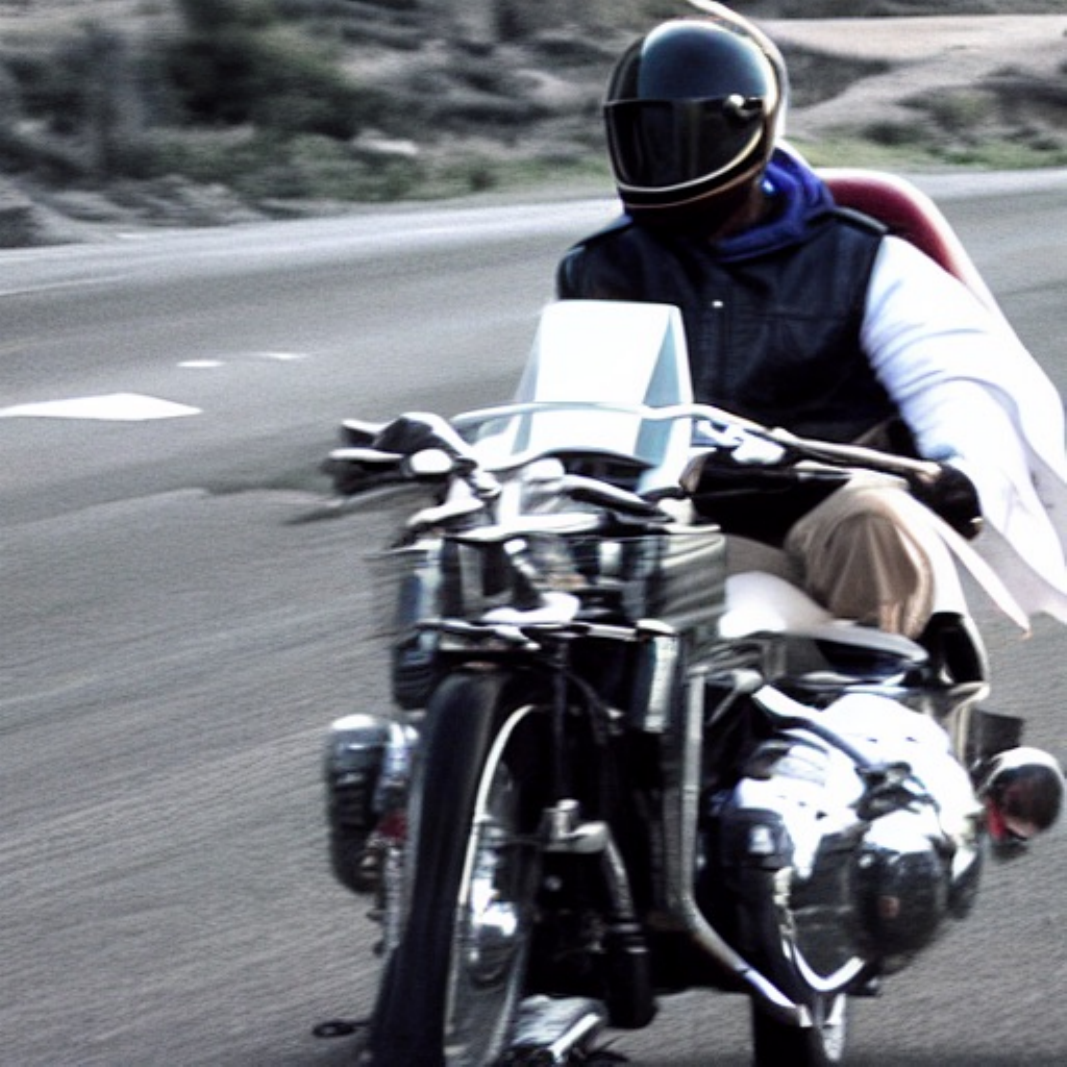}\hfill
    \includegraphics[width=0.24\linewidth]{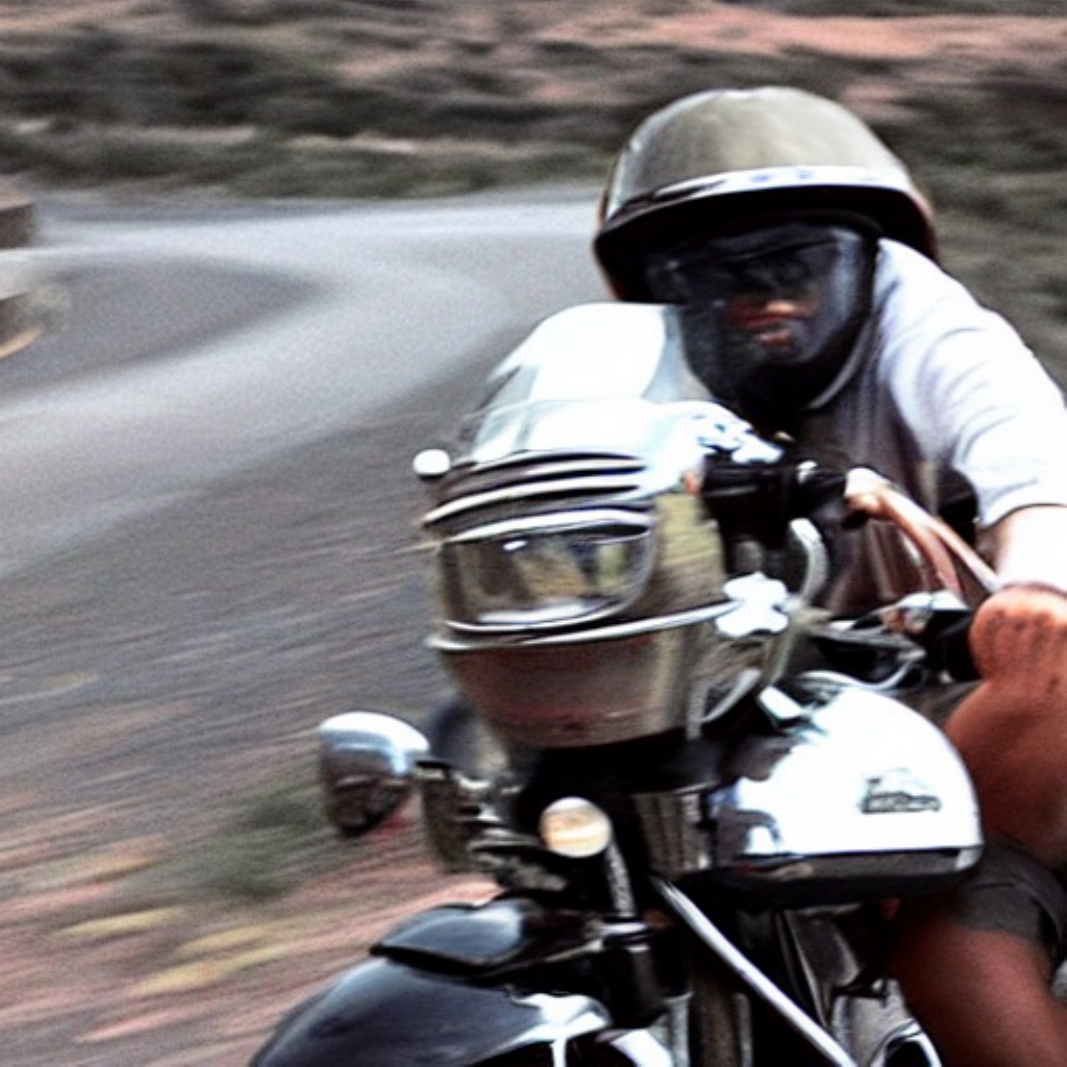}\hfill
    \includegraphics[width=0.24\linewidth]{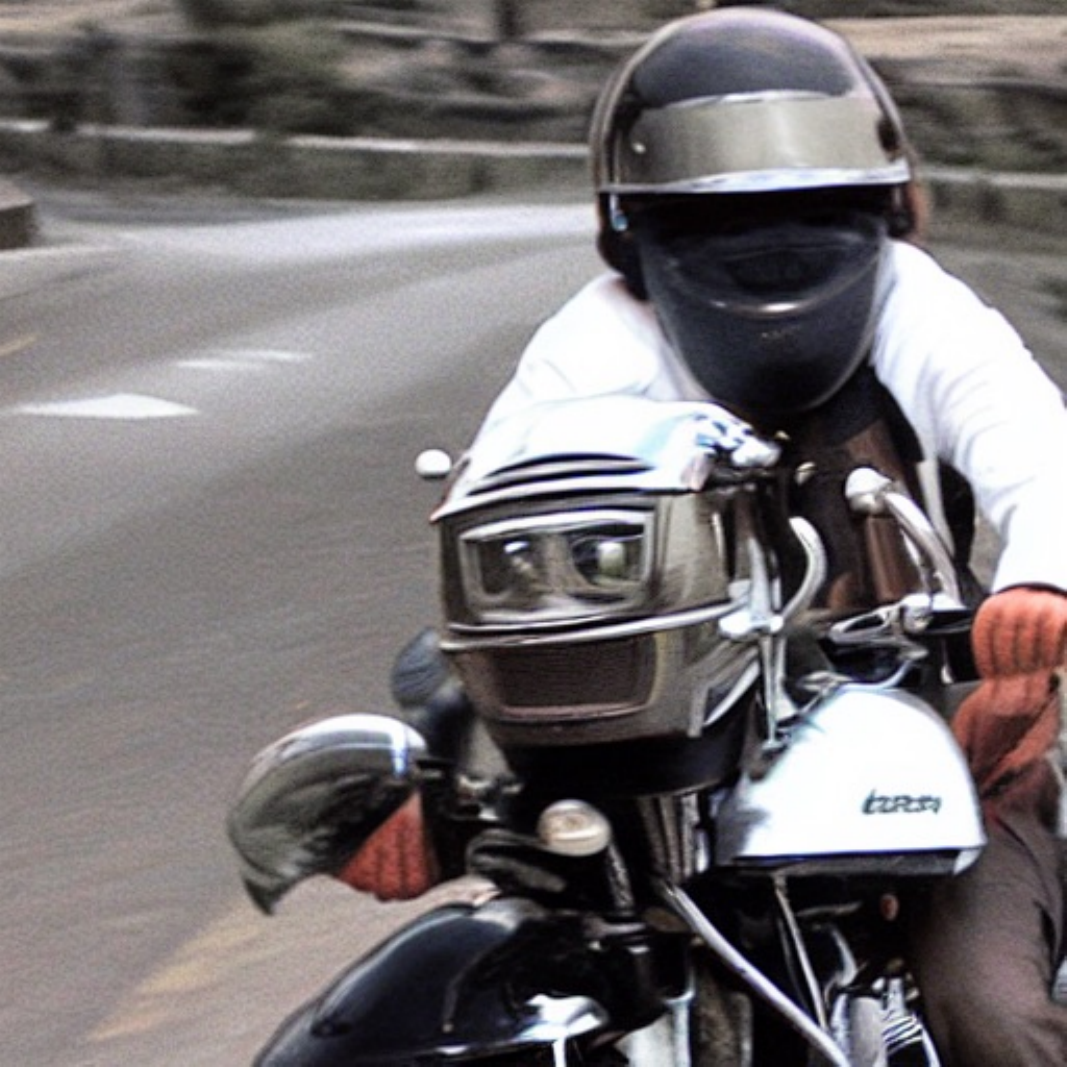}\hfill
    \includegraphics[width=0.24\linewidth]{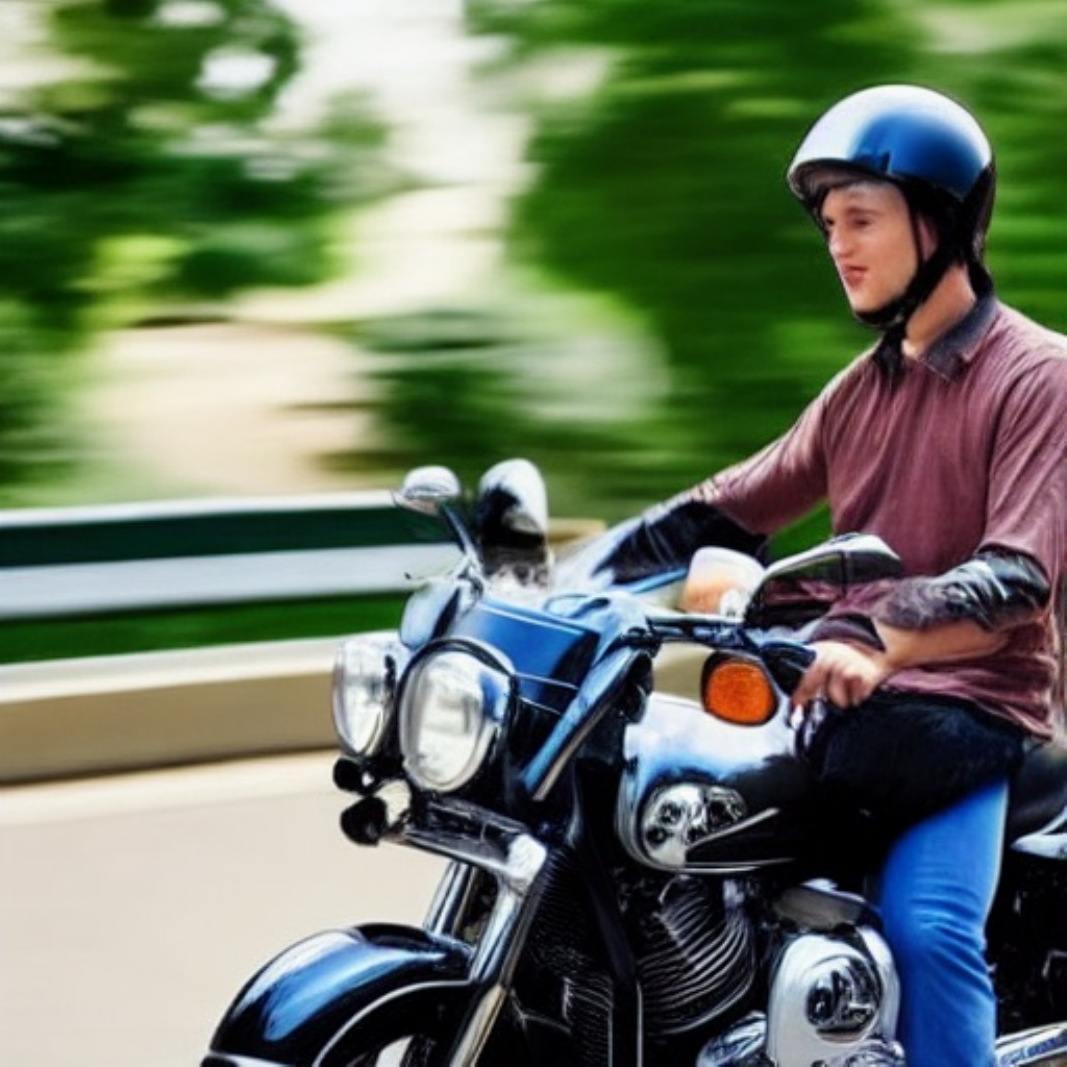}
    
    \vspace{0.5mm}
    \caption*{\small ``A man riding a motorcycle with a helmet on."}
    
    \caption{Steps = 7}
    \label{subfig:nfe7}
  \end{subfigure}
  
  \caption{Side-by-side comparison of selected images generated with Stable Diffusion and DPM-Solver++ in the high NFE regime (Steps $\in \{6, 7\}$).
    Methods (from left to right): DMN, GITS, LD3, and D2PO.}
  \label{fig:sd_dpm++_nfehigh}
\end{figure*}

\section{Additional Qualitative Results}
\label{sec:additional_qualtitative}

In this section, we provide an extensive visual comparison to corroborate the quantitative findings presented in the main paper.
We evaluate Stable Diffusion v1.5 coupled with three advanced ODE solvers (iPNDM, UniPC, DPM-Solver++) with number of steps ranging from 4 to 7.
In the low NFE regime (4-5 steps), D2PO effectively mitigates the structural collapse and artifacts frequently observed in baselines, as shown in \cref{fig:sd_ipndm_nfelow,fig:sd_unipc_nfelow,fig:sd_dpm++_nfelow}.
As the computational budget increases to 6-7 steps, the advantage shifts towards fine-grained details (\cref{fig:sd_ipndm_nfehigh,fig:sd_unipc_nfehigh,fig:sd_dpm++_nfehigh}).
These results confirm that our preference-based optimization is robust across different solver architectures.

\section{Algorithmic Details}
\label{sec:algorithmic_details}

In this section, {Algorithm~\ref{alg:d2po}} outlines the complete training procedure of D2PO (Dynamic Direct Preference Optimization), and {Algorithm~\ref{alg:score_based_distance}} details the computation of our novel score-based distance metric.

\subsection{Training procedure (Algorithm~\ref{alg:d2po})}
The core of D2PO lies in its dynamic preference generation mechanism.
Unlike standard distillation methods that rely on a fixed teacher, D2PO iteratively refines the student policy to create a {dynamic winner} sample $\boldsymbol{x}_w$.
As shown in Algorithm~\ref{alg:d2po}, the winning policy $\phi'$ is derived on-the-fly by refining the current student parameters (e.g., via timestep interpolation).
The model is then optimized with the DPO loss, which favors the winning sample $\boldsymbol{x}_w$ over the degraded losing sample $\boldsymbol{x}_l$, both measured relative to the reference policy $\phi_{\text{ref}}$.

\subsection{Score-based distance (Algorithm~\ref{alg:score_based_distance})}
A critical component of our objective is the energy function used to define the preference.
Instead of relying on pixel-space metrics (e.g., MSE) or external networks (e.g., LPIPS), we leverage the pre-trained diffusion model itself as a critic.
Algorithm~\ref{alg:score_based_distance} describes this procedure.
Given two samples, we perturb both with the same noise $\boldsymbol{\epsilon}$ at a randomly sampled timestep $t$, and pass them through the pretrained network $\epsilon_\theta$. The distance $d_\theta$ is the squared $\ell_2$ difference between the two noise predictions, which (up to the fixed factor $\sigma_t^2$ absorbed by our weighting) equals the score discrepancy at level $t$.
This metric effectively captures the discrepancy between the sample's trajectory and the vector field of the pre-trained diffusion prior, providing a fine-grained signal for structural and textural alignment.

\clearpage

\begin{algorithm}[t]
   \caption{D2PO (Dynamic Direct Preference Optimization)}
   \label{alg:d2po}
   \centering
   \setlength{\abovecaptionskip}{0pt}
   \setlength{\belowcaptionskip}{0pt}
   \begin{algorithmic}[1]
      \STATE {\bfseries Input:} {Learnable params $\phi = \{\mathcal{S}, \boldsymbol{\omega}\}$, $\beta$, learning rate $\eta$, EMA momentum $\lambda$, Sampler $\Phi_\phi(\cdot)$, Degradation operator $\mathcal{G}(\cdot)$}

      \STATE Initialize learnable sampler parameters $\phi$
      \STATE Initialize reference policy $\phi_{\text{ref}} \leftarrow \phi$

      \WHILE{not converged}
         \STATE Sample batch of contexts $(c, \boldsymbol{x}_T)$

         \STATE \textit{\# 1. Generate student and reference outputs}
         \STATE $\boldsymbol{x}_{\phi} \leftarrow \Phi_\phi(c, \boldsymbol{x}_T)$
         \STATE $\boldsymbol{x}_{\phi_{\text{ref}}} \leftarrow \Phi_{\phi_{\text{ref}}}(c, \boldsymbol{x}_T)$

         \STATE \textit{\# 2. Generate dynamic preference (Sec 4.4)}
         \STATE $\phi' \leftarrow \text{Refine}(\phi)$
         \STATE $\boldsymbol{x}_w \leftarrow \Phi_{\phi'}(c, \boldsymbol{x}_T)$
         \STATE $\boldsymbol{x}_l \leftarrow \mathcal{G}(\mathrm{sg}[\boldsymbol{x}_\phi])$

         \STATE \textit{\# 3. Compute final loss}
         \STATE $L_w \leftarrow \beta \left( d_\theta(\boldsymbol{x}_w, \boldsymbol{x}_{\phi_{\text{ref}}}) - d_\theta(\boldsymbol{x}_w, \boldsymbol{x}_{\phi}) \right)$
         \STATE $L_l \leftarrow \beta \left( d_\theta(\boldsymbol{x}_l, \boldsymbol{x}_{\phi_{\text{ref}}}) - d_\theta(\boldsymbol{x}_l, \boldsymbol{x}_{\phi}) \right)$
         \STATE $\mathcal{L}_\text{D2PO} \leftarrow -\log \sigma ( L_w - L_l )$

         \STATE \textit{\# 4. Update Parameters}
         \STATE $\phi \leftarrow \phi - \eta \nabla_\phi \mathcal{L}_\text{D2PO}$
	\STATE $\omega_{\text{ref}} \leftarrow \lambda\,\omega_{\text{ref}} + (1-\lambda)\,\omega$
	\IF{end of epoch}
	   \STATE $\mathcal{S}_{\text{ref}} \leftarrow \mathcal{S}$
	\ENDIF
      \ENDWHILE

      \STATE {\bfseries Return} $\phi$
   \end{algorithmic}
\end{algorithm}

\begin{algorithm}[t]
   \caption{Score-based Distance}
   \label{alg:score_based_distance}

\begin{algorithmic}[1]
   \STATE {\bfseries Input:} {Two samples $(\boldsymbol{x}_a, \boldsymbol{x}_b)$, noise prediction network $\epsilon_\theta(\cdot, \cdot)$, noise schedule $(\alpha(t), \sigma(t))$, uniform range $[t_{\min}, t_{\max}]$, conditioning $c$}
   \STATE Sample a single time step $t \sim \mathcal{U}(t_{\min}, t_{\max})$
   \STATE Sample random noise $\boldsymbol{\epsilon} \sim \mathcal{N}(0, \mathbf{I})$

   \STATE \textit{\# 1. Compute forward-marginal parameters}
   \STATE  $\alpha_t \leftarrow \alpha(t)$ 
   \STATE  $\sigma_t \leftarrow \sigma(t)$

   \STATE \textit{\# 2. Construct noisy samples}
   \STATE  $\boldsymbol{x}_{a,t} = \alpha_t \boldsymbol{x}_a + \sigma_t \boldsymbol{\epsilon}$
   \STATE  $\boldsymbol{x}_{b,t} = \alpha_t \boldsymbol{x}_b + \sigma_t \boldsymbol{\epsilon}$

  \STATE \textit{\# 3. Predict noise}
   \STATE  $\hat{\boldsymbol{\epsilon}}_a = \epsilon_\theta(\boldsymbol{x}_{a,t}, t; c)$ 
   \STATE  $\hat{\boldsymbol{\epsilon}}_b = \epsilon_\theta(\boldsymbol{x}_{b,t}, t; c)$

   \STATE \textit{\# 4. Compute the score-based energy}
   \STATE $d_{\theta} = \|\hat{\boldsymbol{\epsilon}}_a - \hat{\boldsymbol{\epsilon}}_b\|_2^2$ 

   \STATE {\bfseries Return} $d_{\theta}$
\end{algorithmic}
\end{algorithm}

\clearpage

\end{document}